%% file: thesis.tex
\documentclass{uark}

\usepackage{amsmath} %
\usepackage{mathcomp} %
\usepackage{graphicx} %
\usepackage[version=3]{mhchem} %
\usepackage{siunitx} %
\usepackage{wrapfig,lipsum,booktabs}

\input{packages}

\usepackage[margin=1in]{geometry}

\begin{document}

\bibliographystyle{IEEEtran}

\include{Chapters/front_matter}

\pagenumbering{arabic}

\include{Chapters/Chaps/chap-1-introduction}

\include{Chapters/Chaps/chap-2-related-work}
\include{Chapters/Chaps/chap-3-domain-adaptation}

\include{Chapters/Chaps/chap-4-continual-learning}
\include{Chapters/Chaps/chap-5-cross-view-learning}

\include{Chapters/Chaps/chap-6-multimodal-tempoeral-learning}

\include{Chapters/Chaps/chap-7-conclusion}

\bibliography{thesis}

\appendix

\include{Chapters/back_matter}
\end{document}

%% file: packages.tex
\usepackage{url}
\usepackage{footnote}
\usepackage[bottom]{footmisc}
\makeatletter
\g@addto@macro{\UrlBreaks}{\UrlOrds}
\makeatother
\usepackage{epsfig}
\usepackage{graphicx}
\usepackage{amsmath}
\usepackage{amssymb}
\usepackage{bbm}
\usepackage{soul}

\usepackage{float}
\usepackage{pifont}%

\newcommand{\xmark}{\ding{55}}%
\newcommand{\cmark}{\ding{51}}%

\usepackage{xspace}

\def\greycell{\cellcolor[HTML]{C0C0C0}}

\usepackage{multirow}
\usepackage{color, colortbl}

\usepackage{tcolorbox}
\usepackage{tabularx}
\usepackage{arydshln}
\usepackage{caption}

\usepackage{tikz}
\usetikzlibrary{bayesnet}
\usetikzlibrary{arrows}
\usepackage{float}
\newcommand*\rot{\rotatebox{90}}

\newcounter{propositioncounter}
\def\thepropositioncounter{\thesection.\arabic{propositioncounter}}
\newcommand\showpropositioncounter{\refstepcounter{propositioncounter}\thepropositioncounter}

\newcounter{remarkcounter}
\def\theremarkcounter{\thesection.\arabic{remarkcounter}}
\newcommand\showremarkcounter{\refstepcounter{remarkcounter}\theremarkcounter}

\newcounter{pubcounter}
\newcommand\showpubcounter{\refstepcounter{pubcounter}[\thepubcounter] }

%% file: Chapters/front_matter.tex
\title{Towards Robust and Fair Vision Learning in Open-World Environments}

\author{Thanh-Dat Truong}
\degreeyear{2024}

\degreetitle{Doctor of Philosophy in Computer Science}

\input{Chapters/thesis-abstract}

\begin{frontmatter}

\makefrontmatter

\input{Chapters/thesis-acknowledgement}

    \tableofcontents
    \listoffigures  %
    \listoftables   %

\newpage

\input{Chapters/thesis-publications}

\end{frontmatter}

%% file: Chapters/thesis-abstract.tex
\abstract{
The rapid increase of large-scale data and high-performance computational hardware has promoted the development of data-driven machine vision approaches. Advanced deep learning approaches have achieved remarkable performance in various vision problems and are closing the capability gap between artificial intelligence (AI) and humans. However, towards the ultimate goal of AI, which replicates human ability in visual perception tasks, the machine vision learning methods still need to address several ill-posed challenges. First, while the current vision learning methods often rely on large-scale annotated data, the data annotation process is a costly and time-consuming process. Second, the unfaired predictions produced by vision models due to the imbalance of data distribution, known as fairness, pose a significant concern in practical deployment, especially in human-related applications. Third, since human perceptions interpret the world in the open-vocabulary approach with diverse categories and concepts, the current vision machine frameworks should be capable of continually learning new concepts. Fourth, although the current deep learning-based vision approaches achieved impressive performance, their knowledge representations are often uninterpretable. These challenges motivate this dissertation to develop novel approaches toward fairness and robustness in vision learning.
\\
\indent
To address these challenges, the dissertation presents four key contributions toward fairness and robustness in vision learning. First, to address the problem of large-scale data requirements, the dissertation presents a novel Fairness Domain Adaptation approach derived from two major research findings. In particular, the thesis proposes a novel Bijective Maximum Likelihood to Unsupervised Domain Adaptation followed by introducing a novel Fairness Adaptation Learning Framework. Second, to enable the capability of open-world modeling of vision learning, this dissertation presents a novel Open-world Fairness Continual Learning Framework. The success of this research direction is the result of two research lines, i.e., Fairness Continual Learning and Open-world Continual Learning. Third, since visual data are often captured from multiple camera views, robust vision learning methods should be capable of modeling invariant features across views. To achieve this desired goal, the research in this thesis will present a novel Geometry-based Cross-view Adaptation framework to learn robust feature representations across views. 
Finally, with the recent increase in large-scale videos and multimodal data, understanding the feature representations and improving the robustness of large-scale visual foundation models is critical. 
Therefore, this thesis will present novel Transformer-based approaches to improve the robust feature representations against multimodal and temporal data. By introducing new self-attention mechanisms and learning objectives to Transformer networks for multimodal and video understanding, this research line has provided a comprehensive and better understanding of multimodal and temporal feature representations.
Then, I will present a novel Domain Generalization Approach to improve the robustness of visual foundation models. 
My research's theoretical analysis and experimental results have shown the effectiveness of the proposed approaches, demonstrating their superior performance compared to prior studies. I am confident that the contributions in this dissertation have advanced the fairness and robustness of machine vision learning.
}

%% file: Chapters/thesis-acknowledgement.tex
\begin{acknowledgements} 

I would like to thank the Department of Electrical Engineering and Computer Science, University of Arkansas, for giving me the opportunity to accomplish my research study. Besides, I would like to express our gratitude to all those who have supported, influenced, and helped me in the process that ultimately resulted in this dissertation.

First and foremost, I would like to thank my advisor, Prof. Khoa Luu, for being our greatest advisor and providing me with invaluable and uncountable guidance and encouragement. 
Your endless support, insightful feedback, and constant encouragement have been invaluable throughout my research.
Your expertise and passion for research have inspired me to push the boundaries of my work, and I am incredibly fortunate to have had the opportunity to learn from you.

I would also like to thank my doctoral committee members, Dr. Susan Gauch, Dr.  John Gauch, Dr. Ashley Dowling, Dr. Bhiksha Raj, and Dr. Khoa Luu.
Your thoughtful critiques and suggestions have significantly improved the quality of this dissertation.
I deeply appreciate your dedication to my progress and your willingness to share your knowledge and time.
My research and dissertation will never be successful without your support.
It is my honor to have such a distinguished thesis panel.

Special thanks go to my colleagues and friends in the Computer Vision and Image Understanding Lab. The collaborative environment, stimulating discussions, and mutual support have made my Ph.D. journey both productive and enjoyable.
I would like to express my gratitude to Dr. Chi Nhan Duong, Dr. Quach Kha Gia, and Dr. Utsav Prabhu for their mentoring and support during my Ph.D. journey.
I am especially grateful to Xuan-Bac Nguyen, Anh Pha Nguyen, Raviteja NVS Chappa, Manuel Serna-Aguilera, Hoang-Quan Nguyen, Hao Van, and Trong-Thuan Nguyen for your friendship and for being my sounding boards during challenging times.

I would like to acknowledge the financial support provided by the University of Arkansas, NSF DART Project, Google Research, Arkansas Biosciences Institute, Arkansas High Performance Computing Center, and SolaRid LLC. This research would not have been possible without the resources and opportunities made available through your funding.
On a personal note, I would like to thank my family. To my parents and brother, your support and encouragement have been my foundation. Your belief in me has kept me going, even when the path was unclear. 

Finally, I want to express my gratitude to all those who have contributed to my Ph.D. journey, whether through academic, personal, or professional support. This dissertation is as much a product of your contributions as it is of my efforts.

\end{acknowledgements}

%% file: Chapters/thesis-publications.tex
\begin{publications}
\item \showpubcounter
\textbf{Thanh-Dat Truong}, Ashley PG Dowling, Randy J. Sasaka, and Khoa Luu. \textbf{{Sensor-based Smart Insect Monitoring System in the Wild.}} US Patent, No. PCT/US2023/021330.

\item \showpubcounter
\textbf{Thanh-Dat Truong}, Ashley PG Dowling, and Khoa Luu. \textbf{{Smart Insect Control Device via Artificial Intelligence in Realtime Environment.}} US Patent, No. PCT/US2023/021307.

\item \showpubcounter
\textbf{Thanh-Dat Truong}, Utsav Prabhu, Bhiksha Raj, Susan Gauch, and Khoa Luu. {\textbf{EAGLE: Efficient Adaptive Geometry-based Learning in Cross-view Vision-Language Understanding}}. In Proceedings of the Thirty-eighth Conference on Neural Information Processing Systems (NeurIPS), 2024. 

\item \showpubcounter
\textbf{Thanh-Dat Truong} and Khoa Luu. {\textbf{action Recognition Understanding From Exocentric to Egocentric Perspective}}. Journal of Neurocomputing. 2024.

\item \showpubcounter
Hoang-Quan Nguyen$^{*}$, \textbf{Thanh-Dat Truong}$^{*}$, Xuan Bac Nguyen, Ashley Dowling, Xin Li, and Khoa Luu. {\textbf{Insect-Foundation: A Foundation Model and Large-scale 1M Dataset for Visual Insect Understanding}}. In Proceedings of the IEEE/CVF Conference on Computer Vision and Pattern Recognition (CVPR), pp.~21945-21955. 2024. {Highlight Poster.} ($^{*}$Co-first Authors)

\item \showpubcounter
\textbf{Thanh-Dat Truong}, Pierce Helton, Ahmed Moustafa, Jackson Cothren, and Khoa Luu. \textbf{{CONDA: Continual Unsupervised Domain Adaptation Learning in Visual Perception for Self-driving Cars.}} In Proceedings of the IEEE/CVF Conference on Computer Vision and Pattern Recognition Workshops (CVPRW), pp.~5642-5650. 2024.

\item \showpubcounter 
\textbf{Thanh-Dat Truong}, Xin Li, Bhiksha Raj, Jackson Cothren, and Khoa Luu. \textbf{{ED-SAM: An Efficient Diffusion Sampling Approach to Domain Generalization in Vision-Language Foundation Models}}. Under Review of Conference on Neural Information Processing Systems (NeurIPS), 2024.

\item \showpubcounter
\textbf{Thanh-Dat Truong}, Utsav Prabhu, Bhiksha Raj, Jackson Cothren, and Khoa Luu. {\textbf{FALCON: Fairness Learning via Contrastive Attention Approach to Continual Semantic Scene Understanding in Open World}}. Under Review of Conference on Neural Information Processing Systems (NeurIPS), 2024. 

\item \showpubcounter
\textbf{Thanh-Dat Truong}, Hoang-Quan Nguyen, Bhiksha Raj, and Khoa Luu. {\textbf{Fairness Continual Learning Approach to Semantic Scene Understanding in Open-World Environments}}. In Proceedings of Thirty-seventh Conference on Neural Information Processing Systems (NeurIPS), pp.~65456-65467. 2023. 

\item \showpubcounter
\textbf{Thanh-Dat Truong}, Ngan Le, Bhiksha Raj, Jackson Cothren, and Khoa Luu. {\textbf{FREDOM: Fairness Domain Adaptation Approach to Semantic Scene Understanding}}. In Proceedings of the IEEE/CVF Conference on Computer Vision and Pattern Recognition (CVPR), pp.~19988-19997. 2023.

\item \showpubcounter
\textbf{Thanh-Dat Truong}, Chi Nhan Duong, Kha Gia Quach, Ngan Le, Tien Dai Bui, and Khoa Luu. \textbf{{LIAAD: Lightweight Attentive Angular Distillation for Large-scale Age-Invariant Face Recognition.}} Journal of Neurocomputing, Volume 543, p.126198. 2023.

\item \showpubcounter
\textbf{Thanh-Dat Truong}, Quoc-Huy Bui, Chi Nhan Duong, Han-Seok Seo, Son Lam Phung, Xin Li, and Khoa Luu. \textbf{{DirecFormer: A Directed Attention in Transformer Approach to Robust Action Recognition.}} In Proceedings of the IEEE/CVF Conference on Computer Vision and Pattern Recognition (CVPR), pp.~20030-20040. 2022.

\item \showpubcounter
\textbf{Thanh-Dat Truong}, Naga Venkata Sai Raviteja Chappa, Xuan Bac Nguyen, Ngan Le, Ashley Dowling, and Khoa Luu. \textbf{{OTAdapt: Optimal Transport-based Approach For Unsupervised Domain Adaptation}}. In Proceedings of the IEEE International Conference on Pattern Recognition (ICPR), pp.~2850-2856.~2022.

\item \showpubcounter
Anh Pha Nguyen, \textbf{Thanh-Dat Truong}, Miaoqing Huang, Yi Liang, Ngan Le, and Khoa Luu. \textbf{{Self-supervised Domain Adaptation in Crowd Counting}}. In Proceedings of the IEEE International Conference on Image Processing (ICIP), pp.~2786-2790. 2022.

\item \showpubcounter
\textbf{Thanh-Dat Truong}, Chi Nhan Duong, Ngan Le, Son Lam Phung, Chase Rainwater, and Khoa Luu. \textbf{{BiMaL: Bijective Maximum Likelihood Approach to Domain Adaptation in Semantic Scene Segmentation.}} In Proceedings of the IEEE/CVF International Conference on Computer Vision (ICCV), pp.~8548-8557. 2021.

\item \showpubcounter
\textbf{Thanh-Dat Truong}, Chi Nhan Duong, The De Vu, Hoang Anh Pham, Bhiksha Raj, Ngan Le, and Khoa Luu. \textbf{{The Right to Talk: An Audio-Visual Transformer Approach.}} In Proceedings of the IEEE/CVF International Conference on Computer Vision (ICCV), pp.~1105-1114. 2021.

\item \showpubcounter
\textbf{Thanh-Dat Truong}, Chi Nhan Duong, Minh-Triet Tran, Ngan Le, and Khoa Luu. \textbf{{Fast Flow Reconstruction via Robust Invertible $n \times n$ Convolution.}}  Future Internet Journal,  Volume 13, Issue 7, p.179. 2021.

\item \showpubcounter
Chi Nhan Duong, \textbf{Thanh-Dat Truong}, Khoa Luu, Kha Gia Quach, Hung Bui, and Kaushik Roy. \textbf{{Vec2Face: Unveil Human Faces From Their Blackbox Features in Face Recognition.}} Proceedings of the IEEE/CVF Conference on Computer Vision and Pattern Recognition (CVPR), pp.~6132-6141. 2020. {Oral Presentation.}

\item \showpubcounter
\textbf{Thanh-Dat Truong}, Chi Nhan Duong, Khoa Luu, Minh-Triet Tran, and Ngan Le. \textbf{{Domain Generalization via Universal Non-volume Preserving Approach.}} Proceedings of the IEEE Conference on Computer and Robot Vision (CRV), pp.~93-100. 2020.

\item \showpubcounter
\textbf{Thanh-Dat Truong}, Chi Nhan Duong, Ashley Dowling, Son Lam Phung, Jackson Cothren, and Khoa Luu. {\textbf{CROVIA: Seeing Drone Scenes from Car Perspective via Cross-View Adaptation}}. Under Review of IEEE Transactions on Pattern Analysis and Machine Intelligence.

\item \showpubcounter
\textbf{Thanh-Dat Truong}, Chi Nhan Duong, Pierce Helton, Ashley Dowling, Xin Li, and Khoa Luu. \textbf{{CoMaL: Conditional Maximum Likelihood Approach to Self-supervised Domain Adaptation in Long-tail Semantic Segmentation}}. Under Review of  IEEE Transactions on Pattern Analysis and Machine Intelligence.

\item \showpubcounter
\textbf{Thanh-Dat Truong}, Chi Nhan Duong, Hoang Ngan Le, Marios Savvides, and Khoa Luu. \textbf{{Vec2Face-v2: Unveil Human Faces from their Blackbox Features via Attention-based Network in Face Recognition.}}  Under Review of  IEEE Transactions on Pattern Analysis and Machine Intelligence.

\end{publications}

%% file: Chapters/Chaps/chap-1-introduction.tex
\chapter{Introduction}

\begin{figure}[!b]
    \centering
    \includegraphics[width=1.0\linewidth]{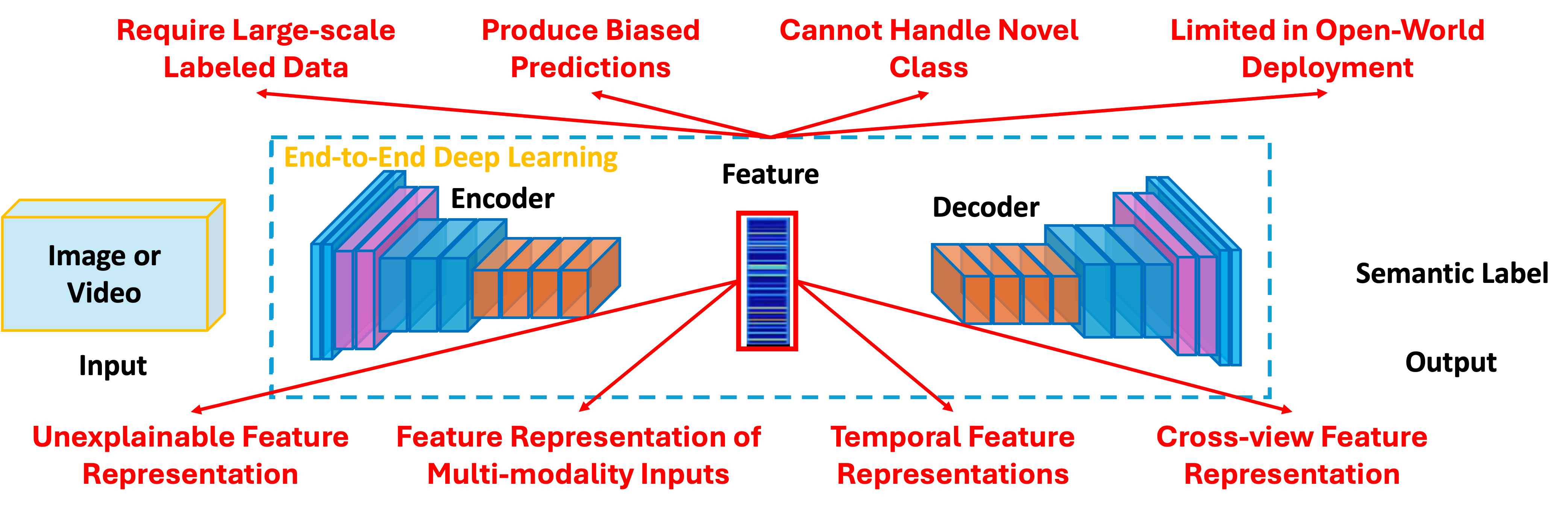}
    \caption{The Challenges in Current Machine Vision Framework.}
    \label{fig:machine_vision_framework}
\end{figure}

Computer Vision, a field in Artificial Intelligence (AI), aims to interpret the visual world by modeling visual information, such as images or videos.
The ultimate goal of computer vision is to automate tasks that replicate human capabilities, e.g., image classification~\cite{imagenet15russakovsky}, object detection~\cite{redmon2018yolov3}, image segmentation~\cite{dat2021bimal_iccv}, human action recognition~\cite{truong2021direcformer}, and then reacting to what they ``see''.
The typical machine vision framework consists of two stages, including (1) an Encoder extracting the features from the visual inputs and (2) a Decoder producing the semantic outputs from the features.
The traditional machine vision model adopts hand-crafted features, e.g., Scale-invariant Feature Transform (SIFT)~\cite{lowe2004distinctive}, Bag of Visual Worlds~\cite{sivic2003video}, Speeded Up Robust Features (SURF)~\cite{bay2006surf}, Local Binary Patterns (LBP)~\cite{he1990texture}, etc, for the encoder. Meanwhile, the decoder is designed via the Support Vector Machine (SVM)~\cite{cortes1995support}, Linear Discriminant Analysis (LDA)~\cite{yu2001direct}, etc.
However, the generalizability of traditional approaches remains limited when the data is scaling up.
Nowadays, AI in general, or deep learning in particular, has achieved phenomenon ability.
In the modern vision learning approaches, both encoder and decoder are designed as an end-to-end deep learning framework (Figure~\ref{fig:machine_vision_framework}), e.g., Convolutional Neural Networks (CNNs)~\cite{resnet}, Transformers~\cite{vit}.

\begin{figure}[!b]
	\centering
        \includegraphics[width=1.0\textwidth]{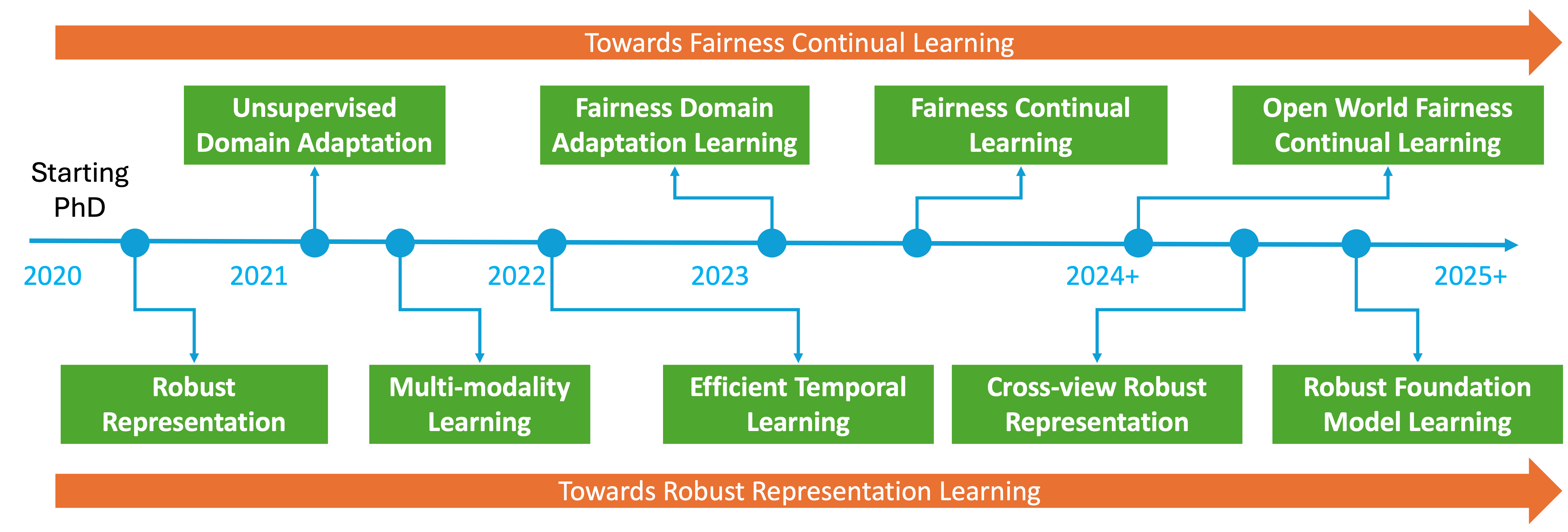}
        \caption
            {
                \textbf{Overview of Research Towards Fairness and Robustness Learning}.
                Unsupervised Domain Adaptation~\cite{dat2021bimal_iccv, truong2022otadapt, nguyen2022self, jalata2022eqadap}. 
                Fairness Domain Adaptation~\cite{Truong:CVPR:2023FREDOM, truong2023comal}. 
                Fairness Continual Learning~\cite{truong2022conda, truong2023fairness}.
                Open-world Fairness Continual Learning~\cite{truong2024falcon}.
                Robust Representation~\cite{duong2020vec2face, truong2023liaad}. Multi-Modality Learning~\cite{truong2021right2talk, nguyen2023insect}. 
                Efficient Temporal Learning~\cite{truong2021direcformer}.
                Cross-View Robust Representation~\cite{truong2023crovia, truong2023cross, truong2024eagle}, and Robust Foundation Model Learning~\cite{truong2024edsam, nguyen2023insect}.
            }
            \label{fig:thesis_goal_timeline}
\end{figure}

\textit{``Artificial Intelligence will likely be smarter than all humans combined by 2029''}, predicted by Elon Musk~\cite{elonmusk_statement}.
The recent success of deep learning approaches is closing the capability gap between humans and AI.
However, toward the goal as predicted by Elon Musk, vision learning methods still need to address several ill-posed challenges.
As shown in Figure~\ref{fig:machine_vision_framework}, the challenges of the current vision learning framework can be divided into two major groups: (1) Learning Approach and (2) Feature Representation.
In the learning approach group, there are four major challenges.
First, vision learning methods often require large-scale annotated data to train the robust model, but data annotation is a costly and time-consuming process.
Second, the AI models typically treat unfairly between classes or objects according to the data distributions, known as the fairness problem. The unfair predictions of the AI models can lead to severe problems in human-related applications.
Third, the data-driven models are often trained on large-scale data having known classes at once.
Then, these models learned on large-scale data may perform poorly in practical deployment as they may counter the new objects or new environments.
In practice, vision learning models should be capable of continually learning new classes without retraining from previous data.
Fourth, current vision learning approaches are often trained on a close-set dataset.
However, deployment in the open world requires the model to be able to model and understand the objects and scenarios that were unknown and unseen during prior training.
Meanwhile, there are also four major concerns in feature representation produced by deep learning-based approaches.
First, although the AI models achieved impressive performance, their knowledge representations are often uninterpretable.
Second, with the recent increase in multimodal data, learning and understanding correlation across input modalities is a challenging task due to the high dimension and complexity of multimodal inputs.
Third, the visual input does not only consist of spatial dimension but also includes temporal dimension. Thus, understanding temporal knowledge is critical to video understanding.
Fourth, the visual inputs could be captured from various camera viewpoints. Learning the model to be robust against camera views is necessary but difficult.

\section{Contributions of the Dissertation}
To address these challenges, the research in this dissertation aims to develop \textbf{novel vision learning toward two goals, i.e., (1) Fairness Continual Learning and (2) Robust Representation Learning (Figure~\ref{fig:thesis_goal_timeline}).}

\noindent
\textbf{Fairness Domain Adaptation}
To address the problem of annotated data, this dissertation introduces a novel bijective maximum likelihood loss to unsupervised domain adaptation~\cite{dat2021bimal_iccv} that can be used to generalize on target domains. This can be further applied to image deblurring~\cite{jalata2022eqadap} and crowd counting~\cite{nguyen2022self}.
To mitigate the unfair predictions, we develop a new fairness domain adaptation~\cite{Truong:CVPR:2023FREDOM} by imposing the fairness constraint based on data distributions.

\noindent
\textbf{Open-world Fairness Continual Learning} 
To equip the continual learning ability of deep learning approaches, this thesis introduces a novel prototypical contrastive learning to continual learning~\cite{truong2022conda, truong2023fairness}.  To enforce the fairness of the model, we propose the fairness constraint~\cite{truong2023fairness} to guarantee fair predictions. 
The ultimate goal of this research is to develop an efficient fairness continual learning approach that can be deployed in open-world environments. 
Toward this ultimate goal, this dissertation introduces a novel contrastive attention approach to fairness and continual learning ~\cite{truong2024falcon}. We propose a new attention-based visual grammar approach to effectively model unknown classes and produce better feature representations for unknown classes in the open world.

\noindent
\textbf{Efficient Geometry-based Approach to Cross-view Learning}
Recent studies have shown the importance of cross-view learning.
Developing cross-view learning to transfer knowledge trained on one view to another view could bring huge benefits to the development of view-invariant understanding. However, this learning direction encounters an ill-posed challenge due to the gap between the two views. 
Toward the goal of cross-view learning, this thesis will use a novel Geometry-based Cross-View Adaptation framework~\cite{truong2024eagle, truong2023crovia} to model structural changes across views via the geodesic flow path. 
Then, we further develop a cross-view learning framework for video understanding by introducing a new cross-view action recognition understanding from exocentric to egocentric perspective~\cite{truong2023cross}.

\noindent
\textbf{Efficient Multimodal and Temporal Learning}
Recent studies have shown that fusing multi-modalities of inputs could potentially improve the performance~\cite{truong2021right2talk, nguyen2023insect}.
However, since the features of different modalities belong to different feature spaces, aligning these features to exploit the correlation among these modalities remains an ill-posed challenge.
Therefore, to study the robust representations in multimodal and temporal learning, this thesis introduces a novel Transformer approach ~\cite{truong2021right2talk} to exploit the correlation of two input modalities (i.e., video and audio).
In addition, while deep learning architectures, e.g., Transformers, have shown outstanding performance in video learning, these approaches still remain limited in learning temporal knowledge due to the non-directed self-attention mechanism. Meanwhile, in temporal learning, direction in the temporal dimension is an important factor since it will form the semantic content of the video. Therefore, to develop efficient temporal learning, this thesis introduces a new Directed Temporal-Spatial Attention mechanism~\cite{truong2021direcformer}. In addition, conditional dependency learning and self-supervised guided loss for directed temporal attention are presented to improve the temporal learning capability.

\noindent
\textbf{Robust Foundation Model Learning}
With the recent development of large-scale vision-language foundation models, these approaches have achieved impressive performance. However, their generalizability remains limited in explaining their predictions and representations. 
The higher goal of my research is to develop a vision learning approach to improve the fairness and robustness of the foundation model. Therefore, this thesis proposes~\cite{truong2024edsam} a novel efficient diffusion sampling approach to improve the robustness of the foundation model. 
Our long-term research will focus on improving the robust representation of the foundation model~\cite{truong2024edsam, nguyen2023insect}.

\section{Summary of Dissertation Organization}

The remaining chapters of this dissertation are organized as follows. 
In Chapter~\ref{chap:related-work}, we present the background and literature review related to the research in this dissertation, including unsupervised domain adaptation, continual learning, fairness learning, multimodal and temporal learning, cross-view learning, robust representation learning, and foundation model.
In Chapter~\ref{chap:adaptation}, we will present our fairness domain adaptation. In particular, we first introduce a novel bijective maximum likelihood approach to unsupervised domain adaptation. Then, we further propose a novel fairness domain adaptation learning framework.
In Chapter~\ref{chap:continual-learning}, we introduce our new fairness continual learning approach via the prototypical contrastive clustering approach. Then, we propose new fairness learning via the contrastive attention approach to continual semantic scene understanding in the open world.
In Chapter~\ref{chap:cross-view}, we propose new geometry-based approaches to cross-view learning. Particularly, we propose a new cross-view geometric correlation modeling via the geodesic flow path applied to semantic segmentation. Then, we present our cross-view learning approach to video understanding from an exocentric to an egocentric perspective.
In Chapter~\ref{chap:multimodal-temporal}, we present new Transformer-based approaches to multimodal and temporal learning. First, we propose a new Transformer-based network to study the correlation of multimodal data. Second, we introduce a new directed attention mechanism in Transformers to improve temporal learning.
Then, we propose a novel domain generalization approach based on diffusion to large-scale vision-language models. 
Finally, Chapter~\ref{chap:conclusions} summarizes the research in this dissertation with the major findings and discusses potential feature research ideas to improve the robustness and fairness of large-scale multimodal models.

%% file: Chapters/Chaps/chap-2-related-work.tex
\chapter{Background and Related Work}\label{chap:related-work}

This chapter will first present the current learning approach to fairness adaptation and continual learning in the open world, followed by the literature reviews of cross-view learning. Then, the background and related work of multimodal and temporal learning will be presented. Finally, this chapter will review the recent studies of large-scale foundation and multimodal models.

\section{Fairness Domain Adaptation in Semantic Scene Segmentation}

\noindent
\textbf{Unsupervised Domain Adaptation.}
Although the deep learning approaches~\cite{chen2018deeplab, le2018segmentation, huynh2021progressive} have illustrated their effectiveness in scene understanding, these methods remain limited when deployed into practical environments due to the gap between the training data (known as source data) and the testing data (known as target data). 
To alleviate this problem, domain adaptation 
has emerged as an efficient approach to aims to adaptively transfer deep learning models into real-world data while reducing the demands of large-scale annotated data. 
Adversarial learning~\cite{chen2018CVPR, hong2018CVPR, tsai2018learning, ganin2015unsupervised, long2015learning, tzeng2017adversarial}, and self-supervised learning~\cite{araslanov2021dasac, zhang2021prototypical, hoyer2022daformer, shahaf2022procst} are common approaches to unsupervised domain adaptation.
The adversarial learning approaches are typically simultaneously trained on source and target data~\cite{hoffman2016fcns, chen2017no, chen2018CVPR} with a generative adversarial framework.
Several other approaches improved adversarial learning by utilizing the generative model~\cite{zhu2017unpaired, murez2018CVPR, hoffman18a}, using additional labels~\cite{lee2018spigan, vu2019dada}, incorporating with  
entropy minimization~\cite{vu2019advent, yan2021pixellevel, dat2021bimal_iccv},
or adopting the curriculum training~\cite{pan2020unsupervised} into the training process. 
Recently, the self-supervised approaches~\cite{zou2018unsupervised, araslanov2021dasac, zhang2021prototypical, hoyer2022daformer, shahaf2022procst} have gained more attention due to their outstanding performance. Prior works have improved the performance of self-supervised approaches by introducing the self-supervised augmentation consistency framework~\cite{araslanov2021dasac} or the prototypical pseudo-label framework~\cite{zhang2021prototypical}. Later, the performance of self-supervised was significantly by using the Transformer network~\cite{hoyer2022daformer} with multi-resolution cropped images~\cite{hoyer2022hrda} and masked image consistency strategy~\cite{hoyer2023mic} to enhance contextual learning. 
Our prior work further improved the performance by promoting the fairness of the model predictions~\cite{Truong:CVPR:2023FREDOM}.

\noindent
\textbf{Fairness and Data Imbalanced Learning.}
The early methods utilized the balanced Softmax loss~\cite{ren2020balanced} to alleviate the impact of imbalanced data distribution.
Later, Wang et al. ~\cite{wang2021seesaw} presented a Seesaw loss to re-balance the contributions of positive and negative instances.
Ziwei et al. ~\cite{liu2019largescale} introduced a dynamic meta-embedding to model the imbalanced classification.
Chu et al. ~\cite{chu2021learning} reduce the bias in the segmentation model via a new stochastic training scheme.
Szabo et al. ~\cite{szabo2021tilted} presented a tilted cross-entropy loss to promote class-relevant fairness.
Truong et al.~\cite{Truong:CVPR:2023FREDOM} presented a fairness domain adaptation approach to semantic segmentation and extended it into a continual learning setting~\cite{truong2023fairness}.
However, these methods~\cite{Truong:CVPR:2023FREDOM, truong2023fairness} rely on the assumption of ideal balanced data, which can not be achieved by nature.

\section{Open-world Fairness Continual Learning}

Although domain adaptation plays a role in bridging the gap between the training data and the deployment data, it remains limited in open-world deployment since domain adaptation is unable to handle the novel unseen classes due to its close-set learning setup. 
This limitation motivates us to develop a new \textbf{Continual Learning} (CL) paradigm, where the deep learning model can continuously learn new knowledge of environments and novel objects from \textbf{\textit{dynamic, open-world}} data.
There are two major challenges in CL, 
i.e., catastrophic forgetting~\cite{douillard2021plop, ssul_neurips_2021, zhang2022representation} and background shift~\cite{cermelli2020modelingthebackground}.
While the former refers to the problem of the model forgetting its knowledge when training on the new data, the latter depicts the problem of labels of different unknown classes collapsed into a single background class. 
Several studies have been introduced to address these challenges~\cite{kirkpatrick2017ewc, lopezpaz2017gem, li2018lwf, douillard2020podnet, douillard2021plop} in both image classification and segmentation problems.
The common approach of CL in semantic segmentation adopts knowledge distillation~\cite{douillard2021plop} and pseudo labels~\cite{cermelli2020modelingthebackground} to model catastrophic forgetting and background shift, respectively.
Later, it is further improved by 
decoupling knowledge representations~\cite{zhang2022representation}, modeling the inter- and intra-class knowledge~\cite{sats_prj_2023}, 
distinguishing the feature representations of the future classes, 
\cite{ssul_neurips_2021}, or modeling knowledge distillation loss as the Geodesic flow path~\cite{simon2021learning}.
The recent approach~\cite{cermelli2023comformer} adopted the mask-based segmentation networks~\cite{cheng2021maskformer, cheng2022mask2former} to significantly improve the performance of CL models. 
Recently, several studies have introduced CL under the unsupervised domain adaptation settings~\cite{volpi2021continual, rostami2021lifelong, saporta2022muhdi}.
Meanwhile, other approaches defined the CL objective as the contrastive clustering problem~\cite{joseph2021towards, truong2023fairness}.

\section{Cross-view Visual Feature Understanding}

The early studies exploited cross-view learning in geo-localization 
by using a polar transform across views~\cite{shi2020where, NIPS2019_9199} or generative networks to cross-view images ~\cite{9010757, Toker_2021_CVPR}. 
Meanwhile, Zhu et al.~\cite{zhu2022transgeo} exploited the correlation between street- and aerial-view data via self-attention.
In semantic segmentation, Coors et al.~\cite{NoVA} first introduced a cross-view adaptation approach utilizing the depth labels and the cross-view transformation between car and truck views. 
However, this change of views in~\cite{NoVA} is not as big a hurdle as the change of views in our problem, i.e., car view to drone view.
Ren et al.~\cite{9878624} presented an adaptation approach across viewpoints using the 3D models of scenes to create pairs of cross-view images.
Vidit et al.~\cite{vidit2023learning} modeled the geometric shift in cross FoV setting for object detection by learning position-invariant homography transform. 
Di Mauro et al.~\cite{SceneAdapt} introduced an adversarial method trained on a multi-view synthetic dataset where images are captured from different pitch and yaw angles at the same altitudes of the camera positions.
Meanwhile, in our problem, the camera views could be placed at different altitudes (e.g., the car and the drone), which reveals large structural differences between the images. 
Truong et al.~\cite{truong2023crovia, truong2023cross} first introduced a simple approach to model the relation across views.
Brady et al.~\cite{zhou2022cross} presented a cross-view transformer that learns the camera-aware positional embeddings.
Although the views are captured from left and right angles, the camera positions in the approach remain at the same altitude.
Similarly, Pan et al.~\cite{pan2020cross} present a View Parsing Network to accumulate features across first-view observations with multiple angles.
Yao et al.~\cite{yao2018multiview} proposed a semi-supervised learning approach to learn the segmentation model from multiple views of an image.
Huang et al.~\cite{huang2021cross} a cross-style regularization for domain adaptation in panoptic segmentation by imposing the consistency of the segmentation between the target images and stylized target images. 
Wang et al.~\cite{wang2021viewpoint} proposed a viewpoint adaptation framework for the person re-identification problem by using the generative model to generate training data across various viewpoints. 
Hou et al.~\cite{hou2015unsupervised} presented a  matching
cross-domain data approach to domain adaptation in visual classification.
Sun et al.~\cite{sun2020unsupervised} proposed a cross-view facial expression adaptation framework to parallel synthesize and recognize cross-view facial expressions.
Goyal et al.~\cite{goyal2022cross} introduced a cross-view action recognition approach to transferring the feature representations to different views. 
Zhang et al.~\cite{zhang2021cross} proposed a multi-view crowd counting approach that adaptively chooses and aggregates multi-cameras and a noise view regularization.
Armando et al.~\cite{armando2023cross} proposed a self-supervised pre-training approach to human understanding learned on pairs of images captured from different viewpoints. Then, the pre-trained models are later used for various downstream human-centric tasks.
In summary, these prior cross-view methods could require either a pair of cross-view images~\cite{armando2023cross} or images captured at the same altitude with different angles~\cite{huang2021cross, zhou2022cross, hou2015unsupervised}.
In addition, these methods lack a theory and a mechanism for cross-view geometric structural change modeling.

\section{Multimodal and Temporal Learning}

\noindent
\textbf{Multimodal Audio-Visual Understanding.} Some early methods~\cite{barzelay2007harmony,hershey1999audio,kidron2005pixels, izadinia2012multimodal} localize the sources of human voice in a video using statistical models and audio-visual correlations. 
Fisher et al.~\cite{fisher2000learning} introduce 
a multi-media fusion method in a complex domain to capture latent audio-visual relationships.
Later, deep learning approaches~\cite{chung2016out,owens2018audio} come into place and exploit the synchronization between visual and audio signals to find the regions in the images that are sensitive to the audio features.
Afouras et al.~\cite{Afouras20b} propose LWTNet that extends the synchronize cues with an optical flow technique to extract and track audio-visual objects for the localization process. %
Zhao et al.~\cite{Zhao_2018_ECCV, zhao2019sound} detect objects in one or multiple frames and use their appearance and motion to differentiate sounds of objects. 
Gao et al.~\cite{gao2019co} propose a co-separation training objective to learn audio-source separation from unlabeled videos containing multiple sources of sounds. 
Ephrat et al.~\cite{ephrat2018looking} contributed a large-scale dataset, namely AVspeech, and proposed an end-to-end audio-visual architecture for this task. %
Afouras et al.~\cite{afouras2018conversation} propose to use the lip regions and consider both audio magnitudes and phases. %
The above-mentioned methods ignore the context of the video, which is a very important cue for the network to improve the quality of separated voices. 
section.

\noindent
\textbf{Video Action Recognition.} 
Many large-scale third-view datasets have been introduced for action recognition tasks, e.g., Kinetics~\cite{kinetic700, kinetics, kinetic600}, Something-Something V2~\cite{ssv2}, Sport1M~\cite{sport1m}, AVA~\cite{ava-kinetic}, etc. 
Many deep learning approaches~\cite{swin, MViTv2, slowfast, i3d, tsn, lin2019tsm, vivit, mvit} have been introduced and achieved remarkable achievements.
The early deep learning approaches~\cite{sport1m, lrcn} have utilized the 2D Convolutional Neural Networks (CNNs)~\cite{resnet, vgg, inceptionv3} to extract the deep spatial representation followed by using Recurrent Neural Networks (RNNs)~\cite{lstm} to learn the temporal information from these extracted spatial features.
Some later approaches have improved the temporal learning capability by introducing the two-stream networks~\cite{tsn, 2streams_simonyan, 2streams_feichtenhofer, spatiotemporal_resnet, spatiotemporal_multiplier} using both RGB video inputs and optical flows for motion modeling. 
Later, the 3D CNN-based approaches~\cite{c3d} and their variants~\cite{i3d, trn} have been introduced, i.e., 
several (2+1)D CNN architectures have been proposed~\cite{slowfast, r21d, x3d, s3d}. Meanwhile, other approaches have used
pseudo-3D CNNs built based on 2D CNNs~\cite{lin2019tsm, p3d}. In addition, to better capture the long-range temporal dependencies among video frames, the non-local operation has also been introduced~\cite{non_local}.
SlowFast~\cite{slowfast} proposes a dual-path network to learn spatiotemporal information at two different temporal rates. X3D~\cite{x3d} progressively expands the networks to search for an optimal network for action recognition.

\noindent
\textbf{Vision Transformer.}~\cite{swin, MViTv2, vivit, mvit, vit, timesformer, truong2021direcformer} has become a dominant backbone in various tasks due to its outstanding performance. The early success of Video Vision Transformer (ViViT)~\cite{vivit} has shown its promising capability in handling spatial-temporal tokens in action recognition.
Then, many variants~\cite{swin, MViTv2, mvit, timesformer, truong2021direcformer, nguyen2024multi, nguyen2023hig} of ViViT have been introduced to improve the accuracy and reduce the computational cost. 
\cite{bulat2021spacetime} presented a space-time mixing attention mechanism to reduce the complexity of the self-attention layers.
TimeSFormer~\cite{timesformer} introduced divided spatial and temporal attention to reduce the computational overhead. Then, it is further improved using the directed attention mechanism~\cite{truong2021direcformer}.
Then,~\cite{mvit} proposed a Multi-scale Vision Transformer (MViT) using multiscale feature hierarchies. Then, MViT-V2~\cite{MViTv2} improves the performance of MViT by incorporating decomposed relative positional embeddings and residual pooling connections.
Swin Video Transformer~\cite{swin} has achieved state-of-the-art performance in action recognition by using shifted windows to limit the self-attention computational cost to local windows and also allow learning attention across windows.
The recent SVFormer~\cite{xing2023svformer} has introduced a temporal warping augmentation to capture the complex temporal variation in videos for semi-supervised action recognition.
Meanwhile, MTV~\cite{yan2022multiview} presented a Multiview Transformer to model different views of the videos with lateral connections to fuse information across views.
Later, M\&M Mix\cite{xiong2022m} further improved MTV by using multimodal inputs.
TADA~\cite{huang2021tada} proposed Temporally-Adaptive Convolutions (TAdaConv) to model complex temporal dynamics in videos.
Inspired by the success of CLIP in vision-language pretraining~\cite{radford2021learning, nguyen2023insect, jia2021scaling}, several studies have adopted this approach to video-language pretraining~\cite{egovlp, wang2023all, sun2022long}.
All-in-One~\cite{wang2023all} presented a unified approach to video-language pretraining by embedding raw video and textual inputs into joint representations
with a unified network.
EgoVLP~\cite{egovlp} introduced Video-Language Pretraining to ego-centric video understanding. 
LF-VILA~\cite{sun2022long} presented a Multimodal Temporal Contrastive and Hierarchical
Temporal Window Attention to model the long-form videos for video-language pretraining.

\noindent
\textbf{Egocentric Video Analysis.}
Apart from third-view videos, egocentric videos provide distinguished viewpoints that pose several challenges in action recognition. Many datasets have been introduced to support the egocentric video analysis tasks, e.g., Charades-Ego~\cite{charades-ego}, EPIC Kitchens~\cite{epic, epic-100}, Ego4D~\cite{Ego4D2022CVPR}, EgoClip~\cite{egovlp}, HOI4D~\cite{Liu_2022_CVPR_HOI4D}. These datasets provide several standard egocentric benchmarks, e.g., action recognition~\cite{charades-ego, epic, Ego4D2022CVPR}, action anticipation~\cite{Ego4D2022CVPR, epic-100}, action detection~\cite{epic-100}, video-text retrieval~\cite{egovlp, Ego4D2022CVPR}.
Many methods have been proposed for egocentric action recognition, including Multi-stream Networks~\cite{ma2016going,li2018eye,epic-fusion,wang2020makes}, RNNs~\cite{furnari2019would,furnari2020rolling,sudhakaran2019lsta}, 3D CNNs~\cite{pirri2019anticipation,lu2019learning}, Graph Neural Networks~\cite{ego-topo}.
Despite the difference in network designs, these prior works are usually pre-trained on the large-scale third-view datasets before fine-tuning them on the first-view dataset. However, there is a significant difference between the first-view and third-view datasets.
Thus, a direct fine-tuning approach without consideration of modeling view changes could result in less efficiency.
Many methods have improved the performance of the action recognition models by using additional egocentric cues or tasks, including gaze and motor attention~\cite{mathe2012dynamic,li2018eye,liu2019forecasting}, object detection~\cite{antonino-next-active,baradel2018object,dessalene2020egocentric,wang2020symbiotic}, hand interactions~\cite{tekin2019h,shan2020understanding,kapidis2019egocentric}.
Ego-Exo~\cite{ego-exo} presented an approach by introducing the auxiliary egocentric tasks into the pre-training phase on the third-view dataset, i.e., ego-score, object-score, and interaction map predictions.
However, these methods usually require the labels of auxiliary egocentric tasks on the third-view datasets or rely on pseudo labels produced by the off-the-shelf pre-trained models on egocentric tasks.

\section{Large-scale Foundation Model}

\noindent
\textbf{Visual Foundation Model.} The contrastive language-image training~\cite{radford2021learning, jia2021scaling, yu2022coca, luo2023lexlip,wang2023equivariant, dehdashtian2024fairerclip} has become a prominent approach in developing the large-scale vision-language model~\cite{radford2021learning, jia2021scaling}. 
By learning on large-scale image-text data, the vision-language foundation model has shown its effectiveness in learning visual representations from the supervision of language. Later, it can be adapted to various downstream tasks with an impressive zero-shot transfer ability.
CLIP~\cite{radford2021learning} and ALIGN~\cite{jia2021scaling}  first introduced contrastive learning to learn strong representations of images and texts for cross-modal alignment.
CoCa~\cite{yu2022coca} proposed an additional decoder and generative image captioning.
SLIP~\cite{mu2022slip}, DeCLIP~\cite{li2021supervision}, FLIP~\cite{li2023scaling} further improve the performance by using self-supervised training techniques.
LaCLIP~\cite{fan2023improving} improved the performance of CLIP by introducing text augmentation via the large language model.
LiT~\cite{zhai2022lit} and BASIC~\cite{pham2023combined} improve the zero-shot transfer ability via further fine-tuning the language encoder.
SimVLM~\cite{wang2021simvlm}, OFA~\cite{wang2022ofa}, and BLIP~\cite{li2022blip} train the vision-language model within an encoder-decoder framework with language generative losses. 
SigLIP~\cite{zhai2023sigmoid} proposed a Sigmoid loss to compute the image-text similarity.

\noindent
\textbf{Large-scale Multimodal Foundation.}
The development of large-scale data processing and large-scale multimodal models (LMMs) has provided a new vehicle to solve complex problems, including multimodal data. Some of them can be accounted including large vision-language models~\cite{liu2024improved, liu2024visual}, large video-language models~\cite{weng2024longvlm, zhao2023learning}, or large audio-language models~\cite{hussain2023m, ghosal2023text}.
By incorporating the power of LLMs, the LMMs have revolutionized the research of large-scale multimodal.
In the design of LMMs, different input modalities are connected to LLMs via the project modules~\cite{liu2024visual}. Then, the alignment across modalities is performed via cross-attention~\cite{liu2024improved}, Q-Formers~\cite{li2023blip}, or MLP~\cite{liu2024visual}.
The training procedure of LMMs typically has two major steps: pretraining and instruction-tuning.
While the first stage learns the alignment of features across modalities, the second stage enables reasoning about concepts in multimodal inputs and tasks.
Recently, Chen and Zhang~\cite{chenfedmbridge} improved LMM learning via multimodal federated learning.
Several benchmarks were introduced to evaluate the LMM performance, e.g., MMMU~\cite{yue2024mmmu}, and
MM-SpuBench~\cite{ye2024mm}.

%% file: Chapters/Chaps/chap-3-domain-adaptation.tex
\chapter{Fairness Domain Adaptation Approach}\label{chap:adaptation}

Semantic segmentation aims to predict pixel-level labels. It has become a popular task in various computer vision applications. While fully supervised segmentation methods have achieved high accuracy on large-scale vision datasets, they are unable to generalize on a new testing environment or a new domain well.
To improve the generalizability, Unsupervised Domain Adaptation in Semantic Scene Segmentation has shown impressive improvement in recent years.
However, the fairness concerns in the domain adaptation have yet to be well defined and addressed. In addition, fairness is one of the most critical aspects when deploying the segmentation models into human-related real-world applications, e.g., autonomous driving, as any unfair predictions could influence human safety. 
In this chapter, first, we present the Bijective Maximum Likelihood (BiMaL) approach to Unsupervised Domain Adaptation in Semantic Segmentation.
Then, we propose a novel Fairness Domain Adaptation (FREDOM) approach to semantic scene segmentation.
Through the ablation studies, the proposed method has shown that the segmentation models perform better and promote fairness in the model predictions.
The experimental results on the two standard benchmarks, i.e., SYNTHIA $\to$ Cityscapes and GTA5 $\to$ Cityscapes, have shown that our method achieved State-of-the-Art (SOTA) performance.

\input{Chapters/Sections/chap-3/chap-3-bimal}

\input{Chapters/Sections/chap-3/chap-3-fredom}

\section{Summary}

In this chapter, we have presented new approaches to address unsupervised domain adaptation and fairness in semantic segmentation.
First, this chapter presented a new Bijective Maximum Likelihood approach to domain adaptation in semantic scene segmentation. Compared to Adversarial Entropy Minimization loss, it is a more generalized form and can work without any assumption about pixel independence. 
A new Unaligned Domain Score metric has been also introduced to measure the efficiency of a segmentation model on a new target domain in an unsupervised manner. 
Second, this chapter presented the new fairness domain adaptation to semantic scene segmentation by analyzing the fairness treatment from class distributions. In particular, the conditional structural constraints have imposed the consistency of the predicted segmentation and modeled the structural information to improve the accuracy of segmentation models. Our ablation studies have analyzed different aspects affecting the fairness of segmentation models. It has also shown the effectiveness of our approach in terms of fairness improvement. Our approaches have achieved SOTA performance compared to prior methods.

%% file: Chapters/Sections/chap-3/chap-3-bimal.tex
\section{Bijective Maximum Likelihood Approach to Domain Adaptation in Semantic Scene Segmentation}\label{sec:bimal-paper}

\setcounter{propositioncounter}{0}
\setcounter{remarkcounter}{0}

Semantic segmentation is one of the most popular  computer vision topics, which aims to assign each pixel in an image to a 
predefined class.
It has various practical applications, especially in autonomous driving where a segmentation model is needed to recognize roads, sidewalks, pedestrians or vehicles in a large variety of urban conditions. A typical supervised segmentation model is usually trained on datasets with labels. However, annotating images for the semantic segmentation task is costly and time-consuming. 
Alternatively, a powerful and cost-effective way to acquire a large-scale training set is to use a simulation, e.g. game engines, to create a synthetic dataset~\cite{Richter_2016_ECCV, Ros_2016_CVPR}. %
However, fully supervised models~\cite{chen2018deeplab, le2018segmentation} trained on the synthetic datasets are often unable to perform well on real images due to the pixel appearance gap between synthetic and real images.

\begin{figure*}[!t]
    \centering
    \includegraphics[width=0.95\textwidth]{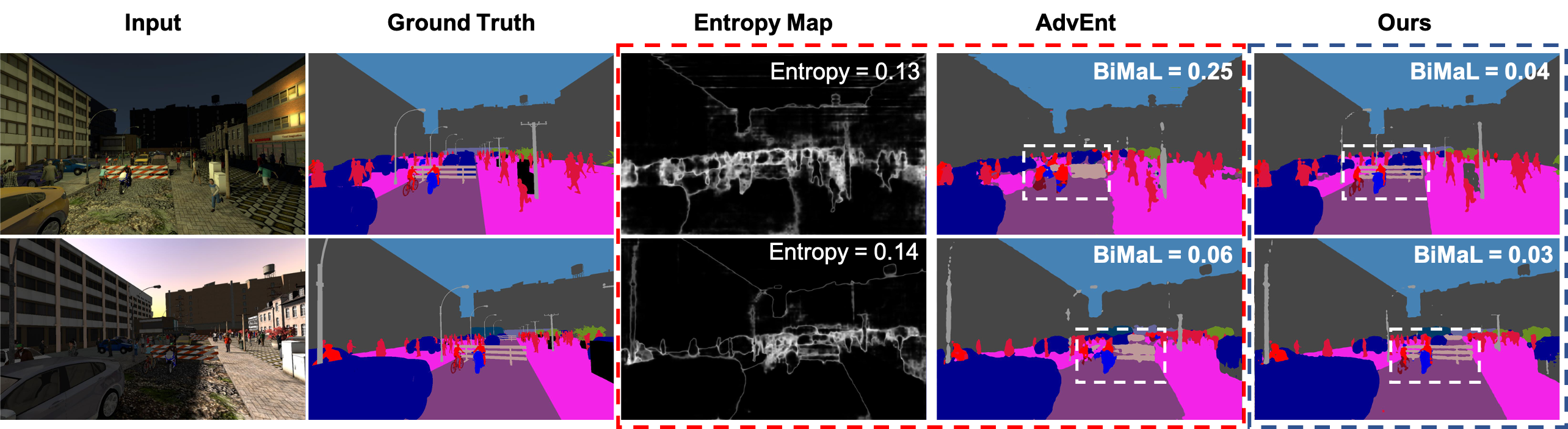}
    \caption{\textbf{Two images have the same entropy but one has a poor prediction (a top image) and one has an better prediction (a bottom image).} Columns 1 and 2 are an input image and a ground truth. Columns 3 and 4 are an entropy map and a prediction of AdvEnt~\cite{vu2019advent}. Column 5 is the results of our proposed method. The two predictions produced by AdvEnt have similar entropy scores ($0.13$ and $0.14$). Meanwhile, the BiMaL value of the bottom prediction ($0.06$) is smaller than the top prediction ($0.14$). Our results in the last column, which have better BiMaL values than AdvEnt, can well model the structure of an image. In particular, our results have sharper results of a barrier and a rider (white dash box), and a clear boundary between road and sidewalk.}
    \label{fig:bimal-same_entropy_1}
\end{figure*}

Unsupervised Domain Adaptation (UDA) aims to train a machine learning model on an annotated dataset, i.e. the source, and guarantee its high performance on a new unlabeled dataset, i.e. the target.
The UDA approaches have been applied to various computer vision tasks such as Semantic Segmentation~\cite{chen2018deeplab, le2018segmentation, li2020content, vu2019advent, vu2019dada, zhang2019category}, Face Recognition~\cite{duong2019shrinkteanet, Luu_BTAS2009, Luu_FG2011, Luu_ROBUST2008, Luu_IJCB2011}. 
The recent UDA methods aim to reduce the cross-domain discrepancy, along with the supervised training on the source domain~\cite{ganin2015unsupervised, long2015learning,pan2020unsupervised, tzeng2017adversarial, vu2019advent}.
In particular, these methods aim to minimize the distribution discrepancy of the 
deep representations extracted from the source and the target domains. This process can be performed at single or multiple levels of deep features using maximum mean discrepancies~\cite{ganin2015unsupervised, long2015learning, tzeng2017adversarial}, or adversarial training~\cite{chen2018CVPR, chen2017no, hoffman18a, hoffman2016fcns, hong2018CVPR, tsai2018learning}. The approaches in this group have shown their potential in aligning the predicted outputs of images from the two domains. However, 
the binary cross-entropy 
label predicted by the learned discriminator is usually a weak indication of structural learning for the segmentation task. Another approach named self-training utilizes the pseudo-labels or generative networks conditioned on target images~\cite{murez2018CVPR, zhu2017unpaired}. Semi-supervised learning is an approach related to UDA where the training set consists of both labeled and unlabeled samples. Thus, it has motivated several UDA approaches such as Class-balanced self-training (CBST)~\cite{zou2018unsupervised},  %
and entropy minimization~\cite{chen2019domain, grandvalet2005semi, pan2020unsupervised, springenberg2015unsupervised, vu2019advent}.
Although metrics such as entropy can be efficiently computed and adopted for training, they tend to rely on easy predictions, i.e. high confident scores, as references for the label transfer from source to target domains. This issue is alleviated in a later approach~\cite{chen2019domain} by preventing learned models from over-focusing on high confident areas. However, this type of metrics is formulated in pixel-wised manner, and, therefore, neglects the structural information presented in the image (see Figure~\ref{fig:bimal-same_entropy_1}).

\input{Tables/chap-3/bimal/method-comparison}

To address these limitations, this chapter presents a new unsupervised domain adaptation approach to tackle the semantic segmentation problem. Table~\ref{tab:bimal-summary} summarizes the difference between our proposed approach and the prior ones.
Firstly, a new Unaligned Domain Score (UDS) is introduced to measure the efficiency of the learned model on a target domain in an unsupervised manner.
Secondly, the presented UDS is further extended as a new loss function, named Bijective Maximum Likelihood (BiMaL) loss, that can be used with an unsupervised deep neural network to generalize on target domains. Indeed, we further demonstrate the BiMaL loss is a generalized form of the Adversarial Entropy Minimization (AdvEnt)~\cite{vu2019advent} without pixel independence assumption. 
Far apart from AdvEnt, which assumes pixel independence, BiMaL loss is formed using a Maximum-likelihood formulation to model the global structure of a segmentation input and a bijective function to map the segmentation structure to a deep latent space.

\subsection{Unaligned Domain Scores}

\begin{figure*}[!t]
    \centering
    \includegraphics[width=0.9\textwidth]{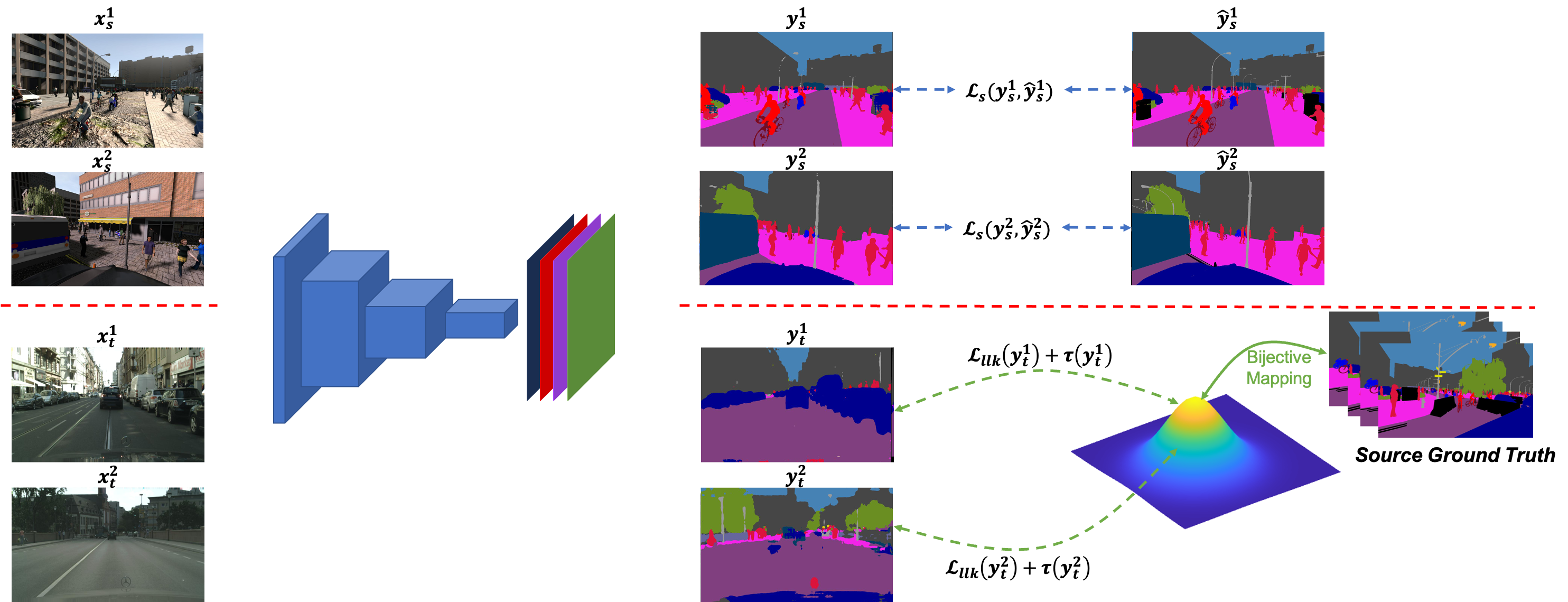}
    \caption{\textbf{The Proposed Framework.} The RGB image input is firstly forwarded to a deep semantic segmentation network to produce a segmentation map. The supervised loss is employed on the source training samples. Meanwhile, the predicted segmentation on target training samples will be mapped to the latent space to compute the Bijective Maximum Likelihood loss. The bijective mapping network is trained on the ground-truth images of the source domain.}
    \label{fig:bimal-proposed_framework}
\end{figure*}

Let $\mathbf{x}_s \in \mathcal{X}_s \subset \mathbb{R}^{H \times W \times 3}$ be an input image of the source domain ($H$ and $W$ are the height and width of an image),  $\mathbf{x}_t \in \mathcal{X}_t \subset \mathbb{R}^{H \times W \times 3}$ be an input image of the target domain, $G: \mathcal{X} \to \mathcal {Y}$ where $\mathcal{X} = \mathcal{X}_s \cup \mathcal{X}_t$ be a semantic segmentation function 
that maps an input image to its corresponding segmentation map $\mathbf{y} \subset \mathbb{R}^{H \times W \times C}$, i.e. $\mathbf{y} = G(\mathbf{x}, \theta)$ ($C$ is the number of semantic classes). 
In general, given $N$ labeled training samples from a source domain 
$\mathcal{D}_s = \{\mathbf{x}_s^i, \hat{\mathbf{y}}_s^i\}_1^N$ 
and $M$ unlabeled samples from a target domain $\mathcal{D}_t = \{\mathbf{x}_t^i\}_1^M$, the 
unsupervised domain adaptation for semantic segmentation can be formulated as in Eqn.~\eqref{eqn:bimal-objective}.
\begin{equation} \label{eqn:bimal-objective}
\small
\begin{split}
    \theta^{*} &= \arg\min_{\theta} \sum_{i,j} \big[\mathcal{L}_{s}(G(\mathbf{x}^i_s, \theta), \hat{\mathbf{y}}^i_s) + \mathcal{L}_{t}(G(\mathbf{x}^j_t, \theta))\big]\\
    &=\arg\min_{\theta} \Big[\mathbb{E}_{\mathbf{x}_s \sim p(\mathbf{x}_s), \hat{\mathbf{y}}_s \sim p(\hat{\mathbf{y}}_s)} \big[\mathcal{L}_{s}(G(\mathbf{x}_s, \theta), \hat{\mathbf{y}}_s)] 
    + \mathbb{E}_{\mathbf{x}_t \sim p(\mathbf{x}_t)} [\mathcal{L}_{t}(G(\mathbf{x}_t, \theta))\big]\Big]\\
    &=\arg\min_{\theta} \Big[\mathbb{E}_{\mathbf{y}_s \sim p(\mathbf{y}_s), \hat{\mathbf{y}}_s \sim p(\hat{\mathbf{y}}_s)} \big[\mathcal{L}_{s}(\mathbf{y}_s, \hat{\mathbf{y}}_s)] + \mathbb{E}_{\mathbf{y}_t \sim p(\mathbf{y}_t)} [\mathcal{L}_{t}(\mathbf{y}_t)\big]\Big]\\
\end{split}
\end{equation}
where $\theta$ is the parameters of $G$, $p(\mathbf{\cdot})$ is the probability density function. As the labels for $\mathcal{D}_s$ are available, $\mathcal{L}_s$ can be efficiently formulated as a supervised cross-entropy loss as in Eqn.~\eqref{eqn:bimal-cross-entropy}.
\begin{equation}\label{eqn:bimal-cross-entropy}
    \small
    \mathcal{L}_{s}(\mathbf{y}_s, \hat{\mathbf{y}}_s) = -
    \sum_{h,w, c} \hat{\mathbf{y}}^{h,w,c}_s\log\left(\mathbf{y}^{h,w,c}_s\right)
\end{equation}
where $\mathbf{y}^{h,w,c}$ and $\hat{\mathbf{y}}^{h,w,c}$ represent the predicted and ground-truth probabilities of the pixel at the location of $(h, w)$  taking the label of $c$, respectively. 
Meanwhile, $\mathcal{L}_t$ handles unlabeled data from
the target domain where the ground-truth labels are not available. To alleviate this label-lacking issue, several forms of $\mathcal{L}_{t}(\mathbf{y}_t)$ have been exploited such as cross-entropy loss with pseudo-labels~\cite{zou2018unsupervised}, Probability Distribution Divergence (i.e. Adversarial loss defined via an additional Discriminator)~\cite{tsai2018learning, tsai2019domain}, %
or entropy formulation~\cite{vu2019advent, pan2020unsupervised}.  

\noindent
\textbf{Entropy minimization revisited.}
By adopting the \mbox{Shannon} entropy formulation to the target prediction and constraining function $G$ to produce a high-confident prediction,
$\mathcal{L}_{t}$ can be formulated as in Eqn.~\eqref{eqn:bimal-entropy_loss}.
\begin{equation} \label{eqn:bimal-entropy_loss}
\small
    \mathcal{L}_{t}(\mathbf{y}_t) = 
    \frac{-1}{\log(C)}\sum_{h,w,c}\mathbf{y}^{h,w,c}_t\log\left(\mathbf{y}^{h,w,c}_t\right)
\end{equation}
Although this form of $\mathcal{L}_{t}$ can give a direct assessment of the predicted segmentation maps, 
it tends to be dominated 
by the high probability areas (since the high probability areas 
produce a higher value 
updated gradient due to  $\lim_{\mathbf{y}^{h,w,c}_t \to 1}\frac{-\partial \mathcal{L}_t(\mathbf{y}_t)}{\partial \mathbf{y}^{h,w,c}_t} = \frac{1}{\log(C)}$ and $\lim_{\mathbf{y}^{h,w,c}_t \to 0}\frac{-\partial \mathcal{L}_t(\mathbf{y}_t)}{\partial \mathbf{y}^{h,w,c}_t} = -\infty$),  %
i.e. easy classes, rather than difficult classes~\cite{vu2019advent}. 
More importantly, this is essentially a pixel-wise formation, where pixels are treated independently of each other.
Consequently, the structural information is usually neglected in this form. This issue could lead to a confusion point during training process where two predicted segmentation maps have similar entropy but different segmentation accuracy, one correct and other incorrect as shown in Figure~\ref{fig:bimal-same_entropy_1}.

\subsubsection{The Proposed Unaligned Domain Score Metric}

In the entropy formulation, the pixel independent constraints are employed to convert the image-level metric to pixel-level metric. In contrast,
we propose an image-level UDS metric that can directly evaluate the structural quality of $\mathbf{y}_t$. 
Particularly, let $p_t(\mathbf{y}_t)$ and $q_t(\mathbf{y}_t)$ be the probability mass functions of the predicted distribution and the real (actual) distribution of the predicted segmentation map $\mathbf{y}_t$, respectively.
UDS metric measuring the efficiency of function $G$ on the target dataset can be expressed as in Eqn.~\eqref{eqn:bimal-UDSEqn}.
\begin{equation}\label{eqn:bimal-UDSEqn}
\small
\begin{split}
    \operatorname{UDS}%
    &= \mathbb{E}_{\mathbf{y}_t \sim p(\mathbf{y}_t)} \mathcal{L_Y}\left(p_t(\mathbf{y}_t),q_t(\mathbf{y}_t)\right) = \int \mathcal{L_Y}\left(p_t(\mathbf{y}_t),q_t(\mathbf{y}_t)\right) p_t(\mathbf{y}_t) d\mathbf{y}_t 
\end{split}
\end{equation}
where 
$\mathcal{L_Y}\left(p_t(\mathbf{y}_t),q_t(\mathbf{y}_t)\right)$ defines the distance between %
two distributions $p_t(\mathbf{y}_t)$ and $q_t(\mathbf{y}_t)$. Since there is no label for sample in the target domain, the direct access to $q_t(\mathbf{y}_t)$ is not available. 
Note that although $\mathbf{x}_s$ and $\mathbf{x}_t$ could vary significantly in image space (e.g. difference in pixel appearance due to lighting, scenes, weather), their segmentation maps $\mathbf{y}_t$ and $\mathbf{y}_s$ share similar distributions in terms of both class distributions as well as global and local structural constraints (sky has to be above roads, trees should be on sidewalks, vehicles should be on roads, etc.). 
Therefore, one can practically adopted the prior knowledge learned from segmentation labels of the source domains for $q_t(\mathbf{y}_t)$ as in Eqn.~\eqref{eqn:bimal-DDS_score_label_extend}.
\begin{equation} \label{eqn:bimal-DDS_score_label_extend}
\small
\begin{split}
    \operatorname{UDS} %
    &\approx \int \mathcal{L_Y}\left(p_t(\mathbf{y}_t),q_s(\mathbf{y}_t)\right) p_t(\mathbf{y}_t) d\mathbf{y}_t 
\end{split}
\end{equation}
where the distribution $q_s(\mathbf{y}_t)$ 
is the probability mass functions of the real distribution learned from ground-truth segmentation maps of $\mathcal{D}_s$. %
As a result, the proposed USD metric can be computed without the requirement of labeled target data for learning the density of segmentation maps in the target domain. 
There are several choices for $\mathcal{L_Y}$ to estimate the divergence between the two distributions $p_t(\mathbf{y}_t)$ and $q_s(\mathbf{y}_t)$. In our approach, we adopt the common metric such as the Kullback–Leibler (KL) formula for $\mathcal{L_Y}$. 
Note that other metrics are also applicable in the proposed UDS formulation.
Moreover, to enhance the smoothness of the predicted semantic segmentation, a regularization term $\tau$ is imposed into $\mathcal{L_Y}$ as in Eqn.~\eqref{eqn:bimal-LYUpdated}.
\begin{equation} \label{eqn:bimal-LYUpdated}
\begin{split}
\small
    \mathcal{L_Y}\left(p_t(\mathbf{y}_t),q_s(\mathbf{y}_t)\right) &= \log\left(\frac{p_t(\mathbf{y}_t)}{q_s(\mathbf{y}_t)}\right) + \tau(\mathbf{y}_t) \\
\end{split}
\end{equation}
By computing UDS, one can measure the quality of the predicted segmentation maps $\mathbf{y}_t$ on the target data. 
In the next sections, we first discuss in detail the learning process of $q_s(\mathbf{y}_t)$, and then derivations of the UDS metric for the novel Bijective Maximum Likelihood loss. 

\subsection{Fundamental of Bijective Mapping to Distribution Modeling} \label{sec:bijective}

Let 
$F: \mathcal{Y} \to \mathcal{Z}$ 
be the bijective mapping function that maps a segmentation $\hat{\mathbf{y}}_s \in \mathcal{Y}$ to the latent space $\mathcal{Z}$, i.e. $\hat{\mathbf{z}}_s = F(\hat{\mathbf{y}}_s, \theta_F)$, where $\hat{\mathbf{z}}_s \sim q_z(\hat{\mathbf{z}}_s)$ is the latent variable, and $q_z$ is the prior distribution.
Then, the probability distribution $q_s(\hat{\mathbf{y}}_s)$ can be formulated via the change of variable formula as in Eqn.~\eqref{eqn:bimal-bijective_mapping}.
\begin{equation} \label{eqn:bimal-bijective_mapping}
\small
    \log(q_s(\hat{\mathbf{y}}_s)) = \log\left(q_{z}(\hat{\mathbf{z}}_s)\right) + \log\left(\left|\frac{\partial F(\hat{\mathbf{y}}_s, \theta_F)}{\partial \hat{\mathbf{y}}_s}\right|\right)
\end{equation}
where $\theta_F$ is the parameters of $F$, $\left|\frac{\partial F(\hat{\mathbf{y}}_s, \theta_F)}{\partial \hat{\mathbf{y}}_s}\right|$ denotes the Jacobian determinant of function $F(\hat{\mathbf{y}}_s, \theta_F)$ with respect to $\hat{\mathbf{y}}_s$. 
To learn the mapping function, the negative log-likelihood
will be minimized as in Eqn.~\eqref{eqn:bimal-BijectiveLearning}.
\begin{equation} \label{eqn:bimal-BijectiveLearning}
\small
\begin{split}
    \theta_F^{*} =& \arg\min_{\theta_F} %
    \mathbb{E}_{\hat{\mathbf{y}}_s \sim q_s(\hat{\mathbf{y}}_s)} \Big[-\log(q_s(\hat{\mathbf{y}}_s))\Big] \\
    =& \arg\min_{\theta_F} \mathbb{E}_{\hat{\mathbf{z}}_s \sim q_z(\hat{\mathbf{z}}_s)} \left[-\log\left(q_{z}(\hat{\mathbf{z}}_s)\right) - \log\left(\left|\frac{\partial F(\hat{\mathbf{y}}_s, \theta_F)}{\partial \hat{\mathbf{y}}_s}\right|\right)\right]
    \raisetag{40pt}
\end{split}
\end{equation}
In general, there are various choices for the prior distribution $q_z$. However, the ideal distribution should satisfy two 
criteria: (1) simplicity in the density estimation, and (2) easy in sampling. 
Considering the two criteria, we choose Normal distribution %
as the prior distribution $q_z$. 
Note that any other distribution is also feasible as long as it satisfies the mentioned criteria.

To enforce the information flow from a segmentation domain to a latent space with different abstraction levels, the bijective function $F$ can be further formulated as a composition of several sub-bijective functions $f_i$ as $F = f_1 \circ f_2 \circ ... \circ f_K$, 
where $K$ is the number of sub-functions. The Jacobian $\frac{\partial F}{\partial \mathbf{y}_s}$ can be derived by $\frac{\partial F}{\partial \hat{\mathbf{y}}_s} = \frac{\partial f_1}{\partial \hat{\mathbf{y}}_s} \cdot \frac{\partial f_2}{ \partial f_1} \cdots \frac{\partial f_K}{ \partial f_{K-1}}$. With this structure, the properties of each $f_i$ will define the properties for the whole bijective mapping function $F$. 
Interestingly, with this form, $F$ becomes a DNN structure when $f_i$ is a non-linear function built from a composition of convolutional layers. Several DNN structures~\cite{dinh2015nice, dinh2017density,Duong_2017_ICCV, duong2020vec2face, glow, duong2019learning, truong2021fastflow} can be adopted for sub-functions.

\subsection{Bijective Maximum Likelihood Loss} %

In this section, we present the proposed Bijective Maximum Likelihood (BiMaL) which can be used as the loss of target domain $\mathcal{L}_t$. 
From Eqns. \eqref{eqn:bimal-DDS_score_label_extend} and \eqref{eqn:bimal-LYUpdated}, %
UDS metric can be rewritten as in Eqn.~\eqref{eqn:bimal-KL_to_llk}.
\begin{equation} \label{eqn:bimal-KL_to_llk}
\small
\begin{split}
        \text{UDS} &= \int \left[\log\left(\frac{p_t(\mathbf{y}_t)}{q_s(\mathbf{y}_t)}\right)+  \tau(\mathbf{y}_t)\right]p_t(\mathbf{y}_t) d\mathbf{y}_t \\
        & = {\mathbb{E}}_{\mathbf{y}_t \sim p_t(\mathbf{y}_t)}\left[
        \log(p_t(\mathbf{y}_t))\right]  - {\mathbb{E}}_{\mathbf{y}_t \sim p_t(\mathbf{y}_t)} \left[\log(q_s(\mathbf{y}_t)) \right] 
        +  {\mathbb{E}}_{\mathbf{y}_t \sim p_t(\mathbf{y}_t)}\left[\tau(\mathbf{y}_t)\right] \\
        &\leq {\mathbb{E}}_{\mathbf{y}_t \sim p_t(\mathbf{y}_t)} \left[-\log(q_s(\mathbf{y}_t)) + \tau(\mathbf{y}_t)\right]  
\end{split}    
\end{equation}
It should be noticed that with any form of the distribution $p_t$, the above inequality still holds as $p_t(\mathbf{y}_t) \in [0,1]$ and $\log(p_t(\mathbf{y}_t)) \leq 0$. Now, we define our Bijective Maximum Likelihood Loss as in Eqn.~\eqref{eqn:bimal-BiMaL}.
\begin{equation} \label{eqn:bimal-BiMaL}
\small
\begin{split}
    \mathcal{L}_t(\mathbf{y}_t) = -\log(q_s(\mathbf{y}_t))+ \tau(\mathbf{y}_t)
\end{split}
\end{equation}
where $\log(q_s(\mathbf{y}_t))$ defines the log-likelihood of $\mathbf{y}_t$ with respect to the density function $q_s$.
Then, by adopting the bijectve function $F$ learned from Eqn.~\eqref{eqn:bimal-BijectiveLearning} using samples from source domain and the prior distribution $q_z$, the first term of $\mathcal{L}_t(\mathbf{y}_t)$ in Eqn.~\eqref{eqn:bimal-BiMaL} can be efficiently computed via log-likelihood formulation as in Eqn.~\eqref{eqn:bimal-define_llk}.
\begin{equation} \label{eqn:bimal-define_llk}
\begin{split}
    \mathcal{L}_{llk}(\mathbf{y}_t) &= -\log(q_s(\mathbf{y}_t)) =  -\log\left(q_{z}(\mathbf{z}_t)\right) - \log\left(\left|\frac{\partial F(\mathbf{y}_t, \theta_F)}{\partial \mathbf{y}_t}\right|\right)
\end{split}
\end{equation}
where $\mathbf{z}_t = F(\mathbf{y}_t, \theta_F)$. Thanks to the bijective property of the mapping function $F$, the minimum negative log-likelihood loss $\mathcal{L}_{llk}(\mathbf{y}_t)$
can be effectively computed via 
the density of the prior distribution $q_z$ and its associated Jacobian determinant $\left|\frac{\partial F(\mathbf{y}_t, \theta_F)}{\partial \mathbf{y}_t}\right|$. For the second term of $\mathcal{L}_t(\mathbf{y}_t)$, we further enhance the smoothness of the predicted semantic segmentation with the pair-wised formulation to encourage similar predictions for neighborhood pixels with similar color as in Eqn.~\eqref{eqn:bimal-define_tau}.
\begin{equation} \label{eqn:bimal-define_tau}
\small
    \tau(\mathbf{y}_t) = \sum_{h,w}\sum_{h', w'} \exp \left\{-\frac{||\mathbf{x}_t^{h,w} -  \mathbf{x}_t^{h', w'}||_2^2}{2\sigma_1^2} - \frac{||\mathbf{y}_t^{h,w} - \mathbf{y}_t^{h', w'}||_2^2}{2\sigma_2^2}\right\}%
\end{equation}
where $(h', w') \in \mathcal{N}_{h, w}$ denotes the neighbourhood pixels of $(h, w)$, $\mathbf{x}^{h,w}$ represents the color at pixel $(h, w)$; and $\{\sigma_1, \sigma_2\}$ are the hyper parameters controlling the scale of Gaussian kernels. 
It should be noted that any regularizers~\cite{chen2018deeplab, duong2029cvpr_automatic} enhancing the smoothness of the segmentation results can also be adopted for $\tau$.   
Putting Eqns. \eqref{eqn:bimal-BiMaL}, \eqref{eqn:bimal-define_llk}, \eqref{eqn:bimal-define_tau} to Eqn \eqref{eqn:bimal-objective}, the objective function can be rewritten as in Eqn.~\eqref{eqn:bimal-final-loss}.
\begin{equation}\label{eqn:bimal-final-loss}
\small
\begin{split}
         \theta^*=\arg\min_{\theta} \Big[&\mathbb{E}_{\mathbf{y}_s \sim p(\mathbf{y}_s), \hat{\mathbf{y}}_s \sim p(\hat{\mathbf{y}}_s)} \big[\mathcal{L}_{s}(\mathbf{y}_s, \hat{\mathbf{y}}_s)] + \mathbb{E}_{\mathbf{y}_t \sim p(\mathbf{y}_t)} [\mathcal{L}_{llk}(\mathbf{y}_t) + \tau(\mathbf{y}_t)\big]\Big]
\end{split}
\end{equation}
Figure~\ref{fig:bimal-proposed_framework} illustrates our proposed BiMaL framework to learn the deep segmentation network $G$. Also, we can prove that direct entropy minimization as Eqn.~\eqref{eqn:bimal-entropy_loss} is just a particular case of our log likelihood maximization. We will further discuss how our maximum likelihood can cover the case of pixel-independent entropy minimization in the next section.

\subsubsection{BiMaL properties}  

\noindent
{\textbf{Global Structure Learning.}} Sharing similar property with~\cite{duong2016dam_cvpr, duong2019dam_ijcv, duong2020vec2face, Duong_2017_ICCV, 9108692}, from Eqn.~\eqref{eqn:bimal-bijective_mapping}, as the learned density function is adopted for the entire segmentation map $\hat{\mathbf{y}}_s$, the global structure in $\hat{\mathbf{y}}_s$ can be efficiently captured and modeled.

\noindent
{\textbf{Tractability and Invertibility.}} Thanks to the designed bijection F, the complex distribution of segmentation maps can be efficiently captured. Moreover, the mapping function is bijective, and, therefore, both inference and generation are exact and tractable.

\subsubsection{Relation to Entropy Minimization} 
The first term of UDS in Eqn.~\eqref{eqn:bimal-KL_to_llk} can be derived as in Eqn.~\eqref{eqn:bimal-proof-upperbound}.
\begin{equation} \label{eqn:bimal-proof-upperbound}
\small
\begin{split}
     \int \log\left(\frac{p_t(\mathbf{y})}{q_s(\mathbf{y})}\right)p_t(\mathbf{y}_t) d\mathbf{y}_t &\geq 0 
     \Leftrightarrow   {\mathbb{E}}_{\mathbf{y}_t \sim p_t(\mathbf{y}_t)}\left[\log(p_t(\mathbf{y}_t)) - \log(q_s(\mathbf{y}_t))\right] \geq 0 \\
    \Leftrightarrow {\mathbb{E}}_{\mathbf{y}_t \sim p_t(\mathbf{y}_t)}\left[-\log(q_s(\mathbf{y}_t))\right] &\geq {\mathbb{E}}_{\mathbf{y}_t \sim p_t(\mathbf{y}_t)}\left[-\log(p_t(\mathbf{y}_t))\right]  \Leftrightarrow  \mathbb{E}_{\mathbf{y}_t \sim p_t(\mathbf{y}_t)}[\mathcal{L}_{llk}(\mathbf{y}_t)] \geq \text{Ent}(\mathbf{Y}_t)%
\end{split}
\end{equation}
where $\mathbf{Y}_t$ is the random variable with possible values $\mathbf{y}_t \sim p_t(\mathbf{y}_t)$, and $\text{Ent}(\mathbf{Y}_t)$ denotes the entropy of the random variable $\mathbf{Y}_t$.
It can be seen that the proposed negative log-likelihood $\mathcal{L}_{llk}$ is an upper bound of the entropy of $\mathbf{Y}_t$. Therefore, minimizing our proposed BiMaL loss will also enforce the entropy minimization process. 
Moreover, by not assuming pixel independence, our proposed BiMaL can model and evaluate structural information at the image-level better than previous pixel-level approaches~\cite{chen2019domain, pan2020unsupervised, vu2019advent}.

\subsection{Experimental Results}

This section will present our experimental results on three different benchmarks, i.e. SYNTHIA $to$ Cityscapes, GTA $to$ Cityscapes, and SYNTHIA $to$ Vistas. First, we overview datasets and network architectures used in our experiments. Second, we present the ablation study to analyze the effectiveness of our proposed BiMaL and the capability of the bijective network. Finally, we present the quantitative and qualitative results of our method compared to prior methods on the three benchmarks.

\subsubsection{Datasets}

\noindent
\textbf{GTA5}~\cite{Richter_2016_ECCV} is a synthetic dataset containing $24,966$ densely labelled images at the resolution of $1,914 \times 1,052$. This dataset was collected from the game Grand Theft Auto V. The ground-truth annotations were automatically generated with 33 categories. In our experiments, we consider 19 categories that are compatible with the Cityscapes~\cite{cordts2016cityscapes}.

\noindent
\textbf{SYNTHIA (SYNHIA-RAND-CITYSCAPES) }\cite{Ros_2016_CVPR} is also synthetic dataset that contains $9,400$ pixel-level labelled RGB images. In our experiments, we use the 16 common categories that overlap with the Cityscapes dataset.

\noindent
\textbf{Cityscapes}~\cite{cordts2016cityscapes} is a real-world dataset including $3,975$ images with fine semantic, dense pixel annotations of 30 classes.
In our experiments, $2,495$ images are used for training and $500$ images are used for testing.

\noindent
\textbf{Vistas (Mapillary Vistas Dataset)}~\cite{MVD2017} is diverse street-level imagery dataset with pixel‑accurate and instance‑specific human annotations for understanding street scenes around the world. Vistas consists of $25,000$ high-resolution images and $124$ semantic object categories. In our experiments, we consider 7 classes that are common to SYNTHIA, Cityscapes and Vistas.

\begin{wrapfigure}{r}{0.5\textwidth}
    \centering
    \includegraphics[width=0.5\textwidth]{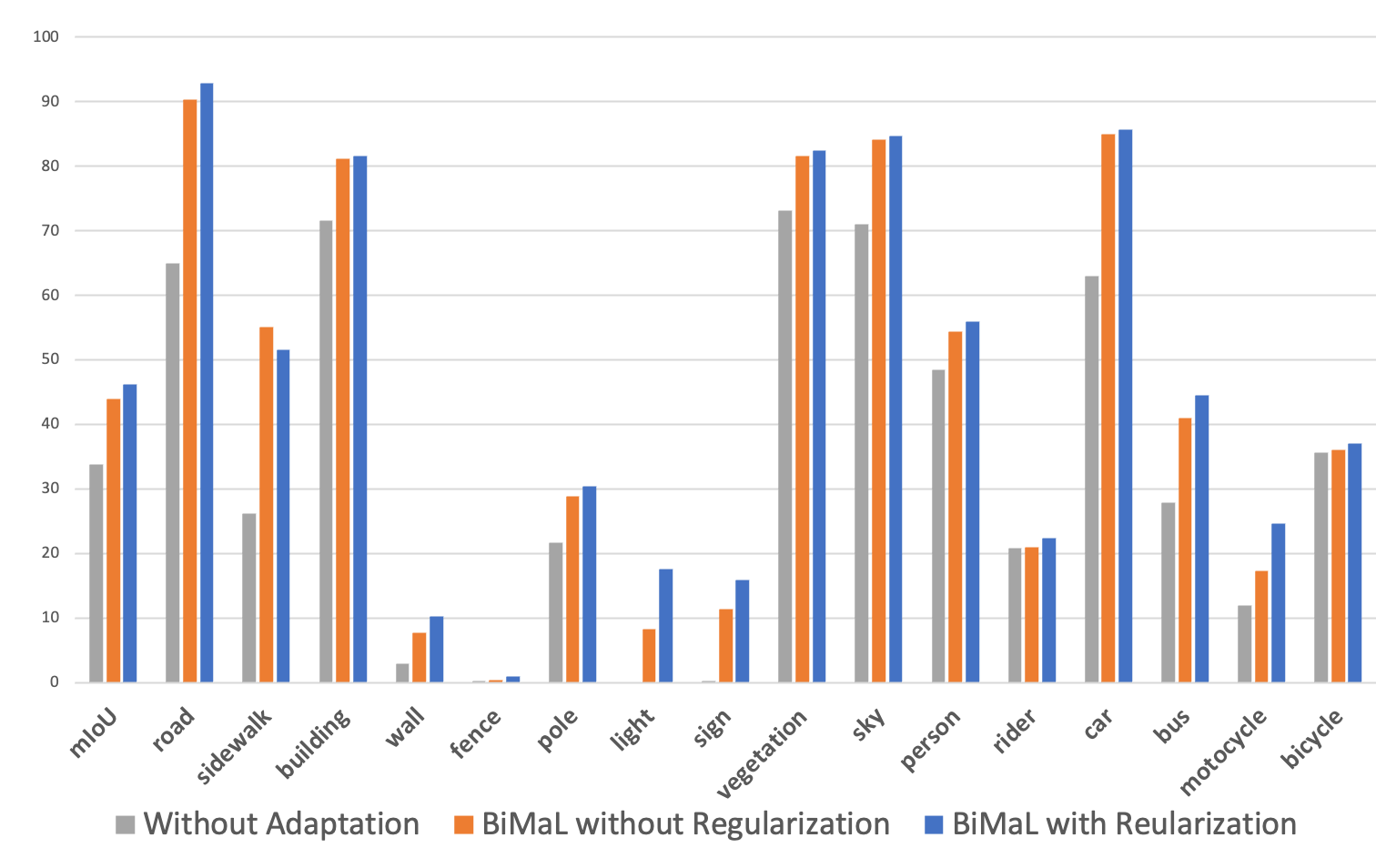}
    \caption{Ablative semantic segmentation performance mIoU (\%) on the effectiveness of the proposed BiMaL loss.}
    \label{fig:bimal-ablation_loss}
\end{wrapfigure}
\noindent
\textbf{Network Architectures.} 
In our experiments, we adopt the DeepLab-V2~\cite{chen2018deeplab} with ResNet-101~\cite{resnet} backbone for the segmentation network $G$. Also, we utilize the Atrous Spatial Pyramid Pooling with sampling rate $\{6, 12, 18, 24\}$. We only use the output of layer \textit{conv5} to predict the segmentation.
In the Bijective network $F$, we use the multi-scale architecture as~\cite{dinh2015nice, dinh2017density, duong2019learning, glow, Duong_2017_ICCV}. For each scale, we have multiple steps of flow, each of which consists of ActNorm, Invertible $1\times 1$ Convolution, and Affine Coupling Layer~\cite{glow, 9108692}. In our experiments, the number of scales and number of flow steps are set to $4$ and $32$, respectively. 
The entire framework is implemented in PyTorch~\cite{paszke2019pytorch}. Training and validating models are conducted on 4 GPUs of NVIDIA Quadpro P8000 with 48GB each GPU. Segmentation and bijective networks are trained by a Stochastic Gradient Descent optimizer~\cite{bottou2010large} with learning rate $2.5\times 10^{-4}$, momentum $0.9$, and weight decay $10^{-4}$. The batch size per GPU is set to $4$ for segmentation network, and $16$ for learning bijective network. 
The image size is set to $1280 \times 720$ pixels in all experiments.

\subsubsection{Ablation Study}

\noindent
\textbf{Effectiveness of Losses.} Figure~\ref{fig:bimal-ablation_loss} reports the semantic performance (mIoU) of BiMaL on the 16 classes of the Cityscape validation set when the model is trained on \mbox{SYNTHIA} dataset. We consider three cases: (1) without adaptation (train with source only), (2) BiMaL without regularization term ($\mathcal{L}_{llk}(\mathbf{y})$ only), and (3) BiMaL with regularization term ($\mathcal{L}_{llk}(\mathbf{y}) + \tau(\mathbf{y})$). Overall, the proposed BiMaL improve the performance of the method. In particular, the mIoU accuracy of the baseline (without adaptation) is $33.7\%$. 
In comparison, BiMaL without regularization and BiMaL with regularization achieve the mIoU accuracy of $43.5\%$ and $46.2\%$, respectively. 
In terms of per-class accuracy, using BiMaL significantly improves the performance on classes of \textit{`road'}, \textit{`sidewalk'}, \textit{`bus'}, and \textit{`motocycle'}.

\begin{wrapfigure}[14]{r}{0.5\textwidth}
    \centering
    \includegraphics[width=0.5\textwidth]{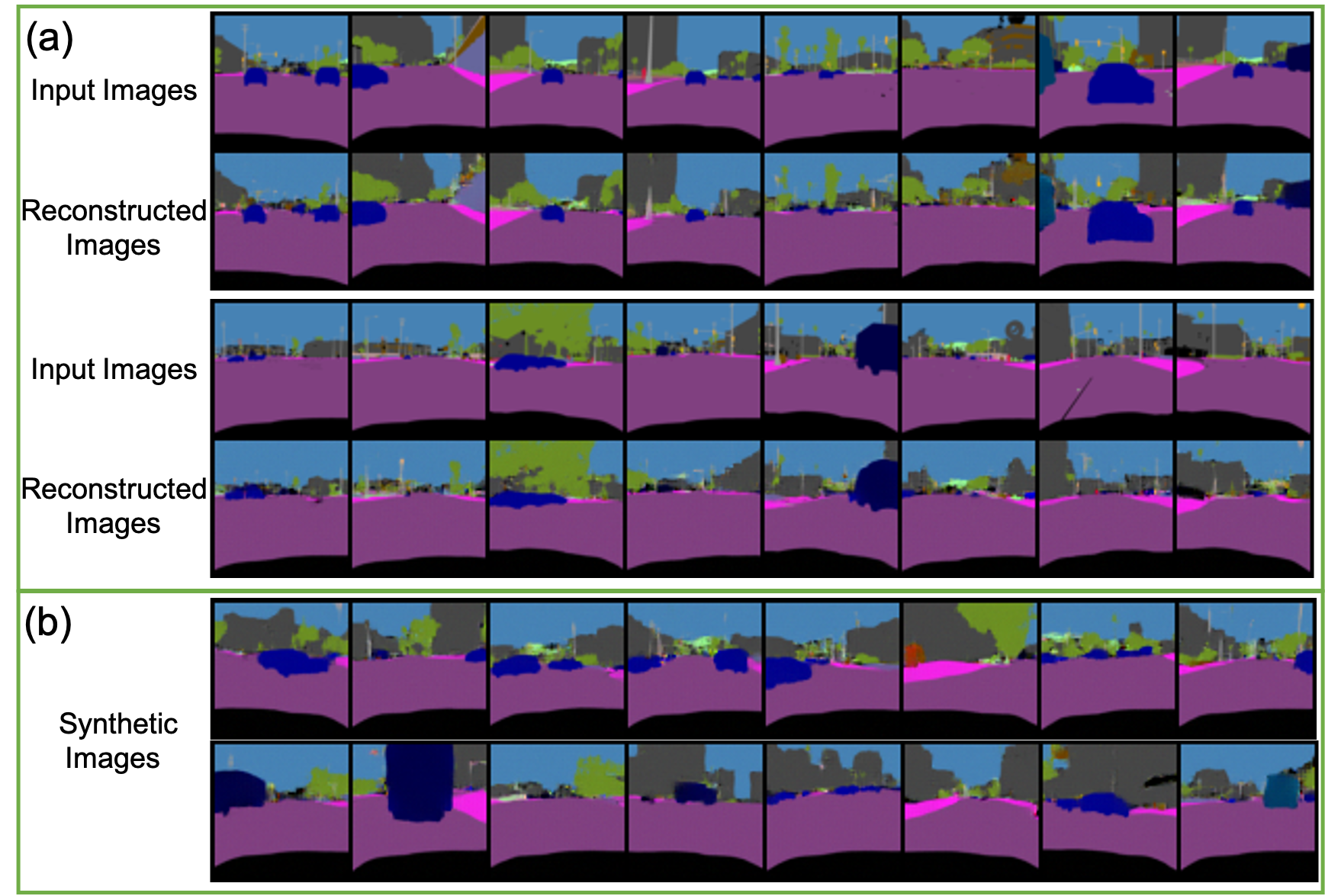}
    \caption{\textbf{Reconstructed Images and Synthetic Images From The Bijective Mapping Function $F$}. (a) Reconstructed images (bottom row) from the corresponding input images (top row). (b) Synthetic images sampled from the latent space.} %
    \label{fig:bimal-rec_images}
\end{wrapfigure}
\noindent
\textbf{Bijective Network Ability. } 
We conduct a pilot experiment of the bijective network on ground-truth semantic segmentation images of the GTA dataset.
This experiment aims to analyze the ability of the bijective network in modeling the image and structure information.
The number of scales and number of flow steps are set to $3$, and $32$, respectively. 
As shown in Figure~\ref{fig:bimal-rec_images}(a), our bijective network can successfully reconstruct good-quality images. 
It also can synthesize images sampled from the latent space as shown in Figure~\ref{fig:bimal-rec_images}(b). These results have shown that the bijective network can model images even with complex structures as scene segmentation.

\subsubsection{Comparisons with State-of-the-Art Methods}

\input{Tables/chap-3/bimal/synthia2city-results}

\input{Tables/chap-3/bimal/gta2city-results}

We present the experimental results of the proposed approach in comparison to other strong baselines.
Comparative experiments are conducted on three benchmarks: 
i.e. SYNTHIA $to$ Cityscapes, GTA5 $to$ Cityscapes, and SYNTHIA $to$ Vistas. In all three benchmarks, our method consistently achieves the SOTA segmentation performance in terms of ``mean Intersection over Union" (mIoU).

\noindent
\textbf{SYNTHIA $to$ Cityscapes.} Table~\ref{tab:bimal-synthia2city} presents the semantic performance (mIoU) on the 16 classes of the Cityscape validation set. Our proposed method achieves better accuracy than the prior methods, i.e. $46.2\%$ higher than DADA~\cite{vu2019dada} by $3.6\%$. Considering per-class results, our method significantly improves the results on classes of \textit{`sidewalk'} ($51.5\%$), \textit{`car'} ($85.7\%$), and \textit{`bus'} ($44.5\%$).

\noindent
\textbf{GTA5 $to$ Cityscapes.} Table~\ref{tab:bimal-gta52city} shows the mIoU of 19 classes of Cityscapes on the validation set. Our approach gains mIoU of $47.3\%$ that is state-of-the-art performance compared to the prior methods. Analysing per-class results,
our method gains the improvement on most classes. In particular, the results on classes of \textit{`terrain'} (+$10.0\%$), \textit{`truck'} ($+9.1\%$), \textit{`bus'} ($+8.0\%$), \textit{`motorbike'} ($+5.6\%$) demonstrate significant improvements compared to AdvEnt. 
For other classes, the proposed method gains moderate improvements, compared to prior SOTA methods.

\input{Tables/chap-3/bimal/synthia2vistas-results}
\noindent
\textbf{SYNTHIA $to$ Vistas.} Table~\ref{tab:bimal-synthia2vistas} reports the mIoU of the Vistas testing set. Our approach gains a mIoU of $67.2\%$, which is the SOTA performance compared to prior methods. 
Moreover, our method also gains moderate improvements in per-class accuracy.%

\begin{figure}[!t]
    \centering
    \includegraphics[width=1.0\textwidth]{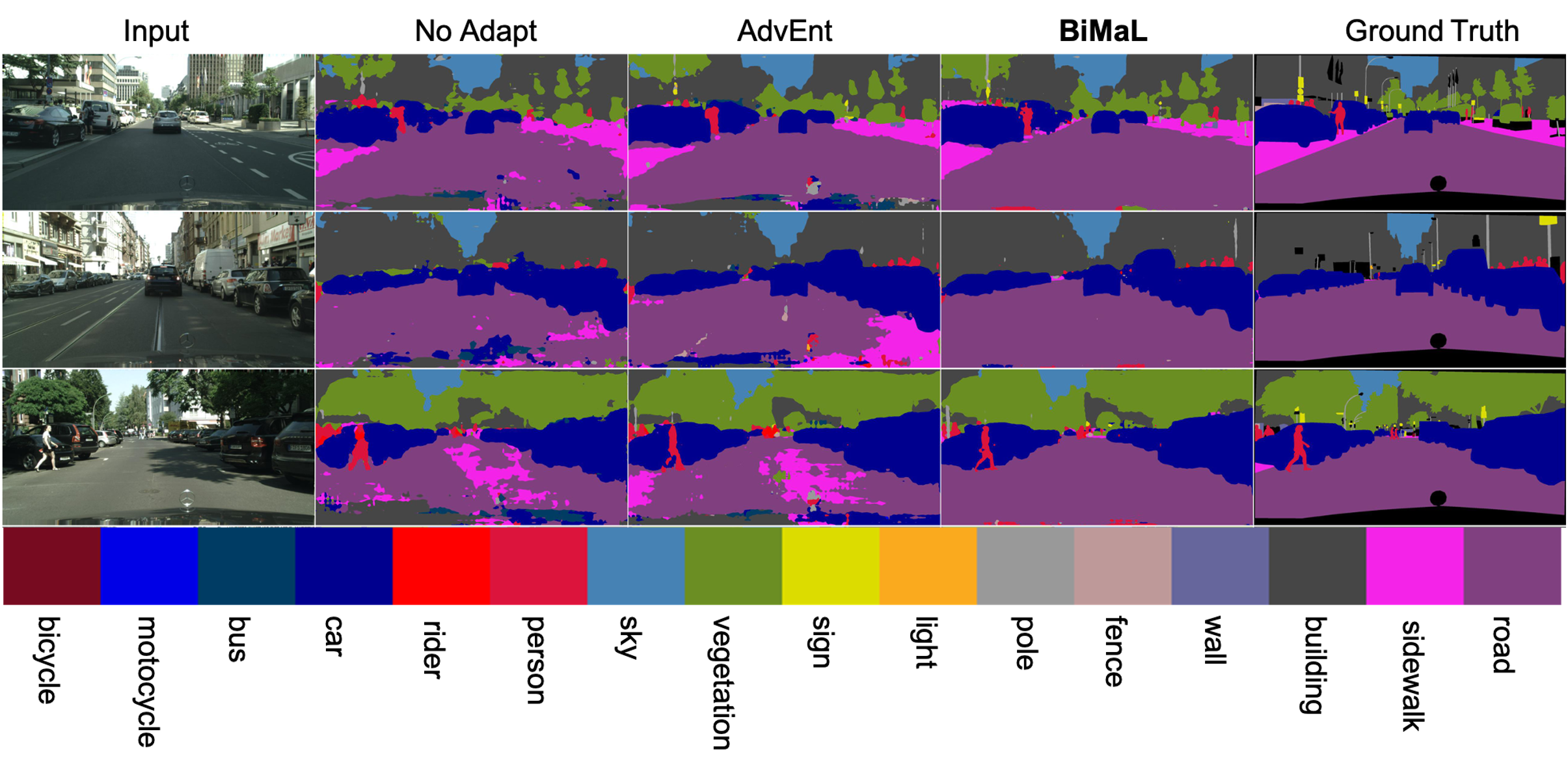}
    \caption{\textbf{Qualitative results of the SYNTHIA $\to$ Cityscapes experiment.} Columns 1 and 5 are the inputs and ground truths. Columns 2, 3 and 4 are the results of the model without adaptation, AdvEnt~\cite{vu2019advent}, and BiMaL.} %
    \label{fig:bimal-qual_res_synthia2cityscape}
\end{figure}

\noindent
\textbf{Qualitative Results.} Figure~\ref{fig:bimal-qual_res_synthia2cityscape} illustrates the qualitative results of the SYNTHIA $to$ Cityscapces experiment. %
Our method gives the better qualitative results compared to a model trained on the source domain and AdvEnt~\cite{vu2019advent}. 
Our method can model well the structure of an image. In particular, our results have a clear border between \textit{`road'} and \textit{`sidewalk'}. Meanwhile, the results of model trained on source only and AdvEnt have an unclear border between \textit{`road'} and \textit{`sidewalk'}. Overall, our qualitative semantic segmentation results are sharper than the results of AdvEnt.

%% file: Tables/chap-3/bimal/method-comparison.tex
\begin{table}[t]
\centering
\caption{ \textbf{Comparison in the properties between our proposed approach and other methods}. Convolutional Neural Network (CNN), Generative Adversarial Net (GAN), Bijective Network (BiN), Entropy Minimization (EntMin), Curriculum Training (CT), Image-wise Weighting (IW), Segmentation Map (Seg), Depth Map (Depth); $\ell_{CE}$: Cross-entropy Loss, $\ell_{adv}$: Adversarial Loss, $\ell_{Huber}$: Huber Loss.}
\resizebox{1.0\textwidth}{!}{
\begin{tabular}{|l|c|c|c|c|c|}
\hline
\textbf{Methods} & \textbf{Architecture}              & \textbf{Source Label} & \textbf{Learning Mechanism} & \textbf{Loss Function}                       & \textbf{Structural Learning} \\ \hline
AdaptSeg \cite{tsai2018learning}        & CNN + GAN                          & Seg  & Domain Adaptation & $\ell_{adv}$                                 & Weak (binary label)                                     \\ \hline
AdaptPatch \cite{tsai2019domain}     & CNN + GAN                          & Seg & Domain Adaptation & $\ell_{adv}$                                  & Weak (binary label)                                     \\ \hline
CBST \cite{zou2018unsupervised}             & CNN                                & Seg &           Self-Training                    &       $\ell_{CE}$                          & Not Applicable                                    \\ \hline
ADVENT \cite{vu2019advent}          & CNN + GAN                          & Seg   & Domain Adaptation & EntMin                         & Weak (binary label)                                     \\ \hline
MaxSquare \cite{chen2019domain}         & CNN + GAN                          & Seg  & Domain Adaptation & Squares loss + IW & Weak (binary label)                                     \\ \hline
IntraDA \cite{pan2020unsupervised}         & CNN + GAN                          & Seg  & Curriculum Learning & EntMin & Weak (binary label)                                     \\ \hline \hline
SPIGAN \cite{lee2018spigan}         & CNN + GAN                          & Seg + Depth & Domain Adaptation & $\ell_{adv}$ + $\ell_{1}$                          & Weak (binary label)                                    \\ \hline
DADA \cite{vu2019dada}            & CNN + GAN                          & Seg + Depth  & Domain Adaptation & $\ell_{adv}$ + $\ell_{Huber}$                          & Depth-aware Label                                   \\ 
\hline \hline
\textbf{BiMaL}  & \textbf{CNN + BiN} & \textbf{Seg}  & \textbf{Domain Adaptation} & \textbf{Maximum   Likelihood}      & \begin{tabular}{@{}c@{}} \textbf{Segmentation Density} \\ \textbf{(Unsupervised)}\end{tabular}                           \\ \hline
\end{tabular}
}
\label{tab:bimal-summary}
\end{table}

%% file: Tables/chap-3/bimal/synthia2city-results.tex
\begin{table}[!t]
        \centering
        \caption{\textbf{Semantic segmentation performance mIoU (\%) on Cityscapes validation set of different models trained on SYNTHIA}. We also show the mIoU (\%) of the $13$ classes (mIoU*) excluding classes with *.}
            \resizebox{\textwidth}{!}{%
            \begin{tabular}{|l|c c c c c c c c c c c c c c c c |c c| c | c c |c c|}
				\multicolumn{19}{c}{ SYNTHIA $\rightarrow$ Cityscapes (16 classes)}\\
				\hline
				Models  & \rotatebox{90}{\textbf{road}} & \rotatebox{90}{\textbf{sidewalk}} & \rotatebox{90}{\textbf{building}} & \rotatebox{90}{\textbf{wall*}} & \rotatebox{90}{\textbf{fence*}} & \rotatebox{90}{\textbf{pole*}} & \rotatebox{90}{\textbf{light}} & \rotatebox{90}{\textbf{sign}} & \rotatebox{90}{\textbf{veg}} & \rotatebox{90}{\textbf{sky}} & \rotatebox{90}{\textbf{person}} & \rotatebox{90}{\textbf{rider}} & \rotatebox{90}{\textbf{car}} & \rotatebox{90}{\textbf{bus}} & \rotatebox{90}{\textbf{mbike}} & \rotatebox{90}{\textbf{bike}} & \rotatebox{90}{\textbf{mIoU}} & \rotatebox{90}{\textbf{mIoU*}}\\
				\hline
				Without Adaptation & 64.9 & 26.1 & 71.5 & 3.0 & 0.2 & 21.7 & 0.1 & 0.2 & 73.1 & 71.0 & 48.4 & 20.7 & 62.9 & 27.9 & 12.0 & 35.6 & 33.7 & 39.6 \\
				SPIGAN-no-PI~\cite{lee2018spigan}&69.5&29.4&68.7&4.4&0.3&32.4&5.8&15.0&81.0&78.7&52.2&13.1&72.8&23.6&7.9&18.7&35.8&41.2\\
				SPIGAN~\cite{lee2018spigan}&71.1&29.8&71.4&3.7&0.3&\textbf{33.2}&6.4&{15.6}&81.2&78.9&52.7&13.1&75.9&25.5&10.0&20.5&36.8&42.4\\
				AdaptSegnet~\cite{tsai2018learning}&79.2&37.2&78.8&-&-&-&9.9&10.5&78.2&80.5&53.5&19.6&67.0&29.5&21.6&31.3&-&45.9\\
				AdaptPatch~\cite{tsai2019domain}&82.2&39.4&79.4&-&-&-&6.5&10.8&77.8&82.0&54.9&21.1&67.7&30.7&17.8&32.2&-&46.3\\
				CLAN~\cite{luo2018taking}&81.3&37.0&80.1&-&-&-&{16.1}&13.7&78.2&81.5&53.4&21.2&73.0&32.9&{22.6}&30.7&-&47.8\\
				AdvEnt~\cite{vu2019advent}&87.0&44.1&79.7&{9.6}&{0.6}&24.3&4.8&7.2&80.1&83.6&{56.4}&\textbf{23.7}&72.7&32.6&12.8&33.7&40.8&47.6\\
				IntraDA \cite{pan2020unsupervised} & 84.3 & 37.7 & 79.5 & 5.3 & 0.4 & 24.9 & 9.2 & 8.4 & 80.0 & 84.1 & \textbf{57.2} & 23.0 & 78.0 & 38.1 & 20.3 & 36.5 & 41.7 & 48.9 \\
				DADA\cite{vu2019dada} &{89.2}&{44.8}&{81.4}&6.8&0.3&26.2&8.6&11.1&{81.8}&{84.0}&54.7&19.3&{79.7}&{40.7}&14.0&{38.8}&{42.6}&{49.8} \\
				\textbf{Our BiMaL} & \textbf{92.8} & \textbf{51.5} & \textbf{81.5} & \textbf{10.2} & \textbf{1.0} & 30.4 & \textbf{17.6} & \textbf{15.9} & \textbf{82.4} & \textbf{84.6} & 55.9 & 22.3 & \textbf{85.7} & \textbf{44.5} & \textbf{24.6} & \textbf{38.8} & \textbf{46.2} & \textbf{53.7} \\
				\hline
			\end{tabular}
			}
			\label{tab:bimal-synthia2city} 
\end{table}

%% file: Tables/chap-3/bimal/gta2city-results.tex
\begin{table}[t]
    \centering
     \caption{\textbf{Semantic segmentation performance mIoU (\%) on Cityscapes validation set of different models trained on GTA5}}
    \resizebox{\textwidth}{!}{
    \begin{tabular}{|l|c c c c c c c c c c c c c c c c c c c |c|}
        \multicolumn{21}{c}{GTA5 $\to$ Cityscapes (19 classes)}\\
        \hline 
        Models & \rotatebox{90}{\textbf{road}} & \rotatebox{90}{\textbf{sidewalk}} & \rotatebox{90}{\textbf{building}} & \rotatebox{90}{\textbf{wall}} & \rotatebox{90}{\textbf{fence}} & \rotatebox{90}{\textbf{pole}} & \rotatebox{90}{\textbf{light}} & \rotatebox{90}{\textbf{sign}} & \rotatebox{90}{\textbf{\textbf{veg}}} & \rotatebox{90}{\textbf{terrain}} & \rotatebox{90}{\textbf{sky}} & \rotatebox{90}{\textbf{\textbf{person}}} & \rotatebox{90}{\textbf{rider}} & \rotatebox{90}{\textbf{car}}& \rotatebox{90}{\textbf{truck}} & \rotatebox{90}{\textbf{bus}} & \rotatebox{90}{\textbf{train}} & \rotatebox{90}{\textbf{mbike}} & \rotatebox{90}{\textbf{bike}} & \rotatebox{90}{\textbf{mIoU}} \\
        \hline
        Without Adaptation~\cite{tsai2018learning}  & 75.8 & 16.8 & 77.2 & 12.5 & 21.0 & 25.5 & 30.1 & 20.1 & 81.3 & 24.6 & 70.3 & 53.8 & 26.4 & 49.9 & 17.2 & 25.9 & \textbf{6.5} & 25.3 & 36.0 & 36.6 \\
        ROAD~\cite{chen2018CVPR}                    & 76.3 & 36.1 & 69.6 & 28.6 & 22.4 & {28.6} & 29.3 & 14.8 & 82.3 & 35.3 & 72.9 & 54.4 & 17.8 & 78.9 & 27.7 & 30.3 & 4.0 & 24.9 & 12.6 & 39.4 \\
        AdaptSegNet~\cite{tsai2018learning}         & 86.5 & 36.0 & 79.9 & 23.4 & 23.3 & 23.9 & {35.2} & 14.8 & 83.4 & 33.3 & 75.6 & 58.5 & 27.6 & 73.7 & 32.5 & 35.4 & 3.9 & 30.1 & 28.1 & 42.4 \\
        MinEnt~\cite{vu2019advent}                  & 84.2 & 25.2 & 77.0 & 17.0 & 23.3 & 24.2 & 33.3 & \textbf{26.4} & 80.7 & 32.1 & 78.7 & 57.5 & {30.0} & 77.0 & {37.9} & 44.3 & 1.8 & 31.4 & \textbf{36.9} & 43.1 \\
        AdvEnt~\cite{vu2019advent}                  & {89.9} & {36.5} & {81.6} & {29.2} & {25.2} & {28.5} & 32.3 & 22.4 & 83.9 & 34.0 & 77.1 & 57.4 & 27.9 & {83.7} & 29.4 & 39.1 & 1.5 & 28.4 & 23.3 & 43.8 \\
        
        \textbf{Our BiMaL} & \textbf{91.2} & \textbf{39.6} & \textbf{82.7} & \textbf{29.4} & \textbf{25.2} & \textbf{29.6} & \textbf{34.3} & 25.5 & \textbf{85.4} & \textbf{44.0} & \textbf{80.8} & \textbf{59.7} & \textbf{30.4 }& \textbf{86.6 }& \textbf{38.5} & \textbf{47.6} & 1.2 & \textbf{34.0} & 36.8 & \textbf{47.3} \\ 
        \hline 
    \end{tabular}
    }
    \label{tab:bimal-gta52city}
    \vspace{-4mm}
\end{table}

%% file: Tables/chap-3/bimal/synthia2vistas-results.tex
\begin{wraptable}{r}{0.5\textwidth}
    \centering
    \caption{\textbf{Semantic segmentation performance mIoU (\%) on Vistas testing set of different models trained on SYNTHIA.} (const. denotes for construction)}
    \resizebox{0.5\textwidth}{!}{
    \begin{tabular}{|l|c c c c c c c|c|}
    \multicolumn{9}{c}{SYNTHIA $\to$ Vistas (7 classes)}\\
    \hline
    Models & \rotatebox{90}{\textbf{flat}} & \rotatebox{90}{\textbf{const.}} & \rotatebox{90}{\textbf{object}} & \rotatebox{90}{\textbf{nature}} & \rotatebox{90}{\textbf{sky}} & \rotatebox{90}{\textbf{human}} & \rotatebox{90}{\textbf{vehicle}} & \rotatebox{90}{\textbf{mIoU}} \\
    \hline
     SPIGAN-no-PI~\cite{lee2018spigan}  &  53.0 & 30.8 & 3.6 & 14.6 & 53.0 & 5.8 & 26.9 & 26.8       \\
     SPIGAN~\cite{lee2018spigan}  &   74.1 & 47.1 & 6.8 & 43.3 & 83.7 & 11.2 & 42.2 & 44.1      \\
    AdvEnt \cite{vu2019advent}  & 86.9 & 58.8 & 30.5 & 74.1 & 85.1 & 48.3 & 72.5 & 65.2 \\
    DADA \cite{vu2019dada} & 86.7 & \textbf{62.1} & 34.9 & 75.9 & \textbf{88.6} & 51.1 & 73.8 & 67.6 \\
    \textbf{Our BiMaL} & \textbf{87.6} & 61.6 &  \textbf{35.3} & \textbf{77.5} &  87.8 &  \textbf{53.3} & \textbf{75.6} & \textbf{68.4} \\
     \hline 
    \end{tabular}
    }
    \label{tab:bimal-synthia2vistas}
\end{wraptable}

%% file: Chapters/Sections/chap-3/chap-3-fredom.tex
\section{Fairness Domain Adaptation Approach to Semantic Scene Understanding}
\label{sec:fredom-paper}

\setcounter{propositioncounter}{0}
\setcounter{remarkcounter}{0}

\begin{figure}[!b]
    \centering
    \includegraphics[width=1.0\textwidth]{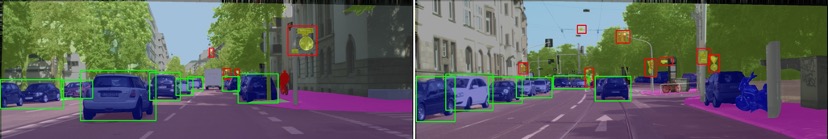}
    \caption{\textbf{Illustration of the Presence of Classes between Major (green boxes) and Minor (red boxes) Groups}. Classes in the minority group typically occupy fewer pixels than the ones in the majority group.}
    \label{fig:fredom-major_minor_fig}
\end{figure}

Despite the phenomenal achievement of advanced semantic segmentation models~\cite{chen2018deeplab, chen2018encoder, lin2017refinenet, xie2021segformer}, these data-driven approaches still need to improve in treating the prediction of each class. In particular, the segmentation models typically treat unfairly between classes in the dataset according to the class distributions. It is known as the fairness problem of semantic segmentation. 

\begin{wrapfigure}{r}{0.5\textwidth}
    \centering
    \includegraphics[width=0.5\textwidth]{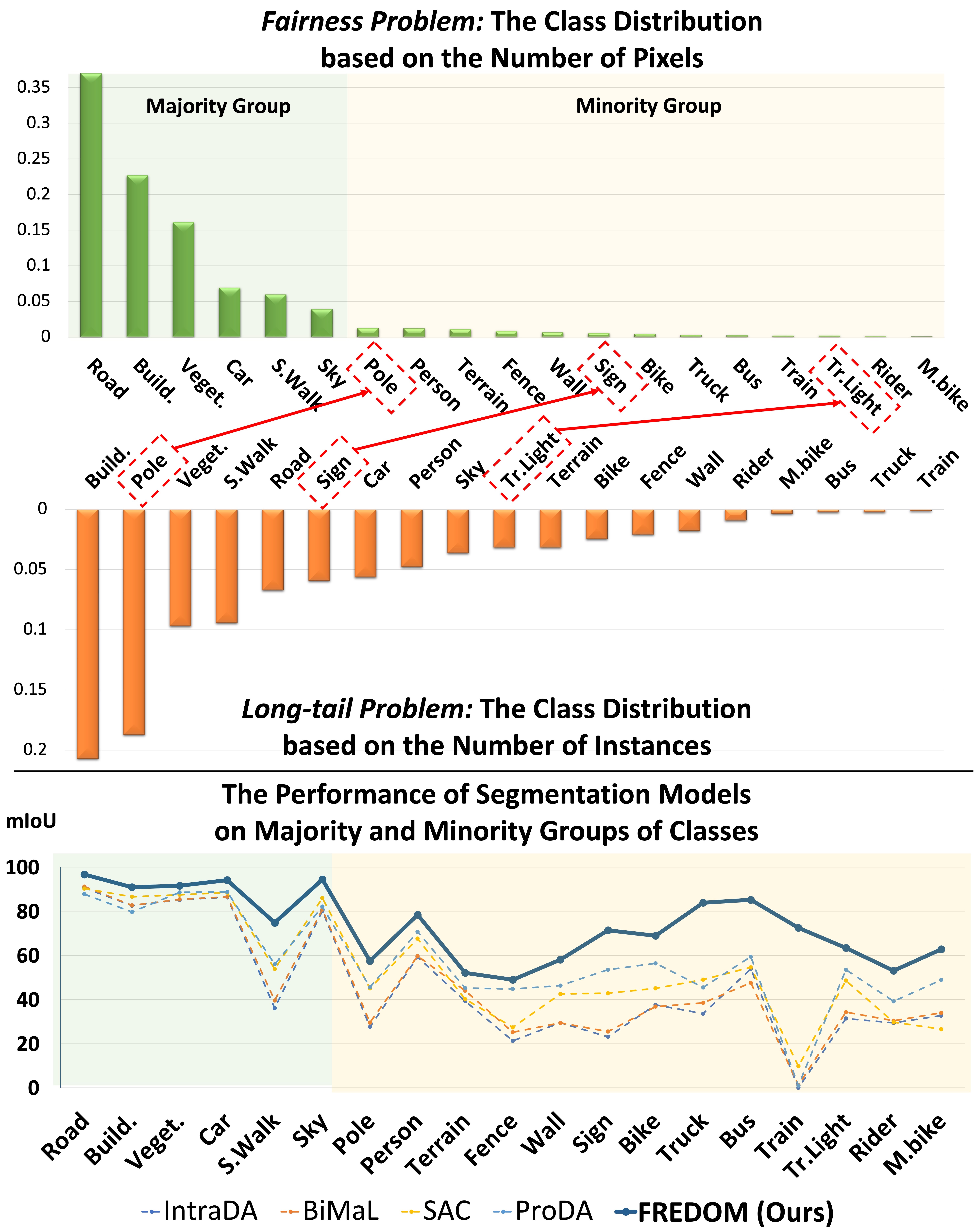}
    \caption{\textbf{The class distributions on Cityscapes are defined for Fairness problem and Long-tail problem.} In \textit{long-tail} problem, several head classes frequently exist in the dataset, e.g., Pole, Traffic Light, or Sign. Still, these classes belong to a minority group in the \textit{fairness} problem as their appearance on images does not occupy too many pixels. 
    Our FREDOM has promoted the fairness of models illustrated by an increase of mIoU on the minority group.}
    \label{fig:fredom-fair_long}
\end{wrapfigure}
The unfair predictions of segmentation models can lead to severe problems, e.g., in autonomous driving, unfair predictions may result in wrong decisions in motion planning control and, therefore, affect human safety. Moreover, the fairness issue of segmentation models is even well observed or exaggerated when the trained models are deployed into new domains. Many prior works alleviate the performance drop on new domains by using unsupervised domain adaptation, but these approaches do not guarantee the fairness property.

There needs to be more attention on addressing the fairness issue in semantic segmentation under the supervised or domain adaptation settings.
Besides, the definition of fairness in semantic segmentation needs to be better defined and, therefore, often needs clarification with the long-tail issue in segmentation. In particular, the \textbf{\textit{long-tail problem}} in segmentation is typically caused by \textbf{\textit{the number of existing instances}} of each class in the dataset~\cite{wang2021seesaw, DBLP:conf/aaai/Ting21}. Meanwhile, the \textbf{\textit{fairness problem}} in segmentation is considered for \textbf{\textit{the number of pixels}} of each class in the dataset.
Although there could be a correlation between fairness and long-tail problems, these two issues are distinct. For example, several objects constantly exist in the dataset, but their presence often occupies only tiny regions of the given image (containing a small number of pixels), 
e.g., the Pole, which is a head class in Cityscapes, accounts for over  $20\%$ of instances while the number of pixels does only less than $0.01\%$ of pixels. 
Hence, upon the fairness definition, it should belong to the minor group of classes as its presence does not occupy many pixels in the image. 
Another example is Person, which accounts for over $5\%$ of instances, while the number of pixels does only less than $0.01\%$ of pixels. 
Traffic Lights or Signs also suffer a similar problem.
Figure~\ref{fig:fredom-major_minor_fig} illustrates the appearance of classes in the majority and minority groups.
Therefore, although instances of these classes constantly exist in the dataset, these are still being mistreated by the segmentation model. Figure~\ref{fig:fredom-fair_long} illustrates the class distributions defined based on long-tail and fairness, respectively.
In the \textit{scope of our work}, we are interested in addressing the fairness problem in semantic segmentation between classes under the unsupervised domain adaptation setting. It should be noted that our interested problem is practical. In real-world applications (e.g., autonomous driving), deep learning models are typically deployed into new domains compared to the training dataset. Then, unsupervised domain adaptation plays a role in bridging the gap between the two domains.

To address the fairness problem in domain adaptation, this chapter presents a novel Unsupervised \textbf{F}ai\textbf{r}n\textbf{e}ss \textbf{Dom}ain Adaptation (FREDOM) approach to semantic segmentation. To the best of our knowledge, this is one of the first works to address the fairness problem in semantic segmentation under the domain adaptation setting. Our contributions can be summarized as follows. First, the new fairness objective is formulated for semantic scene segmentation. Then, based on the fairness metric, we propose a novel fairness domain adaptation approach based on the fair treatment of class distributions. Second, the novel Conditional Structural Constraint is proposed to model the structural consistency of segmentation maps. 
Thanks to our introduced Conditional Structure Network, the spatial relationship and structure information are well modeled by the self-attention mechanism. Significantly, our structural constraint relaxes the assumption of pixel independence held by prior approaches and generalizes the Markovian assumption by considering the structural correlations between all pixels.

\subsection{The Proposed Fairness Domain Adaptation Approach}

\subsubsection{The Fairness Objective Function}

Under the fairness constraint in semantic segmentation, the performance of each class should be equally treated by the deep model. Thus, the goal of fairness in semantic segmentation can be defined as in Eqn.~\eqref{eqn:fredom-fairness}.
\begin{equation} \label{eqn:fredom-fairness}
\small
\begin{split}
    \arg\min_{\theta} \sum_{c_i, c_j}\Big|\mathbb{E}_{\mathbf{x} \in \mathcal{X}}\sum_{k}\mathcal{L}(y^k = c_i) - \mathbb{E}_{\mathbf{x} \in \mathcal{X}}\sum_{k}\mathcal{L}(y^k = c_j)\Big|
\end{split}
\end{equation}
where $y^k$ denotes the $k^{th}$ pixel of the segmentation $\mathbf{y}$, 
$c_i$ and $c_j$ are the class categories, i.e, $c_i, c_j \in [1..C]$ (where $C$ is the number of classes),
$\mathcal{L}$ is the loss function measuring the error rates of predictions. 
Formally, for all pairs of classes in the dataset,  Eqn.~\eqref{eqn:fredom-fairness} aims to minimize the difference in the error rates produced by the model between classes. Therefore, it guarantees all classes in the dataset are treated equally. Eqn.~\eqref{eqn:fredom-fairness} can be further derived as in Eqn.~\eqref{eqn:fredom-fairness_less_adapt}.
\begin{equation} \label{eqn:fredom-fairness_less_adapt}
\scriptsize
\begin{split}
&\sum_{c_i, c_j}\Big|\mathbb{E}_{\mathbf{x} \in \mathcal{X}}\sum_{k}\mathcal{L}(y^k_s = c_i) - \mathbb{E}_{\mathbf{x} \in \mathcal{X}}\sum_{k}\mathcal{L}(y^k_s = c_j)\Big|
\leq 2C\Big[\mathbb{E}_{\mathbf{x}_s, \mathbf{\hat{y}_s} \sim p_s(\mathbf{y}_s, \mathbf{\hat{y}}_s}) \mathcal{L}_s(\mathbf{y}_s, \mathbf{\hat{y}}_s)  + \mathbb{E}_{\mathbf{x}_t \sim p_t(\mathbf{x}_t)} \mathcal{L}_t(\mathbf{y}_t)\Big] 
\end{split}
\end{equation}
From Eqn.~\eqref{eqn:fredom-fairness_less_adapt}, we can observe that the fairness objective in Eqn.~\eqref{eqn:fredom-fairness} is bounded by the standard optimization of domain adaptation in Eqn.~\eqref{eqn:bimal-objective} (presented in Section \ref{sec:bimal-paper}).
Although optimizing the standard domain adaptation could impose the constraint of fairness under the upper bound in Eqn.~\eqref{eqn:fredom-fairness_less_adapt}, the imbalance class distributions of pixels cause the model to behave unfairly between classes when optimizing Eqn.~\eqref{eqn:bimal-objective}. In particular,
Eqn.~\eqref{eqn:bimal-objective} can be rewritten as in Eqn.~\eqref{eqn:fredom-original_optimization_rewrite}.
\begin{equation} \label{eqn:fredom-original_optimization_rewrite}
\scriptsize
    \begin{split}
    &\arg\min_{\theta} \Bigg[\int \mathcal{L}_s(\mathbf{y}_s, \mathbf{\hat{y}}_s)p_s(\mathbf{y}_s) p_s(\mathbf{\hat{y}}_s)d\mathbf{y}_s d\mathbf{\hat{y}}_s 
    + \int \mathcal{L}_t(\mathbf{y}_t)p_t(\mathbf{y}_t)d\mathbf{y}_t\Bigg]\\
    &=\arg\min_{\theta} \Bigg[\int \sum_{k=1}^N\mathcal{L}_s(y^k_s, \hat{y}^k_s) p_s(y^k_s)p_s(\mathbf{y}^{\setminus k}_s | y^k_s)p_s(\mathbf{\hat{y}}_s)d\mathbf{y}_s d\mathbf{\hat{y}}_s  + \int \sum_{k=1}^N\mathcal{L}_t(y^k_t) p_t(y^k_t)p_t(\mathbf{y}^{\setminus k}_t | y^k_t)d\mathbf{y}_t\Bigg]
    \end{split}
\end{equation}
where $N$ is the total number of pixels in the image, 
$y^k_s$ and  $y^k_t$ are the $k^{th}$ pixel of predicted segmentations in source and target domains,
$\mathbf{y}^{\setminus k}_s$ and $\mathbf{y}^{\setminus k}_t$ are predicted segmentations without the $k^{th}$ pixel in source and target domains, $p_s(y^k)$ and $p_t(y^k)$ are the class distributions of pixels in the source and target domains.
The class distributions are computed based on the number of pixels of each class in the dataset.
The terms $p_s(\mathbf{y}^{\setminus k}_s | y^k_s)$ and $p_t(\mathbf{y}^{\setminus k}_t | y^k_t)$ are conditional structure constraints of $\mathbf{y}_s^{\setminus k}$ and $\mathbf{y}_t^{\setminus k}$ on $y^k_s$ and $y^k_t$.

\noindent
\textbf{From imbalance distributions to unfair predictions:}
In practice, the class distributions of pixels $p_s(y_s^k)$ and $p_t(y_t^k)$ suffer imbalance problems as shown in Figure~\ref{fig:fredom-fair_long}.
When the model is learned by the gradient descent method, the model behaves inequitably between classes. In particular, let us consider the behavior of gradients produced by the gradient descent learning method. Formally, let $c_i$ and $c_j$ be the two classes and $p_s(y_s^k=c_i) << p_s(y^k_s=c_j)$. The gradients produced for each class with respect to the predictions can be formed as in Eqn.~\eqref{eqn:fredom-inequality}.
\begin{equation} \label{eqn:fredom-inequality}
\tiny
\begin{split}
    &\left|\left|\frac{\partial \int \sum_{k=1}^{N}\mathcal{L}_s(y^k_s, \hat{y}^k_s) p_s(y^k_s=c_i)p_s(\mathbf{y}^{\setminus k}_s | y^k_s )p_s(\mathbf{\hat{y}}_s)d\mathbf{y}_s d\mathbf{\hat{y}}_s}{\partial \mathbf{y}^{(c_i)}_{s}}\right|\right| \ll  \left|\left|\frac{\partial \int \sum_{k=1}^{N}\mathcal{L}_s(y^k_s, \hat{y}^k_s) p_s(y^k_s=c_j)q_s(\mathbf{y}^{\setminus k}_s | y^k_s )p_s(\mathbf{\hat{y}}_k)d\mathbf{y}_s d\mathbf{\hat{y}}_s}{\partial \mathbf{y}^{(c_j)}_{s}}\right|\right|
\end{split}
\end{equation}
where $||.||$ is the magnitude of the vector,
$\mathbf{y}^{(c_i)}_s$ and $\mathbf{y}^{(c_j)}_s$ represent the predicted probabilities of label $c_i$ and $c_j$, respectively.
As shown in Eqn.~\eqref{eqn:fredom-inequality}, 
the model inclines to produce significant gradient updates of the classes having a large population in the distributions (\textit{a majority group}); meanwhile, the gradient updates of the class having a small population in the distributions (\textit{a minority group}) are minor and dominated by the gradients of majority groups. 
Similar behavior can also be observed in the target domain.

\subsubsection{The Proposed Fairness Adaptation Approach}

\begin{figure}[!b]
    \centering
    \includegraphics[width=1.0\textwidth]{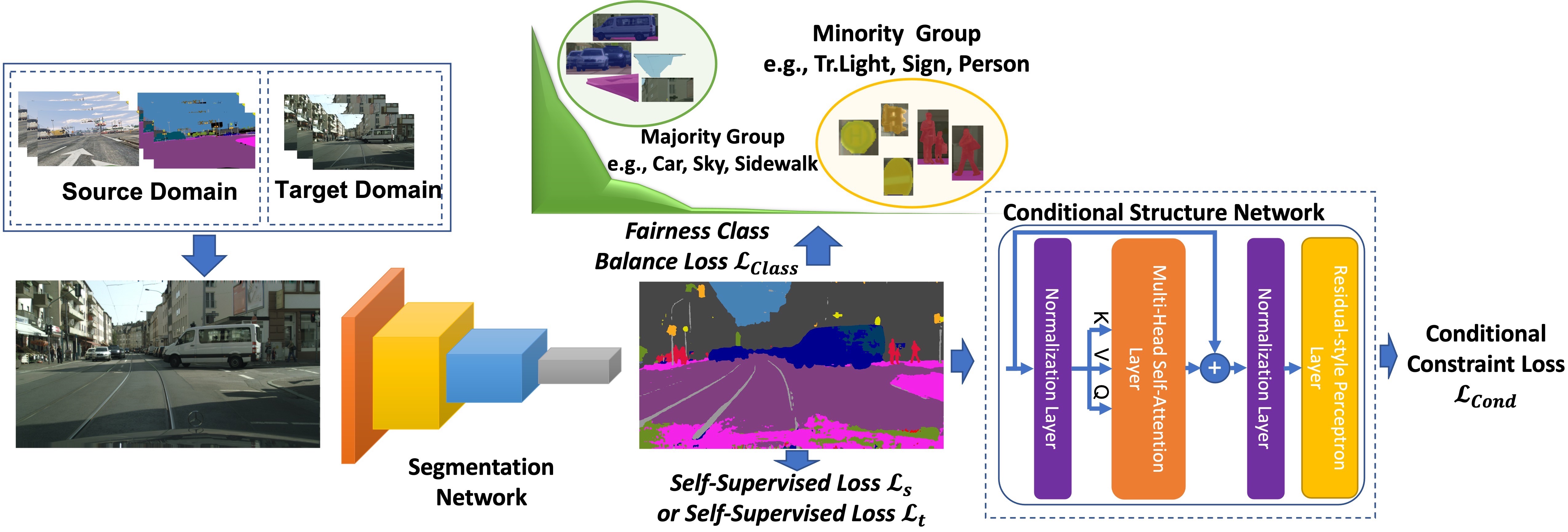}
    \caption{\textbf{The Proposed Fairness Framework.}
    The predictions of the inputs sampled from the source or target domains are penalized by the supervised loss $\mathcal{L}_s$ or the self-supervised loss $\mathcal{L}_t$, respectively. Then, the predictions are imposed by the fairness class balance loss $\mathcal{L}_{Class}$ followed by the Conditional Constraint Loss $\mathcal{L}_{Cond}$ computed via a Conditional Structure Network.
    }
    \label{fig:fredom-fair_da_framework}
\end{figure}

As discussed in the previous section, the fairness problem 
is typically caused by imbalanced class distributions. 
Therefore, to address the fairness problem, we first assume that there exists an ideal distribution $p'_s(\mathbf{y}_s)$ and $p'_t(\mathbf{y}_t)$ so that the model trained on the ideal data distributions behave fairly between classes. 
It should be noted that we assume the ideal data distribution to frame and navigate our proposed approach to the fairness domain adaptation in semantic segmentation.
Then, the ideal data distributions will be relaxed later and there is no requirement for the ideal data distribution during the training process.
Formally, learning the adaptation framework under the ideal data distribution can be formulated as in Eqn.~\eqref{eqn:fredom-opt_from_ideal}.
\begin{equation} \label{eqn:fredom-opt_from_ideal}
\small
\begin{split}
      &\arg\min_{\theta} \Bigg[\mathbb{E}_{\mathbf{x}_s \sim p_s(\mathbf{y}_s), \mathbf{\hat{y}_s} \sim p_s(\mathbf{\hat{y}}_s)} \mathcal{L}_s(\mathbf{y}_s, \mathbf{\hat{y}}_s)\frac{p'_s(\mathbf{y}_s) p'_s(\mathbf{\hat{y}}_s)}{p_s(\mathbf{y}_s)p_s( \mathbf{\hat{y}}_s)}  + \mathbb{E}_{\mathbf{x}_t \sim p_t(\mathbf{x}_t)} \mathcal{L}_t(\mathbf{y}_t)\frac{p'_t(\mathbf{y}_t)}{p_t(\mathbf{y}_t)}\Bigg] 
\end{split}
\end{equation}
The fraction between ideal and real data distributions, i.e. $\frac{p'_s(\mathbf{y}_s) p'_s(\mathbf{\hat{y}}_s)}{p_s(\mathbf{y}_s)p_s( \mathbf{\hat{y}}_s)}$ and $\frac{p'_t(\mathbf{y}_t)}{p_t(\mathbf{y}_t)}$, can be interpreted as the complement of the model needed to be improved to achieve fairness against the imbalanced data.
It should be noted that $p'_s(\mathbf{\hat{y}}_s)$ and $p_s( \mathbf{\hat{y}}_s)$ are constants as they are distributed over segmentation labels, 
so these could be excluded during training. %
Then, Eqn.~\eqref{eqn:fredom-opt_from_ideal} can be further derived as in Eqn.~\eqref{eqn:fredom-opt_from_ideal_sum}.
\begin{equation} \label{eqn:fredom-opt_from_ideal_sum}
\scriptsize
\begin{split}
    & \arg\min_{\theta} \Bigg[\mathbb{E}_{\mathbf{x}_s \sim p_s(\mathbf{y}_s), \mathbf{\hat{y}_s} \sim p_s(\mathbf{\hat{y}}_s)} \sum_{k=1}^{N} \mathcal{L}_s(y^k_s, \hat{y}^k_s) \frac{p'_s(y^k_s)p'_s(\mathbf{y}^{\setminus k}_s|y^k_s)}{p_s(y^k_s)p_s(\mathbf{y}^{\setminus k}_s |y^k_s)}   + \mathbb{E}_{\mathbf{x}_t \sim p_t(\mathbf{x}_t)} \sum_{k=1}^{N}\mathcal{L}_t(y^k_t) \frac{p'_t(y^k_t)p'_t(\mathbf{y}^{\setminus k}_t|y^k_t)}{p_t(y^k_t)p_t(\mathbf{y}^{\setminus k}_t |y^k_t)} \Bigg]
\end{split}
\end{equation}
As shown in Eqn.~\eqref{eqn:fredom-opt_from_ideal_sum}, if the conditional structure fractions $\frac{p'_s(\mathbf{y}^{\setminus k}_s|y^k_s)}{p_s(\mathbf{y}^{\setminus k}_s |y^k_s)}$ and $\frac{p'_t(\mathbf{y}^{\setminus k}_t|y^k_t)}{p_t(\mathbf{y}^{\setminus k}_t |y^k_t)}$ are ignored, Eqn.~\eqref{eqn:fredom-opt_from_ideal_sum} becomes a special case of the weighted class balanced loss~\cite{Cui_2019_CVPR, wang2021seesaw}. %
However, conditional structure plays a vital role in semantic segmentation as it provides the constraints and correlation of structures among objects in images. The ignorance of conditional structure fractions could lower the
performance of segmentation models.
In addition, although the input images of the source and target domains can vary significantly in appearance due to the distribution shift, their segmentation maps between two domains share similar class distributions and structural information~\cite{tsai2018learning, tsai2019domain, dat2021bimal_iccv}. Hence, the distribution of segmentation in the target domain $p_t(\cdot)$ can be practically approximated by distribution in the source domain, i.e., $\frac{p'_t(\mathbf{y}_t)}{p_t(\mathbf{y}_t)} = \frac{p'_s(\mathbf{y}_t)}{p_s(\mathbf{y}_t)}$.
In summary, by taking the log of Eqn.~\eqref{eqn:fredom-opt_from_ideal_sum}, the learning process can be formed as in Eqn.~\eqref{eqn:fredom-take_log}
(the proof of Eqn.~\eqref{eqn:fredom-take_log} can be found in our preliminary work~\cite{Truong:CVPR:2023FREDOM}).
\begin{equation} \label{eqn:fredom-take_log}
\small
\begin{split}
    &\theta^* \simeq \arg\min_{\theta} \Bigg[\mathbb{E}_{\mathbf{x}_s \sim p_s(\mathbf{x}_s), \mathbf{\hat{y}_s} \sim p_s(\mathbf{\hat{y}}_s)} \mathcal{L}_s(\mathbf{y}_s, \mathbf{\hat{y}}_s) + \mathbb{E}_{\mathbf{x}_t \sim p_t(\mathbf{x}_t)} \mathcal{L}_t(\mathbf{y}_t)\\
    &\quad\quad+\frac{1}{N}\sum_{k=1}^{N}
    \Bigg(\mathbb{E}_{\mathbf{x}_s \sim p_s(\mathbf{x}_s)} \log\left(\frac{p'_s(y^k_s)}{p_s(y^k_s)}\right) 
    +\mathbb{E}_{\mathbf{x}_t \sim p_t(\mathbf{x}_t)} \log\left(\frac{p'_s(y^k_t)}{p_s(y^k_t)}\right)\\
    &\quad\quad+\mathbb{E}_{\mathbf{x}_s \sim p_s(\mathbf{x}_s)} \log\left(\frac{p'_s(\mathbf{y}^{\setminus k}_s |y^k_s)}{p_s(\mathbf{y}^{\setminus k}_s |y^k_s)}\right)
    +\mathbb{E}_{\mathbf{x}_t \sim p_t(\mathbf{x}_t)} \log\left(\frac{p'_s(\mathbf{y}^{\setminus k}_t|y^k_t)}{p_s(\mathbf{y}^{\setminus k}_t |y^k_t)}\right)\Bigg)\Bigg]
\end{split}
\end{equation}
In summary, there are three terms in the learning objective of our FREDOM approach. 
Hence, several properties are brought into the learning process that can be observed.

\noindent
\textbf{Domain Adaptation Objective.} The first two terms stand for the objective of domain adaptation. While $\mathcal{L}_s$ learns to a segment on the source domain in the supervised fashion, $\mathcal{L}_t$ aims to unsupervised adapt knowledge to the target domain.

\noindent
\textbf{Fairness Treatment from Class Distributions.} The next two terms, i.e, $\log\left(\frac{p'_s(y^k_t)}{p_s(y^k_t)}\right)$ and $\log\left(\frac{p'_s(y^k_t)}{p_s(y^k_t)}\right)$, denoted as the $\mathcal{L}_{Class}$, impose the behavior of the model with respect to the class distribution. In particular, these constraints aim to regularize the predictions of classes so that the model should behave fairly between classes with respect to the class distribution. Under the ideal data distribution assumption, the model is expected to equally treat predictions of all classes. Thus, to achieve the desired goal, the distributions of pixel classes should be uniformly distributed. Therefore, we adopt the uniform distribution of the class distribution $p'_s(y^k_s)$, i.e., $p'_s(y^k_s) = \frac{1}{C}$ where C is the number of classes.

\noindent
\textbf{Conditional Structure Constraint.}
The last two terms, i.e., $\log\left(\frac{p'_s(\mathbf{y}^{\setminus k}_s |y^k_s)}{p_s(\mathbf{y}^{\setminus k}_s |y^k_s)}\right)$ and $\log\left(\frac{p'_s(\mathbf{y}^{\setminus k}_t|y^k_t)}{p_s(\mathbf{y}^{\setminus k}_t |y^k_t)}\right)$, denoted as $\mathcal{L}_{Cond}$, impose the conditional structure of the predicted semantic segmentation.
This condition plays a role as a metric to measure the structural consistency of predicted segmentation maps with respect to the one under the ideal distributions where the model behaves fairly.
Modeling the conditional structure, i.e., $p_s(\mathbf{y}^{\setminus k}_s |y^k_s)$, is a challenging problem. Several prior works 
modeled structural constraints by adopting the Markovian assumption~\cite{Zheng_2015_ICCV, chen2018deeplab}
where the models only 
consider the correlation between %
the current pixel with its neighbor pixels.
However, the smoothness of predicted segmentation maps is highly dependent on the window size used in Markovian approaches (the number of neighbor pixels being selected). 
In our work, to sufficiently capture the conditional structural constraint, instead of modeling only neighborhood dependencies as Markovian approaches, we generalize it by modeling $p_s(\mathbf{y}^{\setminus k}_s |y^k_s)$ via a conditional structure network (detailed in Sec. \ref{sec:Cond_Network}) 
to consider the correlation between all pixels in the segmentation. 

\noindent
\textbf{Relaxation of Ideal Data Distribution.}
One of the key challenging problems in optimizing Eqn.~\eqref{eqn:fredom-take_log} is that the conditional ideal data distributions $p'_s(\mathbf{y}^{\setminus k}_s |y^k_s)$ and $p'_s(\mathbf{y}^{\setminus k}_t|y^k_t)$ are not available.
Therefore, instead of directly optimizing these terms, let us consider the tight bound as in Eqn.~\eqref{eqn:fredom-bound}.
\begin{equation}
\label{eqn:fredom-bound}
\small
\begin{split}
    &\mathbb{E}_{\mathbf{x}_s \sim p_s(\mathbf{x}_s)} \log\left(\frac{p'_s(\mathbf{y}^{\setminus k}_s |y^k_s)}{p_s(\mathbf{y}^{\setminus k}_s |y^k_s)}\right) 
    + \mathbb{E}_{\mathbf{x}_t \sim p_t(\mathbf{x}_t)} \log\left(\frac{p'_s(\mathbf{y}^{\setminus k}_t|y^k_t)}{p_s(\mathbf{y}^{\setminus k}_t |y^k_t)}\right)\\
    &\leq -\Big[\mathbb{E}_{\mathbf{x}_s \sim p_s(\mathbf{x}_s)} \log p_s(\mathbf{y}^{\setminus k}_s |y^k_s) 
    + \mathbb{E}_{\mathbf{x}_t \sim p_t(\mathbf{x}_t)} \log p_s(\mathbf{y}^{\setminus k}_t |y^k_t)\Big]
\end{split}
\end{equation}
With any form of ideal distribution $p'_s(\cdot)$, Eqn.~\eqref{eqn:fredom-bound} always hold due to $\log p'_s(\cdot) \leq 0$. Hence, optimizing Eqn.~\eqref{eqn:fredom-bound} also ensure the conditional structural constraint in Eqn.~\eqref{eqn:fredom-take_log} imposed due to the upper bound of Eqn.~\eqref{eqn:fredom-bound}. Therefore, the demand for ideal data distribution is relaxed.
Figure~\ref{fig:fredom-fair_da_framework} illustrates our proposed fairness domain adaptation framework.

\subsection{The Conditional Structure Network} \label{sec:Cond_Network}

The conditional structural constraint $p_s(\mathbf{y}^{\setminus k}_s |y^k_s)$ can be learned on the source dataset due to the availability of the ground-truth segmentation in the source domain.
Formally, let $p_s(\mathbf{y}^{\setminus k}_s |y^k_s)$ be modeled by the conditional structure network $G$ with parameters $\Theta$. Then the conditional structure network can be auto-regressively formed as in Eqn.~\eqref{eqn:fredom-auto_erg_rnn}.
\begin{equation} \label{eqn:fredom-auto_erg_rnn}
\small
\begin{split}
    &\arg\min_{\Theta} \mathbb{E}_{\mathbf{y}_s \in \mathcal{Y}_s} -\log p_s(\mathbf{y}^{\setminus k}_s |y^k_s, \Theta) = \arg\min_{\Theta} \mathbb{E}_{\mathbf{y}_s \in \mathcal{Y}_s} \sum_{i=1}^{N-1} -\log p_s(y^{\sigma^k_i} | y^{\sigma^k_{i-1}}, ..., y^{\sigma^k_1}, y^k_s, \Theta)
\end{split}
\end{equation}
where $\sigma^k$ is the permutation of $\{1...N\} \setminus \{k\}$. Eqn.~\eqref{eqn:fredom-auto_erg_rnn} could be modeled by Recurrent Neural Networks~\cite{oord2016pixel}. However, directly adopting recurrent approaches remains some potential limitations.
Particularly, as the recurrent approaches use a pre-defined permutation of regressive orders, it requires different conditional structure models for different initial pixel conditions, e.g., $p_s(\mathbf{y}^{\setminus k_1}_s |y^{k_1}_s)$ and $p_s(\mathbf{y}^{\setminus k_2}_s |y^{k_2}_s)$ should be modeled two different models.
This problem could be alleviated by considering the permutation of regressive order
as an network's input. %
However, learning a single network to model conditional structural constraints of different permutations is a heavy task and ineffective.

Instead of regressively forming $p_s(\mathbf{y}^{\setminus k}_s |y^k_s)$, we propose to model $p_s(\mathbf{y}^{\setminus k}_s |y^k_s)$ in the parallel fashion. 
Particularly, let $\mathbf{m}$ be the binary masked matrix of $\mathbf{y}_s$, where
the values of one and zero indicate a given pixel (unmasked pixel) and an unknown pixel (masked pixel), respectively. 
Then, the conditional structure $p_s(\mathbf{y}^{\setminus k}_s |y^k_s)$ can be rewritten as $p_s(\mathbf{y}_s \odot (\mathbf{1} - \mathbf{m}) | \mathbf{y}_s \odot \mathbf{m})$, 
where $\odot$ is the element-wise product and the mask $\mathbf{m}$ contains only one unmasked pixel, i.e., the given $k^{th}$ pixel ($m^k = 1$).
Learning the conditional structure constraint via binary mask $\mathbf{m}$ can be formed as in Eqn.~\eqref{eqn:fredom-bert_gpt}.
\begin{equation} \label{eqn:fredom-bert_gpt}
\begin{split}
    \arg\min_{\Theta} \mathbb{E}_{\mathbf{y}_s \in \mathcal{Y}_s, \mathbf{m} \in \mathcal{M}} -\log p_s(\mathbf{y}_s \odot (\mathbf{1} - \mathbf{m}) | \mathbf{y}_s \odot \mathbf{m})
\end{split}
\end{equation}
where $\mathcal{M}$ is the set of possible binary masks. 
Through Eqn.~\eqref{eqn:fredom-bert_gpt}, modeling the conditional structural constraint $p_s(\mathbf{y}^{\setminus k}_s |y^k_s)$
can be equivalently interpreted as learning the condition of \textit{masked pixels} on the given \textit{unmask pixel}.
To increase the modeling capability of the conditional structure network, three different strategies of the binary mask are adopted during training.
First, the binary mask only contains one unmasked pixel to model the condition structural constraint $p_s(\mathbf{y}^{\setminus k}_s |y^k_s)$.
Second, the binary mask does not contain any unmasked pixels (a zero mask). In this case, the model is going to learn the likelihood of the segmentation map $p_s(\mathbf{y}_s)$.
Third, the binary mask contains more than one unmasked pixel that aims to increase the generalizability of the conditional structure network in modeling segmentation structures conditioned on the unmasked pixels.

To model conditional structure network $G$ in a parallel fashion, the network $G$ is designed as a Transformer.
In particular, considering each pixel as a token, the network $G$ is formed as the Transformer %
with $L$ self-attention blocks where each block is designed in a residual style and the layer norms are applied to both the multi-head self-attention and multi-perceptron layers.
By this design, the spatial relationship and structural dependencies 
can be modeled by the self-attention mechanism.
To effectively optimize the network $G$, we adopt the learning tactic of Image-GPT~\cite{chen2020generative}.

\subsection{Experimental Results}

In this section, we present our experimental results on two standard benchmarks, i.e., SYNTHIA $\to$ Cityscapes and GTA5 $\to$ Cityscapes. First, we review datasets and our implementation, followed by analyzing the effectiveness of our approach to fairness improvement in ablation studies. Finally, we compare our experimental results with prior SOTA domain adaptation approaches. The performance of segmentation models is evaluated using the mean Intersection over Union (mIoU) and the IoU's standard deviation.

\noindent \textbf{Implementation.} Our proposed FREDOM approach is implemented based on the implementation of SAC~\cite{araslanov2021dasac}, DAFormer~\cite{hoyer2022daformer}, and ImageGPT~\cite{chen2020generative}.
Two different segmentation architectures are used in our experiments, i.e., (1) DeepLab-V2~\cite{chen2018deeplab} with the Resnet-101 backbone and (2) Transformer with the MiT-B3 backbone~\cite{xie2021segformer}. The Transformer design of~\cite{chen2020generative} has been adapted to our conditional network structure $G$. The Conditional Structure Network $G$ is trained on the segmentation labels of the source domain.
Our framework is implemented in PyTorch and trained on four 48GB-VRAM NVIDIA Quadro P8000 GPUs. The model is optimized by the SGD optimizer with learning rate $2.5 \times 10^{-4}$, momentum $0.9$, weight decay $10^{-4}$, and batch size of $4$ per GPU. The image size is set to $1280 \times 720$ pixels. In the proposed FREDOM framework, the learning strategies and sampling techniques of~\cite{hoyer2022daformer, araslanov2021dasac} are adopted for the self-supervised loss $\mathcal{L}_t$ to train our model. 
Our training procedure and augmentation methods are implemented based on the implementation of SAC~\cite{araslanov2021dasac} and DAFormer~\cite{hoyer2022daformer}. 
We also adopt the data sampling technique of SAC~\cite{araslanov2021dasac} in our training. 
For the mask sampling technique, during training the conditional structure network, for each image in each iteration, we randomly generate a binary mask within three cases, as aforementioned.

\subsubsection{Ablation Study}

Our ablation studies evaluate DeepLab-V2 models on two benchmarks under two settings, i.e., With and Without Adaptation.
Each setting has three configs, i.e., (A) Model without $\mathcal{L}_{Class}$ and $\mathcal{L}_{Cond}$, (B) \textbf{\textit{Fairness model}} with only $\mathcal{L}_{Class}$, and (C) \textbf{\textit{Fairness model}} with $\mathcal{L}_{Class}$ and $\mathcal{L}_{Cond}$. 

\noindent \textbf{Does Adaptation Improve the Fairness?} 
We evaluate the impact of Domain Adaptation in improving the fairness of classes in the minor group. As shown in Table~\ref{tab:fredom-ablation}, domain adaptation significantly improves fairness. In particular, without adaptation, the segmentation models trained only on the source data remain low performance in classes in the minor group, i.e., Traffic Light, Sign, and Fence. However, with our fairness domain adaptation approach, the overall accuracy and individual IoU of classes in the minor group are significantly boosted. In particular, the mIoU accuracy of segmentation models has been improved by $+22.4\%$ and $+21.6\%$ on SYNTHIA $\to$ Cityscapes and GTA5 $\to$ Cityscapes benchmarks. The model's fairness has been improved. Meanwhile, the IoU's STD of classes has been reduced by $1.4\%$ and $4.5\%$ on two benchmarks, respectively.

\input{Tables/chap-3/fredom/fredom-ablation}

\noindent \textbf{Does Class Distributions Matter to Fairness Improvement?}
As shown in Table~\ref{tab:fredom-ablation}, the fairness treatment from the class distribution loss $\mathcal{L}_{Class}$ contributes a significant improvement to both the overall performance and accuracy of classes in the minority group. In particular,  the IoU accuracy of each class in configuration (B) is improved compared to the one in configuration (A) in both with and without adaptation settings. Specifically, in the adaptation setting on benchmark SYNTHIA $\to$ Cityscapes, the class distribution loss $\mathcal{L}_{Class}$ has boosted the performance of classes in the minority group, e.g., Traffic Light (from $48.9\%$ to $51.9\%$),  Sign (from $40.5$ to $43.0\%$), Pole (from $45.8\%$ to $48.6\%$).
Without adaptation, improvement is also observed. Moreover, the standard deviation of IoU over classes has been reduced. It shows that the model's fairness has been promoted. Similarly, our performance on benchmark GTA5 $\to$ Cityscapes is also consistently improved. 

\begin{wrapfigure}{r}{0.5\textwidth}
    \centering
    \includegraphics[width=0.5\textwidth]{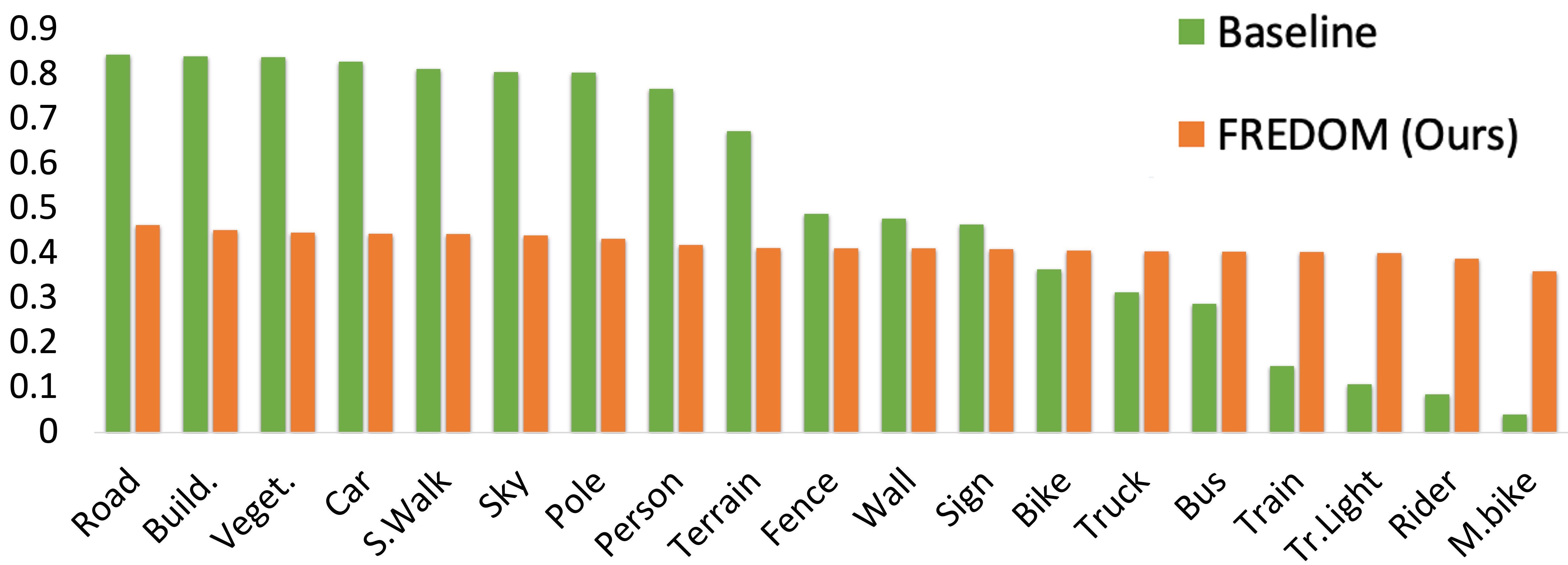}
    \caption{\textbf{The Mean Magnitude of Normalized Gradients Updated for Each Class.} Configuration (A) is the baseline.}
    \label{fig:fredom-grad_per_class}
\end{wrapfigure}
\noindent
\textbf{Does the Conditional Structure Constraint Contribute to Fairness Improvement?}
Configuration (C) in Table~\ref{tab:fredom-ablation} reports experimental results of our model using conditional structure constraint loss $\mathcal{L}_{Cond}$. Results in Table~\ref{tab:fredom-ablation} have shown the de facto role of the conditional structure constraint in performance improvement. Indeed, it enhances the IoU accuracy of each class in the minority group. For example, the average IoU accuracy of Fences, Pole, Traffic Light, and Sign has been improved by $2.3\%$. Overall, the performance of segmentation models has been improved by a notable margin, i.e., $+2.1\%$ and $2.7\%$ on SYNTHIA $\to$ Cityscapes and GTA5 $\to$ Cityscapes, respectively. The difference in performance between classes is reduced, illustrated by the decrease of the IoU's standard deviation, which means the model's fairness is improved notably.

\noindent \textbf{Does the Network Design Improve the Fairness?} Table~\ref{tab:fredom-sota} illustrates the results of our approach using DeepLab-V2 and Transformer networks. As in our results, the performance of segmentation models using a more powerful backbone, i.e., Transformer, outperforms the models using DeepLab-V2. The performance of classes in the minority group has been improved notably, e.g., the performance of classes Fence, Traffic Light,  Sign, and Pole has been improved to $9.3\%$, $65.1\%$, $60.1\%$, and $57.3\%$ on the SYNTHIA $\to$ Cityscapes benchmark. The major improvements in the performance of overall and individual classes are also perceived in the GTA5 $\to$ Cityscapes benchmark. Also, the standard deviation of IoU over classes has been majorly reduced by $3.3\%$, illustrating that fairness has been promoted.

\noindent \textbf{Does the Model Fairly Treat all Class During Training?} Figure~\ref{fig:fredom-grad_per_class} visualizes the gradients produced w.r.t each class in the domain adaptation setting. In particular, we take a subset in Cityscapes and compute the normalized gradients updated for each class. The model with our proposed approach tends to update gradients for each class fairly. Meanwhile, without using our fairness method, the gradients of classes in the minority group are dominated by the ones in the majority group, which could result in models' unfair behaviors.

\noindent
\textbf{Does Mask Sampling Approach Improve Fairness?}
To further illustrate the effectiveness of mask sampling, we conduct additional ablation studies using DeepLabV2 (DL-V2) under the domain adaptation setting trained on the GTA5 $\to$ Cityscapes benchmark.
We consider experiments of the pre-defined weight-balancing different classes [13] ($\mathcal{L}_{CB}$), normalizing gradients ($\mathcal{L}_{NG}$).
In addition, we evaluate the impact of the mask sampling approaches on fairness improvement. There are three different strategies of binary mask samplings that will be evaluated, i.e., (1) If $\mathbf{ m}$ contains only one unmasked pixel (denoted as \textbf{M1}), $G$ learns to capture structural information of segmentation conditioned on a given pixel, 
(2) If $\mathbf{m}$ contains more than one unmasked pixel (denoted as \textbf{M2}), it increases the flexibility of $G$ on learning segmentation structures conditioned on unmasked pixels,
(3) If $\mathbf{m}$ does not contain any unmasked pixels (denoted as \textbf{M3}), it is equivalent to learning the log-likelihood of segmentation maps.
The experimental results in Table~\ref{tab:fredom-mask_ab} show the advantages of our method. 
We found that $\mathcal{L}_{NG}$ stabilizes the training procedure and $\mathcal{L}_{CB}$ brings a minor improvement. 
Also, while $\mathcal{L}_{Cond}$ with simple binary masks sampled as M1 is not powerful enough to model the conditional structures, 
combining three strategies of mask samplings brings a significant performance improvement of the segmentation model, especially in classes of the minority group, and promotes fairness in the model.

\input{Tables/chap-3/fredom/fredom-sampling-ablation}

\subsubsection{Comparison with SOTA Approaches}

\input{Tables/chap-3/fredom/fredom-main-results}

\noindent
\textbf{SYNTHIA $\to$ Cityscapes.}
Table~\ref{tab:fredom-sota} presents our experimental results using  DeepLab-V2 and Transformer compared to prior SOTA approaches. Our proposed approach achieves SOTA performance and outperforms prior methods using the same network backbone. 
Specifically, the mIoU accuracy of our approach using Transformer  is $67.0\%$ and higher than DAFormer~\cite{hoyer2022daformer} by $+6.1\%$.
Although the results of several individual classes are slightly lower than prior methods, overall, the mIoU accuracy and performance of individual classes in the minor group have been significantly promoted.
Analyzing the mIoU accuracy of classes in the minor group, 
our results have been significantly improved compared to the prior SOTA method (i.e., DAFormer~\cite{hoyer2022daformer}). In particular, the performance of Rider, Fence, Pole, Traffic Light, and Sign classes has been improved by $4.1\%$, $+2.8\%$, $+7.3\%$, $+10.1\%$, and $+5.5\%$, respectively. In addition, the IoU accuracy of classes in the major group is also slightly enhanced. For example, the IoU accuracy of Building, Car, Sidewalk, and Sky has been improved to $87.8\%$, $89.7\%$, $54.1\%$, and $89.5\%$, respectively. 
It is vital to highlight that, to enhance the performance of classes in the minority group, the model does not sacrifice its ability to identify classes in the majority group. Instead, to promote the model's fairness, our approach enhances its ability to segment classes in the minor group to reduce the difference in performance between classes in minor and major groups.

\noindent \textbf{GTA5 $\to$ Cityscapes.}
As shown in Table~\ref{tab:fredom-sota}, on the same network backbone, our FREDOM approach performs better than previous SOTA methods. In particular, our approach using Transformer achieves the mIoU accuracy of $73.6\%$, which is the SOTA result; meanwhile, the result of the prior method~\cite{hoyer2022daformer} is $68.3\%$. Noticeably, the performance results have been significantly enhanced in the classes of the minority group, e.g., in comparison with DAFormer~\cite{hoyer2022daformer}, the IoU accuracy of Rider, Motorbike, Pole, Traffic Light, and Sign has been increased by $+8.4$, $+6.9\%$, $7.9\%$, $+7.6\%$, and $+12.0\%$. The performance accuracy has also improved in the majority group classes. For example, the accuracy of Building, Car, Sidewalk, and Sky is brought up to $90.9\%$, $94.1\%$, $74.8\%$, and $94.4\%$. Our FREDOM approach has strengthened the model's ability to segment classes in the minor group to lessen the performance gap between minor and major groups. In addition, the IoU's standard deviation over classes has been decreased compared to prior methods, which means that fairness has been promoted. 

\begin{figure}[!t]
    \centering
    \includegraphics[width=1.0\textwidth]{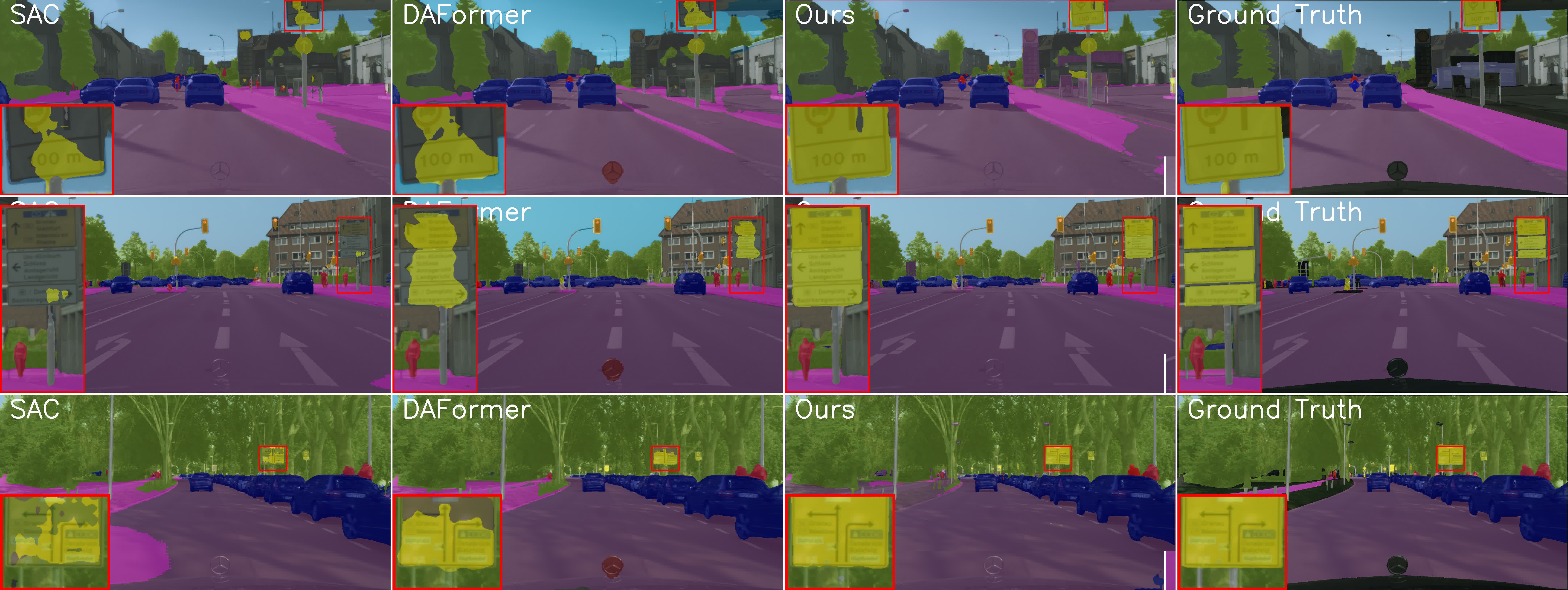}
    \caption{\textbf{Qualitative Results on SYNTHIA $\to$ Cityscapes}
    Columns 1-4 are the results of 
    SAC~\cite{araslanov2021dasac}, DAFormer~\cite{hoyer2022daformer}, FREDOM, and ground truths.
    }
    \label{fig:fredom-syn2city_result}
\end{figure}

\noindent \textbf{Qualitative Results.}
Figure~\ref{fig:fredom-syn2city_result} illustrates our results of the SYNTHIA $\to$ Cityscapces experiment. Our approach produces better quality results than prior UDA methods. Particularly, a significant improvement can be observed from the predictions of classes in the minority group, e.g., the predicted segmentation of signs, persons, and poles is sharper. The model can well segment the classes in the minor group cogently and minimize the region of classes being erroneously classified. The borders between classes are accurately identified and segmentation continuity has improved compared to prior works. Although our predictions contain some noise, the boundaries are still clear and correspond to the labels.

%% file: Tables/chap-3/fredom/fredom-ablation.tex
\begin{table*}[!t]
    \centering
    \caption{Effectiveness of our FREDOM (DeepLab-V2) approach to fairness improvement. Three configurations:
    (A) Model without $\mathcal{L}_{Class}$ and $\mathcal{L}_{Cond}$. 
    (B) \textbf{Fairness Model} with $\mathcal{L}_{Class}$ only. 
    (C) \textbf{Fairness Model} with $\mathcal{L}_{Class}$ and $\mathcal{L}_{Cond}$. 
    }
    \label{tab:fredom-ablation}
    \setlength{\tabcolsep}{2.5pt}
    \resizebox{\textwidth}{!}{%
    \begin{tabular}{|c c | c c c c c c | c c c c c c c c c c c c c | c c|}
\hline

\multicolumn{2}{|c|}{\multirow{2}{*}{Configuration}}  & \multicolumn{6}{c|}{Majority Group} & \multicolumn{13}{c|}{Minority Group} & \multirow{2}{*}{mIoU}          & \multirow{2}{*}{STD}  \\
 \cline{3-21}
   & & Road          & Build.        & Veget.        & Car           & S.Walk        & Sky           & Pole          & Person        & Terrain       & Fence         & Wall          & Sign          & Bike          & Truck         & Bus           & Train         & Tr.Light      & Rider         & M.bike        &           &          \\
\hline
\multicolumn{23}{|c|}{SYNTHIA $\to$ Cityscapes}        \\
\hline

\multirow{3}{*}{\begin{tabular}{@{}c@{}} Without \\ Adaptation \end{tabular} }
    & (A)          & 64.9          & 71.5          & 73.1          & 62.9          & 26.1          & 71.0          & 21.7          & 48.4          & $-$           & 0.2           & 3.0           & 0.2           & 35.6          & $-$           & 27.9          & $-$           & 0.1           & 20.7          & 12.0          & 33.7          & 27.8          \\
    & (B)          & 65.0          & 72.1          & 64.9          & 65.8          & 31.9          & 66.6          & 23.2          & 49.6          & $-$           & 0.2           & 5.0           & 2.5           & 31.7          & $-$           & 26.8          & $-$           & 2.4           & 21.3          & 18.7          & 34.4          & 26.1          \\
    & (C)          & 65.2          & 73.3          & 65.4          & 69.0          & 32.2          & 67.7          & 34.5          & 50.0          & $-$           & 0.3           & 17.5          & 3.5           & 39.9          & $-$           & 27.0          & $-$           & 3.9           & 21.9          & 18.5          & 36.7          & 25.4          \\
\cline{3-23}

\multirow{3}{*}{\begin{tabular}{@{}c@{}} With \\ Adaptation \end{tabular} }
    & (A)          & 84.9          & 85.7          & 86.4          & 86.8          & 44.9          & 88.6          & 45.8          & 69.3          & $-$           & 2.5           & 31.0          & 40.5          & 57.1          & $-$           & 45.9          & $-$           & 48.9          & 31.4          & 47.4          & 56.1          & 25.3          \\
    & (B)          & 84.8          & 85.8          & 86.4          & 86.8          & 45.2          & 88.9          & 47.6          & 70.1          & $-$           & 2.6           & 31.3          & 43.0          & 58.5          & $-$           & 46.0          & $-$           & 51.9          & 34.1          & 49.2          & 57.0          & 24.9          \\
    & \textbf{(C)} & \textbf{86.0} & \textbf{87.0} & \textbf{87.1} & \textbf{87.1} & \textbf{46.3} & \textbf{89.1} & \textbf{48.7} & \textbf{71.2} & \textbf{$-$}  & \textbf{5.3}  & \textbf{33.3} & \textbf{46.8} & \textbf{59.9} & \textbf{$-$}  & \textbf{54.6} & \textbf{$-$}  & \textbf{53.4} & \textbf{38.1} & \textbf{51.3} & \textbf{59.1} & \textbf{24.0} \\                                                    
\hline                              
\multicolumn{23}{|c|}{GTA5 $\to$ Cityscapes}        \\
\hline

 \multirow{3}{*}{\begin{tabular}{@{}c@{}} Without \\ Adaptation \end{tabular} }
 & (A)          & 75.8          & 77.2          & 81.3          & 49.9          & 16.8          & 70.3          & 25.5          & 53.8          & 24.6          & 21.0          & 12.5          & 20.1          & 36.0          & 17.2          & 25.9          & 6.5           & 30.1          & 26.4          & 25.3          & 36.6          & 24.0          \\
 & (B)          & 76.2          & 77.7          & 83.0          & 51.2          & 17.5          & 71.5          & 26.0          & 52.5          & 28.5          & 21.7          & 13.7          & 22.6          & 37.7          & 18.4          & 26.5          & 7.1           & 40.7          & 27.1          & 26.3          & 38.2          & 23.6          \\
 & (C)          & 77.1          & 79.4          & 84.7          & 52.9          & 18.5          & 72.3          & 28.6          & 54.4          & 33.8          & 22.5          & 15.6          & 23.7          & 38.9          & 19.7          & 27.1          & 7.9           & 41.6          & 28.6          & 28.0          & 39.7          & 23.6          \\

\cline{3-23}
\multirow{3}{*}{\begin{tabular}{@{}c@{}} With \\ Adaptation \end{tabular} }
 & (A)          & 90.3          & 87.2          & 88.1          & 88.6          & 53.5          & 87.3          & 44.4          & 67.3          & 42.2          & 28.5          & 41.1          & 50.1          & 54.4          & 52.5          & 56.9          & 33.7          & 48.9          & 33.1          & 42.6          & 57.4          & 20.9          \\
 & (B)          & 90.6          & 87.3          & 88.1          & 88.8          & 53.7          & 87.4          & 44.9          & 67.7          & 42.3          & 28.6          & 41.9          & 52.9          & 57.6          & 55.2          & 57.5          & 47.6          & 50.8          & 36.9          & 44.9          & 59.2          & 19.8          \\
 & \textbf{(C)} & \textbf{90.9} & \textbf{87.8} & \textbf{88.6} & \textbf{89.7} & \textbf{54.1} & \textbf{89.5} & \textbf{45.2} & \textbf{68.8} & \textbf{42.6} & \textbf{32.6} & \textbf{44.1} & \textbf{57.1} & \textbf{58.1} & \textbf{58.4} & \textbf{62.6} & \textbf{55.3} & \textbf{51.4} & \textbf{40.0} & \textbf{47.7} & \textbf{61.3} & \textbf{19.1} \\
 
\hline
\end{tabular}
    }
\end{table*}

%% file: Tables/chap-3/fredom/fredom-sampling-ablation.tex
\begin{table}[!t]
\caption{\textbf{Effectiveness of our Mask Sampling Approach to Fairness Improvement Using DeepLab-V2 (DL-V2).}}
\label{tab:fredom-mask_ab}
\setlength{\tabcolsep}{2.5pt}
\resizebox{\textwidth}{!}{%
\begin{tabular}{|c |c|c c c|c c c c c c|c c c c c c c c c c c c c|c c|}
\hline
                            &  \multirow{2}{*}{$\mathcal{L}_{Class}$}                & \multicolumn{3}{c|}{$\mathcal{L}_{Cond}$}                       & \multicolumn{6}{c|}{Majority Group} &  \multicolumn{13}{c|}{Minority Group} & \multirow{2}{*}{mIoU} & \multirow{2}{*}{STD}  \\ \cline{3-24}
                            &                       & M1                    & M2                    & M3                    & Road     & Build. & Veget. & Car  & S.Walk & Sky  & Pole     & Person & Terrain & Fence & Wall & Sign & Bike & Truck & Bus  & Train & Tr.Light & Rider & M.bike &      &      \\ \hline
              DL-V2 & \xmark & \xmark & \xmark & \xmark & 90.3     & 87.2   & 88.1   & 88.6 & 53.5   & 87.3 & 44.4     & 67.3   & 42.2    & 28.5  & 41.1 & 50.1 & 54.4 & 52.5  & 56.9 & 33.7  & 48.9     & 33.1  & 42.6   & 57.4 & 20.9 \\ %
$+\mathcal{L}_{NG}$ & \xmark & \xmark & \xmark & \xmark & 90.4     & 87.1   & 88.0   & 88.6 & 53.6   & 87.2 & 44.7     & 67.4   & 42.3    & 28.4  & 41.2 & 49.8 & 54.9 & 53.0  & 57.2 & 37.8  & 48.8     & 33.0  & 42.5   & 57.7 & 20.7 \\ %
$+\mathcal{L}_{CB}$   & \xmark & \xmark & \xmark & \xmark & 90.5 & 87.2 & 88.2 & 88.7 & 53.5 & 87.3 & 44.8 & 67.6 & 42.2 & 28.5 & 41.6 & 51.6 & 53.8 & 54.3 & 57.6 & 37.5 & 49.2 & 33.6 & 43.5 & 58.0 & 20.6 \\ \hline
DL-V2              & \cmark & \xmark & \xmark & \xmark & 90.6     & 87.3   & 88.1   & 88.8 & 53.7   & 87.4 & 44.9     & 67.7   & 42.3    & 28.6  & 41.9 & 52.9 & 57.6 & 55.2  & 57.5 & 47.6  & 50.8     & 36.9  & 44.9   & 59.2 & 19.8 \\ %
DL-V2              & \cmark & \cmark & \xmark & \xmark & 90.6     & 87.3   & 88.2   & 88.8 & 53.7   & 87.5 & 44.9     & 67.8   & 42.2    & 29.0  & 41.9 & 53.0 & 57.7 & 55.2  & 57.7 & 48.8  & 50.8     & 37.5  & 45.1   & 59.4 & 19.7 \\ %
DL-V2              & \cmark & \cmark & \cmark & \xmark & 90.8     & 87.5   & 88.5   & 88.9 & 54.0   & 87.6 & 45.1     & 68.4   & 42.3    & 30.4  & 42.1 & 53.6 & 57.8 & 55.3  & 58.7 & 53.7  & 50.8     & 39.5  & 46.1   & 60.1 & 19.3 \\ %
DL-V2              & \cmark & \cmark & \cmark & \cmark & \textbf{90.9}     & \textbf{87.8}   & \textbf{88.6}   & \textbf{89.7} & \textbf{54.1}   & \textbf{89.5} & \textbf{45.2}     & \textbf{68.8}   & \textbf{42.6}    & \textbf{32.6}  & \textbf{44.1} & \textbf{57.1} & \textbf{58.1} & \textbf{58.4}  & \textbf{62.6} & \textbf{55.3}  & \textbf{51.4}     & \textbf{40.0}    & \textbf{47.7}   & \textbf{61.3} & \textbf{19.1} \\ \hline
\end{tabular}}
\end{table}

%% file: Tables/chap-3/fredom/fredom-main-results.tex
\begin{table}[!t]
        \centering
        \caption{Comparison of Semantic Segmentation Performance with UDA Methods Using DeepLab-V2 (\textbf{DL-V2}) and Transformer (\textbf{Trans.}).}
        \label{tab:fredom-sota}
        \setlength{\tabcolsep}{2.5pt}
        \resizebox{\textwidth}{!}{%
        \begin{tabular}{|l c|cccccc|ccccccccccccc|cc|}
        \hline
         \multicolumn{2}{|c|}{\multirow{2}{*}{Approach $\quad$ Network}}   & \multicolumn{6}{c|}{Majority Group} & \multicolumn{13}{c|}{Minority Group} & \multirow{2}{*}{mIoU}          & \multirow{2}{*}{STD}  \\
         \cline{3-21}
        & &  Road          & Build.        & Veget.        & Car           & S.Walk        & Sky           & Pole          & Person        & Terrain       & Fence         & Wall          & Sign          & Bike          & Truck         & Bus           & Train         & Tr.Light      & Rider         & M.bike        &          &    \\
         
        \hline
        \multicolumn{23}{|c|}{SYNTHIA $\to$ Cityscapes} \\
        \hline
        
    IntraDA \cite{pan2020unsupervised}&  DL-V2  & 84.3 & 79.5   & 80.0   & 78.0 & 37.7   & 84.1 & 24.9 & 57.2   & $-$     & 0.4   & 5.3  & 8.4  & 36.5 & $-$   & 38.1 & $-$   & 9.2      & 23.0  & 20.3   & 41.7 & 31.0 \\
    BiMaL \cite{dat2021bimal_iccv} &   DL-V2 & \textbf{92.8} & 81.5   & 82.4   & 85.7 & 51.5   & 84.6 & 30.4 & 55.9   & $-$     & 1.0   & 10.2 & 15.9 & 38.8 & $-$   & 44.5 & $-$   & 17.6     & 22.3  & 24.6   & 46.2 & 30.9 \\
    SAC \cite{araslanov2021dasac} &   DL-V2   & 89.3 & 85.6   & 87.1   & 87.0 & \textbf{47.3}   & 89.1 & 43.1 & 63.7   & $-$     & 1.3   & 26.6 & 32.0 & 52.8 & $-$   & 35.6 & $-$   & 45.6     & 25.3  & 30.3   & 52.6 & 27.9 \\
    
    ProDA \cite{zhang2021prototypical} &   DL-V2  & 87.8 & \textbf{84.6}   & \textbf{88.1}   & 88.2 & 45.7   &  84.4    & 44.0 & \textbf{74.2}   & $-$     & 0.6   & \textbf{37.1} & 37.0 & 45.6 & $-$   & 51.1 & $-$   & \textbf{54.6}     & 24.3  & 40.5   & 55.5 & 26.4 \\
    
    \textbf{FREDOM} & DL-V2 & 86.0 & \textbf{87.0}   & 87.1   & 87.1 & 46.3   & \textbf{89.1} & \textbf{48.7} & 71.2   & $-$     & \textbf{5.3}   & 33.3 & \textbf{46.8} & \textbf{59.9} & $-$   & \textbf{54.6} & $-$   & 53.4     & \textbf{38.1}  & \textbf{51.3}   & \textbf{59.1} & \textbf{24.0} \\
    
    \hline
    TransDA \cite{transda} & Trans.  & \textbf{90.4} & 86.4   & \textbf{90.3}   & \textbf{92.3} & \textbf{54.8}   & 93.0 & 53.8 & 71.2   & $-$     & 1.7   & 31.1 & 37.1 & 49.8 & $-$   & 66.0 & $-$   & 61.1     & 25.3  & 44.4   & 59.3 & 27.3 \\
    ProCST \cite{shahaf2022procst} & Trans.   & 84.3 & 87.7   & 86.1   & 87.6 & 41.1   & 87.9 & 50.7 & 74.7   & $-$     & 6.1   & 42.6 & 54.2 & 62.5 & $-$   & 61.4 & $-$   & 55.5     & 47.2  & 53.3   & 61.4 & 22.6 \\
    DAFormer  \cite{hoyer2022daformer}& Trans. & 84.5 & 88.4   & 86.0   & 87.2 & 40.7   & 89.8 & 50.0 & 73.2   & $-$     & 6.5   & 41.5 & 54.6 & 61.7 & $-$   & 53.2 & $-$   & 55.0     & 48.2  & 53.9   & 60.9 & 22.8 \\
    
    \textbf{FREDOM} & Trans.  & 89.4 & \textbf{89.3}   & 89.9   & 90.5 & 50.8   & \textbf{93.7} & \textbf{57.3} & \textbf{79.4}   & $-$     & \textbf{9.3}   & \textbf{48.8} & \textbf{60.1} & \textbf{68.1} & $-$   & \textbf{66.0} & $-$   & \textbf{65.1}     & \textbf{51.6}  & \textbf{62.3}   & \textbf{67.0} & \textbf{22.0} \\

        \hline
        \multicolumn{23}{|c|}{GTA5 $\to$ Cityscapes}\\
        \hline
        
    IntraDA \cite{pan2020unsupervised} & DL-V2  & 90.6 & 82.6   & 85.2   & 86.4 & 36.1   & 80.2 & 27.6 & 59.3   & 39.3    & 21.3  & 29.5 & 23.1 & 37.6 & 33.6  & 53.9 & 0.0   & 31.4     & 29.4  & 32.7   & 46.3 & 26.7 \\
    BiMaL \cite{dat2021bimal_iccv} &  DL-V2   & \textbf{91.2} & 82.7   & 85.4   & 86.6 & 39.6   & 80.8 & 29.6 & 59.7   & 44.0    & 25.2  & 29.4 & 25.5 & 36.8 & 38.5  & 47.6 & 1.2   & 34.3     & 30.4  & 34.0   & 47.3 & 25.9 \\
    SAC \cite{araslanov2021dasac} &   DL-V2    & 90.3 & 86.6   & 87.5   & 88.5 & 53.9   & 86.0 & 45.1 & 67.6   & 40.2    & 27.4  & 42.5 & 42.9 & 45.1 & 49.0  & 54.6 & 9.8   & 48.6     & 29.7  & 26.6   & 53.8 & 24.2 \\
    ProDA \cite{zhang2021prototypical} &  DL-V2  & 87.8 & 79.7   & 88.6   & 88.8 & \textbf{56.0}   & 82.1 & \textbf{45.6} & \textbf{70.7}   & \textbf{45.2}    & \textbf{44.8}  & \textbf{46.3} & \textbf{53.5} & 56.4 & 45.5  & 59.4 & 1.0   & 53.5     & 39.2  & 48.9   & 57.5 & 21.7 \\
    
    \textbf{FREDOM} & DL-V2 & 90.9 & \textbf{87.8}   & \textbf{88.6}   & \textbf{89.7} & 54.1   & \textbf{89.5} & 45.2 & 68.8   & 42.6    & 32.6  & 44.1 & \textbf{57.1} & \textbf{58.1} & \textbf{58.4}  & \textbf{62.6} & \textbf{55.3}  & 51.4     & \textbf{40.0}  & \textbf{47.7}   & \textbf{61.3} & \textbf{19.1} \\
    
    \hline
    TransDA \cite{transda} & Trans. & 94.7 & 89.2   & 90.4   & 92.5 & 64.2   & 93.7 & 50.1 & 76.7   & 50.2    & 45.8  & 48.1 & 40.8 & 55.4 & 56.8  & 60.1 & 47.6  & 60.2     & 47.6  & 49.6   & 63.9 & 19.1 \\
    ProCST \cite{shahaf2022procst} &  Trans.  & 95.8 & 89.8   & 90.2   & 92.3 & 69.6   & 93.0 & 49.8 & 72.2   & 50.3    & 45.0  & 55.8 & 63.3 & 63.1 & 72.2  & 78.8 & 65.1  & 56.8     & 44.9  & 56.4   & 68.7 & 17.1 \\
    DAFormer \cite{hoyer2022daformer} & Trans. & 95.7 & 89.4   & 89.9   & 92.3 & 70.2   & 92.5 & 49.6 & 72.2   & 47.9    & 48.1  & 53.5 & 59.4 & 61.8 & 74.5  & 78.2 & 65.1  & 55.8     & 44.7  & 55.9   & 68.3 & 17.3 \\
    \textbf{FREDOM} & Trans. & \textbf{96.7} & \textbf{90.9}   & \textbf{91.6}   & \textbf{94.1} & \textbf{74.8}   & \textbf{94.4} & \textbf{57.5} & \textbf{78.4}   & \textbf{52.1}    & \textbf{49.0}  & \textbf{58.1} & \textbf{71.4} & \textbf{68.9} & \textbf{83.9}  & \textbf{85.2} & \textbf{72.5}  & \textbf{63.4}     & \textbf{53.1}  & \textbf{62.8}   & \textbf{73.6} & \textbf{15.8} \\
        
        \hline
        
        \end{tabular}
        }
\end{table}

%% file: Chapters/Chaps/chap-4-continual-learning.tex
\chapter{Open-world Fairness Continual Learning}\label{chap:continual-learning}

Continual Learning in semantic scene segmentation aims to continually learn new unseen classes in dynamic environments while maintaining previously learned knowledge. Prior studies focused on modeling the catastrophic forgetting and background shift challenges in continual learning. However, fairness, another major challenge that causes unfair predictions leading to low performance among major and minor classes, still needs to be well addressed. In addition, prior methods have yet to model the unknown classes well, thus resulting in producing non-discriminative features among unknown classes. This chapter presents novel Fairness Continual Learning approaches in semantic scene understanding.
Through our experiments, our proposed approach achieves State-of-the-Art performance on different continual learning benchmarks, i.e., ADE20K, Cityscapes, and Pascal VOC. The results have illustrated that our approach promotes the fairness of the continual semantic segmentation model.

\input{Chapters/Sections/chap-4/faircl}

\input{Chapters/Sections/chap-4/falcon}

\section{Summary}

This paper chapter has presented novel fairness continual learning approaches in open-world environments.
First, a new learning paradigm of continual learning, i.e., the prototypical Contrastive Clustering loss, is proposed to sufficiently address the catastrophic forgetting and the background shift problems.
Second, the fairness contrastive clustering loss has been introduced to address fairness problems efficiently without the assumption of ideal distributions. 
Third, the visual grammar model was presented to model the unknown classes efficiently in open-world environments.
The experimental results on different benchmarks have shown the state-of-the-art performance and fairness promotion of our approaches.

%% file: Chapters/Sections/chap-4/faircl.tex
\section{Fairness Continual Learning Approach to Semantic Scene Understanding in Open-World Environments}
\label{sec:faircl-paper}

\setcounter{propositioncounter}{0}
\setcounter{remarkcounter}{0}

Convolutional Neural Networks (CNNs)~\cite{chen2018deeplab, chen2018encoder} and Transformers~\cite{xie2021segformer, transda} have been introduced to approach semantic segmentation tasks where the models learn from the large-scale data having known classes at once.
These segmentation models learned on large-scale data may perform poorly as they may counter the new objects or new environments.
In practice, the segmentation models should be capable of learning new classes continually without re-training from the previous data.
This paradigm is defined as \textbf{Continual Semantic Segmentation} (CSS)~\cite{douillard2021plop, zhang2022representation}.
The current continual semantic segmentation approaches~\cite{douillard2021plop, zhang2022representation} concentrate on addressing two main challenges, i.e., (1) catastrophic forgetting~\cite{robins1995catastrophicforgetting,french1999catastrophicforgetting,thrun1998lifelonglearning} and (2)  background shift~\cite{douillard2021plop, zhang2022representation}. 
The former problem indicates the forgetting issue of the model about the previous knowledge when learning on new training data ~\cite{douillard2021plop, zhang2022representation, ssul_neurips_2021}. Meanwhile, the latter problem refers to the classes of previous or future data that have collapsed into a background class~\cite{cermelli2020modelingthebackground}.
However, another critical problem that needs attention is the fairness issue in semantic segmentation.

\begin{wrapfigure}{r}{0.5\textwidth}
    \centering
    \includegraphics[width=0.5\textwidth]{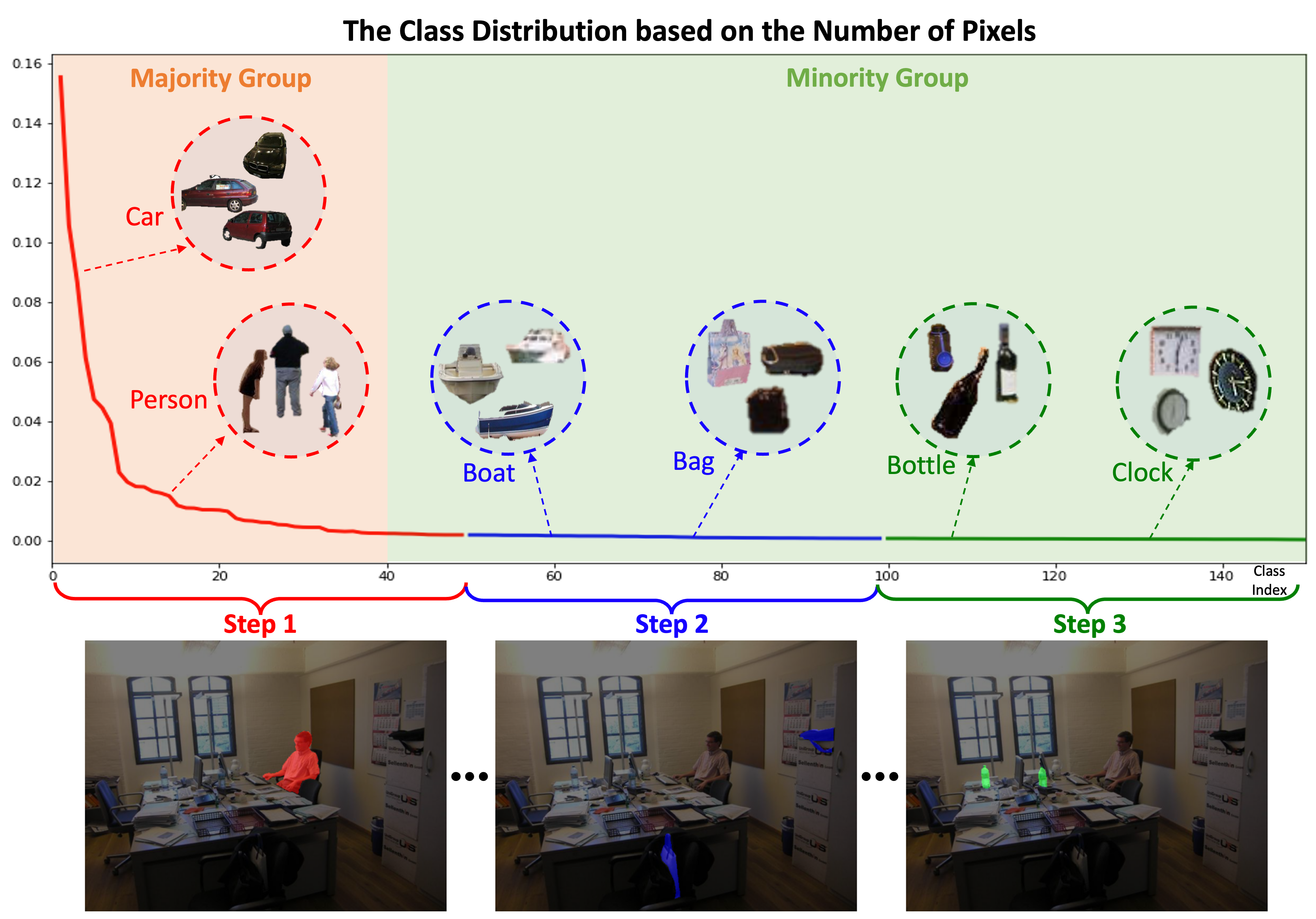} 
    \caption{\textbf{The Class Distribution of ADE20K.} In the ADE20K 50-50 (3 steps) benchmark, the distribution of classes in the majority group in the early step dominates the ones in the minority groups in the later steps. The distributions of classes gradually decrease over steps.}
    \label{fig:faircl-distribution}
\end{wrapfigure}
As presented in Section \ref{sec:fredom-paper}, the unfair predictions caused by imbalanced data class distribution (Figure~\ref{fig:faircl-distribution}) could result in severe problems, especially in human-related applications that could influence human safety. 
Moreover, the fairness problem could even be well observed in the context of continual learning when the model encounters new classes without accessing previous training data. The prior work of continual learning in image classification~\cite{zhao2020maintaining} has also considered this problem. Several prior studies~\cite{araslanov2021dasac, hoyer2022daformer, DBLP:conf/aaai/Ting21} in semantic segmentation have tried to reduce the effect of the class imbalance by introducing the weighted cross entropy~\cite{Cui_2019_CVPR, wang2021seesaw, DBLP:conf/aaai/Ting21}, focal loss~\cite{araslanov2021dasac}, over-sampling techniques~\cite{hoyer2022daformer, zhang2021prototypical, araslanov2021dasac}.
However, the fairness problem in continual semantic segmentation has yet to be well-defined and directly addressed. Therefore, there should be more attention on addressing the fairness issue in continual semantic segmentation. 

In this chapter, we present a novel \textbf{Fair}ness \textbf{C}ontinual \textbf{L}earning (\textbf{FairCL}) approach to semantic scene segmentation. 
First, under the perspective of fairness learning, the new metric is formulated to measure the fairness of the model via the error rate among classes. Then, the metric is further derived into the three main objectives, i.e., (1) the \textit{Task-specific Objective} that handles the catastrophic forgetting problem, (2) the \textit{Fairness Objective} that maintains the fairness of predictions produced by the model based on the class distribution, and (3) the \textit{Conditional Structural Constraint} that imposes the consistency of the segmentation predictions. 
Second, to sufficiently model the continual learning problem, the novel \textit{Prototypical Contrastive Clustering} loss is presented to address the catastrophic forgetting and the background shifting problems. Moreover, the proposed Prototypical Contrastive Clustering loss has been proven to be a generalized paradigm of knowledge distillation approaches commonly adopted in continual learning.

\subsection{The Proposed Fairness Continual Learning Approach}

Let $F$ parameterized by $\theta$ be the deep semantic segmentation model that maps an input image $\mathbf{x} \in \mathcal{X}$ to the segmentation map $\mathbf{y} \in \mathcal{Y}$, $\mathbf{y} = F(\mathbf{x}, \theta)$. 
Continual Semantic Segmentation (CSS) aims to learn a model in $T$ steps. 
In each step, the segmentation model $F$ encounters a dataset $\mathcal{D}^t = \{\mathbf{x}^{t}, \mathbf{\hat{y}}^t\}$ where $\mathbf{x}^t$ is the input image and $\mathbf{\hat{y}}^t$ is the ground-truth segmentation at time $t \in [1..T]$. The current ground-truth segmentation map $\mathbf{\hat{y}}^t$ only contains the labels of the current classes $\mathcal{C}^t$ and all the class labels of prevision steps, $\mathcal{C}^{1..t-1}$, or the future steps, $\mathcal{C}^{t+1..T}$ are collapsed into a background class or ignored. Formally, learning the semantic segmentation at time step $t$ can be formulated as in Eqn.~\eqref{eqn:faircl-continual-learning-general}.
\begin{equation} \label{eqn:faircl-continual-learning-general}
\small
\theta_t^* = \arg\min_{\theta_t} \mathbb{E}_{\mathbf{x}^t, \mathbf{\hat{y}}^t \in \mathcal{D}^t} \mathcal{L}\left(F(\mathbf{x}^t, \theta_t),  \mathbf{\hat{y}}^t\right) 
\end{equation}
where $\theta_t^*$ is the parameters of $F$ at time step $t$, $\mathcal{L}$ is the objective learning of the continual learning task.
In CSS learning, at the current time step $t$, the segmentation model $F$ is expected to not only predict all the classes $\mathcal{C}^{1..t-1}$ learned in the previous steps but also predict the current new classes $\mathcal{C}^t$.
Three significant challenges have been identified in this learning setting and should be addressed.
\begin{itemize}
    \item \textbf{\textit{Background Shift.}} At time step $t$, the labels of previous and future steps have been ignored. Thus, the pixels of these classes are ambiguous, which means these could contain either the class of previous or future steps. During learning $\mathcal{C}^t$, the model could consider these classes as negative samples. As a result, the model tends to learn non-discriminative features for these pixels, leading to difficulty learning new classes or forgetting the old ones.
    
    \item \textbf{\textit{Catastrophic Forgetting.}} cause the model may partially or completely forget the knowledge of classes $\mathcal{C}^{1..t-1}$ when learning the new classes $\mathcal{C}^t$. This problem could be caused by the background shift and the learning mechanism. Since classes in $\mathcal{C}^{1..t-1}$ are considered as the background class at time step $t$, the model tends to update the knowledge of the new classes while the predictions of classes incline to be suppressed.
    
    \item \textbf{\textit{Fairness.}} While the prior approaches~\cite{douillard2021plop, zhang2022representation, ssul_neurips_2021} focus on addressing the two above challenges, the fairness issue has received less attention and has not been well addressed yet. However, fairness is one of the most important criteria as it guarantees the model behaves fairly among not only classes in $\mathcal{C}^t$ but also classes in $\mathcal{C}^{1..t}$ that have been learned. The fairness in CSS is typically caused by 
    the imbalance distribution between classes as several classes occupy the larger portion or exist more frequently than other classes (Figure~\ref{fig:faircl-distribution}). In addition, the appearance of training samples at the training step $t$ also exaggerates the bias due to the new classes since the classes in the previous training steps have collapsed.
\end{itemize}

To address these challenges in CSS, we first reconsider the semantic segmentation problem from the fairness viewpoint followed by introducing a novel approach to alleviate the fairness issue based on the ideal class distribution.
Then, we introduce a novel Prototypical Contrastive Clustering loss to model background classes and catastrophic forgetting. 

\subsubsection{Fairness Learning Approach to Continual Semantic Segmentation}

To address the fairness problem in continual learning, we adopt our fairness learning approach presented in Section \ref{sec:fredom-paper}.
In particular, we first assume that there exists an ideal distribution $q(\cdot)$ where the class distributions $q(y_{i, j})$ are equally distributed. Under this assumption, the model learned is expected to behave fairly among classes as there is no bias toward any groups of classes. \textit{It should be noted that our assumption is used to derive our learning objective and is going to be relaxed later. In other words, the ideal data is not required at training.} Then, our continual learning objective could be formed as Eqn.~\eqref{eqn:faircl-opt_from_idealum}.
\begin{equation} \label{eqn:faircl-opt_from_idealum}
\small
\begin{split}
    \theta^* = \arg\min_{\theta} \left[\mathbb{E}_{\mathbf{x} \sim p(\mathbf{y}), \mathbf{\hat{y}} \sim p(\mathbf{\hat{y}})} \sum_{i,j} \mathcal{L}(y_{i,j}, \hat{y}_{i,j}) \frac{q(y_{i,j})q(\mathbf{y}_{\setminus (i,j)}|y_{i,j})}{p(y_{i,j})p(\mathbf{y}_{\setminus (i,j)}|y_{i,j})} \right]
\end{split}
\end{equation}
The fraction between the ideal distribution $q(\cdot)$ and the actual data distribution $p(\cdot)$ is the residual learning objective for the model to achieve the desired fairness goal. 
Let us further derive Eqn.~\eqref{eqn:faircl-opt_from_idealum} by taking the log as in Eqn.~\eqref{eqn:faircl-take_log}.
\begin{equation} \label{eqn:faircl-take_log}
    \small
    \begin{split}
        &\theta^* = \arg\min_{\theta} \mathbb{E}_{\mathbf{x} \sim p(\mathbf{x}), \mathbf{\hat{y}} \sim p(\mathbf{\hat{y}})}\left\{\mathcal{L}(\mathbf{y}, \mathbf{\hat{y}}) 
        +\frac{1}{N}\sum_{i, j}
        \left[\log\left(\frac{q(y_{i,j})}{p(y_{i,j})}\right) 
        +\log\left(\frac{q(\mathbf{y}_{\setminus (i, j)} |y_{i, j})}{p(\mathbf{y}_{\setminus (i, j)} |y_{i, j})}\right)\right]\right\}
    \end{split}
\end{equation}
The proof of our Eqn.~\eqref{eqn:faircl-opt_from_idealum}  can be found in our preliminary work~\cite{truong2023fairness}. As shown in Eqn.~\eqref{eqn:faircl-take_log}, there are three learning objectives as follows:
\begin{itemize}
    \item \textbf{The Continual Learning Objective.} The first term, i.e., $\mathcal{L}({\mathbf{y}, \mathbf{\hat{y}}})$, represents the task-specific loss which is the continual learning objective. 
    This objective aims to address the catastrophic forgetting and background shift problems.  To achieve this desired goal, we introduce a novel Prototypical Contrastive Clustering Loss that will be discussed in Section \ref{sec:clustering_loss}. 

    \item \textbf{The Fairness Objective.}  The second term, i.e., $\mathcal{L}_{class} = \log\left(\frac{q(y_{i,j})}{p(y_{i,j})}\right)$, maintains the fairness in the predictions produced by the model. This objective penalizes the prediction of classes forcing the model to behave fairly based on the class distribution.
    Under the ideal distribution assumption where the model is expected to perform fairly, the $q(y_{i,j})$ will be considered as a uniform distribution $\mathcal{U}$, i.e., $q(y_{i,j}) \sim \mathcal{U}(C)$ ($C$ is the number of classes).
    
    \item \textbf{The Conditional Structural Consistency Objective.} The third term, i.e., $\mathcal{L}_{cons} = \log\left(\frac{q(\mathbf{y}_{\setminus (i, j)} |y_{i, j})}{p(\mathbf{y}_{\setminus (i, j)} |y_{i, j})}\right)$, regularizes the structural consistency of the prediction.  This objective acts as a metric to constrain the structure of the predicted segmentation under the ideal distribution assumption. To model this conditional structure consistency, we introduce a  conditional structural constraint based on the Markovian assumption discussed in Section \ref{sec:cond_structure}.
    
\end{itemize}

\subsubsection{Prototypical Contrastive Clustering Loss for Handling Unknown Classes}
\label{sec:clustering_loss}

\begin{figure}[!t]
    \centering
    {\includegraphics[width=0.9\textwidth]{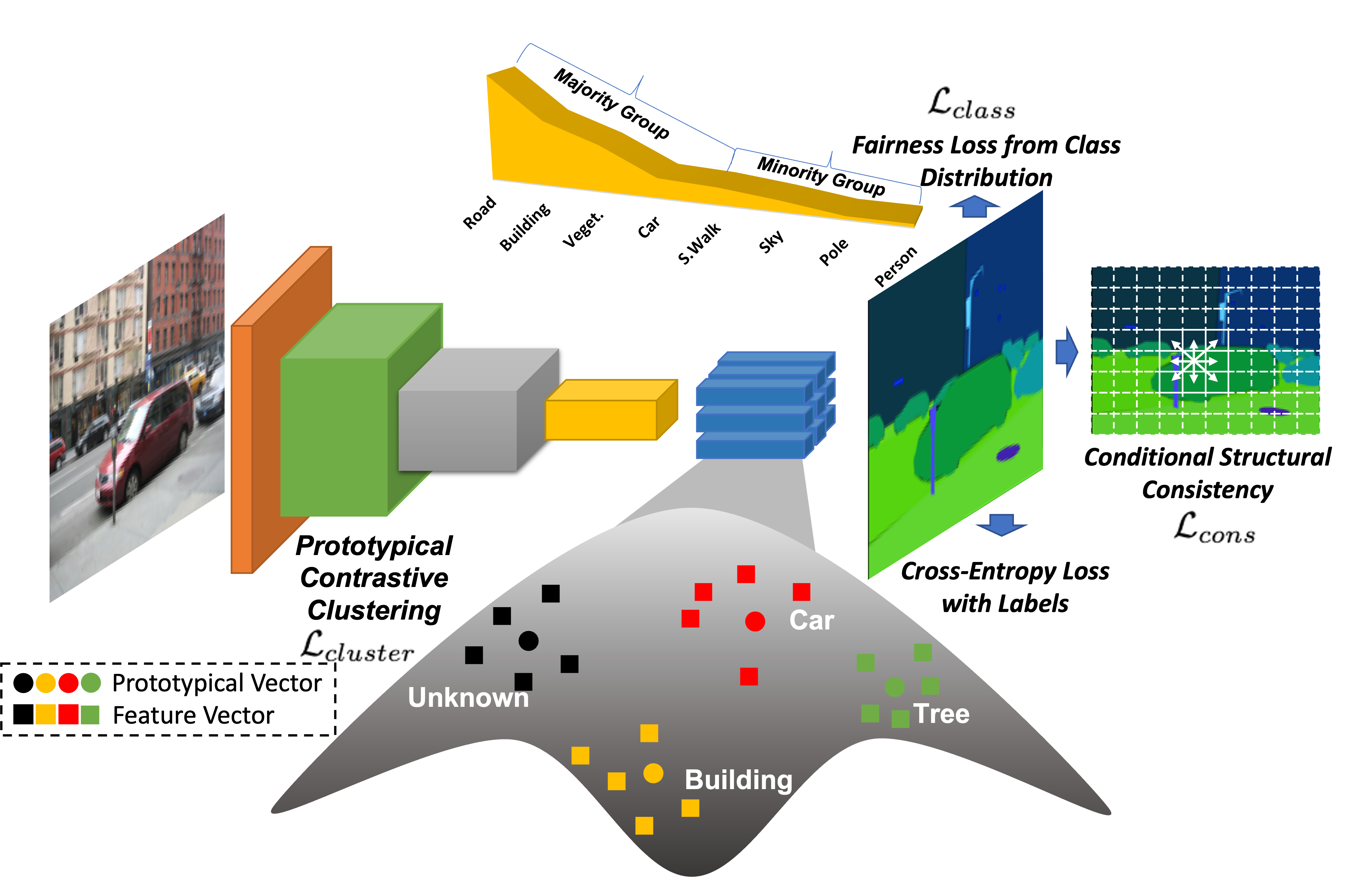}}
    \caption{\textbf{The Proposed Fairness Continual Learning Framework.}
    The predicted segmentation maps are imposed the cross-entropy loss, the Prototypical Contrastive Clustering loss ($\mathcal{L}_{cluster}$), the Fairness Loss from Class Distribution ($\mathcal{L}_{class}$), and the Conditional Structural Consistency loss ($\mathcal{L}_{cons}$)}\label{fig:faircl-proposed_framework}
\end{figure}

A successful continual learning approach should be able to model background classes without explicit supervision and confront the forgetting problem of previously learned classes when the labels of the task are provided to update the knowledge of the model~\cite{joseph2021towards}. 
A straightforward adoption of Softmax could not be enough to handle. Indeed, the unlabeled pixels will be ignored during training. Thus, it results in these unannotated pixels could be treated as negative samples. Consequently, the segmentation model tends to produce indiscriminative features for these unknown classes that limit the capability of learning new classes in future tasks or recognizing the classes learned previously.

To address this limitation, in addition to the Softmax loss, we introduce a novel Prototypical Contrastive Clustering Loss. 
In particular, the semantic segmentation pixel belonging to each class can be represented in latent space.
Inspired by~\cite{joseph2021towards, cen2021deep, li2023open}, the features representing the classes can be separated by defining it as a contrastive clustering problem~\cite{joseph2021towards} where features of the same class would be pulled closer while features of different classes would be pushed far away. 
In addition, the deep representations of unknown classes will be grouped into the same cluster of unknown classes to produce discriminative features against other classes.

Formally, for each class $c \in \mathcal{C}^{1..t}$, it is represented by a prototypical vector $\mathbf{p}_c$. In addition, the additional prototypical vector $\mathbf{p}_0$ represents a cluster of unknown classes.
Let $\mathbf{f}^t_{i,j}$ be a feature representation of pixel at location $(i, j)$ of the input $\mathbf{x}^t$.
Then, the Prototypical Contrastive Clustering Loss $\mathcal{L}_{cluster}$ can be defined via a distance $\mathcal{D}$ as in Eqn.~\eqref{eqn:faircl-clustering_loss}.
\begin{equation}
\small
\label{eqn:faircl-clustering_loss}
\begin{split}
    \mathcal{L}_{cluster}(\mathbf{x}^t, F, \theta_t) & = 
    \sum_{i, j}\sum_{c} \mathcal{D}(\mathbf{f}^t_{i, j}, \mathbf{p}_c) \\ %
    \mathcal{D}(\mathbf{f}^t_{i, j}, \mathbf{p}_c) & = 
    \begin{cases} 
        \ell(\mathbf{f}^t_{i, j}, \mathbf{p}_c) & \text{If } \hat{y}^t_{i, j} = c\\
        \max\{0, \Delta - \ell(\mathbf{f}^t_{i, j}, \mathbf{p}_c\} & \text{otherwise}
    \end{cases}
\end{split}
\end{equation}
where $\ell$ is a distance metric, $\Delta$ is a margin between the feature vectors of different classes, 
and $\hat{y}^t_{i, j}$ is the label of pixel $(i, j)$. Minimizing this loss separates the classes represented in the latent space. For step $t > 1$, $\hat{y}^t_{i, j}$ of an unknown-class pixel will utilize a pseudo label where its assigned label is computed based on the closest cluster. In addition,  since the prototypical vectors of classes $c \in \mathcal{C}^{1..t-1}$ have been well learned to represent for classes, these vectors $\mathbf{p}_c$ (where $c \in \mathcal{C}^{1..t-1}$) will be frozen at step $t$ to maintain its learned knowledge of classes $\mathcal{C}^{1..t-1}$. 

The set of prototypical vectors at current step $t$, i.e., $\mathbf{p}_0$ and $\mathbf{p}_c$ where $c \in \mathcal{C}^t$ are updated gradually with respect to the growth of feature vectors. In particular, the prototypical vector $\mathbf{p}_c$ will be periodically updated (after every $M$ iterations) with momentum $\eta$ based on the set of features $\mathbf{f}_{i,j}$ of class $c$. Following common practices~\cite{joseph2021towards, he2020momentum}, to effectively support the updating step and memory efficiency, for each class $c$, we only maintain a set of features $\mathcal{S}_c$ with a fixed length of $L$.
Figure~\ref{fig:faircl-proposed_framework} illustrates our proposed FairCL framework.

\subsubsection{Prototypical Constrative Clustering Loss to Catastrophic Forgetting}

Knowledge Distillation is a common continual learning approach~\cite{shmelkov2017incremental, cermelli2020modelingthebackground, douillard2021plop, zhang2022representation} where the knowledge of the previous model will be distilled into the current model. This mechanism prevents the segmentation model from diverging knowledge learned previously and avoiding the catastrophic forgetting problem. This continual learning paradigm has been widely adopted due to its efficiency in computation. In addition, this approach also does not require data rehearsal, i.e., storing the data samples of previous tasks. In this paper, we demonstrate that our Prototypical Constrative Clustering approach is a comprehensive upper limit of the Knowledge Distillation approach. In particular, the common knowledge distillation approach can be formulated as in Eqn.~\eqref{eqn:faircl-ell_distill}.
\begin{equation}\label{eqn:faircl-ell_distill}
\small
    \mathcal{L}_{distill}(\mathbf{x}^t, F, \theta_t, \theta_{t-1}) = \mathcal{D}(\mathbf{f}^{t-1}, \mathbf{f}^t)
\end{equation}
where $\mathbf{f}^{t}$ and $\mathbf{f}^{t-1}$ are the features of the input $\mathbf{x}^t$ produced by the segmentation model at step $t$ and step $t-1$, respectively; and the distance metric $\mathcal{D}$ measure the knowledge gap between $\mathbf{f}^{t}$ and $\mathbf{f}^{t-1}$.

\begin{tcolorbox}[colback=blue!5!white,colframe=blue!75!black,title=\textbf{Proposition \showpropositioncounter\label{pro:clustering_upperbound}}]
The Prototypical Constrative Clustering Approach is the generalized upper bound of the Knowledge Distillation Approach.
\begin{equation}
   \small
   \mathcal{L}_{distill}(\mathbf{x}^t, F, \theta_t, \theta_{t-1})  = \mathcal{O}\left(\mathcal{L}_{cluster}(\mathbf{x}^t, F, \theta_t)\right)
\end{equation}
\end{tcolorbox}

The proof of \textbf{Proposition \ref{pro:clustering_upperbound}} can be found in our preliminary work~\cite{truong2023fairness}.
Intuitively, under this upper bound of \textbf{Proposition \ref{pro:clustering_upperbound}}, by only optimizing our prototypical contrastive clustering loss, the knowledge distillation constraint has also been implicitly imposed.
Beyond the property of generalized upper bound stated in \textbf{Proposition \ref{pro:clustering_upperbound}}, our approach offers other benefits over the knowledge distillation approach. In particular, our approach is computationally efficient, where our method only requires a single forward pass of the segmentation model. Meanwhile, the knowledge distillation demands two forward passes for both the current and previous models, which also requires additional computational memory for the previous model. Moreover, our approach provides a better representation of each class $c$ through the prototypical vector $\mathbf{p}_c$. This mechanism helps to effectively maintain the knowledge of classes learned previously while allowing the model to update the new knowledge without rehearsing the old data.

\subsubsection{Learning Conditional Structural Consistency}
\label{sec:cond_structure}

The conditional structural constraint plays an important role as it will ensure the consistency of the predicted segmentation map. However, modeling the conditional structural constraint $\log\left(\frac{q(\mathbf{y}_{\setminus (i, j)} |y_{i, j})}{p(\mathbf{y}_{\setminus (i, j)} |y_{i, j})}\right)$ in Eqn.~\eqref{eqn:faircl-take_log} is a quite challenging problem due to two factors, i.e., (1) the unknown ideal conditional distribution $q(\mathbf{y}_{\setminus (i, j)} |y_{i, j})$, and the complexity of the distribution $q(\mathbf{y}_{\setminus (i, j)} |y_{i, j})$. 
To address the first limitation of unknown ideal distribution, let us consider the following tight bound as in Eqn.~\eqref{eqn:faircl-ideal_relax}. 
\begin{equation}\label{eqn:faircl-ideal_relax}
\small
    \mathbb{E}_{\mathbf{x} \sim p(\mathbf{x})} \log\left(\frac{q(\mathbf{y}_{\setminus (i, j)} |y_{i, j})}{p(\mathbf{y}_{\setminus (i, j)} |y_{i, j})}\right) \leq -\mathbb{E}_{\mathbf{x} \sim p(\mathbf{x})} \log p(\mathbf{y}_{\setminus (i, j)} |y_{i, j})
\end{equation}
The inequality in Eqn.~\eqref{eqn:faircl-ideal_relax} always hold with respect to any form ideal distribution $q(\cdot)$. Thus, optimizing the negative log-likelihood of $\log p(\mathbf{y}_{\setminus (i, j)} |y_{i, j})$ could also regularize the conditional structural constraint due to the upper bound of Eqn.~\eqref{eqn:faircl-ideal_relax}. More importantly, \textit{the requirement of ideal data distribution during training has also been relaxed.} 
However, up to this point, the second limitation of modeling the complex distribution $q(\mathbf{y}_{\setminus (i, j)} |y_{i, j})$ has still not been solved. 
To address this problem, we adopt the Markovian assumption~\cite{chen2018deeplab, dat2021bimal_iccv} to model conditional structural consistency. In particular, we propose a simple yet effective approach to impose the consistency of the segmentation map through the prediction at location $(i, j)$ and predictions of its neighbor pixels. Formally, the conditional structure consistency can be formed via the Gaussian kernel as in Eqn.~\eqref{eqn:faircl-cond_loss_simple}.
\begin{equation} \label{eqn:faircl-cond_loss_simple}
\small
    - \log p(\mathbf{y}_{\setminus (i, j)} |y_{i, j})  \propto  \sum_{(i', j') \in \mathcal{N}(i, j)} \operatorname{exp}{\left\{-\frac{||x_t^{i, j} -  x^t_{i', j'}||_2^2}{2\sigma_1^2} - \frac{||y_t^{i, j} -  y^t_{i', j'}||_2^2}{2\sigma_2^2}\right\}}%
\end{equation}
where $\mathcal{N}(i, j)$ is the set of neighbor pixels of $(i, j)$, $\{\sigma_1, \sigma_2\}$ are the scale hyper-parameters  of the Gaussian kernels. The conditional structural consistency loss defined in Eqn.~\eqref{eqn:faircl-cond_loss_simple} enhance the smoothness and maintain the consistency of the predicted segmentation map by imposing similar predictions of  neighbor pixels with similar contextual colors.

\subsection{Experimental Results}

In this section, we first describe the datasets and metrics used in our experiments. Then, we present the ablation studies to illustrate the effectiveness of our proposed method. Finally, we compare our approach with prior CSS methods to demonstrate our SOTA performance. 

\subsubsection{Datasets and Evaluation Protocols}

\noindent
\textbf{Datasets.} \textbf{\textit{ADE20K}}~\cite{ade20k_challenge} is a semantic segmentation dataset that consists of more than 20K scene images of 150 semantic categories. Each image has been densely annotated with pix-level objects and objects parts labels. 
\textbf{\textit{Cityscapes}}~\cite{cordts2016cityscapes} is a real-world autonomous driving dataset collected in European. This dataset includes $3,975$ urban images with high-quality, dense labels of 30 semantic categories. 
\textbf{\textit{PASCAL VOC}}~\cite{Everingham15} is a common dataset that consists of more than $10$K images of $20$ classes. 

\noindent
\textbf{Implementation.}
Two segmentation network architectures have been used in our experiments, i.e., (1) DeepLab-V3~\cite{chen2018deeplab} with the ResNet-101 backbone, and (2) SegFormer~\cite{xie2021segformer} with MiT-B3 backbone.
Our framework is implemented in PyTorch and trained on four 40GB-VRAM NVIDIA A100 GPUs.
The model is optimized by the SGD optimizer~\cite{bottou2010large} with momentum 0.9, weight decay $10^{-4}$, and batch size of $6$ per GPU. 
The learning rate is set individually for each step and dataset. 
In particular, the learning rate for the initial step and the continual steps of the ADE20K dataset is $10^{-2}$ and $10^{-3}$ respectively, while the learning rate for the Cityscapes experiment is $2 \times 10^{-2}$ and $2 \times 10^{-3}$.
The feature vectors from the last layer of the decoder are used for the prototypical clustering loss.
For each class, the number of feature vectors in each set $\mathcal{S}_c$ for computing the prototypes is $500$ features.
Following common practices in contrastive learning~\cite{joseph2021towards, he2020momentum}, we adopt the Euclidean distance for our $\ell$ in the Prototypical Contrastive Clustering loss $\mathcal{L}_{cluster}$ and the margin $\nabla$ between features of different classes is set to $10$. The momentum $\eta$ to update the prototypical vectors is set to $0.99$. 
Following~\cite{chen2018deeplab, dat2021bimal_iccv}, in the conditional structural consistency loss, the number of neighbor pixels is within a window size of $3 \times 3$.

\noindent
\textbf{Evaluation Protocols.} 
Following~\cite{douillard2021plop}, we focus on the overlapped CSS evaluation. Our proposed method is evaluated on several settings for each dataset, i.e.,  ADA20K 100-50 (2 steps), ADA20K  100-10 (6 steps), ADA20K 100-5 (11 steps), Cityscapes 11-5 (3 steps),  Cityscapes 11-1 (11 steps), Cityscapes 1-1 (21 steps), Pascal VOC 15-1 (3 steps), and Pascal VOC 10-1 (11 steps).
The mean Intersection over Union (mIoU) metric is used in our experiments.
The mIoU is computed after the last step for the classes learned from the first step, the later continual classes, and all classes. 
The mIoU for the initial classes shows the robustness of the model to catastrophic forgetting, while the metric for the later classes reflects the ability to learn new classes.
To measure the fairness of the model, we also report the standard deviation (STD) of IoUs over all classes.

\subsubsection{Ablation Study}

Our ablative experiments study the effectiveness of our proposed FairCL approach on the performance of the CSS model and fairness improvement on the ADE20K 100-50 benchmark (Table~\ref{tab:faircl-ablation}).

\input{Tables/chap-4/faircl/ablation_study}

\noindent
\textbf{Effectiveness of the Network Backbone.}
Table~\ref{tab:faircl-ablation} illustrates the results of our approach using the DeepLab-V3~\cite{chen2018deeplab} with the Resnet101 backbone and the SegFormer~\cite{xie2021segformer} with a Transformer backbone, i.e. MiT-B3~\cite{xie2021segformer}.
As shown in our results, the performance of segmentation models using a more powerful backbone, i.e., Transformer, outperforms the models using the Resnet backbone.
The capability of learning new classes has been improved notably, i.e., the mIoU of classes 101-150 in the full configuration has been improved from $19.86\%$ to $25.46\%$ 
while the model keeps robust to catastrophic forgetting, the mIoU has been increased from $41.96\%$ to $43.56\%$ in the classes 0-100.
Additionally, fairness between classes has been promoted when the standard deviation of the IoU over classes has been reduced from $21.67\%$ to $21.10\%$.

\noindent
\textbf{Effectiveness of the Prototypical Contrastive Clustering Loss.}
We evaluate the impact of the Prototypical Contrastive Clustering Loss ($\mathcal{L}_{cluster}$) in improving  the performance in the continual learning problem compared to the fine-tuning approach.
As shown in Table~\ref{tab:faircl-ablation}, the clustering loss has significant improvements in the catastrophic forgetting robustness compared to using only the Softmax loss.
In particular, the mIoU of classes 0-100 for both DeepLab-V3 and SegFormer backbones has been improved by $+41.63\%$ and $+43.30\%$ respectively that makes the overall mIoU increase by $+26.45\%$ and $+28.44\%$,
and the average mIoU between classes increases by $+13.17\%$ and $+14.03\%$.
Although the STD of IoUs has slightly increased in this setting, the major target of our $\mathcal{L}_{cluster}$ is used to model the catastrophic forgetting and background shift problems in CSS illustrated by the significant performance improvement of mIoU.

\noindent
\textbf{Effectiveness of the Fairness Treatment Loss.}
As reported in Table~\ref{tab:faircl-ablation}, the fairness treatment from the class distribution loss $\mathcal{L}_{class}$ significantly improves the overall performance and the accuracy of classes.
In detail, the STD of IoU from classes has been reduced by $0.96\%$ and $0.46\%$ for both backbones while the mIoU has been improved from $32.97\%$ to $34.40\%$ and from $36.18\%$ to $36.78\%$, respectively. The results have shown that our approach has promoted the fairness of the model.

\noindent
\textbf{Effectiveness of the Conditional Structural Consistency.}
Table~\ref{tab:faircl-ablation} shows experimental results of our model using conditional structure constraint loss $\mathcal{L}_{cons}$.
As illustrated in our results, the conditional structure constraint demonstrates effective improvement.
Indeed, it promotes the accuracy of the initial classes and the novel classes when the mIoU has been increased from $43.35\%$ to $43.56\%$ and from $23.50\%$ to $25.46\%$ respectively with the Transformer backbone.
The fairness of classes is also improved as the standard deviation of the IoU of classes 0-100 and classes 101-150 is reduced from $19.03\%$ to $18.71\%$ and from $20.75\%$ to $19.99\%$. %

\subsubsection{Comparison with State-of-the-Art Methods}

\input{Tables/chap-4/faircl/cityscapes}
\noindent
\textbf{Cityscapes.}
As shown in Table~\ref{tab:faircl-cityscapes_domain}, our FairCL outperforms previous SOTA methods evaluated on Cityscapes benchmarks.
In particular, in the 11-5 task, our method using Resnet and Transformer achieves the mIoU of $66.96\%$ and $67.85\%$ respectively which shows better performance than prior methods.
Meanwhile, the results for the 11-1 task are $66.61\%$ and $67.09\%$ w.r.t. the Resnet and Transformer backbones.
For the 1-1 task, the mIoU of our method is $49.22\%$ and $55.68\%$. 

\input{Tables/chap-4/faircl/ade20k}

 \input{Tables/chap-4/faircl/voc}

\noindent
\textbf{ADE20K.}
Table~\ref{tab:faircl-adeota} presents our experimental results using ResNet and Transformer backbones compared to prior SOTA approaches.
Our approach achieves SOTA performance and outperforms prior methods.
In particular, our approach achieves the final mIoU of $36.99\%$ for Resnet and $37.56\%$ for Transformer in the 100-50 tasks.
For the 50-50 tasks, the model reaches the final mIoU of $34.55\%$ and $35.15\%$ for the Resnet and Transformer backbones, respectively while the result of the prior method~\cite{goswami2023attribution} is $33.50\%$.
Meanwhile, the overall results of our method for the 100-10 task are $34.65\%$ and $35.49\%$ which shows outperforming prior methods.

\noindent
\textbf{Pascal VOC.}
As shown in Table~\ref{tab:faircl-voc}, the proposed method outperforms the prior approaches evaluated on the Pascal VOC 2012 dataset.
In detail, our method achieves the overall mIoU of $61.5\%$ in the 15-1 task while the result of the previous method~\cite{zhang2022representation} is $59.4\%$. Meanwhile, the mIoU in the 10-1 task is $36.6\%$ which shows better performance than the prior methods.

\noindent
\textbf{Qualitative Results.}
Figure~\ref{fig:faircl-visualize} visualizes the qualitative result of our method compared to PLOP~\cite{douillard2021plop}. 
Initially, the ground truth contains the class ``car'' in the first step and the class ``minibike'' in the third step.
Then, in the fourth step, the class ``bicycle'' is included.
As a result, PLOP~\cite{douillard2021plop} partly forgets the ``minibike'' information when learning the class ``bicycle'' information.
Meanwhile, our method consistently maintains the information of ``minibike'' and predicts segmentation correctly.

%% file: Tables/chap-4/faircl/ablation_study.tex
\begin{table}[!b]
\setlength{\tabcolsep}{3.5pt}
\caption{
Effectiveness of our approach on the ADE20K 100-50 benchmark. Two different networks, i.e., DeepLab-V2 \cite{chen2018deeplab} and SegFormer \cite{xie2021segformer}, and three different losses, i.e., Prototypical Contrastive Clustering ($\mathcal{L}_{cluster}$), Fairness Loss ($\mathcal{L}_{class}$), and Conditional Structural Consistency ($\mathcal{L}_{cons}$). 
}
\label{tab:faircl-ablation}
\resizebox{1.0\textwidth}{!}{
\begin{tabular}{|c|ccc|cc|cc|cc|cc|}
\hline
\multirow{2}{*}{\textbf{Backbone}}    & \multirow{2}{*}{$\mathcal{L}_{cluster}$} & \multirow{2}{*}{$\mathcal{L}_{class}$} & \multirow{2}{*}{$\mathcal{L}_{cons}$} & \multicolumn{2}{c}{\textbf{0-100}} & \multicolumn{2}{|c}{\textbf{100-150}} & \multicolumn{2}{|c}{\textbf{all}} & \multicolumn{2}{|c|}{\textbf{avg}} \\
                             &                             &                      &                       & \textbf{mIoU}        & \textbf{STD}         & \textbf{mIoU}         & \textbf{STD}          & \textbf{mIoU}       & \textbf{STD}        & \textbf{mIoU}       & \textbf{STD}        \\
\hline
\multirow{4}{*}{DeepLab-V3}      &                             &                      &                       &  0.08  & 0.84  & 19.52 & 20.18 & 6.52  & 13.14 & 24.41 & 13.14      \\
                             & \cmark                           &                      &                       & 41.71 & 19.90 & 15.33 & 21.96 & 32.97 & 23.03 & 37.58 & 23.03      \\
                             & \cmark                           & \cmark                    &                       & 42.25 & 19.31 & 18.55 & 20.52 & 34.40 & 22.07 & 38.35 & 22.07      \\
                             & \cmark                           & \cmark                    & \cmark                     & \textbf{43.40} & \textbf{19.08} & \textbf{24.04} & \textbf{19.12} & \textbf{36.99} & \textbf{21.67} & \textbf{40.45} & \textbf{21.67}      \\
                             \hline
                             \hline
\multirow{4}{*}{SegFormer} &                             &                      &                       &  0.10  & 0.84  & 23.18 & 19.83 & 7.74  & 15.74 & 25.82 & 15.74      \\
                             & \cmark                           &                      &                       & 43.40 & 19.35 & 21.60 & 22.06 & 36.18 & 22.32 & 39.85 & 22.32      \\
                             & \cmark                           & \cmark                    &                       & 43.35 & 19.03 & 23.50 & 20.75 & 36.78 & 21.86 & 40.34 & 21.86      \\
                             & \cmark                           & \cmark                    & \cmark                     & \textbf{43.56} & \textbf{18.71} & \textbf{25.46} & \textbf{19.99} & \textbf{37.56} & \textbf{21.10} & \textbf{40.73} & \textbf{21.10}      \\
\hline
\end{tabular}
}
\end{table}

%% file: Tables/chap-4/faircl/cityscapes.tex
\begin{wraptable}{r}{0.5\textwidth}
\centering
\small
\setlength{\tabcolsep}{5pt}
\caption{Final mIoU (\%) for Continual Semantic Segmentation on Cityscapes.}
\label{tab:faircl-cityscapes_domain}
\resizebox{0.5\textwidth}{!}{
\begin{tabular}{|l|c|c|c|}
\hline
\multirow{2}{*}{\textbf{Method}} & \textbf{11-5} & \textbf{11-1}  &  \textbf{1-1} \\
 & 3 steps &  11 steps & 21 steps \\
\hline
Joint & 79.30 & 79.30 & 79.30 \\
\hline
LWF-MC \cite{rebuffi2017icarl} & 58.90 & 56.92 & 31.24\\
ILT \cite{michieli2019ilt} & 59.14 & 57.75 & 30.11 \\
MiB \cite{cermelli2020modelingthebackground} & 61.51 & 60.02 & 42.15 \\
PLOP \cite{douillard2021plop} & 63.51 & 62.05 & 45.24 \\
RCIL \cite{zhang2022representation} & 64.30 & 63.00 & 48.90 \\
\hline
FairCL + DeepLab-V3 & 66.96 & 66.61 & 49.22 \\
FairCL + SegFormer & \textbf{67.85} &\textbf{ 67.09} &\textbf{ 55.68} \\
\hline
\end{tabular}
}

\end{wraptable} 

%% file: Tables/chap-4/faircl/ade20k.tex
\begin{table}[!b]
\centering
\setlength{\tabcolsep}{3pt}
\caption{Continual Semantic Segmentation results on ADE20K in mIoU (\%).}
\label{tab:faircl-adeota}
 \resizebox{1.0\textwidth}{!}{
\begin{tabular}{|l|cccc|cccc|cccc|}
\hline
\multirow{2}{*}{\textbf{Method}} & \multicolumn{4}{c|}{\textbf{100-50} (2 steps)} & \multicolumn{4}{c|}{\textbf{50-50} (3 steps)} & \multicolumn{4}{c|}{\textbf{100-10} (6 steps)}\\
 & 0-100 & 101-150 & \textit{all} & \textit{avg} & 0-50 & 51-150 & \textit{all} & \textit{avg}  & 0-100 & 101-150 & \textit{all} & \textit{avg} \\
\hline
Joint & 44.30 & 28.20 & 38.90 & - & 51.10 & 32.80 & 38.90 & - & 44.30 & 28.20 & 38.90 & - \\
\hline
ILT \cite{michieli2019ilt} & 18.29  & 14.40  & 17.00  & 29.42  &  3.53  & 12.85  &  9.70 & 30.12 &  0.11  &  3.06 &  1.09 & 12.56 \\ 
MiB \cite{cermelli2020modelingthebackground} & 40.52  & 17.17  & 32.79  & 37.31  & 45.57  & 21.01  & 29.31 & 38.98 & 38.21 & 11.12 & 29.24 & 35.12 \\
PLOP \cite{douillard2021plop} & {41.87}  & 14.89  & {32.94}  & {37.39}  & {48.83}  & {20.99}  & {30.40} & {39.42} & {40.48}  & {13.61} & {31.59} & {36.64}\\
RCIL \cite{zhang2022representation} & 42.30 & 18.80 & 34.50 & 38.48 & 48.30 & 25.00 & 32.50 & - & 39.30 & 17.60 & 32.10 & - \\
MiB + AWT \cite{goswami2023attribution} & 40.90 & 24.70 & 35.60 &  -  &  46.60 &  26.85 & 33.50 & - & 39.10 & 21.28 & 33.20 & - \\
SSUL \cite{ssul_neurips_2021} & 41.28 & 18.02 & 33.58 & - & 48.38 & 20.15 & 29.56 & - & 40.20 & 18.75 & 33.10 & - \\
SATS \cite{sats_prj_2023} & - & - & - & - & - & - & - & - & 41.42 & 19.09 & 34.18 & - \\
\hline
FairCL + DeepLab-V3 & 43.40 & 24.04 & 36.99 & 40.45 & \textbf{49.65} & 26.84 & 34.55 & 41.68 & 41.73 & 20.36 & 34.65 & 39.01 \\ 
FairCL + SegFormer & \textbf{43.56} & \textbf{25.46} & \textbf{37.56} & \textbf{40.73} & {49.62} & \textbf{27.78} & \textbf{35.15} & \textbf{42.25} & \textbf{42.21} & \textbf{21.91} & \textbf{35.49} & \textbf{39.36} \\
\hline
\end{tabular}
}

\end{table}

%% file: Tables/chap-4/faircl/voc.tex
\begin{table*}[!t]
\begin{minipage}[b]{0.55\textwidth}
    \centering
    \setlength{\tabcolsep}{3pt}
    \resizebox{1.0\textwidth}{!}{
    \begin{tabular}{|l| ccc | ccc|}
    \hline
           \multirow{2}{*}{\textbf{Method}} & \multicolumn{3}{c|}{\textbf{15-1 (6 steps)}} & \multicolumn{3}{c|}{\textbf{10-1 (11 steps)}} \\
            & 0-15         & 16-20         & all          & 0-10          & 11-20         & all          \\
           \hline
    Joint & 79.8 & 72.4 & 77.4 & 78.4 & 76.4 & 77.4 \\
    \hline
    LWF \cite{rebuffi2017icarl}    & 6.0          & 3.9           & 5.5          & 8.0           & 2.0           & 4.8          \\
    ILT \cite{michieli2019ilt}    & 9.6          & 7.8           & 9.2          & 7.2           & 3.7           & 5.5          \\
    MiB \cite{cermelli2020modelingthebackground}   & 38.0         & 13.5          & 32.2         & 20.0          & 20.1          & 20.1         \\
    SDR \cite{sdr}    & 47.3         & 14.7          & 39.5         & 32.4          & 17.1          & 25.1         \\
    PLOP \cite{douillard2021plop}   & 65.1         & 21.1          & 54.6         & 44.0          & 15.5          & 30.5         \\
    RCIL \cite{zhang2022representation}   & 70.6         & \textbf{23.7}          & 59.4         & 55.4          & 15.1          & 34.3         \\
    \hline
    FairCL + DeepLab-V3 & {72.0}          & {22.7}          & {60.3}          & {42.3}          & \textbf{25.6}          & {34.4}        \\
    FairCL + SegFormer & \textbf{73.5}          & {22.8}          & \textbf{61.5}          & \textbf{57.1}          & {14.2}          & \textbf{36.6}        \\
    \hline
    \end{tabular}
    }
     \caption{The mIoU Performance (\%) of CSS on Pascal VOC.}
    \label{tab:faircl-voc}
\end{minipage} 
\hspace{5mm}
\begin{minipage}[b]{0.40\textwidth}
    \centering
    \includegraphics[width=1.0\textwidth]{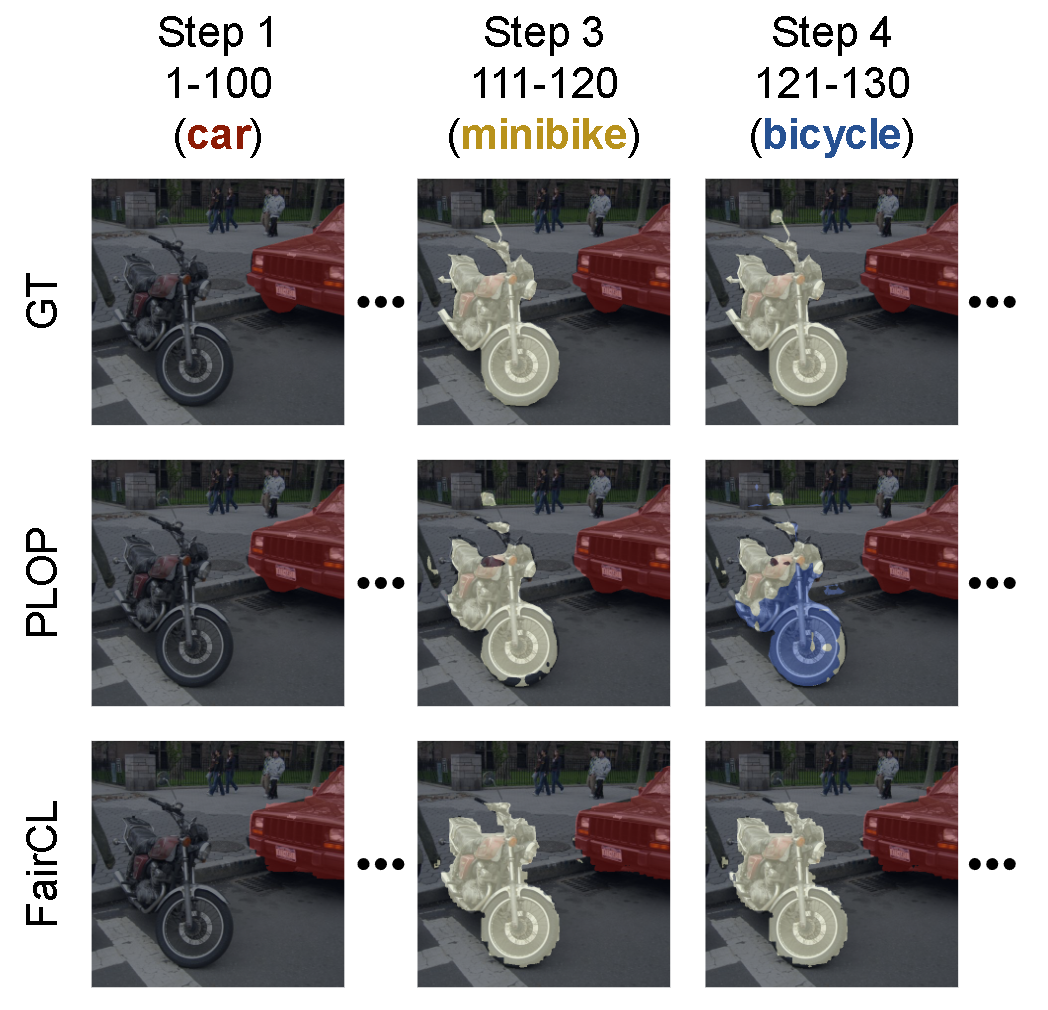} 
    \captionsetup{type=figure}
    \caption{Qualitative results of Our Approach and PLOP \cite{douillard2021plop} on ADE20K 100-10.}
    \label{fig:faircl-visualize}
\end{minipage}
\end{table*}

%% file: Chapters/Sections/chap-4/falcon.tex
\section{Fairness Learning via Contrastive Attention Approach to Continual Semantic Scene Understanding}
\label{sec:falcon-paper}

\setcounter{propositioncounter}{0}
\setcounter{remarkcounter}{0}

\begin{figure}[!b] %
    \centering
    \includegraphics[width=0.9\textwidth]{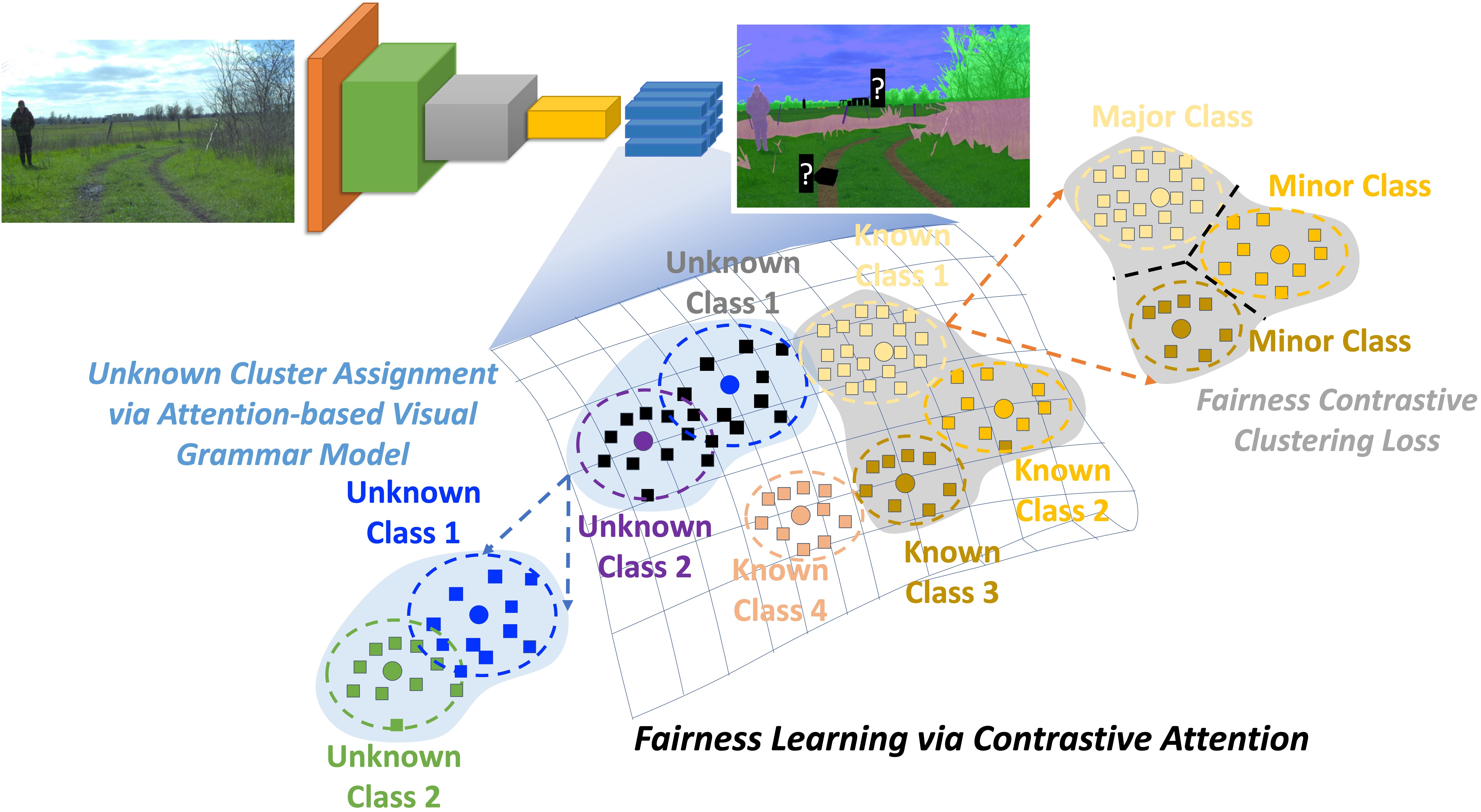}
    \caption{\textbf{Our Fairness Learning via Contrastive Attention to Continual Semantic Segmentation.} The \textit{Fairness Contrastive Clustering Loss} promotes the fairness of the model while the \textit{Attention-based Visual Grammar} models the unknown classes. 
    } \label{fig:falcon-highlight}
\end{figure}

In the previous section, we have presented the approach to address the fairness problem in continual semantic segmentation. However, the FairCL approach still suffers several limitations.
First, the FairCL approach relies on the assumption of ideal balanced data which is impractical and could not be achieved by nature.
Second, FairCL cannot handle unknown classes since they either consider these unknown classes as a background class or assign unknown pixels by a pseudo label of prior known classes.
This problem will limit the ability of the model to handle unknown classes, especially in the \textbf{open-world} or \textbf{open-set} context, where the model may encounter new, unseen objects.
To address these limitations, this section introduces a novel approach to effectively model the fairness problem and unknown classes in the continual learning setting.

In this chapter, we present a novel {\textit{\textbf{Fa}irness \textbf{L}earning via \textbf{Con}trastive Attention Approach (\textbf{FALCON})}} to Continual Semantic Segmentation (as shown in Figure~\ref{fig:falcon-highlight}). 
First, we introduce a novel \textit{\textbf{Contrastive Clustering Paradigm}} approach to Continual Learning that models the catastrophic forgetting problem. Second, by analyzing the limitation of vanilla Contrastive Clustering in biased data, we introduce a novel \textit{\textbf{Fairness Contrastive Clustering}} loss to model the fairness problem in continual learning efficiently. Third, to effectively model the background shift problem, we introduce a new \textit{\textbf{Attention-based Visual Grammar}} 
that model the topological structures of feature distribution to handle the unknown classes effectively.

\subsection{Fundamental of Contrastive Learning to Continual Learning}

Following the continual learning framework presented in Section \ref{sec:faircl-paper}, learning the CSS model at step $t$ can be reformed as Eqn.~\eqref{eqn:falcon-general_obj}.
\begin{equation} 
\small
\label{eqn:falcon-general_obj}
\theta_t^* = \arg\min_{\theta_t} \mathbb{E}_{\mathbf{x}^t, \mathbf{\hat{y}}^t \in \mathcal{D}^t} \left[\mathcal{L}_{CE}\left(\mathbf{y}^t,  \mathbf{\hat{y}}^t\right) + \lambda_{CL}\mathcal{L}_{CL}\left(F(\mathbf{x}^t)\right)\right]
\end{equation}
where, $\mathbf{y}^t = F(\mathbf{x}^t, \theta_t)$, 
$\theta_t$ is the parameter of model $F$ at current learning step $t$,
$\mathcal{L}_{CE}$ is the cross-entropy loss,
$\lambda_{CL}$ is the balanced weight.
and $\mathcal{L}_{CL}$ is the CSS objective.
Prior methods~\cite{shmelkov2017incremental, cermelli2020modelingthebackground, douillard2021plop, zhang2022representation, cermelli2023comformer} adopt knowledge distillation to design $\mathcal{L}_{CL}$.
However, this method prevents the CSS model from diverging knowledge learned previously, therefore resulting in limiting the ability to adopt new knowledge~\cite{truong2023fairness}.
In addition, these methods have not addressed fairness and background shift problems due to their dedicated design for maintaining knowledge via distillation~\cite{douillard2021plop, cermelli2023comformer, cswkd_cvpr_2022}. Therefore, to address these problems, we introduce a novel \textbf{\textit{Fairness Learning via Contrastive Attention Approach}} to CSS.

\subsubsection{Continual Learning via Contrastive Clustering}

\begin{figure}[t]
    \centering
    \includegraphics[width=1.0\textwidth]{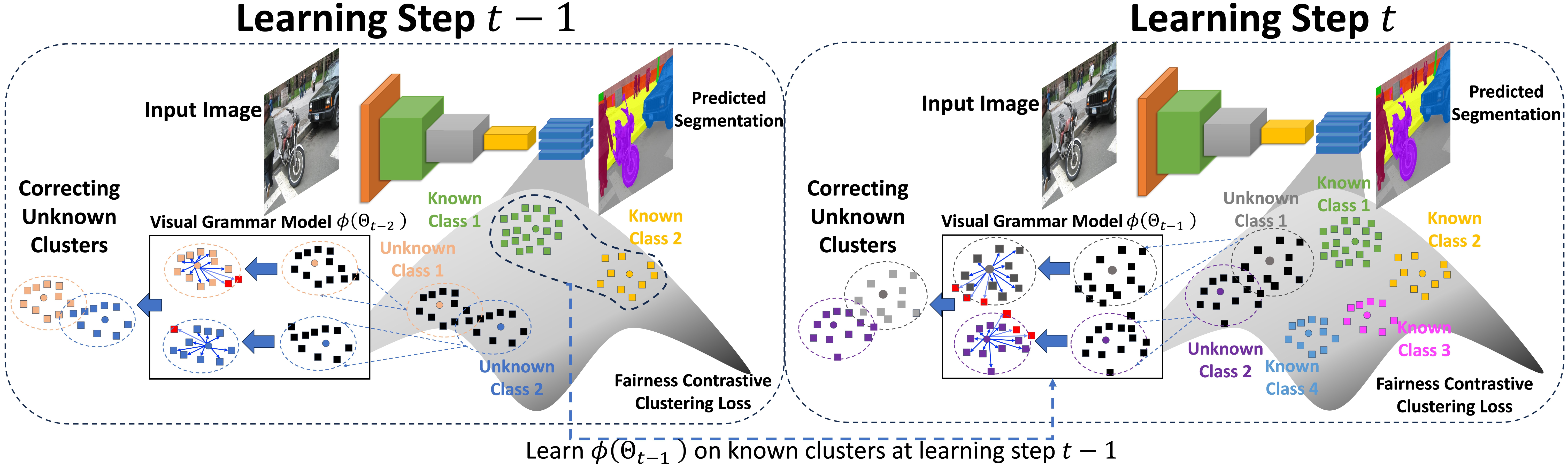}
    \caption{\textbf{The Proposed Fairness Continual Learning Framework.}}\label{fig:falcon-css_framework}
\end{figure}

Apart from prior methods~\cite{douillard2021plop, cermelli2023comformer, cswkd_cvpr_2022}, our CSS is defined as Contrastive Clustering Learning.
Given a set of centroid vectors $\{\mathbf{c}_i\}_{i=1}^{N_K + N_U}$ where $N_K = |\mathcal{C}^{1..t}|$ and $N_U$ is the number of known and unknown classes up to current learning tasks.
Prior work~\cite{joseph2021towards, truong2023fairness, cermelli2023comformer} often defined the number of unknown classes as $1$ where background classes are considered as a single unknown class. 
Formally, our Contrastive Clustering Learning for CSS can be defined as Eqn.~\eqref{eqn:falcon-general_clustering}.
\begin{equation} 
\small
\label{eqn:falcon-general_clustering}
\begin{split}
\mathcal{L}_{CL}\left(F(\mathbf{x}^t)\right) &= \sum_{\mathbf{c}_i}\mathcal{L}_{Cont}(\mathbf{F}^t, \mathbf{c}_i) =  \sum_{\mathbf{c}_i}\sum_{h, w}-\phi(\mathbf{f}^t_{h,w}, \mathbf{c}_i)\log \frac{\exp(\mathbf{f}^t_{h,w} \times \mathbf{c}_i)}{\sum_{\mathbf{f}'}\exp(\mathbf{f}' \times \mathbf{c}_i)}
\end{split}
\end{equation}
where $\mathbf{F}^t \in \mathbb{R}^{H \times W \times D}$ 
is the feature maps extracted from the input image $\mathbf{x}^t$ by the segmentation network $F$, 
$\mathbf{f}^t_{h,w} \in \mathbb{R}^{D}$ is the feature at the pixel location $(h , w)$ of features $\mathbf{F}^t$,
$\sum_{\mathbf{f}'}$ means the summation over all feature representations $\mathbf{f}' \in \mathbb{R}^{D}$, 
and $\phi: \mathbb{R}^{D} \times \mathbb{R}^{D}  \to [0, 1]$ is the function that determines either $\mathbf{f}^t_{h,w}$ belongs to the cluster $\mathbf{c}_i$ or not.

\begin{wrapfigure}[16]{r}{0.5\textwidth}
    \centering
    \includegraphics[width=0.5\textwidth]{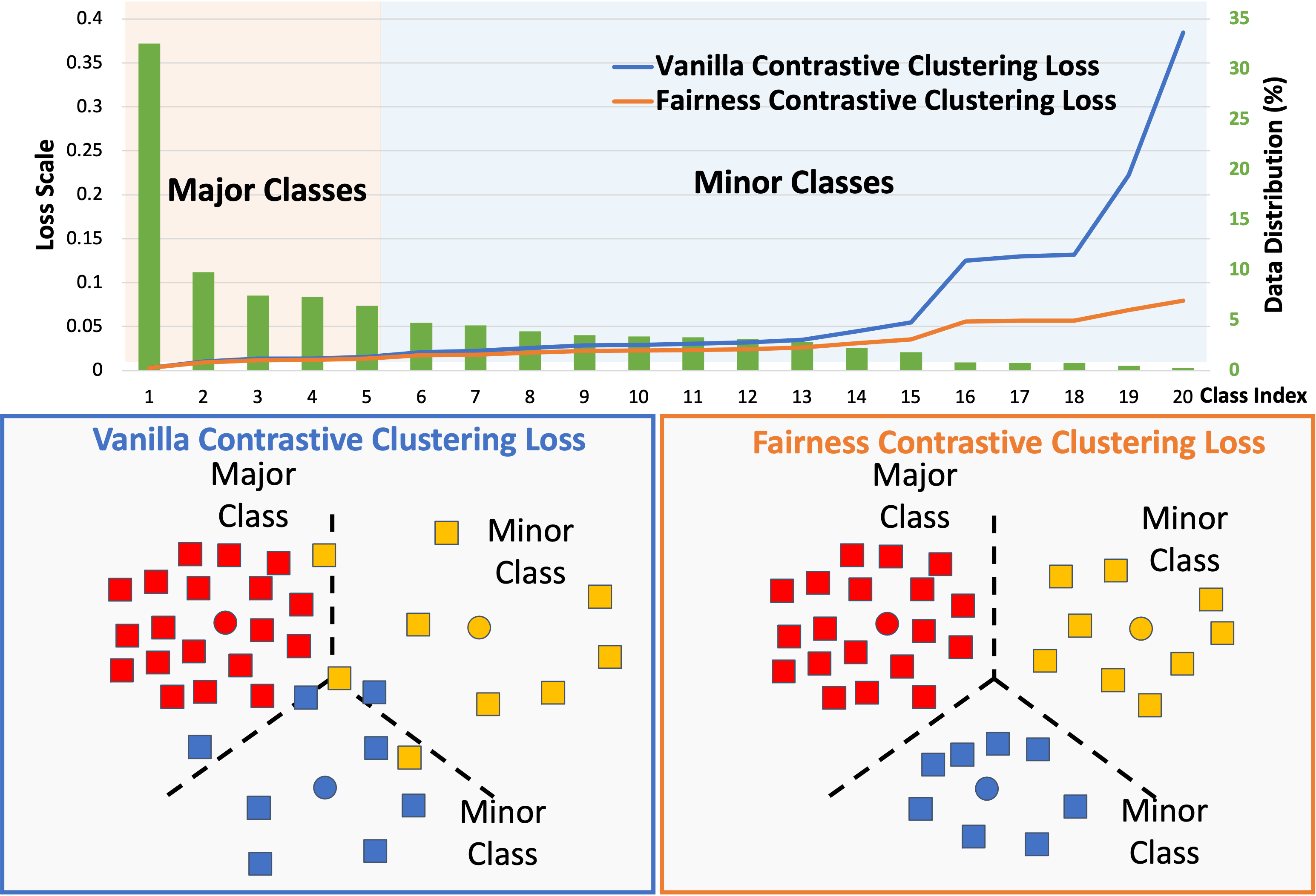} 
    \caption{\textbf{The Enforcement Loss of Contrastive Clustering $\mathcal{L}_{Cont}$ and Fairness Contrastive Clustering $\mathcal{L}^{\alpha}_{Cont}$ on Pascal VOC}.
    Since $\mathcal{L}_{Cont}$ suffers severe biased among classes, its clusters of minor classes remain scattered.
    Our $\mathcal{L}^{\alpha}_{Cont}$ produces a more uniform loss among classes, thus promoting fairness and compactness of clusters.
    }
    \label{fig:falcon-impact_biased_data}
\end{wrapfigure}
By defining CSS as contrastive clustering learning, the knowledge of the segmentation model has been well maintained via the cluster vectors $\mathbf{c}$ to avoid the catastrophic forgetting problem, while the learning objective allows the cluster vectors to adapt to new, unseen environments and classes. 

\subsection{The Proposed Fairness Learning via Contrastive Attention}

Two major problems are identified in Eqn.~\eqref{eqn:falcon-general_clustering}.
First, the $\mathcal{L}_{CL}$ objective in  Eqn.~\eqref{eqn:falcon-general_clustering} suffers severe biased among classes because several classes appear more frequently than others due to the imbalanced training data. 
Second, as the function $\phi$ requires the labels to determine the features belonging to clusters, it limits the ability to model the unknown classes where their labels are not available. 
Therefore, the following sections will present a novel approach to tackle these problems.

\subsubsection{Fairness Contrastive Clustering Learning}
\label{sec:falcon-fairness}

While contrastive clustering learning defined in Eqn.~\eqref{eqn:falcon-general_clustering} promotes the compact representations of features around their clusters, inspired by~\cite{zhu2022balanced, cui2021parametric, chen2022perfectly}, we observe that the imbalanced class distribution will influence unfair behaviors among classes. 
In particular, for simplicity, we consider $\{\mathbf{f}^t_i\}_{i=1}^{L}$ is the set of features that belong to the cluster $\mathbf{c}$ at learning step $t$ (i.e., $\phi(\mathbf{f}_i^{t}, \mathbf{c}) = 1$) and $L$ is the number of features (in this case, $L$ is the total number of pixels belong to the class of cluster $\mathbf{c}$). Let us define the enforcement between the feature $\mathbf{f}^t_t$ and the cluster $\mathbf{c}$ as $\ell_i = \frac{\exp(\mathbf{f}^t_{i} \times \mathbf{c})}{\sum_{\mathbf{f}'}\exp(\mathbf{f}' \times \mathbf{c})}$. Hence, 
the lower the value of the enforcement $\ell_i$ is, the more compact the representation of visual features and clusters is.
Then, the contrastive clustering learning loss in Eqn.~\eqref{eqn:falcon-general_clustering} of entire cluster $\mathbf{c}$ can be defined as Eqn.~\eqref{eqn:falcon-loss_for_one_cluster}.
\begin{equation}\label{eqn:falcon-loss_for_one_cluster}
\small
    \mathcal{L}_{Cont}(;, \mathbf{c}) = -\sum_{i=1}^L\log\frac{\exp(\mathbf{f}^t_{i} \times \mathbf{c})}{\sum_{\mathbf{f}'}\exp(\mathbf{f}' \times \mathbf{c})} = -\sum_{i=1}^L\log \ell_i 
\end{equation}

\begin{tcolorbox}[colback=blue!5!white,colframe=blue!75!black,title=\textbf{Proposition \showpropositioncounter\label{pro:unfair_contrastive_loss}}]
If the contrastive clustering loss $\mathcal{L}_{Cont}(;, \mathbf{c})$ achieves the optimal value, the enforcement $\ell_i$ between the feature and the cluster will converge to $\ell_i = L^{-1}$.
\end{tcolorbox}

\noindent
\textbf{Proposition \ref{pro:unfair_contrastive_loss}} has implied that the class with more samples will result in a lower value of the enforcement and produce a more compact representation, while the class having fewer samples will be more scattered in the feature space due to the higher value of the enforcement. In particular, let $L_{major}$ and $L_{minor}$ be the number of samples of the major and minor class where $L_{major} >> L_{minor}$. Then, based on \textbf{Proposition \ref{pro:unfair_contrastive_loss}}, the enforcement between features and the cluster of the major class will be significantly lower than the one of the minor class, i.e., $L_{major}^{-1} << L_{minor}^{-1}$.
Therefore, a direct adoption of the contrastive clustering loss in Eqn.~\eqref{eqn:falcon-general_clustering} will result in an unfair CSS model.
In addition, for classes in the minority group, the weak enforcement results in the feature presentations of classes being far away from their clusters. Thus, the model will produce non-discriminative features compared to the ones in the majority group.  Moreover, if the loss is applied to the cases of unknown labels, these feature representations can be scattered in the latent space and pulled into the incorrect clusters due to weak enforcement between features and clusters (Figure~\ref{fig:falcon-impact_biased_data}).

To address the unfair problem in contrastive clustering learning, 
inspired by~\cite{zhu2022balanced, cui2021parametric, chen2022perfectly}, 
we introduce a scaling factor $\alpha$ and a learnable transition vector $\mathbf{v}$ for each cluster $\mathbf{c}$ (all clusters have the same value of $\alpha$ but different vector $\mathbf{v}$).
\textbf{\textit{ Our Fairness Contrastive Clustering Learning Loss}} for the entire cluster in Eqn.~\eqref{eqn:falcon-loss_for_one_cluster} can be re-formed as in Eqn.~\eqref{eqn:falcon-loss_for_one_cluster_alpha}.
\begin{equation}\label{eqn:falcon-loss_for_one_cluster_alpha}
\small
    \mathcal{L}^{\alpha}_{Cont}(;, \mathbf{c}) = -\alpha\sum_{i=1}^L\log\frac{\exp(\mathbf{f}^t_{i} \times \mathbf{c})}{\sum_{\mathbf{f}'}\exp(\mathbf{f}' \times \mathbf{c})} -\log\frac{\exp(\mathbf{v} \times \mathbf{c})}{\sum_{\mathbf{f}'}\exp(\mathbf{f}' \times \mathbf{c})}
\end{equation}
Intuitively, the scaling factor $\alpha$ will help to re-scale the impact of the enforcement in learning, and the transitive vector $\mathbf{v}$ assists in translating the center cluster into the proper position of the latent space. 
This action promotes the compactness of clusters in the minority group.

\begin{tcolorbox}[colback=blue!5!white,colframe=blue!75!black,title={\textbf{Proposition \showpropositioncounter\label{pro:fair_contrastive_loss}}}]
If the fairness contrastive clustering loss $\mathcal{L}^{\alpha}_{Cont}(;, \mathbf{c})$ achieves the optimal value, the enforcement $\ell_i$ between the feature and the cluster will converge to $\ell_i = (\alpha^{-1} + L)^{-1}$.
\end{tcolorbox}

The proof of propositions can be found in our preliminary work~\cite{truong2024falcon}. 
Under the \textbf{Proposition \ref{pro:fair_contrastive_loss}}, when the value of $\alpha$ is small, the divergence of the enforcement between major and minor classes will be smaller, i.e., $||(\alpha^{-1}+L_{major})^{-1} - (\alpha^{-1}+L_{minor})^{-1}|| < || L^{-1}_{major} - L^{-1}_{minor}|| $. Figure~\ref{fig:falcon-impact_biased_data} has illustrated the impact of fairness contrastive clustering loss. Therefore, our designed proposed fairness contrastive loss has effectively addressed the fairness issue in Eqn.~\eqref{eqn:falcon-general_clustering}. 
It should be noted that although the smaller $\alpha$ results in the fairer enforcement varied from major to minor classes. However, if the value of scaling factor $\alpha$ is too small, the contrastive clustering loss will rely more on the enforcement of the transitive vector $\mathbf{v}$, and the distribution of features $\mathbf{f}^t_i$ around its cluster $\mathbf{c}$ will be scattered due the weak enforcement caused by small $\alpha$.
Therefore, the value of scaling factor $\alpha$ in practice should be carefully selected.

\subsubsection{An Efficient Unknown Class Modeling}
\label{sec:falcon-unknown_class_modeling}

An ideal CSS approach must be able to model the unknown classes without supervision, especially in open-set contexts~\cite{joseph2021towards, truong2023fairness} where there could be multiple unknown classes or objects.
Prior studies have adopted the pseudo-label strategies~\cite{douillard2021plop, cermelli2023comformer} based on the model predictions to assign labels for seen classes while unseen classes have been ignored, thus resulting in non-discriminative features.  
\cite{joseph2021towards, truong2023fairness} improved the background modeling by using an additional prototypical representation for unknown classes.
However, these approaches consider different unknown classes as one (i.e., $N_U = 1$) resulting in non-distinguished representations of different unknown classes.
Thus, modeling function $\phi$ in Eqn.~\eqref{eqn:falcon-general_clustering} without supervision of different unknown classes (i.e., $N_U > 1$) is challenging.

\begin{figure}[!b] %
    \centering
    \includegraphics[width=1.0\textwidth]{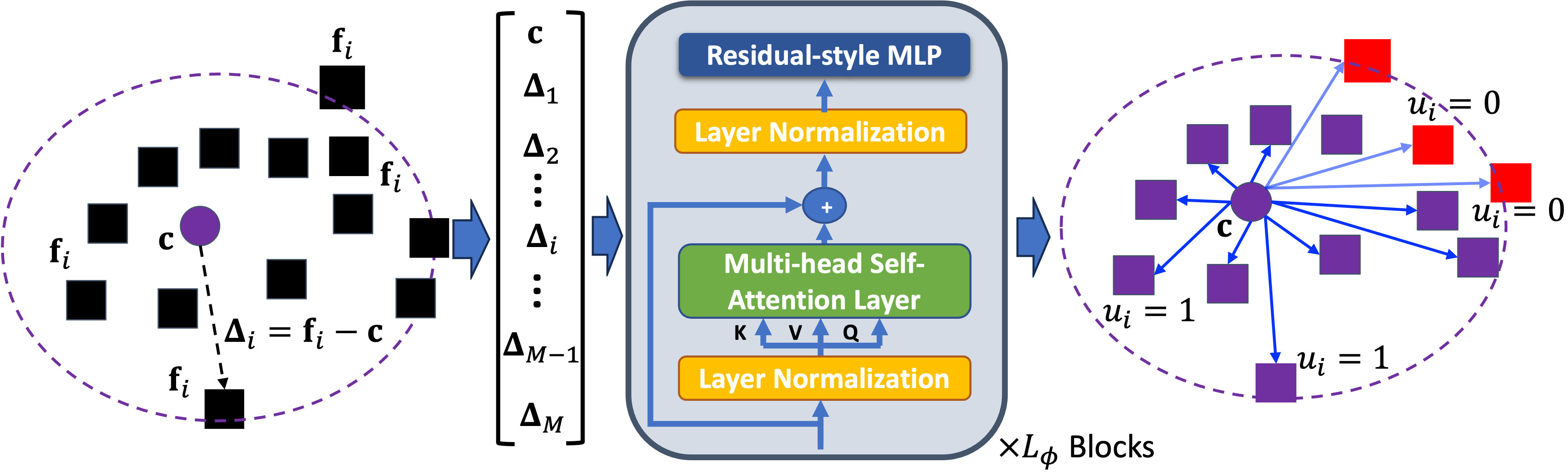}
    \caption{\textbf{The Proposed Visual Grammar Model.}}
    \label{fig:falcon-visual_grammar_model}
\end{figure}

Although modeling $\phi$ to determine the single feature $\mathbf{f}$ belonging to the cluster $\mathbf{c}$ is challenging, 
prior studies in clustering~\cite{nguyen2021clusformer, yang2020learning, yang2019learning} have suggested that determine a set of features $\{\mathbf{f}^t_i\}_{i=1}^M$ belonging to cluster $\mathbf{c}$ should be easier.
This derives from the fact that even though the feature representations of different classes are different, \textit{the distributions of features around its cluster} (termed as \textbf{\textit{Visual Grammar}}) \textit{in the feature space should be similar among classes or clusters}.
As a result, by learning the distribution of features and their clusters, the model $\phi$ can determine whether a feature belongs to a cluster.
Then, by learning the model $\phi$ on prior known clusters, the knowledge of  $\phi$ can be adaptively applied to unknown clusters.
Figure~\ref{fig:falcon-visual_grammar_model} illustrates our visual grammar model.

\noindent
\textbf{Limitations of Prior Clustering Methods.}
The traditional methods in clustering, e.g., KNN or density-based clustering~\cite{ester1996density}, remain limited to noisy features leading to producing the incorrect cluster assignment.
Meanwhile, the modern clustering methods, e.g., Graph Neural Networks (GNNs)~\cite{yang2020learning, yang2019learning}, require a large memory to build the affinity graph for clusters. 
In addition, GNNs often learn the local structures of graphs (or clusters) and accumulate them via the aggregation layers.
Hence, the global structures of the clusters, i.e., \textit{visual grammar}, are not well modeled by GNNs~\cite{nguyen2021clusformer}.
Therefore, to address these limitations, we introduced a new \textbf{\textit{Attention-based Visual Grammar}} approach to efficiently model the distribution of features and their clusters via self-attention~\cite{vaswani2017attention}.

\begin{tcolorbox}[colback=black!5!white,colframe=black!75!black,title=\textbf{Remark \showremarkcounter\label{rmk:visual_grammar}}]
Given a center $\mathbf{c}$ and a set of $M$ features $\{\mathbf{f}^{\mathbf{c}}_{i}\}_{i=1}^M$ where $\mathbf{f}^{\mathbf{c}}_{i}$ denotes the feature $\mathbf{f}_{i}$ belonging to the cluster $\mathbf{c}$, and $ \forall i \in [1..M-1]: \cos(\mathbf{f}_i^{\mathbf{c}}, \mathbf{c}) \geq \cos(\mathbf{f}_{i+1}^{\mathbf{c}}, \mathbf{c})$
the \textbf{Visual Grammar} of the cluster $\mathbf{c}$ parameterized by $\Theta$ can be defined as Eqn.~\eqref{eqn:falcon-visual_grammar}.
\begin{equation} \label{eqn:falcon-visual_grammar}
\small
\begin{split}
    \Theta^{*} &= \arg\min_{\Theta} \mathbb{E}_{\mathbf{c}, \{\mathbf{f}^{\mathbf{c}}_{i}\}_{i=1}^M} \left[-\log p(\mathbf{f}^{\mathbf{c}}_1, \mathbf{f}^{\mathbf{c}}_2, ..., \mathbf{f}^{\mathbf{c}}_M, \mathbf{c}, \Theta)\right] \\
    \Leftrightarrow \Theta^{*} &= \arg\min_{\Theta} \mathbb{E}_{\mathbf{c}, \{\mathbf{f}^{\mathbf{c}}_{i}\}_{i=1}^M} \left[-\log p(\mathbf{\Delta}^{\mathbf{c}}_1, \mathbf{\Delta}^{\mathbf{c}}_2, ..., \mathbf{\Delta}^{\mathbf{c}}_M, \mathbf{c}, \Theta)\right]
\end{split}
\end{equation}
where $\Delta^{\mathbf{c}}_i = \mathbf{f}^{\mathbf{c}}_i - \mathbf{c}$.
\end{tcolorbox}

\textbf{Remark \ref{rmk:visual_grammar}} defines the visual grammar of the cluster by modeling the feature distribution of $\mathbf{f}_i^{\mathbf{c}}$ and its cluster center $\mathbf{c}$.
Let $\phi: \mathbb{R}^{(M + 1) \times D} \to [0, 1]^{M}$ be a function receiving a center $\mathbf{c}$ and a set of $M$ features $\{\mathbf{f}_i\}_{i=1}^M$ $\left(\cos(\mathbf{f}_i, \mathbf{c}) \geq \cos(\mathbf{f}_{i+1}, \mathbf{c})\right)$ to determine whether $\mathbf{f}_i$ belonging to $\mathbf{c}$, i.e., 
$\mathbf{u} = \phi(\mathbf{\Delta}_1, \mathbf{\Delta}_2 , ..., \mathbf{\Delta}_{M}, \mathbf{c})$  
where $\mathbf{\Delta}_i = \mathbf{f}_i - \mathbf{c}$, $\mathbf{u} = [u_1, u_2, ..., u_M]$ and $u_i = 1$ denotes $\mathbf{f}_i$ belong to cluster $\mathbf{c}$ and vice versa. 
Hence, the visual grammar model in Eqn.~\eqref{eqn:falcon-visual_grammar} can be modeled by the network $\phi$ with parameter $\Theta$ as in Eqn.~\eqref{eqn:falcon-visual_grammar_phi}.
\begin{equation} \label{eqn:falcon-visual_grammar_phi}
\small
\begin{split}
    \Theta^{*} &= \arg\min_{\Theta} \mathbb{E}_{\mathbf{c}, \{\mathbf{f}_{i}\}_{i=1}^M} \left[-\log p(\mathbf{u} | \mathbf{\Delta}_1, \mathbf{\Delta}_2, ..., \mathbf{\Delta}_M, \mathbf{c}, \Theta)\right]
\end{split}
\end{equation}

Eqn.~\eqref{eqn:falcon-visual_grammar_phi} aims to model the distribution of features around its cluster by learning the correlation of relatively topological structures $\mathbf{\Delta}_i$ of features $\mathbf{f}_i$ around cluster $\mathbf{c}$.
Then, based on knowledge of the cluster distribution, the model $\phi$ is able to determine whether a feature $\mathbf{f}_i$ belongs to cluster $\mathbf{c}$.
Hence, it is essential that the model $\phi$ has the ability to exploit the correlation between features $\mathbf{f}_i$ and cluster $\mathbf{c}$ to learn the topological structure of visual grammar. 
Therefore, 
we adopt the self-attention mechanism~\cite{vaswani2017attention, nguyen2021clusformer} to efficiently model these feature correlations.
Particularly, the model $\phi$ is formed by $L_{\phi}$ blocks of self-attention as in Eqn.~\eqref{eqn:falcon-attnetion_grammar_model}.
\begin{equation}\label{eqn:falcon-attnetion_grammar_model}
\small
\begin{split}
    \mathbf{z}_0 &= \operatorname{LN([\mathbf{\Delta}_1, ..., \mathbf{\Delta}_M, \mathbf{c}])} + \boldsymbol{\beta}, \quad
    \mathbf{a}_l = \mathbf{z}_l + \operatorname{MHSA}(\mathbf{z})) \\ 
    \mathbf{z}_{l+1} &= \mathbf{a}_l + \operatorname{MLP}(\operatorname{LN}(\mathbf{a}_l)),  \quad\quad\quad\quad \mathbf{u} = \operatorname{Proj}(\mathbf{z}_{L_{\phi}})
\end{split}
\end{equation}
where $\boldsymbol{\beta}$ is the positional embedding, $\operatorname{LN}$ is Layer Normalization, $\operatorname{MHSA}$ is multi-head self-attention, $\operatorname{MLP}$ is the multi-layer perception, and $\operatorname{Proj}$ is the linear projection.
By using Transformers, the correlation of cluster distributions can be well modeled by the self-attention mechanism.

\noindent
\textbf{Cluster Assignment via Visual Grammar.} 
Instead of assigning the clusters based on the model prediction~\cite{douillard2021plop, cermelli2023comformer, cswkd_cvpr_2022} or nearest cluster~\cite{truong2023fairness, joseph2021towards} that are less effective,
the cluster assignment in our approach will be performed by the visual grammar model, i.e., the visual grammar model will consider the $M$ closest features around cluster $\mathbf{c}$ to assign the cluster for these features.
Then, the cluster assignments are used to compute our Fairness Contrastive Clustering loss. 
In addition, following common practices~\cite{douillard2021plop, cermelli2023comformer, cswkd_cvpr_2022}, we improve background shift modeling by using 
the cluster assignments of features as the pseudo labels of pixels.
Theoretically, although there is a possibility that a feature could not be assigned to a cluster via the visual grammar model, we have empirically observed that this issue rarely happens in our approach. 
Indeed, since we initialize the known clusters via the DB-SCAN, it guarantees that for each feature, there is at least one cluster nearby that the feature representation should belong to.
However, to preserve the integrity of our approach, for the outlier features in cases that cannot be assigned clusters via the visual grammar model, 
these outliers will be heuristically assigned to their closest clusters as similar to~\cite{joseph2021towards, truong2023fairness}.

\noindent
\textbf{Unknown Cluster Initialization.} Prior work~\cite{truong2023fairness, joseph2021towards} initialized a single unknown cluster ($N_U=1$), thus resulting in producing non-discriminative class-wise features.
However, there should be more than a single unknown cluster ($N_U>1$) to produce discriminative features for different unknown classes. 
Therefore, our approach first initializes a list of potential unknown clusters at each learning step via DB-SCAN~\cite{ester1996density} on the features of unknown classes extracted by the current CSS model.
In addition, to reduce the noise clusters and isolated clusters, we also merge several close clusters, i.e., if the distance between two clusters is less than the margin $2\nabla$, these will be merged into a single cluster where the new cluster center will be the means of these two merging cluster centers.
By empirical observation, we have noticed that the number of unknown clusters initialized at each learning step, i.e., $N_U$ at the current learning step $t$, is not greater than 1.5$\times$ times of the remaining classes (i.e., $|\mathcal{C}^{t+1..T}|$) in the dataset, e.g., in our ADE20K 100-50 experiments, at the first learning step of $100$ classes, there are $68$ unknown clusters that have been initialized while there are $50$ remaining unknown classes in the dataset. 
For the new known class $\mathcal{C}^{t}$, we initialize these clusters based on the mean of their feature representations. Meanwhile, the clusters of known classes learned in previous steps are maintained.

\subsubsection{Continual Learning Procedure}

Figure~\ref{fig:falcon-css_framework} illustrates the training procedure of our continual learning approach. At each learning step $t$, the CSS model $F$ with $\theta_t$ is trained with the \textbf{\textit{Fairness Contrastive Clustering}} loss defined in Eqn.~\eqref{eqn:falcon-loss_for_one_cluster_alpha} and the previous visual grammar model $\phi$ with $\Theta_{t-1}$. In addition, we introduce a cluster regularizer $\mathcal{R}_{C}$ to avoid the clusters of different classes collapsing into a single cluster. Therefore, the entire CSS learning objective in our approach can be formed as in Eqn.~\eqref{eqn:falcon-final-falcon-training}.
\begin{equation}\label{eqn:falcon-final-falcon-training}
\small
\begin{split}
    \arg\min_{\theta_t} \mathbb{E}_{\mathbf{x}^t, \mathbf{\hat{y}}^t} \Big[\mathcal{L}_{CE}\left(\mathbf{y}^t,  \mathbf{\hat{y}}^t\right) + \lambda_{CL}\sum_{\mathbf{c}_i}\mathcal{L}^{\alpha}_{Cont}\left(\mathbf{F}^t, \mathbf{c}_i \right)  + \lambda_{C}\mathcal{R}_{C}(\mathbf{c})\Big]
\end{split}
\end{equation}
where $\mathcal{R}_{C}(\mathbf{c}) = \sum_{\mathbf{c}_i, \mathbf{c}_j}\{\max(0, 2\nabla - ||\mathbf{c}_i -\mathbf{c}_j||)\}^2$ is the regularizer to avoid the cluster collapsing,
$\lambda_C$ is the balanced weight, 
and $\nabla$ is the margin between clusters. 

\noindent
\textbf{Training Procedure of Visual Grammar Model.} At CSS learning step $t$, we adopt the visual grammar model trained on the previous learning step, i.e., $\phi$ with $\Theta_{t-1}$, to perform the cluster assignment for the contrastive clustering loss defined in Eqn.~\eqref{eqn:falcon-general_clustering}.
Then, the visual grammar model at learning step $t$, i.e., $\phi$ with $\Theta_t$, will be learned (initialized from $\Theta_{t-1}$) on the features extracted from the dataset and the set of known clusters $\mathbf{c}$ up to the current learning step. 
Following~\cite{nguyen2021clusformer}, we sample a center $\mathbf{c}$ from the known clusters and its $M$ closest features %
to train the visual grammar model. 

\noindent
\textbf{Initial Visual Grammar Model.} At the first learning step $t=1$, since no clusters have been learned at initial, the visual grammar model $\phi$ with $\Theta_{0}$ is not available.
However, as common practices in CSS~\cite{douillard2021plop, cermelli2023comformer, ssul_neurips_2021}, the segmentation model is typically trained from a pre-trained backbone on ImageNet~\cite{deng2009imagenet}.
As a result, the features extracted at the first learning step are characterized by the ImageNet features. Therefore, we adopt this philosophy to initialize our visual grammar model ($\phi$ with $\Theta_0$) by pre-training the visual grammar model on the ImageNet dataset.
Then, during CSS training, we will progressively train our visual grammar model at each learning step as aforementioned.

\subsection{Experimental Results}

\subsubsection{Implementations and Evaluation Protocols}

\input{Tables/chap-4/falcon/ablation_fairness_loss_and_grammar_model}

\noindent
\textbf{Implementation.}
Following common practices~\cite{douillard2021plop, truong2023fairness, cermelli2020modelingthebackground}, we adopt DeepLab-V3~\cite{chen2017rethinking} with ResNet-101~\cite{resnet} and SegFormer~\cite{xie2021segformer} with MiT-B3~\cite{xie2021segformer} in our experiments.
For the Visual Grammar model, we adopt the design of~\cite{nguyen2021clusformer} with $L_{\phi}=12$ blocks of multi-head self-attention layers.
The feature vectors from the last layer of the decoder are used for our $\mathcal{L}^{\alpha}_{Cont}$ loss. 
The value $\alpha$ is set individually for each dataset, i.e., $\alpha = 5\times10^{-2}$ for ADE20K, $\alpha = 10^{-2}$ for VOC for Cityscapes.
To update the cluster vectors $\mathbf{c}$, following prior work~\cite{truong2023fairness, joseph2021towards, he2020momentum}, we maintain a set of $500$ features for each cluster and update the clusters after $K = 100$ steps with a momentum $\eta = 0.99$. 
In our domain incremental experiments, all clusters are updated at each learning step by momentum update.
The number of features selected for each cluster in the visual grammar model is set to $M = 128$.
The balanced weights of CSS objective $\lambda_{CL}$ and $\lambda_{C}$ are set to $1$, and the margin $\nabla$ is set to 10.

\input{Tables/chap-4/falcon/ablation_scaling_and_loss_contribution}

\noindent
\textbf{Evaluation Protocols.} 
Following common practices~\cite{truong2023fairness, cermelli2023comformer, ssul_neurips_2021}, our experiments are conducted on the overlapped CSS settings, including  ADE20K 100-50 (2 steps), ADE20K 100-10 (6 steps), and ADE20K 100-5 (11 steps), VOC 15-5 (2 steps), VOC 15-1 (6 steps), and VOC 10-1 (11 steps).
On Cityscapes, we conduct domain incremental experiments with three settings, i.e., Cityscapes 11-5 (3 steps), Cityscapes 11-1 (11 steps), and Cityscapes 1-1 (21 steps).
Following~\cite{douillard2021plop, cermelli2023comformer}, the mean Intersection over Union (mIoU) metric is adopted in our comparison, including mIoU of the last learning step on initial classes, incremental classes, and all classes. 
In addition, to illustrate the fairness improvement, we report the mIoU of major and minor classes.

\subsubsection{Ablation Study}

\begin{wrapfigure}{r}{0.5\textwidth}
    \centering
    \includegraphics[width=0.5\textwidth]{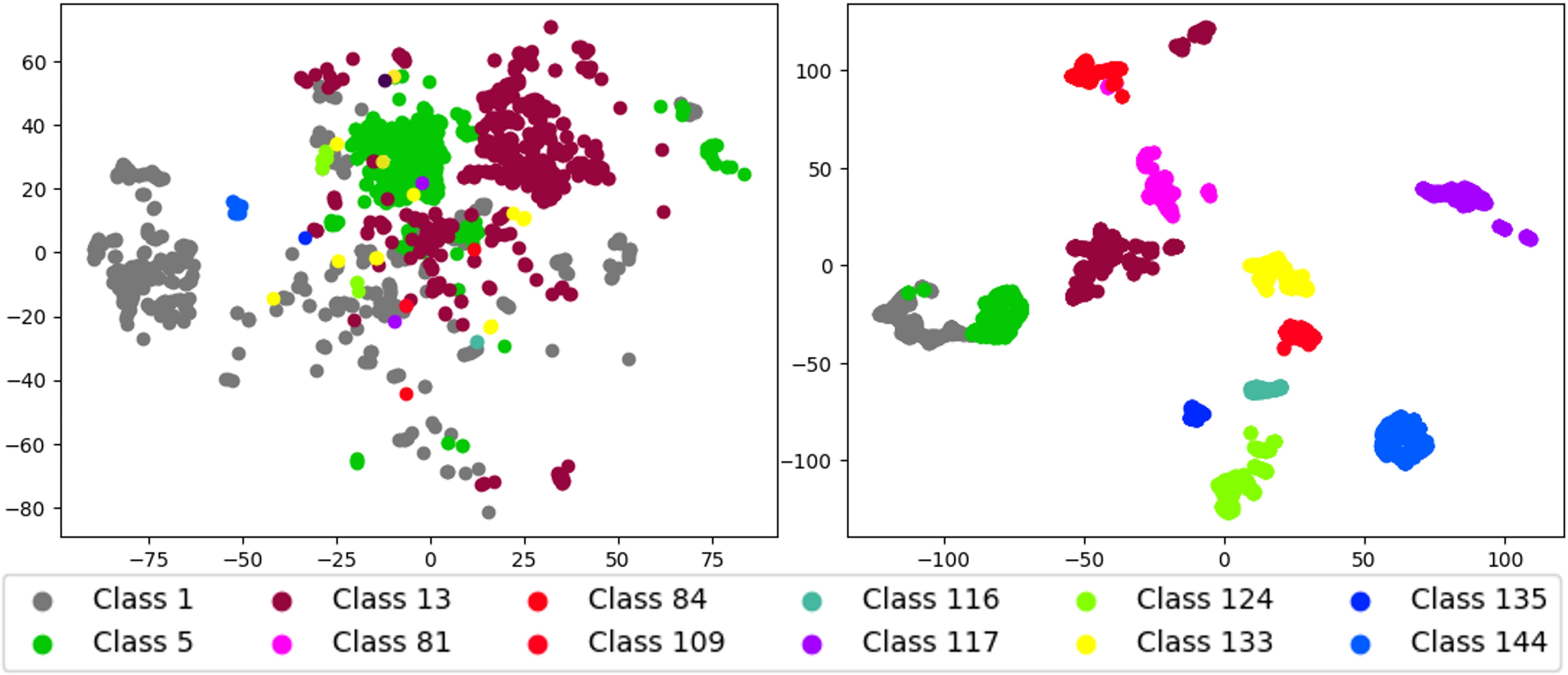}
    \caption{Cluster Distribution at Learning Step $t=1$ of ADE20K 100-50 (classes 109-144 are future classes) without (left) and with (right) Our Fairness Learning.
    }
    \label{fig:falcon-unknown_class_distribution}
\end{wrapfigure}
\noindent
\textbf{Effectiveness of Fairness Contrastive Clustering.}
Table~\ref{tab:falcon-abl_fairness_loss} presents our results using DeepLab-V3~\cite{chen2018deeplab} with Resnet101 on ADE20K 100-50 and ADE20K 100-10 benchmarks.
We evaluate the impact of the fairness contrastive clustering loss $\mathcal{L}^{\alpha}_{Cont}$ by comparing it with the vanilla contrastive clustering loss $\mathcal{L}_{Cont}$.
As shown in our results, the overall performance has been significantly improved to $37.9\%$ and $36.4\%$ on ADE20K 100-50 and ADE20K 100-10, respectively.
The fairness of the model has also been promoted since the mIoU performance of major and minor groups was enhanced.

\input{Tables/chap-4/falcon/ablation_margin}
\noindent
\textbf{Effectiveness of Scaling Factor of Cluster.}
Table~\ref{tab:falcon-abl_alpha} illustrates the experimental results of the impact of different scaling factor $\alpha$ on ADE20K 100-50 and Pascal VOC 15-5 benchmarks.
As shown in Table~\ref{tab:falcon-abl_alpha}, when the value of scaling factor $\alpha$ gradually decreases, the performance of our proposed approach is improved accordingly since the fairness contrastive loss in Eqn \eqref{eqn:falcon-loss_for_one_cluster_alpha} tends to be more uniform across major and minor classes.
However, when the scaling factor is too small ($\alpha = 0.005$), the impact of the loss enforcement becomes weaker leading to the weaker enforcement of the fairness contrastive clustering, resulting in lower overall performance. 
In addition, we have observed that the higher the number of classes demands the higher the value of $\alpha$ since it will increase the compactness of more clusters.

\noindent
\textbf{Effectiveness of Loss Contributions.}
\noindent
Table~\ref{tab:falcon-abl_loss_contribute} illustrates the contributions of proposed learning objectives.
For the model without using visual grammar, we only use a single unknown cluster ($N_U = 1$) and adopt the nearest cluster strategies to assign clusters of unknown pixels.
By using only cross-entropy loss, the mIoU performance remains low due to catastrophic forgetting and background shift problems.
Meanwhile, with our fairness clustering loss $\mathcal{L}^{\alpha}_{Cont}$, visual grammar model $\phi$, and the cluster regularizer $\mathcal{R}$, the mIoU performance has been significantly improved to $37.9\%$ and $36.4\%$ on ADE20K 100-50 and ADE20K 100-10, respectively.
Moreover, FALCON has significantly promoted the fairness of segmentation models illustrated by the mIoU improvement of major and minor groups.

\input{Tables/chap-4/falcon/ablation_m_grammar_model}

\noindent
\textbf{Effectiveness of Visual Grammar.}
We evaluate FALCON under three settings, i.e., Nearest Cluster, Fixed $\phi$ pretrained ImageNet (without updating on each learning step), and Adaptive $\phi$ (with updating on each learning step).
As in Table~\ref{tab:falcon-abl_visual_grammar}, the mIoU result using only the nearest cluster remains ineffective.
Meanwhile, the adaptive visual grammar model updated at each learning step further boosts the mIoU performance and promotes fairness, i.e., increased by $4.5\%$ and $4.9\%$ on ADE20K 100-50 and ADE20K 100-10 compared to the nearest cluster approach.
Figure~\ref{fig:falcon-unknown_class_distribution} illustrates the feature distributions of unknown classes (future class).
As a result, our FALCON approach is able to model features of unknown classes into different clusters and produce better and more compact clusters compared to the one without our fairness learning.

\noindent
\textbf{Effectiveness of Choosing Margin $\nabla$.}
Table~\ref{tab:falcon-abl_margin} studies the effectiveness of the value of margin $\nabla$ to the performance of our approach.
As shown in the results, the change of $\nabla$ also slightly influences the performance of the model. Since the margin defines the distance between two clusters, while the smaller value of the margin $\nabla$ could cause the incorrect cluster assignment of the features, the larger value of the margin $\nabla$ could produce the less compact clusters.

\noindent
\textbf{Effectiveness of Choosing Number of Features $M$.}
We study the impact of choosing the number of features $M$ in the visual grammar model. As in shown Table~\ref{tab:falcon-abl_visual_grammar_m}, the optimal performance of our approach is $M = 128$.
When the number of features selected is small ($M = 96$), it does not have enough number of features to form the visual grammar so the model is hard to exploit the correlation among features and the cluster. 
Meanwhile, when we increase the number of selected features ($M = 256$), the clusters will consist of many outlier features (the ones that do not belong to the cluster), thus being challenging for the visual grammar model to exploit the topological structures of the feature distribution.

\input{Tables/chap-4/falcon/ablation_different_seg} 
\noindent
\textbf{Effectiveness of Different Segmentation Networks.}
To illustrate the flexibility of our proposed approach, we evaluate our proposed approach with different network backbones. Table~\ref{tab:falcon-abl_different_backbones} illustrates the results of our approach using DeepLab-V3~\cite{chen2018deeplab}, SegFormer~\cite{xie2021segformer} with different backbones, i.e., ResNet-50, ResNet-101, MiT-B2, and MiT-B3.
As shown in the performance, the more powerful the segmentation model is, the better performance of the model is.

\noindent
In particular, our approach has shown its flexibility since it consistently improves the performance of the segmentation model and achieves the SOTA performance on two different benchmarks, i.e., the performance of Transformer models achieves $41.9\%$, and $40.3\%$ on ADE20K 100-50, ADE20K 100-10, respectively.

\input{Tables/chap-4/falcon/ade20k}

\subsubsection{Comparison with Prior SOTA Methods}

\begin{wrapfigure}{r}{0.5\textwidth}
    \centering
    \includegraphics[width=0.5\textwidth]{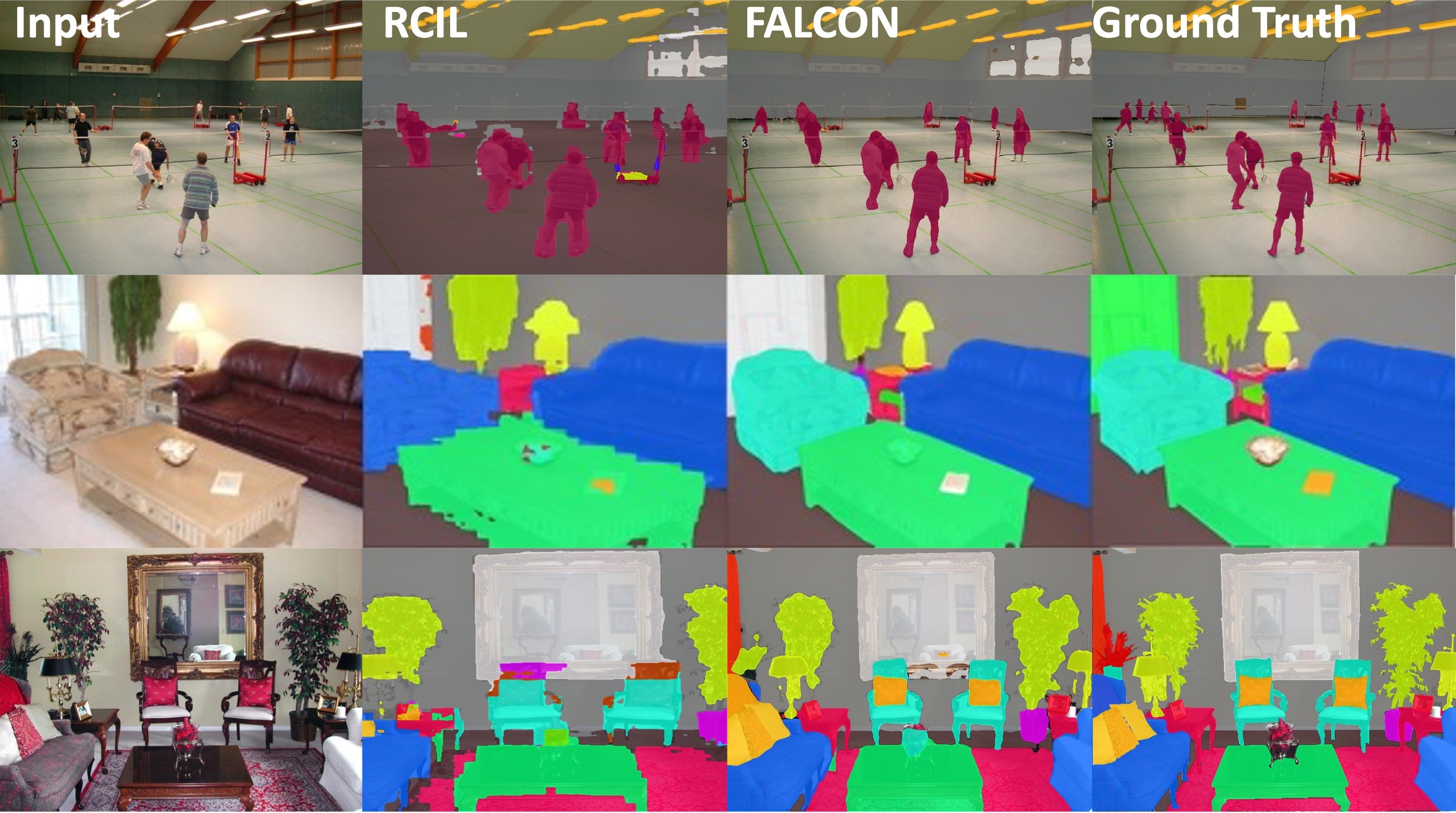}
    \caption{Our Results on ADE20K.}
    \label{fig:falcon-visualization}
\end{wrapfigure}

\noindent
\textbf{ADE20K.}
Table~\ref{tab:falcon-ade20k} presents our experimental results using DeepLab-V3 and Transformer networks compared to prior CSS methods. Overall, our proposed approach has achieved the SOTA performance compared to prior methods.
In particular, by using DeepLab-V3, 
our approach has achieved SOTA performance, i.e., the mIoU results of $37.9\%$ and $+36.4\%$ on ADE20K 100-50 and ADE20K 100-10 benchmarks, higher than prior FairCL~\cite{truong2023fairness}. 
Meanwhile, our approach using Transformer has outperformed the prior SOTA CoMFormer~\cite{cermelli2023comformer} model by $+3.5\%$, $+8.0\%$, and $+2.6\%$ on ADE20K 100-50, ADE20K 100-10, and ADE20K 100-5 respectively.
In addition, our mIoU results on the initial classes remain competitive with the upper-bounded results because our method is able to well handle the fairness problem compared to the fully supervised learning approach.
As in Figure~\ref{fig:falcon-visualization}, FALCON produces better segmentation maps compared to prior methods.

\input{Tables/chap-4/falcon/voc_city}

\noindent
\textbf{Pascal VOC.}
Table~\ref{tab:falcon-voc} presents our results on Pascal VOC benchmarks.
Our proposed approach has consistently achieved the SOTA performance on three benchmarks. In particular, compared to the prior FairCL~\cite{truong2023fairness} approach, our methods using DeepLab-V3 have improved the mIoU performance up to $73.50\%$, $69.83\%$, and $62.41\%$ on  Pascal VOC 15-5, Pascal VOC 15-1, and Pascal VOC 10-1, respectively. Additionally, by using the better network backbone, i.e., Transformer, the performance of the segmentation model is also further improved
and reduced the gap with the supervised result.

\noindent
\textbf{Cityscapes.}
Table~\ref{tab:falcon-cityscape} reports the performance of our approach using DeepLab-V3 compared to prior methods on three different settings of Cityscapes benchmarks, i.e., Cityscapes 11-5, Cityscapes 11-1, and Cityscapes 1-1.
As shown in the experimental results, the performance of our methods has consistently outperformed prior FairCL~\cite{truong2023fairness} approach by $+3.78\%$, $+3.14\%$, and $+6.02\%$ on three benchmarks. Similar to our experiments on ADE20K and VOC, the better network brings higher results.

%% file: Tables/chap-4/falcon/ablation_fairness_loss_and_grammar_model.tex
\begin{table}[!b]
    \begin{minipage}[t]{.49\textwidth}
        \centering
        \caption{Effectiveness of Fairness Contrastive Learning Loss.} \label{tab:falcon-abl_fairness_loss}
        \resizebox{1.0\textwidth}{!}{  
        \begin{tabular}{|c  c | c c c c c |}
        \hline
        \multicolumn{7}{|c|}{(a) ADE20K 100-50}                 \\
        
        \hline
        $\mathcal{L}_{Cont}$ & $\mathcal{L}^{\alpha}_{Cont}$ & 0-100 & 100-150 & all  & Major & Minor \\
        \hline
        
        \cmark &    & 44.6  & 15.2    & 34.8 & 51.5  & 26.4  \\
         & \cmark   & \textbf{44.6} & \textbf{24.5} & \textbf{37.9} & \textbf{52.1} & \textbf{30.8} \\
        
        \hline
        \multicolumn{7}{|c|}{(b) ADE20K 100-10}                 \\
        \hline
        
        $\mathcal{L}_{Cont}$ & $\mathcal{L}^{\alpha}_{Cont}$ & 0-100 & 100-150 & all  & Major & Minor \\
        
        \hline
        
        \cmark & &  41.9  & 16.0    & 33.2 & 49.9  & 24.9  \\
        & \cmark  &  \textbf{44.4}  & \textbf{20.4}    & \textbf{36.4} & \textbf{51.8}  & \textbf{28.7}  \\
        
        \hline
        \end{tabular}
        }
    \end{minipage}%
    \hfill
    \begin{minipage}[t]{.49\textwidth}
        \centering
        \caption{Effectiveness of Visual Grammar.} \label{tab:falcon-abl_visual_grammar}
        \resizebox{1.0\textwidth}{!}{  
        \begin{tabular}{|l| c c c c c |}
        \hline
        \multicolumn{6}{|c|}{(a) ADE20K 100-50}                    \\
        \hline
                        & 0-100 & 101-150 & all  & Major & Minor \\
        \hline
        
        Nearest Cluster & 44.3	& 11.5	& 33.4	& 51.5	& 24.3 \\
        Fixed $\phi$       & 44.6  & 17.6    & 35.6 & 52.0  & 27.4  \\
        Adaptive $\phi$     & \textbf{44.6} & \textbf{24.5} & \textbf{37.9} & \textbf{52.1} & \textbf{30.8} \\
        
        \hline
        \multicolumn{6}{|c|}{(b) ADE20K 100-10}                    \\
        \hline
        
                        & 0-100 & 101-150 & all  & Major & Minor \\
        \hline

        Nearest Cluster & 40.1	& 14.3	& 31.5	& 48.7	& 22.9 \\
        Fixed $\phi$       & 43.0  & 18.5    & 34.9 & 50.6  & 27.0  \\
        Adaptive $\phi$     &  \textbf{44.4}  & \textbf{20.4}    & \textbf{36.4} & \textbf{51.8}  & \textbf{28.7}  \\
        \hline
        \end{tabular}
        }
    \end{minipage}
\end{table}

%% file: Tables/chap-4/falcon/ablation_scaling_and_loss_contribution.tex
\begin{table}[!b]
    \begin{minipage}[t]{.49\textwidth}
        \centering
        \caption{Effectiveness of Scaling Factor $\alpha$.} \label{tab:falcon-abl_alpha}
        \resizebox{1.0\textwidth}{!}
        {  
        \begin{tabular}{|l| c c c c c |} %
        \hline
        \multicolumn{6}{|c|}{(a) ADE20K 100-50} \\
        \hline
        $\alpha$  & 0-15 & 16-20 & all & Major & Minor \\ 
        \hline
        $\alpha = 0.1$    & 43.1 & 19.8 & 35.3 & 50.6 & 27.7 \\
        $\alpha = 0.05$   & \textbf{44.6} & \textbf{24.5} & \textbf{37.9} & \textbf{52.1} & \textbf{30.8} \\
        $\alpha = 0.01$   & 43.6 & 21.3 & 36.2 & 51.0 & 28.7 \\
        $\alpha = 0.005$  & 42.4 & 18.6 & 34.5 & 50.1 & 26.6 \\
        \hline
        \multicolumn{6}{|c|}{(b) Pascal VOC 15-5} \\
        \hline
        $\alpha$  & 0-15 & 16-20 & all & Major & Minor \\ 
        \hline
        $\alpha = 0.1$    & 74.8 & 51.6 & 69.3 & 76.9 & 63.5 \\
        $\alpha = 0.05$   & 76.2 & 51.3 & 70.3 & 79.0 & 63.8 \\
        $\alpha = 0.01$    & \textbf{79.4} & \textbf{54.8} & \textbf{73.5} & \textbf{81.3} & \textbf{67.7} \\
        $\alpha = 0.005 $ & 74.6 & 48.9 & 68.5 & 77.6 & 61.6 \\
        \hline
        \end{tabular}
        }
    \end{minipage}%
    \hfill
    \begin{minipage}[t]{.49\textwidth}
        \centering
        \caption{Effectiveness of Our Proposed Losses.}
        \label{tab:falcon-abl_loss_contribute}
        \setlength{\tabcolsep}{1.9pt}
        \resizebox{1.0\textwidth}{!}{  
        \begin{tabular}{|cccc|ccccc|}
        \hline
        \multicolumn{9}{|c|}{(a) ADE20K 100-50}                                                                                       \\
        \hline
        $\mathcal{L}_{CE}$                 & $\mathcal{L}^{\alpha}_{Cont}$ & $\phi$   & $\mathcal{R}_C$ & 0-100 & 101-150 & all  & Major & Minor \\
        \hline
        \cmark &        &        & & 0.0	& 18.9	& 6.3	& 0.0	& 9.4 \\
        \cmark & \cmark &        & & 44.0  & 7.9     & 31.9 & 51.6  & 22.1  \\
        \cmark & \cmark & \cmark & & 43.8 & 21.8 & 36.4 & 51.1 & 29.1 \\
        \cmark & \cmark & \cmark & \cmark & \textbf{44.6} & \textbf{24.5} & \textbf{37.9} & \textbf{52.1} & \textbf{30.8} \\
        \hline
        \multicolumn{9}{|c|}{(b) ADE20K 100-10}                                                                                       \\
        \hline
        $\mathcal{L}_{CE}$                 & $\mathcal{L}^{\alpha}_{Cont}$ & $\phi$ & $\mathcal{R}_C$ & 0-100 & 101-150 & all  & Major & Minor \\
        \hline
        \cmark &        &        & & 0.0	& 3.5	& 1.2	& 0.0	& 1.8 \\
        \cmark & \cmark &        &  & 39.0  & 13.1    & 30.4 & 47.8  & 21.6  \\
        \cmark & \cmark & \cmark & & 43.4 & 18.5 & 35.1 & 51.2 & 27.1 \\ 
        \cmark & \cmark & \cmark & \cmark & \textbf{44.4}  & \textbf{20.4}    & \textbf{36.4} & \textbf{51.8}  & \textbf{28.7}  \\
        \hline
        \end{tabular}
        }
    \end{minipage}
\end{table}

%% file: Tables/chap-4/falcon/ablation_margin.tex
\begin{wraptable}{r}{0.5\textwidth}
\centering
\caption{Effectiveness of Choosing Margin $\nabla$.} 
\label{tab:falcon-abl_margin}
\resizebox{0.5\textwidth}{!}{
\begin{tabular}{|l| c c c c c |}
\hline
\multicolumn{6}{|c|}{(a) ADE20K 100-50}                    \\
\hline
                & 0-100 & 101-150 & all  & Major & Minor \\
\hline
$\nabla=5$     &  44.4 & 21.8 & 36.9 & 51.9 & 29.4 \\
$\nabla=10$     & \textbf{44.6} & \textbf{24.5} & \textbf{37.9} & \textbf{52.1} & \textbf{30.8} \\
$\nabla=20$     &  44.7 & 22.2 & 37.2 & 51.7 & 29.9  \\
\hline
\multicolumn{6}{|c|}{(b) ADE20K 100-10}                    \\
\hline

                & 0-100 & 101-150 & all  & Major & Minor \\
\hline
$\nabla=5$     &  43.2 & 18.7 & 35.0 & 50.5 & 27.3 \\
$\nabla=10$     &  \textbf{44.4}  & \textbf{20.4}    & \textbf{36.4} & \textbf{51.8}  & \textbf{28.7}  \\
$\nabla=20$     &  43.5 & 19.9 & 35.7 & 51.2 & 27.9 \\
\hline
\end{tabular}
}
\end{wraptable}

%% file: Tables/chap-4/falcon/ablation_m_grammar_model.tex
\begin{wraptable}{r}{0.5\textwidth}
\centering
\caption{Effectiveness of Number of Features $M$ in a Cluster of Visual Grammar Model.} \label{tab:falcon-abl_visual_grammar_m}
\resizebox{0.5\textwidth}{!}{
\begin{tabular}{|l| c c c c c |}
\hline
\multicolumn{6}{|c|}{(a) ADE20K 100-50}                    \\
\hline
                & 0-100 & 101-150 & all  & Major & Minor \\
\hline
$M=96$     &  43.0 &	19.6 &	35.2 & 50.5 & 27.5  \\
$M=128$     & \textbf{44.6} & \textbf{24.5} & \textbf{37.9} & \textbf{52.1} & \textbf{30.8} \\
$M=256$     &  43.6 &	21.6 &	36.3 & 51.0 & 28.9 \\ 
\hline
\multicolumn{6}{|c|}{(b) ADE20K 100-10}                    \\
\hline

                & 0-100 & 101-150 & all  & Major & Minor \\
\hline
$M=96$     &  42.2 &	16.4 &	33.6 & 50.2 & 25.3 \\
$M=128$     &  \textbf{44.4}  & \textbf{20.4}    & \textbf{36.4} & \textbf{51.8}  & \textbf{28.7}  \\
$M=256$     &  42.7 &	17.1 &	34.2 & 50.6 & 26.0 \\
\hline
\end{tabular}
}
\end{wraptable}

%% file: Tables/chap-4/falcon/ablation_different_seg.tex
\begin{wraptable}{r}{0.5\textwidth}
\small
\centering
\caption{Effectiveness of Different Backbones on ADE20K.} \label{tab:falcon-abl_different_backbones}
\resizebox{0.5\textwidth}{!}{
\begin{tabular}{|l | c | c c c c c |}
\hline
\multicolumn{7}{|c|}{(a) ADE20K 100-50}                    \\
\hline
   & Backbone             & 0-100 & 101-150 & all  & Major & Minor \\
\hline

\multirow{2}{*}{DeepLab-V3}  &R-50       &  44.3 & 15.2 & 34.7   & 51.5 & 26.4    \\
 & R-101     & \textbf{44.6} & \textbf{24.5} & \textbf{37.9} & \textbf{52.1} & \textbf{30.8} \\
 \hdashline
\multirow{2}{*}{Transformer} & MiT-B2       & 44.5 & 27.4 & 38.8  & 52.4 & 32.2      \\
 & MiT-B3      & \textbf{47.5} & \textbf{30.6} & \textbf{41.9}  & \textbf{53.8} & \textbf{35.8}   \\

\hline
\multicolumn{7}{|c|}{(b) ADE20K 100-10}                    \\
\hline
    
    & Backbone            & 0-100 & 101-150 & all  & Major & Minor \\
\hline

\multirow{2}{*}{DeepLab-V3} & R-50       &  43.5	& 16.5 &	34.5 & 51.1 & 26.2     \\
 & R-101     &  \textbf{44.4}  & \textbf{20.4}    & \textbf{36.4} & \textbf{51.8}  & \textbf{28.7}  
\\
\hdashline
\multirow{2}{*}{Transformer} & MiT-B2       &  45.4 &	22.7 &	37.8 & 52.6 & 30.4    \\
 & MiT-B3       & \textbf{47.3} & \textbf{26.2} & \textbf{40.3} & \textbf{54.0} & \textbf{33.4}   \\
\hline
\end{tabular}
}
\end{wraptable}

%% file: Tables/chap-4/falcon/ade20k.tex
\begin{table}[!t]
\centering
\caption{Comparison with Prior Methods on ADE20K Benchmarks (Note: The results of MiB \cite{cermelli2020modelingthebackground}, PLOP \cite{douillard2021plop}, and FairCL \cite{truong2023fairness} using Transformer on ADE20K 100-5 were not reported in prior studies. The upper bound results are not trained with fairness objective).}
\label{tab:falcon-ade20k}
\setlength{\tabcolsep}{2pt}
\resizebox{1.0\textwidth}{!}{  
\begin{tabular}{| l | l|c c c c|c c c c|c c c c |}
\hline
\multirow{2}{*}{Network}   & \multirow{2}{*}{Method} & \multicolumn{4}{c|}{ADE20K 100-50}     & \multicolumn{4}{c|}{ADE20K 100-10}& \multicolumn{4}{c}{ADE20K 100-5} \\ \cline{3-14} 
                            &                         & 0-100  & 101-150 & all  & avg  & 0-100  & 101-150 & all  & avg  & 0-100 & 101-150 & all  & avg  \\ 
\hline

\multirow{7}{*}{DeepLab-V3} 
                            & PLOP  \cite{douillard2021plop}                  & 41.9   & 14.9    & 32.9 & 37.4 & 40.5   & 14.1    & 31.6 & 36.6 & 39.1  & 7.8     & 28.8 & 35.3 \\ %
                            & RCIL \cite{zhang2022representation}                   & 42.3   & 18.8    & 34.5 & $-$  & 39.3   & 17.6    & 32.1 & $-$  & 38.5  & 11.5    & 29.6 & $-$  \\ %
                            & REMINDER \cite{cswkd_cvpr_2022} & 41.6 & 19.2 & 34.1 & $-$ & 39.0 & 21.3 & 33.1 & $-$ & 36.1 & 16.4 & 29.5 & $-$ \\
                            & RCIL+LGKD \cite{yang2023label} & 43.3 & 25.1 & 37.2 & $-$ & 42.2 & 20.4 & 34.9 & $-$ & $-$ & $-$ & $-$ & $-$ \\
                            & FairCL \cite{truong2023fairness}                 & 43.4   & {24.0}    & 37.0 & 40.5 & 41.7   & \textbf{20.4}    & 34.7 & 39.0 & $-$   & $-$     & $-$  & $-$  \\ %
                            & \textbf{FALCON} & \textbf{44.6}	& \textbf{24.5}	& \textbf{37.9} & \textbf{41.3}  & \textbf{44.4}   & \textbf{20.4}    & \textbf{36.4} &  \textbf{40.1}    & \textbf{38.0}  & \textbf{16.1}    & \textbf{30.7} & \textbf{37.6}  \\ 
                            \cdashline{2-14} 
                            & Upper Bound & 44.3 & 28.2 & 38.9 & $-$ & 44.3 & 28.2 & 38.9 & $-$ & 44.3 & 28.2 & 38.9 & $-$ \\
                            \hline

 & MiB \cite{cermelli2020modelingthebackground}                    & 37.0   & 24.1    & 32.6 & 38.3 & 23.5   & 10.6    & 26.6 & 29.6 & 21.0  & 6.1     & 16.1 & 27.7 \\ %
Mask2Former & PLOP \cite{douillard2021plop}                   & 44.2   & 26.2    & 38.2 & 41.1 & 34.8   & 15.9    & 28.5 & 35.2 & 33.6  & 14.1    & 27.1 & 33.6 \\ %
 & CoMFormer \cite{cermelli2023comformer}               & 44.7   & 26.2    & 38.4 & 41.2 & 40.6   & 15.6    & 32.3 & 37.4 & 39.5  & 13.6    & 30.9 & 36.5 \\ %
 \hline 
 
 & MiB \cite{cermelli2020modelingthebackground} &  43.4 & \textbf{30.6} & 39.2 & 38.7 & 39.1 & 20.4 & 34.2 & 39.5 & $-$ & $-$ & $-$ & $-$ \\
 & PLOP \cite{douillard2021plop} & 43.8 & 26.2 & 38.0 & 38.1 & 43.3 & 24.1 & 36.2 & 40.3 & $-$ & $-$ & $-$ & $-$ \\
Transformer & FairCL \cite{truong2023fairness}                 & 43.6 & 25.5 & 37.6 & 40.7 & 42.2 & 21.9 & 35.5 & 39.4 & $-$   & $-$ & $-$  & $-$ \\
 & \textbf{FALCON}                    & \textbf{47.5}   & \textbf{30.6}    & \textbf{41.9} & \textbf{43.5}  & \textbf{47.3}   & \textbf{26.2}    & \textbf{40.3} & \textbf{42.8}  & \textbf{40.8}  & \textbf{18.9}    & \textbf{33.5} & \textbf{38.1}  \\
\cdashline{2-14}
& Upper Bound & 48.7 & 39.0 & 45.5 & $-$ & 48.7 & 39.0 & 45.5 & $-$ & 48.7 & 39.0 & 45.5 & $-$ \\

\hline
\end{tabular}
}
\end{table}

%% file: Tables/chap-4/falcon/voc_city.tex
\begin{table}[!t]
	\begin{minipage}[t]{0.57\textwidth}
	   \centering
\caption{Comparisons with Prior Methods on Pascal VOC.}
\label{tab:falcon-voc}
\setlength{\tabcolsep}{4pt}
\resizebox{1.0\textwidth}{!}{  
\begin{tabular}{|l | l|ccc|ccc|ccc|}
\hline
\multirow{2}{*}{}  & \multirow{2}{*}{Method} & \multicolumn{3}{c|}{Pascal VOC 15-5} & \multicolumn{3}{c|}{Pascal VOC 15-1} & \multicolumn{3}{|c}{Pascal VOC 10-1 } \\
\cline{3-11}
                       & & 0-15       & 16-20     & all       & 0-15       & 16-20     & all       & 0-10       & 11-20      & all       \\
\hline
\multirow{7}{*}{\rot{DeepLab-V3}} & MiB \cite{cermelli2020modelingthebackground}             & 76.37      & 49.97     & 70.08     & 38.00      & 13.50     & 32.20     & 20.00      & 20.10      & 20.10     \\
 & PLOP \cite{douillard2021plop}           & 75.73      & 51.71     & 70.09     & 65.10      & 21.10     & 54.60     & 44.00      & 15.50      & 30.50     \\
& RCIL \cite{zhang2022representation}           & $-$        & $-$       & $-$       & 70.60      & 23.70     & 59.40     & 55.40      & 15.10      & 34.30     \\
& FairCL  \cite{truong2023fairness}     & $-$        & $-$       & $-$       & 72.00      & 22.70     & 60.30     & 42.30      & 25.60      & 34.40     \\
& SSUL  \cite{ssul_neurips_2021}                  & 77.82      & 50.10     & 71.22     & 77.31      & 36.59     & 67.61     & 71.31      & 45.98      & 59.25     \\
& \textbf{FALCON}        & 
\textbf{79.35}      & \textbf{54.77}     & \textbf{73.50}     & \textbf{78.34}      & \textbf{42.57}     & \textbf{69.83}     &   \textbf{73.94}	& \textbf{49.73}	& \textbf{62.41}\\
\cdashline{2-11}
& Upper Bound                    & 79.77      & 72.35     & 77.43     & 79.77      & 72.35     & 77.43     & 78.41      & 76.35      & 77.43    \\
\hline
\multirow{5}{*}{\rot{Transformer}} & PLOP \cite{douillard2021plop} & 72.51 & 48.37 & 66.76 & 64.59 & 37.23 & 58.08 & 48.53 & 33.71 & 41.47 \\
& SSUL \cite{ssul_neurips_2021} & 79.91 & {56.83} & {74.41} & \textbf{79.91} & 40.56 & 70.54 & 74.06 & 51.85 & 63.48 \\
 & FairCL   \cite{truong2023fairness}    & $-$        & $-$       & $-$       & 73.50      & 22.80     & 61.50     & 57.10      & 14.20      & 36.60     \\
 & \textbf{FALCON} & \textbf{81.20} &	\textbf{58.04} &	\textbf{75.69} & {78.71}      & \textbf{47.54}     & \textbf{71.28}     & \textbf{74.92}	& \textbf{52.54}	& \textbf{64.26}     \\
\cdashline{2-11}
& Upper Bound & 80.84 & 74.97 & 79.44 & 80.84 & 74.97 & 79.44 & 80.84 & 74.97 & 79.44 \\ 
\hline
\end{tabular}
}
	   
	\end{minipage} 
	\hspace{0.1mm}
	\begin{minipage}[t]{0.4\textwidth}

 \centering
\caption{Comparisons on Cityscapes.}
\label{tab:falcon-cityscape}
\setlength{\tabcolsep}{5pt}
\resizebox{0.92\textwidth}{!}{  
\begin{tabular}{|l | l |c c c|}
\hline
\multirow{1}{*}{} & \multirow{1}{*}{Method} & 11-5    & 11-1     & 1-1      \\ 
\cline{1-5} 
\hline
\multirow{8}{*}{\rot{DeepLab-V3}} & LWF-MC \cite{rebuffi2017icarl}         & 58.90   & 56.92    & 31.24    \\ %
 & ILT \cite{michieli2019ilt}            & 59.14   & 57.75    & 30.11    \\ %
& MİB \cite{cermelli2020modelingthebackground}             & 61.51   & 60.02    & 42.15    \\ %
 & PLOP \cite{douillard2021plop}           & 63.51   & 62.05    & 45.24    \\ %
 & RCIL \cite{zhang2022representation}           & 64.30   & 63.00    & 48.90    \\ %
& FairCL \cite{truong2023fairness}     & 66.96   & 66.61    & 49.22    \\ %
& \textbf{FALCON}         & \textbf{70.74}   & \textbf{69.75}    & \textbf{55.24}    \\ %
\cdashline{2-5}
& Upper Bound & 79.30 & 79.30 & 79.30 \\
\hline
 \multirow{3}{*}{\rot{Trans.}} & FairCL \cite{truong2023fairness}      & 67.85   & 67.09    & 55.68    \\ %
 & \textbf{FALCON}       & \textbf{71.33}   & \textbf{70.14}    & \textbf{58.79}    \\ 
\cdashline{2-5}
& Upper Bound  & 83.80 & 83.80 & 83.80 \\
\hline
\end{tabular}
}
   	
	\end{minipage}
\end{table}

%% file: Chapters/Chaps/chap-5-cross-view-learning.tex
\chapter{Efficient Geometry-based Approach to Cross-view Learning}\label{chap:cross-view}

Unsupervised Domain Adaptation has been an efficient approach to transferring deep learning-based models across data distributions. 
Meanwhile, the recent open-vocabulary learning approach based on large-scale vision language models is effective in open-set settings because it can learn diverse concepts and categories. 
However, these prior methods fail to generalize across different camera views due to the lack of cross-view geometric modeling. At present, there are limited studies analyzing cross-view learning. 
In this chapter, we introduce a novel Unsupervised Cross-view Adaptation Learning approach to modeling the geometric structural change across views in Semantic Scene Understanding and Video Understanding. 
In particular, we introduce a novel Cross-view Geometric Constraint on Unpaired Data to model structural changes across cameras.
The experiments on different cross-view adaptation benchmarks have shown the effectiveness of our approach in cross-view modeling, demonstrating that we achieve State-of-the-Art (SOTA) performance compared to prior unsupervised domain adaptation and open-vocabulary learning methods.

\input{Chapters/Sections/chap-5/eagle}

\input{Chapters/Sections/chap-5/cvar}

\section{Summary}

In this chapter, we have presented novel approaches to cross-view in visual semantic segmentation and video understanding.
First, this chapter has presented a novel unsupervised cross-view adaptation approach that models the geometric correlation across views.
We have introduced the Geodesic Flow-based metric to better model geometric structural changes across camera views.
In addition, a new view-condition prompting mechanism has been presented to further improve the cross-view modeling.
Second, this chapter has presented a novel approach for cross-view learning in action recognition.  Using our proposed cross-view self-attention loss, our approach has effectively transferred the knowledge learned from the exocentric to the egocentric view. Moreover, our approach does not require pairs of videos across views, which increases the flexibility of our learning approaches in practice. 
Through our theoretical analysis and state-of-the-art performance in standard benchmarks of semantic segmentation and video understanding, our approach has shown its effectiveness in cross-view modeling and improved robustness of deep models across views.

%% file: Chapters/Sections/chap-5/eagle.tex
\section{Efficient Adaptive Geometry-based Learning in Cross-view Understanding}
\label{sec:paper-eagle}

\setcounter{propositioncounter}{0}
\setcounter{remarkcounter}{0}

Modern segmentation models~\cite{chen2018deeplab, chen2017rethinking, xie2021segformer} have achieved remarkable results on the close-set training with a set of pre-defined categories and concepts.
To work towards human-level perception where the scenes are interpreted with diverse categories and concepts, the open-vocabulary (open-vocab) perception model~\cite{qin2023freeseg, rao2021denseclip} based on the power of large vision-language models~\cite{li2022grounded, radford2021learning} has been introduced to address the limitations of close-set training.
By using the power of language as supervision, the large-scale vision language model is able to learn the more powerful representations where languages offer better reasoning mechanisms and open-word concept representations compared to prior methods~\cite{chen2018deeplab, xie2021segformer, cheng2021maskformer}.

\begin{wrapfigure}{r}{0.5\textwidth}
    \centering
    \includegraphics[width=0.5\textwidth]{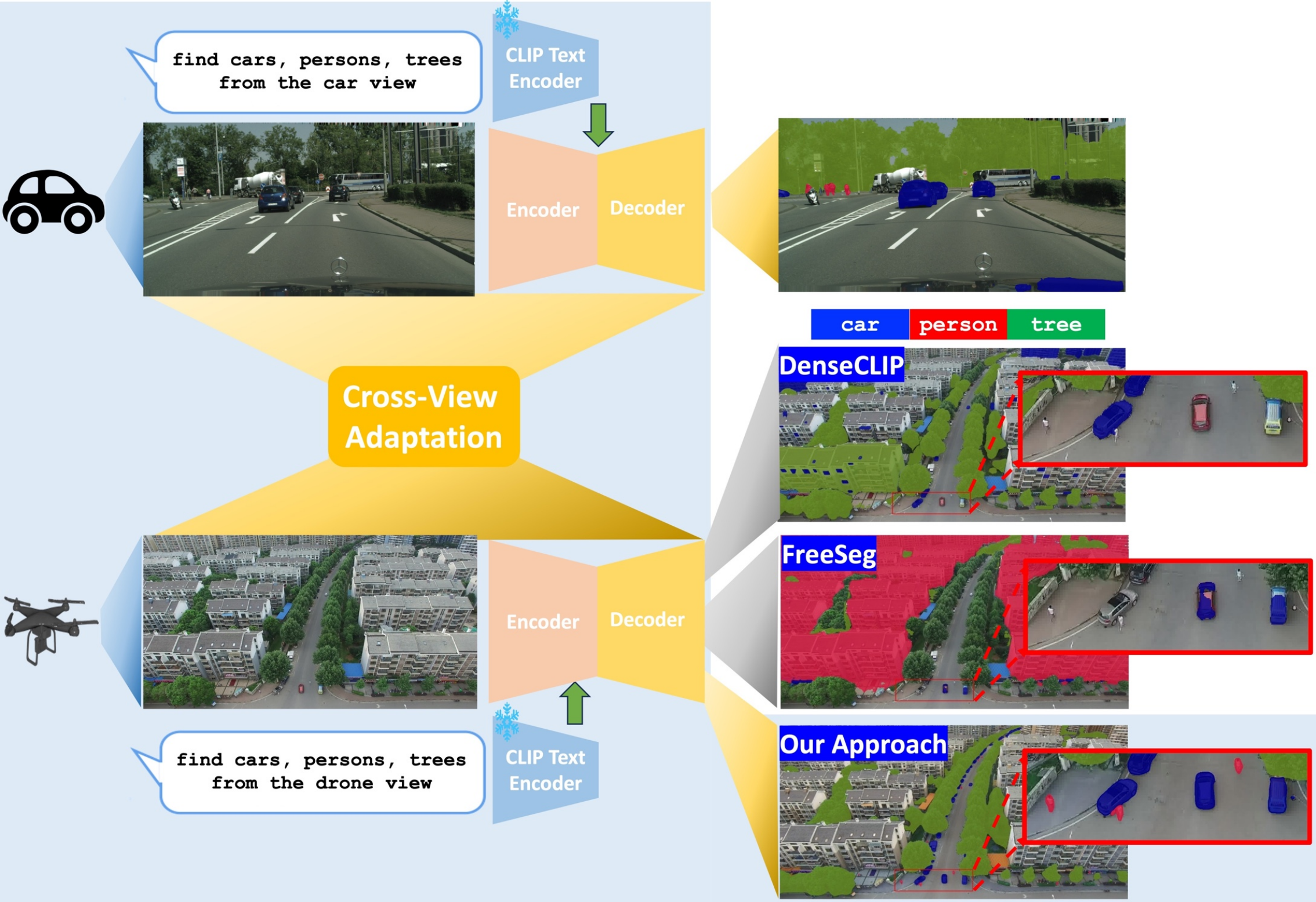}
    \caption{\textbf{\textit{Our Proposed Cross-view Adaptation Learning Approach.}}
    Prior models, e.g., FreeSeg~\cite{qin2023freeseg}, DenseCLIP~\cite{rao2021denseclip}, trained on the car view could not perform well on the drone-view images. Meanwhile, our cross-view adaptation approach is able to generalize well from the car to drone view.}
    \label{fig:eagle-cross_view_prompt_learning}
\end{wrapfigure}
Recent work is inspired by the success of large vision-language models~\cite{radford2021learning, jia2021scaling} that are able to learn informative feature representations of both visual and textual inputs from large-scale image-text pairs.  These have been adopted to further develop open-vocab semantic segmentation models~\cite{qin2023freeseg, rao2021denseclip, liang2023open, li2022languagedriven} that can work well in open-world environments.
However, the open-vocab perception models remain unable to generalize across camera viewpoints.
As shown in Figure~\ref{fig:eagle-cross_view_prompt_learning}, the open-vocab model trained on car views is not able to perform well on the images captured from unmanned aerial vehicles (UAVs) or drones.
While this issue can be improved by training the segmentation model on drone-view data, the annotation process of high-resolution UAV data is costly and time-consuming.
At present, there exist many large-scale datasets with dense labels captured from camera views on the ground, e.g., car views (SYNTHIA~\cite{Ros_2016_CVPR}, GTA~\cite{Richter_2016_ECCV}, Cityscapes~\cite{cordts2016cityscapes}, BDD100K~\cite{yu2020bdd100k}). They have been widely adopted to develop robust perception models.
Since these car view and drone view datasets have many common objects of interest, incorporating knowledge from car views with drone views benefits the learning process by reusing large-scale annotations and saving efforts of manually labeling UAV images.

Unsupervised domain adaptation (UDA)~\cite{vu2019advent, hoyer2022daformer, araslanov2021dasac, dat2021bimal_iccv, Truong:CVPR:2023FREDOM} is one of the potential approaches to transfer the knowledge from the car view (i.e., source domain) to the drone view (i.e., target domain). 
While UDA approaches have shown their effectiveness in transferring knowledge across domains, e.g., environment changes or geographical domain shifts, these methods remain limited in the cases of changing camera viewpoints.

\begin{wrapfigure}{r}{0.5\textwidth}
    \centering
    \includegraphics[width=0.5\textwidth]{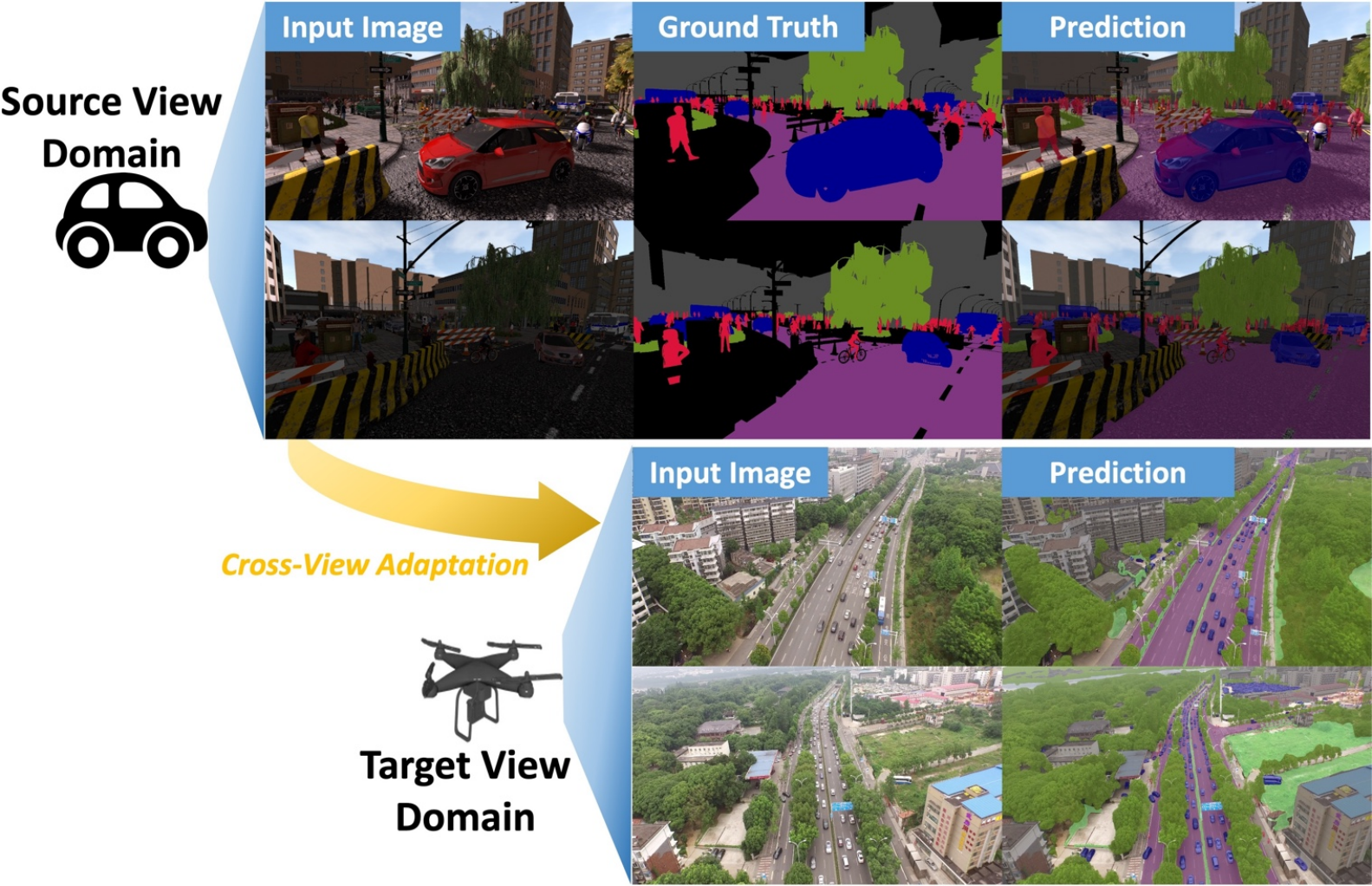}
    \caption{Illustration of Cross-View Adaptation.}
    \label{fig:eagle-cross_view_tasks}
\end{wrapfigure}
Indeed, the changes in camera positions, e.g., from the ground of cars to the high positions of drones, bring a significant difference in structures and topological layouts of scenes and objects ({Figure~\ref{fig:eagle-cross_view_tasks}}).
Therefore, UDA is not a complete solution to this problem due to its lack of cross-view structural modeling.
Additionally, although the open-vocab segmentation models have introduced several prompting mechanisms, e.g., context-aware prompting~\cite{rao2021denseclip} or adaptive prompting~\cite{qin2023freeseg}  to improve context learning across various open-world concepts, they are unable to model the cross-view structure due to the lack of view-condition information in prompts and geometric modeling.
To the best of our knowledge, there are limited studies that have exploited this cross-view learning. 
These limitations motivate us to develop a new adaptation learning paradigm, i.e., \textbf{\textit{Unsupervised Cross-view Adaptation}}, that addresses prior methods to improve the performance of semantic segmentation models across views.

To address these prior limitations, this work introduces a novel \textit{\textbf{E}fficient \textbf{A}daptive \textbf{G}eometry-based \textbf{L}earning \textbf{(EAGLE)}}
to \textbf{\textit{Unsupervised Cross-view Adaptation}} that can adaptively learn and improve the performance of semantic segmentation models across camera viewpoints.
First, by analyzing the geometric correlations across views, we introduce a novel \textbf{\textit{cross-view geometric constraint}} on \textbf{\textit{unpaired data}} of structural changes in images and segmentation masks. 
Second, to efficiently model \textbf{\textit{cross-view geometric structural changes}}, we introduce a new \textbf{\textit{Geodesic Flow-based Metric}} to measure the structural changes across views via their manifold structures.
In addition, to further improve the prompting mechanism of the open-vocab segmentation network in cross-view adaptation learning, we introduce a new \textbf{\textit{view-condition prompting}}.
Then, our \textbf{\textit{cross-view geometric constraint}} is also imposed on its feature representations of view-condition prompts to leverage its geometric knowledge embedded in our prompting mechanism.
Our proposed method holds a promise to be an effective approach to addressing the problem of cross-view learning and contributes to improving UDA and open-vocab segmentation in cross-view learning. 
Thus, it increases the generalizability of the segmentation models across camera views.

\subsection{The Proposed Geometry-based Learning Approach to Cross-view Adaptation}
\label{sec:proposed_method}

In this study, we consider cross-view adaptation learning as UDA where the images of the source and target domains are captured from different camera positions ({Figure~\ref{fig:eagle-cross_view_tasks}}).
Formally, 
let $\mathbf{x}_s, \mathbf{x}_t$ be the input images in the source and target domains, 
$\mathbf{p}_s, \mathbf{p}_t$ be the the corresponding prompts, 
and $\mathbf{y}_s, \mathbf{y}_t$ be the segmentation masks of $\mathbf{x}_s, \mathbf{x}_t$.  
Then, the open-vocab segmentation model $F$ maps the input $\mathbf{x}$ and the prompt $\mathbf{p}$ to the corresponding output $\mathbf{y} = F(\mathbf{x}, \mathbf{p})$.
It should be noted that in the case of traditional semantic segmentation, the prompt $\mathbf{p}$ will be ignored, i.e., $\mathbf{y} = F(\mathbf{x})$
The cross-view adaptation learning can be formulated as Eqn.~\eqref{eqn:eagle-general_opt}.
\begin{equation}
\label{eqn:eagle-general_opt}
\begin{split}
    \arg\min_{\theta}\Big[\mathbb{E}_{\mathbf{x}_s, \mathbf{p}_s, \mathbf{\hat{y}}_s} \mathcal{L}_{Mask}(\mathbf{y}_s, \mathbf{\hat{y}}_s) + \mathbb{E}_{\mathbf{x}_t, \mathbf{p}_t}\mathcal{L}_{Adapt}(\mathbf{y}_t)\Big]
\end{split}
\end{equation}
where 
$\theta$ is the parameters of $F$, 
$\mathbf{\hat{y}}_s$ 
is the ground truth,
$\mathcal{L}_{Mask}$ is the supervised (open-vocab) segmentation loss with ground truths, 
and $\mathcal{L}_{Adapt}$ is unsupervised adaptation loss from the source to the target domain.
In the open-vocab setting, we adopt the design of Open-Vocab Mask2Former~\cite{cheng2022mask2former, qin2023freeseg} to our network $F$.
Prior UDA methods defined the adaptation loss $\mathcal{L}_{Adapt}$ via the adversarial loss~\cite{lee2018spigan, chen2018domain}, entropy loss~\cite{dat2021bimal_iccv, vu2019advent}, or self-supervised loss~\cite{hoyer2022daformer, hoyer2023mic}.
Although these prior results have illustrated their effectiveness in UDA, these losses remain limited in cross-view adaptation setup.
Indeed, the adaptation setting in prior studies~\cite{vu2019advent, araslanov2021dasac, hoyer2022daformer, fahes2023poda} is typically deployed in the context of environmental changes (e.g., simulation to real~\cite{vu2019advent, vu2019dada, fahes2023poda}, day to night~\cite{hoyer2023mic, fahes2023poda}, etc) where the camera positions between domains remain similar. 
Meanwhile, in cross-view adaptation, the camera position of the source and target domain remains largely different (as shown in Figure~\ref{fig:eagle-cross_view_tasks}). 
This change in camera positions leads to significant differences in the geometric layout and topological structures between the source and target domains.
As a result, direct adoption of prior UDA approaches to cross-view adaptation would be ineffective due to the lack of cross-view geometric correlation modeling. 
To effectively address cross-view adaptation, the adaptation loss $\mathcal{L}_{Adapt}$ should be able to model (1) \textbf{\textit{the geometric correlation between two views of source and target domains}} and (2) \textbf{\textit{the structural changes across domains}}.

\subsubsection{Cross-View Geometric Modeling}

To efficiently address the cross-view adaptation learning task, it is essential to explicitly model cross-view geometric correlations by analyzing the relation between two camera views.
Therefore, we first re-reconsider the cross-view geometric correlation. 
In particular,
let $\mathbf{\bar{x}}_{t}$ be the corresponding image of $\mathbf{x}_{s}$ captured from the target view, 
$\mathbf{y}_{s}$ and $\mathbf{\bar{y}}_{t}$ be the semantic segmentation outputs of source image $\mathbf{x}_{s}$ and target image $\mathbf{\bar{x}}_{t}$, 
$\mathbf{\bar{p}}_t$ be the corresponding prompt of $\mathbf{p}_s$ in target view, respectively.
Formally, the images captured from the source and the target views can be modeled as Eqn.~\eqref{eqn:eagle-render}.
\begin{equation} \label{eqn:eagle-render}
\begin{split}
    \mathbf{x}_{s} &= \mathcal{R}(\mathbf{K}_{s}, [\mathbf{R}_{s},\mathbf{t}_{s}], \Theta), \quad       
    \mathbf{\bar{x}}_{t} = \mathcal{R}(\mathbf{K}_{t}, [\mathbf{R}_{t},\mathbf{t}_{t}], \Theta)
\end{split}
\end{equation}
where $\mathcal{R}$ is the rendering function, $\mathbf{K}_{s}$ and $\mathbf{K}_{t}$ are the intrinsic matrices,  $[\mathbf{R}_{s}, \mathbf{t}_{s}]$ and $[\mathbf{R}_{t}, \mathbf{t}_{t}]$ are the extrinsic matrices,
and $\Theta$ represents the capturing scene.
In addition, as the camera parameters of both source and target views are represented by matrices, there should exist linear transformations of camera parameters between two views as in Eqn.~\eqref{eqn:eagle-cam_equi}.
\begin{equation} \label{eqn:eagle-cam_equi}
\small
    \begin{split}
        \mathbf{K}_{t} &= \mathbf{T}_{\mathbf{K}} \times \mathbf{K}_{s}, \quad [\mathbf{R}_{t}, \mathbf{t}_{t}] = \mathbf{T}_{\mathbf{Rt}} \times [\mathbf{R}_{s}, \mathbf{t}_{s}]
    \end{split}
\end{equation}
where $\mathbf{T}_{\mathbf{K}}$ and $\mathbf{T}_{\mathbf{Rt}}$ are the transformation matrices.

\begin{tcolorbox}[colback=black!5!white,colframe=black!75!black,title=\textbf{Remark \showremarkcounter\label{rmk:camera_camera}}]
\textbf{The Geometric Transformation Between Camera Views.}
From Eqn.~\eqref{eqn:eagle-render} and Eqn.~\eqref{eqn:eagle-cam_equi}, we argue that there should exist a geometric transformation $\mathcal{T}$ of images between two camera views as: $\mathbf{\bar{x}}_{t} = \mathcal{T}(\mathbf{x}_{s}; \mathbf{T_K}, \mathbf{T_{Rt}})$.
\end{tcolorbox}

\begin{tcolorbox}[colback=black!5!white,colframe=black!75!black,title=\textbf{Remark \showremarkcounter\label{rmk:image_segmentation}}]
\textbf{The Equivalent Transformation Between Image and Segmentation Output.}
As RGB images and segmentation maps are pixel-wised corresponding, the same geometric transformation $\mathcal{T}$ in the image space can be adopted for segmentation space as: $\mathbf{\bar{y}}_{t} = \mathcal{T}(\mathbf{y}_{s}; \mathbf{T_K}, \mathbf{T_{Rt}})$
\end{tcolorbox}

\textbf{Remarks \ref{rmk:camera_camera}-\ref{rmk:image_segmentation}} have depicted that the geometric transformation of both image and segmentation from the source to the target view can be represented by shared transformation $\mathcal{T}$ with camera transformation matrices $\mathbf{T_K}, \mathbf{T_{Rt}}$. 
Let $\mathcal{D}_x(\mathbf{x}_{s}, \mathbf{\bar{x}}_{t})$ and $\mathcal{D}_y(\mathbf{y}_{s}, \mathbf{\bar{y}}_{t})$ {\textbf{be the metrics measure the cross-view structures changes}} of images and segmentation maps from the source to target domains.

We argue that the cross-view geometric correlation in the image space, i.e., $\mathcal{D}_x(\mathbf{x}_{s}, \mathbf{\bar{x}}_{t})$, is theoretically proportional to the one in the segmentation space, i.e., $\mathcal{D}_y(\mathbf{y}_{s}, \mathbf{\bar{y}}_{t})$. 
Since the camera transformations between the two views are linear (Eqn.~\eqref{eqn:eagle-cam_equi}) and the images $\mathbf{x}$ and outputs $\mathbf{y}$ are pixel-wised corresponding, we hypothesize that the cross-view geometric correlation in the image space $\mathcal{D}_x(\mathbf{x}_{s}, \mathbf{\bar{x}}_{t})$ and the segmentation space $\mathcal{D}_y(\mathbf{y}_{s}, \mathbf{\bar{y}}_{t})$ can be modeled by a linear relation with linear scale $\alpha$ as in Eqn.~\eqref{eqn:eagle-cross_view_condition}.
\begin{equation} \label{eqn:eagle-cross_view_condition}
\small
\begin{split}
    \mathcal{D}_x(\mathbf{x}_{s}, \mathbf{\bar{x}}_{t}) &\propto \mathcal{D}_y(\mathbf{y}_{s}, \mathbf{\bar{y}}_{t}) \Leftrightarrow
    \mathcal{D}_x(\mathbf{x}_{s}, \mathbf{\bar{x}}_{t})  = \alpha \mathcal{D}_y(\mathbf{y}_{s}, \mathbf{\bar{y}}_{t})
\end{split}
\end{equation}

\subsubsection{Cross-view Geometric Learning on Unpaired Data}

Eqn.~\eqref{eqn:eagle-cross_view_condition} defines a necessary condition to explicitly model the cross-view geometric correlation. Therefore, cross-view adaptation learning in Eqn.~\eqref{eqn:eagle-general_opt} can be re-formed as follows:
\begin{equation} 
\small
\label{eqn:eagle-obj_with_cond_pair}
\begin{split}
     \theta^* = \arg\min_{\theta} \Big[&\mathbb{E}_{\mathbf{x}_{s}, \mathbf{p}_s, \mathbf{\hat{y}}_{s}} \mathcal{L}_{Mask}(\mathbf{y}_{s}, \mathbf{p}_s, \hat{\mathbf{y}}_{s}) + \mathbb{E}_{\mathbf{x}_{s}, \mathbf{p}_s, \mathbf{\bar{x}}_{t}, \mathbf{\bar{p}}_t} ||\mathcal{D}_x(\mathbf{x}_{s}, \mathbf{\bar{x}}_{t})  - \alpha \mathcal{D}_y(\mathbf{y}_{s}, \mathbf{\bar{y}}_{t}) || \Big]\\
\end{split}
\end{equation}
where, $\mathcal{L}_{Adapt}(\mathbf{y}_{s}, \mathbf{\bar{y}}_{t}) = ||\mathcal{D}_x(\mathbf{x}_{s}, \mathbf{\bar{x}}_{t})  - \alpha \mathcal{D}_y(\mathbf{y}_{s}, \mathbf{\bar{y}}_{t}) ||$ is the cross-view geoemtric adaptation loss, $|| \cdot ||$ is the mean squared error loss.
However, in practice, the pair data between source and target views are inaccessible as data from these two views are often collected independently. 
Thus, optimizing Eqn.~\eqref{eqn:eagle-obj_with_cond_pair} without cross-view pairs of data remains an ill-posed problem.
To address this limitation, instead of learning Eqn.~\eqref{eqn:eagle-obj_with_cond_pair} on paired data, we proposed to model this correlation on unpaired data.
Instead of solving the cross-view geometric constraint of Eqn.~\eqref{eqn:eagle-obj_with_cond_pair} on pair data, let us consider all cross-view unpaired samples $(\mathbf{x}_{s}, \mathbf{x}_{t})$.
Formally, learning the \textbf{\textit{Cross-view Geometric Constraint}} between unpaired samples can be formulated as in Eqn.~\eqref{eqn:eagle-loss_for_lt}.
\begin{equation} 
\small
\label{eqn:eagle-loss_for_lt}
\begin{split}
     \theta^* = \arg\min_{\theta} \Big[&\mathbb{E}_{\mathbf{x}_{s}, \mathbf{\hat{y}}_{s}} \mathcal{L}_{Mask}(\mathbf{y}_{s}, \mathbf{p}_{s}, \hat{\mathbf{y}}_{s}) + \mathbb{E}_{\mathbf{x}_{s}, \mathbf{p}_{s}, \mathbf{{x}}_{t}, \mathbf{p}_{t}} ||\mathcal{D}_x(\mathbf{x}_{s}, \mathbf{{x}}_{t})  - \alpha \mathcal{D}_y(\mathbf{y}_{s}, \mathbf{{y}}_{t}) || \Big]
\end{split}
\end{equation}

where $\mathbf{x}_{s}$ and $\mathbf{x}_{t}$ are unpaired data, and $\mathcal{L}_{Adapt}(\mathbf{y}_s, \mathbf{y}_t) = ||\mathcal{D}_x(\mathbf{x}_{s}, \mathbf{{x}}_{t})  - \alpha \mathcal{D}_y(\mathbf{y}_{s}, \mathbf{{y}}_{t}) ||$ is the \textbf{\textit{Cross-view Geometric Adaptation}} loss on unpaired data.
{
Intuitively, although the cross-view pair samples are not available, the cross-view geometric constraints on paired samples between two views can be indirectly imposed by modeling the cross-view geometric structural constraint among unpaired samples.
}
Then, by modeling the cross-view structural changes in the image and segmentation spaces, the structural change on images of unpaired data could be considered as the reference for the cross-view structural change in the segmentation space during the optimization process. 
This action promotes the structures of segmentation that can be effectively adapted from the source view to the target view.
Importantly, the cross-view geometric constraint imposed on unpaired data can be mathematically proved as an upper bound of the cross-view constraint on paired data as in \textbf{Proposition~\ref{pro:upper_bound_eagle_constraint}}.

\begin{tcolorbox}[colback=blue!5!white,colframe=blue!75!black,title={\textbf{Proposition \showpropositioncounter\label{pro:upper_bound_eagle_constraint}}}]
The cross-view geometric constraint imposed on unpaired data is an upper bound of the one on paired data as in Eqn. \eqref{eqn:eagle-upper_bound}.
\begin{equation} \label{eqn:eagle-upper_bound}
\small
\begin{split}
    || \mathcal{D}_x(\mathbf{x}_{s}, \mathbf{\bar{x}}_{t})  - \alpha \mathcal{D}_y(\mathbf{y}_{s}, \mathbf{\bar{y}}_{t})|| = \mathcal{O}\left(\mathcal{D}_x(||\mathbf{x}_{s}, \mathbf{{x}}_{t})  - \alpha \mathcal{D}_y(\mathbf{y}_{s}, \mathbf{{y}}_{t})||\right)
\end{split}
\end{equation}
where $\mathcal{O}$ is the Big O notation.
\end{tcolorbox}

The upper bound in \textbf{Proposition~\ref{pro:upper_bound_eagle_constraint}} can be proved by using the properties of triangle inequality and our correlation metrics $\mathcal{D}_{x}$ and $\mathcal{D}_{y}$ (Sec. \ref{sec:geodesic_flow_metric}).  The detailed proof can be found in our preliminary work \cite{truong2024eagle}.
 
\begin{wrapfigure}{!r}{0.6\textwidth}
    \centering
    \includegraphics[width=0.6\textwidth]{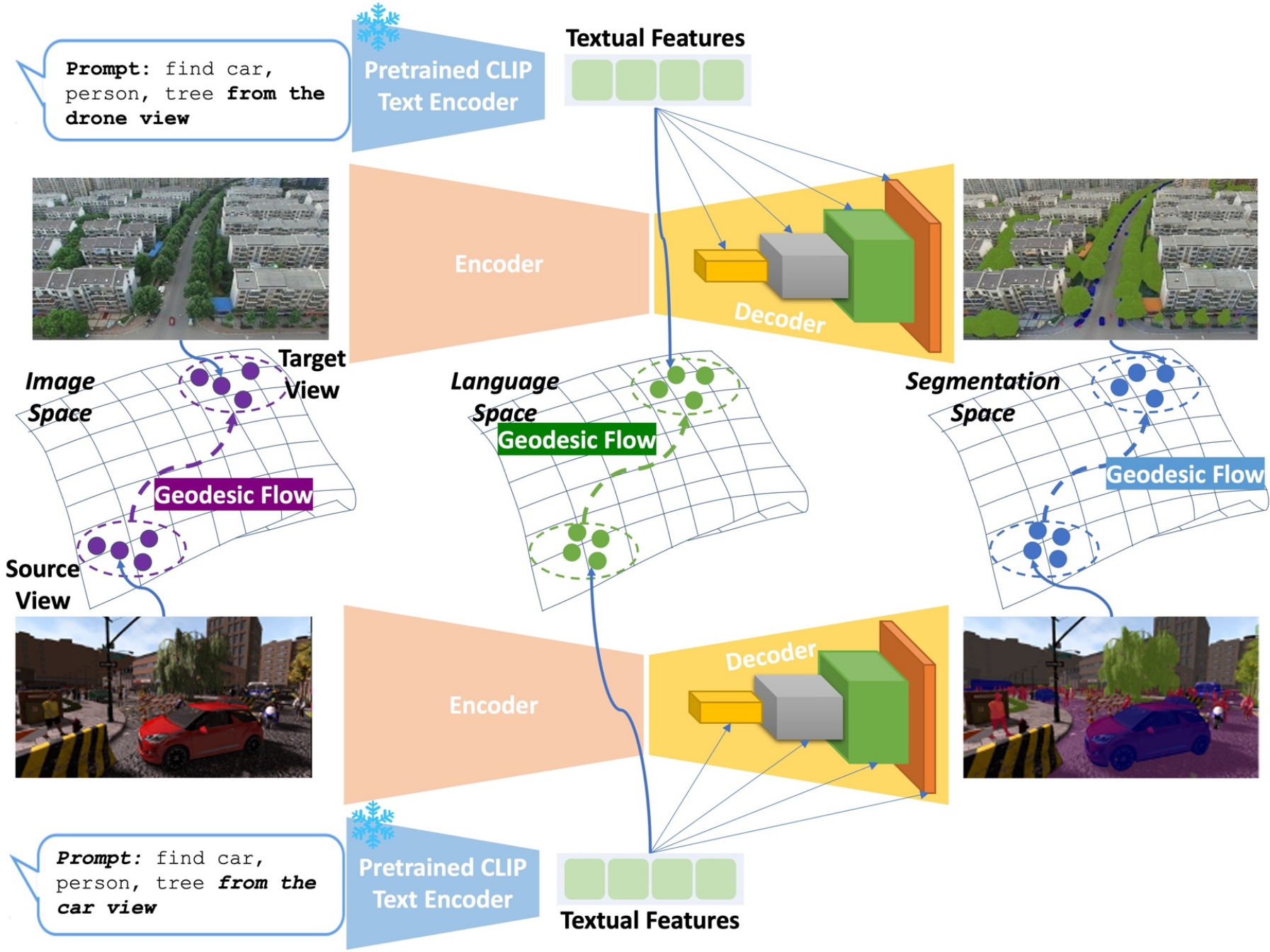}
    \caption{\textbf{Our Cross-View Learning Framework.}} 
    \label{fig:eagle-cross_view_framework}
\end{wrapfigure}

Eqn.~\eqref{eqn:eagle-upper_bound} has illustrated that by minimizing the cross-view geometric constraint on unpaired samples in Eqn.~\eqref{eqn:eagle-loss_for_lt}, the cross-view constraint on paired samples in Eqn.~\eqref{eqn:eagle-obj_with_cond_pair} is also maintained due to the upper bound. 
Therefore, our proposed Cross-view Geometric Constraint loss \textbf{\textit{does NOT require the pair data between source and target views}} during training.
Figure~\ref{fig:eagle-cross_view_framework} illustrates our cross-view adaptation learning framework.
The proof of upper bound of Eqn.~\eqref{eqn:eagle-upper_bound} can be found in our preliminary work ~\cite{truong2024eagle}.

\subsubsection{Cross-view Structural Change Modeling via Geodesic Flow Path}
\label{sec:geodesic_flow_metric}

Modeling the correlation metrics $\mathcal{D}_x$ and $\mathcal{D}_y$ is an important task in our approach.
Indeed, the metrics should be able to model the structure changes from the source to the target view.
Intuitively, the changes from the source to the target view are essentially the geodesic flow between two subspaces on the Grassmann manifold.
Then, the images (or segmentation) of two views can be projected along the geodesic flow path to capture the cross-view structural changes.
Therefore, to model $\mathcal{D}_x$ and $\mathcal{D}_y$, we adopt the \textbf{\textit{Geodesic Flow}} path to measure the cross-view structural changes by modeling the geometry in the latent space.

\begin{tcolorbox}[colback=black!5!white,colframe=black!75!black,title=\textbf{Remark \showremarkcounter\label{rmk:grassman_manifold}}]
\textbf{Grassmann Manifold} is 
the set of $N$-dimensional linear subspaces of $\mathbb{R}^{D} (0 < N < D)$, i.e, $\mathcal{G}(N, D)$. 
A  matrix with orthonormal columns $\mathbf{P} \in \mathbb{R}^{D \times N}$ define a subspace of $\mathcal{G}(N, D)$, i.e., $\mathbf{P} \in \mathcal{G}(N, D) \Rightarrow \mathbf{P}^\top \mathbf{P} = \mathbf{I}_{N}$ where $\mathbf{I}_N$ is the $N \times N$ identity matrix.
\end{tcolorbox}
\noindent

For simplicity, we present our approach to model the cross-view structural change $\mathcal{D}_x$ in the image space.
Formally, let $\mathbf{P}_{s}$ and $\mathbf{P}_{t}$ be the basis of the source and target domains. These bases can be obtained by the PCA algorithm. 
The geodesic flow between $\mathbf{P}_{s}$ and $\mathbf{P}_{t}$ in the manifold can be defined via the function $\boldsymbol{\Pi}: \nu \in [0..1] \to \boldsymbol{\Pi}(\nu)$, where $\boldsymbol{\Pi}(\nu) \in \mathcal{G}(N, D)$ is the subspace lying on the geodesic flow path from the source to the target view as in Eqn.~\eqref{eqn:eagle-pi-source-target}.
\begin{equation}\label{eqn:eagle-pi-source-target}
\small
    \boldsymbol{\Pi}(\nu) = [\mathbf{P}_{s} \;\;\; \mathbf{R}] 
    [\mathbf{U}_1 \boldsymbol{\Gamma}(\nu) \;\;\; -\mathbf{U}_2 \boldsymbol{\Sigma}(\nu)]^\top 
\end{equation}
where $\mathbf{R} \in \mathbb{R}^{D \times (D-N)}$ is the orthogonal complement of $\mathbf{P}_{s}$, i.e., $\mathbf{R}^\top\mathbf{P}_s = \boldsymbol{0}$. 
$\boldsymbol{\Gamma}(\nu)$ and $\boldsymbol{\Sigma}(\nu)$ are the diagonal matrices whose diagonal element at row $i$ 
can be defined as $\gamma_i = \cos(\nu\omega_i)$ and $\sigma_i = \sin(\nu\omega_i)$.
The list of $\omega_i$ is the principal angles between source and target subspaces, i.e., $0 \leq \omega_1 \leq ... \leq \omega_N \leq \frac{\pi}{2}$.
$\mathbf{U}_1$ and $\mathbf{U}_2$ are the orthonormal matrices obtained by the following pair of SVDs as in Eqn.~\eqref{eqn:eagle-svd-decompose}.
\begin{equation}\label{eqn:eagle-svd-decompose}
\small
    \mathbf{P}_s^\top\mathbf{P}_T = \mathbf{U}_1\boldsymbol{\Gamma}(1)\mathbf{V}^\top \quad\quad \mathbf{R}^\top\mathbf{P}_T= -\mathbf{U}_2\boldsymbol{\Sigma}(1)\mathbf{V}^\top
\end{equation}
Since $\mathbf{P}_s^\top \mathbf{P}_{t}$ and $\mathbf{R}^\top  \mathbf{P}_{t}$ share the same singular vectors $\mathbf{V}$, we adopt the generalized Singular Value Decomposition (SVD)~\cite{gong2012geodesic, simon2021learning} to decompose the matrices. In our approach, we model the cross-view structural changes $\mathcal{D}_x$ by modeling the cosine similarity between projections along the geodesic flow $\boldsymbol{\Pi}(\nu)$. In particular, given a subspace $\boldsymbol{\Pi}(\nu)$ on the geodesic flow path from the source to the target view, 
the cross-view geometric correlation of images between the source and target views can formulated by the inner product $g_{\boldsymbol{\Pi}(\nu)}(\mathbf{x}_s, \mathbf{x}_t)$ along the geodesic flow  $\boldsymbol{\Pi}(\nu)$ as in Eqn.\eqref{eqn:eagle-geodesic_flow_s_t}.
\begin{equation}\label{eqn:eagle-geodesic_flow_s_t}
\begin{split}
    g(\mathbf{x}_{s}, \mathbf{x}_{t}) &=  \int_0^1 g_{\boldsymbol{\Pi}(\nu)}(\mathbf{x}_s, \mathbf{x}_t)d\nu = \int_0^1 \mathbf{x}_{s}^\top \boldsymbol{\Pi}(\nu)\boldsymbol{\Pi}(\nu)^\top \mathbf{x}_{t}d\nu \\
    &= \mathbf{x}_{s}^\top \left(\int_0^1  \boldsymbol{\Pi}(\nu)\boldsymbol{\Pi}(\nu)^\top d\nu\right) \mathbf{x}_{t} = \mathbf{x}_{s}^\top \mathbf{Q} \mathbf{x}_{t} 
\end{split}
\end{equation}
where $\mathbf{Q} = \int_0^1  \boldsymbol{\Pi}(\nu)\boldsymbol{\Pi}(\nu)^\top d\nu$. 
Intuitively, the matrix $\mathbf{Q}$ represents the manifold structure between the source to the target view. Then, Eqn.~\eqref{eqn:eagle-geodesic_flow_s_t} measures the cross-view structural changes between the source and the target domain based on their manifold structures.
The matrix $\mathbf{Q}$ can be obtained in a closed form~\cite{gong2012geodesic, simon2021learning} as in Eqn.~\eqref{eqn:eagle-manifold-q}.
\begin{equation}\label{eqn:eagle-manifold-q}
\small
    \mathbf{Q} = \left[ \mathbf{P}_s\mathbf{U}_1 \quad \mathbf{R}\mathbf{U}_2\right]\begin{bmatrix}
        \boldsymbol{\Lambda}_1 & \boldsymbol{\Lambda}_2 \\
        \boldsymbol{\Lambda}_2 & \boldsymbol{\Lambda}_3 
    \end{bmatrix}
    \begin{bmatrix}
        \mathbf{U}_1^\top\mathbf{P}_s^\top \\ 
        \mathbf{U}_2^\top\mathbf{R}^\top
    \end{bmatrix}
\end{equation}
where $\boldsymbol{\Lambda}_1$, $\boldsymbol{\Lambda}_2$, and $\boldsymbol{\Lambda}_3$ are the diagonal matrices, whose diagonal elements at row $i$ can be defined as in Eqn.~\eqref{eqn:eagle-lambda_diag}.
\begin{equation}\label{eqn:eagle-lambda_diag}
\begin{split}
    \lambda_{1,i} = 1+\frac{\sin(2\omega_i)}{2\omega_i}, \; \lambda_{2,i} = \frac{\cos(2\omega_i)-1}{2\omega_i}, \; \lambda_{3,i} = 1-\frac{\sin(2\omega_i)}{2\omega_i}
\end{split}
\end{equation}
In practice, we model the cross-view structural changes $\mathcal{D}_x$ via the cosine similarity along the geodesic flows. Finally, the cross-view structural changes $\mathcal{D}_x$ can be formulated as in Eqn.~\eqref{eqn:eagle-final_metric_dx}.
\begin{equation}\label{eqn:eagle-final_metric_dx}
    \mathcal{D}_x(\mathbf{x}_{s}, \mathbf{x}_{t}) = 1 - \frac{\mathbf{x}_{s}^\top \mathbf{Q} \mathbf{x}_{t}}{||\mathbf{Q}^{1/2}\mathbf{x}_s||||\mathbf{Q}^{1/2}\mathbf{x}_t||}
\end{equation}
Similarly, we can model the cross-view geometric correlation of segmentation $\mathcal{D}_{y}$ via Geodesic Flow.

\subsubsection{View-Condition Prompting to Cross-View Learning}

\noindent
\textbf{View-Condition Prompting.} 
Previous efforts~\cite{rao2021denseclip, qin2023freeseg, gao2023clip, zhou2022learning} in open-vocab segmentation have shown that a better prompting mechanism can provide more meaningful textual and visual knowledge.
Prior work in open-vocab segmentation designed the prompt via the class names~\cite{xu2022simple, ding2022decoupling, qin2023freeseg}, e.g., ``$\texttt{class}_1$, $\texttt{class}_1$, ..., $\texttt{class}_K$''. Meanwhile, other methods improve the prompting mechanism by introducing the learnable variables into the prompt~\cite{rao2021denseclip} or adding the task information~\cite{qin2023freeseg}. This action helps to improve the context learning of the vision-language model. 
In our approach, we also exploit the effectiveness of designing prompting to cross-view learning. In particular, 
describing the view information can further improve the visual context learning, e.g., ``$\texttt{class}_1$, $\texttt{class}_1$, ..., $\texttt{class}_K$ \texttt{captured from the} \texttt{[domain]} \texttt{view}'', where \texttt{[domain]} could be car (source domain) or drone (target domain). Therefore, we introduce a view-condition prompting mechanism by introducing the view information, i.e., \texttt{captured from the} \texttt{[domain]} \texttt{view}'', into the prompt.
Our view-condition prompt offers the context specific to visual learning, thus providing better transferability in cross-view segmentation.

\noindent
\textbf{Cross-view Correlation of View-Condition Prompts.}
We hypothesize that the correlation of the input prompts across domains also provides the cross-view geometric correlation in their deep representations. In particular, let $\mathbf{f}^p_s$ and $\mathbf{f}^p_t$ be the deep textual embeddings of view-condition prompts $\mathbf{p}_s$ and $\mathbf{p}_t$, and $\mathcal{D}_p$ be metric measuring the correlation between $\mathbf{f}^p_s$ and $\mathbf{f}^p_t$. 
In addition, since the textual encoder has been pre-trained on large-scale vision-language data~\cite{radford2021learning, jia2021scaling}, the visual and the textual representations have been well aligned.
Then, we argue that the correlation of textual feature representations across views, i.e., $\mathcal{D}_p(\mathbf{f}^p_s, \mathbf{f}^p_t)$, also provides the cross-view geometric correlation due to the embedded view information in the deep representation of prompts aligned with visual representations.
Therefore, similar to Eqn.~\eqref{eqn:eagle-cross_view_condition}, we hypothesize the cross-view correlation of segmentation masks and textual features can be modeled as a linear relation with a scale factor $\gamma$ as in Eqn.~\eqref{eqn:eagle-cross_view_condition_text}.
\begin{equation} \label{eqn:eagle-cross_view_condition_text}
\begin{split}
    \mathcal{D}_p(\mathbf{f}^p_{s}, \mathbf{f}^p_{t})  \propto \mathcal{D}_y(\mathbf{y}_{s}, \mathbf{y}_{t}) \Leftrightarrow \mathcal{D}_p(\mathbf{f}^p_{s}, \mathbf{f}^p_{t})  &= \gamma \mathcal{D}_y(\mathbf{y}_{s}, \mathbf{y}_{t})
\end{split}
\end{equation}
Then, 
learning the cross-view adaptation with view-condition prompts can be formulated as in Eqn.~\eqref{eqn:eagle-final_loss}.
\begin{equation} 
\small
\label{eqn:eagle-final_loss}
\begin{split}
     \theta^* = \arg\min_{\theta} \Big[\mathbb{E}_{\mathbf{x}_{s}, \mathbf{p}_{s}, \mathbf{\hat{y}}_{s}} \mathcal{L}_{Mask}(\mathbf{y}_{s}, \hat{\mathbf{y}}_{s}) &+  \mathbb{E}_{\mathbf{x}_{s}, \mathbf{x}_{s}, \mathbf{{x}}_{t}, \mathbf{p}_{t}} \big(\lambda_{I}||\mathcal{D}_x(\mathbf{x}_{s}, \mathbf{{x}}_{t})  - \alpha \mathcal{D}_y(\mathbf{y}_{s}, \mathbf{{y}}_{t}) \\
    &+  \lambda_{P}||\mathcal{D}_p(\mathbf{f}^p_{s}, \mathbf{f}^p_{t})  - \gamma \mathcal{D}_y(\mathbf{y}_{s}, \mathbf{{y}}_{t}) ||\big)
    \Big]
\end{split}
\end{equation}
where $\lambda_{I}$ and $\lambda_{P}$ are the balanced-weight of losses.
Similar to metrics $\mathcal{D}_x$ and $\mathcal{D}_y$, we also adopt the geodesic flow path to model the cross-view correlation metric~$\mathcal{D}_p$.

\subsection{Experimental Results}

\subsubsection{Datasets, Benchmarks, and Implementation}

To efficiently evaluate cross-view adaptation, the cross-view benchmarks are set up from the car to the drone view. 
Following common practices in UDA~\cite{hoyer2022daformer, vu2019advent}, we choose SYNTHIA~\cite{Ros_2016_CVPR}, GTA~\cite{Richter_2016_ECCV}, and BDD100K~\cite{yu2020bdd100k} as the source domains while UAVID~\cite{uavid_dataset} is chosen as the target domain. 
We chose to adopt these datasets because they share a class of interests and are commonly used in UDA and segmentation benchmarks~\cite{hoyer2022daformer, UNetFormer}.

\noindent
\textbf{SYNTHIA $\to$ UAVID Benchmark.}
SYNTHIA and UAVID share five classes of interest, i.e., \texttt{Road}, \texttt{Building}, \texttt{Car}, \texttt{Tree}, and \texttt{Person}.
Since the UAVID dataset annotated cars, trucks, and buses as a class of \texttt{Car}, we collapse these classes in SYNTHIA into a single class of \texttt{Car}.

\noindent
\textbf{GTA $\to$ UAVID Benchmark}
consists of five classes in the SYNTHIA $\to$ UAVID benchmark and includes one more class of \texttt{Terrain}.
Therefore, the GTA $\to$ UAVID benchmark has six classes of interest, i.e., \texttt{Road}, \texttt{Building}, \texttt{Car}, \texttt{Tree}, \texttt{Terrain}, and \texttt{Person}.

\noindent
\textbf{BDD $\to$ UAVID Benchmark.} is a real-to-real cross-view adaptation setting.
Similar to GTA $\to$ UAVID benchmark, there are six classes of interest between BDD100K and UAVID. 
In our experiments, we adopt the mean Intersection over Union (mIoU) metric to measure the performance.

\noindent
\textbf{Implementation.}
We follow the implementation of Mask2Former~\cite{cheng2022mask2former} and FreeSeg~\cite{qin2023freeseg} with ResNet~\cite{resnet} and Swin backbones~\cite{liu2021swin} for our segmentation network.
In particular, we adopt Mask2Former with Semantic Context Interaction of FreeSeg~\cite{qin2023freeseg} for our open-vocab segmentation network.
We use the pre-trained text encoder of CLIP~\cite{radford2021learning}.
The textual features $\mathbf{f}_s^p$ and $\mathbf{f}_t^p$ are obtained by the CLIP textual encoder.
Following common practices~\cite{qin2023freeseg, liang2023open}, we adopt the open-vocab segmentation loss of FreeSeg~\cite{qin2023freeseg} to our supervised loss $\mathcal{L}_{Mask}$.
For experiments without prompting, we use the Mask2Former network.
Following the UAV protocol of~\cite{UNetFormer}, the image size is set to $512 \times 512$. 
The linear scale factors $\alpha$ and $\gamma$ are set to $\alpha=1.5$ and $\gamma=1.0$, respectively.
For the Geodesic Flow modeling, we adopt the implementation of generalized SVD decomposition~\cite{gong2012geodesic, simon2021learning} in the framework.
The subspace dimension in our geodesic flow-based metrics is set to $D=256$.
The batch size and the base learning rate in our experiments are set to $16$ and $2.5\times10^{-4}$. 
The balanced weights of losses in our experiments are set to $\lambda_{I} = 1.0$ and $\lambda_P = 0.5$.
During training, the classes in the prompts are generated similarly for both view images.
In our Geodesic Flow-based metrics, the subspaces of images and ground-truth segmentation of the source domain are pre-computed on the entire data.
For the language space, we compute the subspaces of each view based on the textual feature representations of all possible prompts in each domain.
Meanwhile, the subspaces of the segmentation on the target domain are computed based on the current batch of training.
For the implementation of DenseCLIP~\cite{rao2021denseclip} and FreeSeg~\cite{qin2023freeseg} with AdvEnt~\cite{vu2019advent}, we perform the adaptation process on the mask predictions. Meanwhile, we adopt the pseudo labels and the self-supervised framework of SAC\cite{araslanov2021dasac} for   DenseCLIP~\cite{rao2021denseclip} and FreeSeg~\cite{qin2023freeseg}.

\subsubsection{Ablation Study}

\input{Tables/chap-6/eagle/ablation_different_seg}
\noindent
\textbf{Effectiveness of Cross-view Correlation Metrics and Network Backbones.} Table~\ref{tab:eagle-ablation_diff_seg} studies the impact of choosing metrics and network backbones.
We consider two options, i.e., Euclidean Metric and our Geodesic Flow-based Metric, for correlation metrics $\mathcal{D}_x$, $\mathcal{D}_y$, and $\mathcal{D}_p$. As shown in Table~\ref{tab:eagle-ablation_diff_seg}, our Geodesic Flow-based metrics significantly improve the performance of our cross-view adaptation. It has shown that our approach is able to measure the structural changes across views better than using the Euclidean metrics.
In addition, by using the more powerful backbone (Swin), the performance of cross-view adaptation is further improved.

\input{Tables/chap-6/eagle/ablation_prompting}

\noindent
\textbf{Effectiveness of Cross-view Adaptation and Prompting Mechanisms.} Table~\ref{tab:eagle-ablation_prompting} analyzes the effectiveness of prompting mechanisms, i.e., i.e., with and without \textbf{\textit{Prompting}}, with and without \textbf{\textit{Cross-view Adaptation}} (in Eqn.~\eqref{eqn:eagle-loss_for_lt}), with and without \textbf{\textit{View-Condition Prompting}} (in Eqn.~\eqref{eqn:eagle-final_loss}).
For supervised results, we train two different models on UAVID with and without the Terrain class on two benchmarks.
As in Table~\ref{tab:eagle-ablation_prompting}, the cross-view adaptation loss in Eqn.~\eqref{eqn:eagle-loss_for_lt} significantly improve the performance of segmentation models.
With prompting and cross-view adaptation, the mIoU performance is further boosted, i.e., the mIoU performance achieves $48.6\%$ and $40.1\%$ on two benchmarks.
By further using the view-condition prompting mechanism with our cross-view loss in Eqn.~\eqref{eqn:eagle-final_loss}, 
the mIoU results are slightly improved by $+1.1\%$ and $+1.7\%$ on two benchmarks compared to the one without view-condition prompting.
Our results have closed the gap with the supervised upper-bound results.

\input{Tables/chap-6/eagle/ablation_alpha}

\noindent
\textbf{Effectiveness of Cross-view Learning Parameters.} Table~\ref{tab:eagle-ablation_alpha} illustrates the impact of the linear scaling factors $\alpha$ and $\beta$.
As in Table~\ref{tab:eagle-ablation_alpha}, the mIoU performance has been majorly affected by the relation between images and segmentation.
The best performance is gained at the optimal value of $\alpha=1.5$.
Since the variation of RGB images is higher than the segmentation, the small value $\alpha$ could not correctly scale the relation between images and segmentation while the higher value of $\alpha$ exaggerates the structural change of segmentation masks.
Additionally, the change of $\gamma$ slightly affects the mIoU performance.
Since the textual features are well-aligned with the image, the performance of segmentation models when changing $\gamma$ also behaves similarly to the changes of $\alpha$. 
However, the linear scale factor $\alpha$ is more sensitive to mIoU results since the images play a more important role in the segmentation results due to the pixel-wise corresponding of images and segmentation.

\begin{wrapfigure}{r}{0.55\textwidth}
    \includegraphics[width=0.55\textwidth]{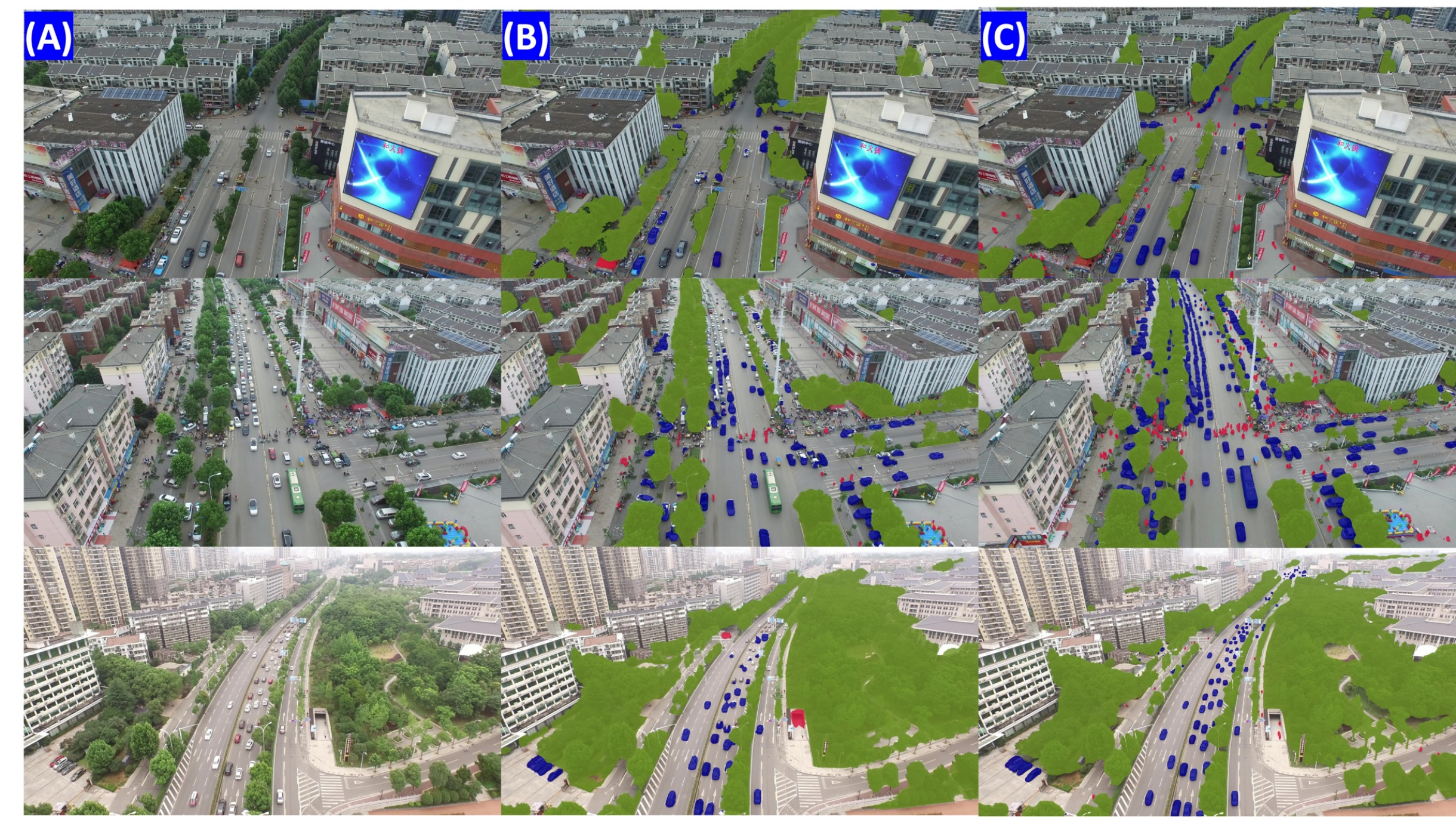}
    \centering
    \caption{Results of Segmenting \texttt{Cars}, \texttt{Trees}, \texttt{Persons}. (A) Input, (B) FreeSeg~\cite{qin2023freeseg}, and (C) Our EAGLE.}
    \label{fig:eagle-qual_res}
\end{wrapfigure}
\noindent
\textbf{Effectiveness of Subspace Dimension in Geodesic Flow.} 
Table~\ref{tab:eagle-ablation_alpha} reveals the importance of choosing the subspace dimension.
The cross-view geometric structural change is better modeled by increasing the dimension of the subspaces.
As in Table~\ref{tab:eagle-ablation_alpha}, the performance is improved when the dimension is increased from $96$ to $256$. However, beyond that point, the mIoU performance tends to be dropped.
We have observed that low dimensionality cannot model the structural changes across views since it captures small variations in structural changes.
Conversely, higher dimensionality includes more noise in the cross-view structural changes and increases the computational cost.

\subsubsection{Comparisons with Prior UDA Methods}

\input{Tables/chap-6/eagle/domain_adaptation_exp}

\noindent
\textbf{SYNTHIA $\to$ UAVID.}
As shown in Table~\ref{tab:eagle-uda_exp}, our EAGLE has achieved SOTA results and outperforms prior UDA methods by a large margin.
For fair comparisons, we adopt the DeepLab~\cite{chen2018deeplab} and DAFormer~\cite{hoyer2022daformer} for the segmentation network.
In particular, our mIoU results using DeepLab and DAFormer are $45.2\%$ and $50.8\%$.
In the DAFormer backbone, the mIoU results of our approach are higher than CROVIA~\cite{truong2023crovia} and MIC~\cite{hoyer2023mic} by $+4.8\%$ and $+9.0\%$. The IoU result of each class also consistently outperformed the prior methods.
Highlighted that although our approach does NOT use depth labels, our results still outperform the one using depths, i.e., DADA~\cite{vu2019dada}. It has emphasized that our approach is able to better capture the cross-view structural changes compared to prior methods.

\noindent
\textbf{GTA $\to$ UAVID.}
As shown in Table~\ref{tab:eagle-uda_exp}, our effectiveness outperforms prior domain adaptation approaches when measured by both mIoU performance and the IoU accuracy of each class.
In particular, our mIoU performance using DeepLab and DAFormer network achieves $36.7\%$ and $40.7\%$, respectively.
Our results have substantially closed the performance gap with the supervised results.
By using a better segmentation-based network, Mask2Former with ResNet, our performance is further improved to $40.1\%$ compared to DeepLab with ResNet.

\subsubsection{Comparisons with Open-vocab Segmentation}

We compare EAGLE with the prior open-vocab segmentation methods, i.e., DenseCLIP~\cite{rao2021denseclip} and an adaptive prompting FreeSeg~\cite{qin2023freeseg}
with four settings, i.e., Source Only, with AdvEnt~\cite{vu2019advent}, and with SAC~\cite{araslanov2021dasac}, and our Cross-View Adaptation in Eqn.~\eqref{eqn:eagle-loss_for_lt} (without view-condition).

\input{Tables/chap-6/eagle/openvocab-seenunseen}

\noindent
\textbf{Open-vocab Semantic Segmentation.}
As in Table~\ref{tab:eagle-open_vocab_exp}, the mIoU performance of our proposed approach with cross-view adaptation outperforms prior DenseCLIP by a large margin on SYNTHIA $\to$ UAVID.
By using our cross-view geometric adaptation loss, the performance of DenseCLIP and FreeSeg is further enhanced, i.e., higher than DenseCLIP and FreeSeg with SAC by +3.7\% and +5.0\%.
While FreeSeg~\cite{qin2023freeseg} with our cross-view adaptation slightly outperforms EAGLE due to its adaptive prompting, our EAGLE approach with the better view-condition prompting achieves higher mIoU performance.
Similarly, our proposed cross-view loss consistently improves the performance of DenseCLIP and FreeSeg on GTA $\to$ UAVID.
The mIoU results of DenseCLIP and FreeSeg using our cross-view loss achieve 37.3\% and 44.4\%.
By further using the view-condition prompting mechanism, our mIoU result is considerably higher than FreeSeg with our cross-view adaptation by +1.3\%.
Figure~\ref{fig:eagle-qual_res} visualizes the qualitative results of our approach.

\noindent
\textbf{Open-vocab Segmentation on Unseen Classes.} Table~\ref{tab:eagle-open_vocab_unseen_exp} illustrates the experimental results of our cross-view adaptation approach on unseen classes. 
In this experiment, we consider classes of {\texttt{Tree} and \texttt{Person}} as the unseen classes.
As shown in the results, our cross-view adaptation approach with a view-condition prompting mechanism has achieved the best mIoU performance on unseen classes on both benchmarks, i.e., $39.3\%$ and $39.6\%$ on two benchmarks. 
Our experimental results have further confirmed the effectiveness and the generalizability of our cross-view geometric modeling and view-condition prompting approach to the open-vocab segmentation across camera views.

\input{Tables/chap-6/eagle/real2real_exp}
\noindent
\textbf{Real-to-Real Cross-view Adaptation Setting.}
We evaluated our approach in the real-to-real setting, i.e., BDD $\to$ UAVID. 
Our approach is evaluated in two different settings, i.e., Unsupervised Domain Adaptation and Open-Vocab Semantic Segmentation.
As shown in Table~\ref{tab:eagle-real2real},  our results have shown a significant improvement in our approach in real-to-real settings in both unsupervised domain adaptation and open-vocab semantic semantic segmentation.
While the results of prior unsupervised domain adaptation, i.e., BiMaL~\cite{dat2021bimal_iccv} and DAFormer~\cite{hoyer2022daformer}, gain limited performance due to their limits in cross-view learning, our method outperforms other methods these prior methods by a large margin.

%% file: Tables/chap-6/eagle/ablation_different_seg.tex
\begin{wraptable}{r}{0.5\textwidth}
\setlength{\tabcolsep}{2pt}
\centering
\caption{Effectiveness of Backbones and Cross-view Metrics.}
\label{tab:eagle-ablation_diff_seg}
\resizebox{0.5\textwidth}{!}{  
\begin{tabular}{|c|c|ccccccc|}
\hline
\multicolumn{9}{|c|}{SYNTHIA $\to$ UAVID} \\
\hline
Network                 & Metric    & Road          & Building      & Car           & Tree          & Terrain       & Person        & mIoU          \\
\hline 
\multirow{2}{*}{ResNet} & Euclidean & 23.7          & 31.2          & 33.2          & 36.7          & -             & 11.5          & 27.2          \\
                        & Geodesic  & \textbf{38.4} & \textbf{76.1} & \textbf{62.8} & \textbf{62.1} & \textbf{-}    & \textbf{21.8} & \textbf{52.2} \\
\hline
\multirow{2}{*}{Swin}   & Euclidean & 24.7          & 31.9          & 41.2          & 39.7          & -             & 14.1          & 30.3          \\
                        & Geodesic  & \textbf{40.8} & \textbf{76.4} & \textbf{65.8} & \textbf{62.7} & \textbf{-}    & \textbf{27.9} & \textbf{54.7} \\
\hline
\multicolumn{9}{|c|}{GTA $\to$ UAVID} \\
\hline
\multirow{2}{*}{ResNet} & Euclidean & 21.7          & 30.0          & 26.2          & 39.7          & 31.7          & 9.5           & 26.5          \\
                        & Geodesic  & \textbf{29.2} & \textbf{67.1} & \textbf{45.2} & \textbf{56.6} & \textbf{48.5} & \textbf{27.9} & \textbf{45.7} \\
\hline
\multirow{2}{*}{Swin}   & Euclidean & 24.3          & 33.7          & 28.5          & 40.1          & 32.8          & 9.7           & 28.2          \\
                        & Geodesic  & \textbf{31.0} & \textbf{67.1} & \textbf{46.8} & \textbf{56.9} & \textbf{48.7} & \textbf{31.9} & \textbf{47.1} \\
\hline
\end{tabular}
}
\end{wraptable}

%% file: Tables/chap-6/eagle/ablation_prompting.tex
\begin{table*}[!b]
\setlength{\tabcolsep}{2pt}
\centering
\caption{Effectiveness of Our Cross-view Adaptation Losses and Prompting Mechanism.}
\label{tab:eagle-ablation_prompting}
\resizebox{1.0\textwidth}{!}{  
\begin{tabular}{|ccc|cccccc|ccccccc|}
\hline
\multirow{2}{*}{\begin{tabular}[c]{@{}c@{}}With\\Prompt\end{tabular}}
& \multirow{2}{*}{\begin{tabular}[c]{@{}c@{}}Cross-View\\Adapt\end{tabular}}
& \multirow{2}{*}{\begin{tabular}[c]{@{}c@{}}View\\Condition\end{tabular}} 
& \multicolumn{6}{c|}{SYNTHIA $\to$ UAVID}          & \multicolumn{7}{c|}{GTA $\to$ UAVID}                         \\ \cline{4-16} 
                        &                             &                                 & Road & Building & Car  & Tree & Person & mIoU & Road & Building & Car  & Tree & Terrain & Person & mIoU \\ \hline
\xmark                      & \xmark                          & \xmark                              & 8.1  & 19.1     & 7.4  & 30.3 & 1.3    & 13.2 & 7.5  & 13.0     & 2.7  & 26.8 & 26.6    & 1.0    & 12.9 \\
\xmark                      & \cmark                         & \xmark                              & \textbf{31.4} & \textbf{75.1}     & \textbf{57.5} & \textbf{59.2} & \textbf{19.5}   & \textbf{48.6} & \textbf{22.9} & \textbf{64.6}     & \textbf{37.8} & \textbf{52.8} & \textbf{48.5}    & \textbf{13.8}   & \textbf{40.1} \\
\multicolumn{3}{|c|}{\greycell Supervised}                                    & \greycell 75.5 & \greycell 91.6     & \greycell 79.1 & \greycell 77.7 & \greycell 42.1   & \greycell 73.2 & \greycell 76.8 & \greycell 91.8     & \greycell 81.1 & \greycell 77.6 & \greycell 67.8    & \greycell 43.4   & \greycell 73.1 \\ \hline
\cmark                     & \xmark                          & \xmark                              & 15.7 & 27.8     & 15.7 & 34.1 & 7.7    & 20.2 & 16.6 & 26.8     & 7.2  & 30.0 & 21.7    & 6.0    & 18.1 \\
\cmark                     & \cmark                         & \xmark                              & 36.8 & 75.5	& 61.3 & 60.8 & 21.2 & 51.1 & 27.3	& 66.8	& 42.3	& 55.5	& 47.1	& 25.1	& 44.0 \\ 
\cmark                     & \cmark                         & \cmark                             & \textbf{38.4} & \textbf{76.1}     & \textbf{62.8} & \textbf{62.1} & \textbf{21.8}   & \textbf{52.2} & \textbf{29.2} & \textbf{67.1}     & \textbf{45.2} & \textbf{56.6} & \textbf{48.5}    & \textbf{27.9}   & \textbf{45.7} \\
\multicolumn{3}{|c|}{\greycell Supervised}                                    & \greycell 79.8 & \greycell 92.6     & \greycell 82.9 & \greycell 79.1 & \greycell 48.0   & \greycell 76.5 & \greycell 80.5 & \greycell 93.3     & \greycell 82.7 & \greycell 79.2 & \greycell 71.3    & \greycell 49.9   & \greycell 76.1 \\
\hline
\end{tabular}
}
\end{table*}

%% file: Tables/chap-6/eagle/ablation_alpha.tex
\begin{table}[!t] %
\centering
\caption{Effectiveness of Linear Scale Factors, i.e., $\alpha$ and $\gamma$, and Subspace dimension $D$.}
\label{tab:eagle-ablation_alpha}
\resizebox{1.0\textwidth}{!}{  
\begin{tabular}{|l|cccccc|ccccccc|}
\hline
\multirow{2}{*}{Factor}    & \multicolumn{6}{c|}{SYNTHIA $\to$ UAVID} & \multicolumn{7}{c|}{GTA $\to$ UAVID} \\
\cline{2-14}
  & Road          & Building      & Car           & Tree          & Person        & mIoU          & Road          & Building      & Car           & Tree          & Terrain       & Person        & mIoU          \\
\hline
$\alpha =$ 0.5                           & 35.9          & 73.6          & 59.6          & 57.2          &  20.8          & 49.4          & 25.4          & 63.9          & 40.5          & 44.6          & 46.8          & 25.6          & 41.1          \\
$\alpha =$ 1.0                           & 37.8          & 75.8          & 61.0          & 60.7          &  21.6          & 51.4          & 26.9          & 64.3          & 41.8          & 48.0          & 47.2          & 26.3          & 42.4          \\
$\alpha =$ 1.5                           & \textbf{38.4} & \textbf{76.1} & \textbf{62.8} & \textbf{62.1} & \textbf{21.8} & \textbf{52.2} & \textbf{29.2} & \textbf{67.1} & \textbf{45.2} & \textbf{56.6} & \textbf{48.5} & \textbf{27.9} & \textbf{45.7} \\
$\alpha =$ 2.0                           & 36.9          & 74.8          & 60.7          & 59.4          &  21.2          & 50.6          & 28.1          & 66.0          & 44.2          & 51.8          & 48.1          & 27.3          & 44.2          \\
\hline
$\gamma =$ 0.5                           & 37.6          & 75.5          & 60.6          & 60.0          &  21.4          & 51.1          & 27.8          & 65.1          & 42.7          & 51.7          & 47.7          & 26.8          & 43.6          \\
$\gamma =$ 1.0                           & \textbf{38.4} & \textbf{76.1} & \textbf{62.8} & \textbf{62.1} & \textbf{21.8} & \textbf{52.2} & \textbf{29.2} & \textbf{67.1} & \textbf{45.2} & \textbf{56.6} & \textbf{48.5} & \textbf{27.9} & \textbf{45.7} \\
$\gamma =$ 1.5                           & 36.2          & 75.3          & 61.6          & 58.5          &  20.5          & 50.4          & 28.5          & 66.0          & 44.2          & 54.7          & 47.9          & 27.5          & 44.8          \\
$\gamma =$ 2.0                           & 36.0          & 74.2          & 60.0          & 58.1          &  20.7          & 49.8          & 26.8          & 64.6          & 42.8          & 52.5          & 47.0          & 26.8          & 43.4         \\
\hline
$D=96$              & 36.6 & 72.0     & 60.6 & 57.7 & 21.3   & 49.6 & 26.8 & 60.6     & 42.2 & 50.7 & 46.5    & 27.0   & 42.3 \\
$D=128$             & 37.1 & 72.7     & 61.3 & 58.9 & 21.4   & 50.3 & 27.7 & 62.4     & 43.0 & 52.9 & 47.1    & 27.3   & 43.4 \\
$D=256$             & \textbf{38.4} & \textbf{76.1}     & \textbf{62.8} & \textbf{62.1} & \textbf{21.8}   & \textbf{52.2} & \textbf{29.2} & \textbf{67.1}     & \textbf{45.2} & \textbf{56.6} & \textbf{48.5}    & \textbf{27.9}   & \textbf{45.7} \\
$D=512$             & 37.9 & 75.8     & 62.4 & 61.1 & 21.4   & 51.7 & 28.2 & 64.9     & 44.2 & 54.1 & 47.9    & 27.6   & 44.5 \\
\hline
\end{tabular}
}
\end{table}

%% file: Tables/chap-6/eagle/domain_adaptation_exp.tex
\begin{table}[!b]
\setlength{\tabcolsep}{1.5pt}
\centering
\caption{Comparisons with Domain Adaptation Approaches (Without Prompting).}
\label{tab:eagle-uda_exp}
\resizebox{1.0\textwidth}{!}{  
\begin{tabular}{|c|l|cccccc|ccccccc|}
\hline
\multirow{2}{*}{Network}     & \multirow{2}{*}{Method} & \multicolumn{6}{c|}{SYNTHIA $\to$ UAVID} & \multicolumn{7}{c|}{GTA $\to$ UAVID} \\
\cline{3-15} 
                             &                         & Road          & Building      & Car           & Tree          & Person        & mIoU          & Road          & Building      & Car           & Tree          & Terrain       & Person        & mIoU          \\
\hline
\multirow{8}{*}{DeepLab}     & AdvEnt \cite{vu2019advent} & 4.7           & 63.2          & 31.7          & 48.6          & 11.4          & 31.9          & 2.0           & 30.3          & 14.9          & 29.8          & 41.5          & 1.8           & 20.0          \\
                             & DADA \cite{vu2019dada}                    & 10.7          & 63.1          & 32.9          & 50.0          & 16.2          & 34.6          & -             & -             & -             & -             & -             & -             & -             \\
                             & BiMaL \cite{dat2021bimal_iccv}                   & 5.4           & 62.1          & 34.8          & 50.7          & 12.7          & 33.1          & 1.3           & 44.6          & 10.1          & 49.2          & 20.0          & 10.9          & 22.7          \\
                             & SAC \cite{araslanov2021dasac}                     & 13.9          & 64.0          & 18.7          & 48.0          & 15.6          & 32.0          & 4.5           & 36.9          & 7.8           & 47.9          & 44.1          & 7.8           & 24.8          \\
                             & ProDA \cite{zhang2021prototypical}                   & 10.6          & 64.7          & 34.1          & 44.5          & 17.0          & 34.2          & 6.9           & 50.6          & 28.4          & 25.5          & 38.7          & 4.5           & 25.8          \\
                             & CROVIA \cite{truong2023crovia}                  & 10.6          & 65.7          & 51.7          & 55.6          & 17.0          & 40.1          & 18.2          & 49.8          & 10.4          & 48.1          & 44.0          & 8.0           & 29.7 \\
                             & \textbf{EAGLE}           & \textbf{29.9} & \textbf{65.7} & \textbf{55.5} & \textbf{56.8} & \textbf{18.3} & \textbf{45.2} & \textbf{20.5} & \textbf{53.0} & \textbf{37.6} & \textbf{50.7} & \textbf{45.3} & \textbf{13.0} & \textbf{36.7} \\

 & \greycell Supervised              & \greycell 67.2          & \greycell 90.7          & \greycell 74.0          & \greycell 76.3          & \greycell 36.8          & \greycell 69.0          & \greycell 68.1          & \greycell 91.0          & \greycell 77.5          & \greycell 75.7          & \greycell 62.2          & \greycell 35.8          & \greycell 68.4          \\
\hline
\multirow{4}{*}{DAFormer}    & DAFormer \cite{hoyer2022daformer}                & 7.3           & 75.1          & 51.7          & 48.0          & 15.1          & 39.4          & 
15.3          & 51.6          & 33.6          & 27.8          & 38.5          & 4.0           & 28.5              \\
& MIC \cite{hoyer2023mic}	& 10.8	& 76.4	& 53.3	& 52.7	&	16.0	& 41.8	& 20.7	& 51.9	& 13.3	& 55.2	& 44.8	& 9.3	& 32.5 \\
                             & CROVIA \cite{truong2023crovia}                  & 16.3          & 75.1          & 59.6          & 60.0          & 19.1          & 46.0          & 20.5          & 56.1          & 37.6          & 50.7          & 45.3          & 10.9          & 36.8          \\
                             & \textbf{EAGLE}           & \textbf{30.6} & \textbf{75.3} & \textbf{59.7} & \textbf{63.1} & \textbf{25.3} & \textbf{50.8} & \textbf{23.9} & \textbf{65.0} & \textbf{38.5} & \textbf{53.5} & \textbf{49.3} & \textbf{14.1} & \textbf{40.7} \\
              
                             & \greycell Supervised              & \greycell 78.0          & \greycell 91.2          & \greycell 79.7          & \greycell 77.5          & \greycell 44.2          & \greycell 74.1          & \greycell 79.0          & \greycell 92.8          & \greycell 81.9          & \greycell 78.4          & \greycell 70.3          & \greycell 45.7          & \greycell 74.7          \\
\hline
\multirow{2}{*}{Mask2Former} & \textbf{EAGLE}           & \textbf{31.4} & \textbf{75.1} & \textbf{57.5} & \textbf{59.2} & \textbf{19.5} & \textbf{48.6} & \textbf{22.9} & \textbf{64.6} & \textbf{37.8} & \textbf{52.8} & \textbf{48.5} & \textbf{13.8} & \textbf{40.1} \\
                    
                             & \greycell  Supervised              & \greycell 75.5          & \greycell 91.6          & \greycell 79.1          & \greycell 77.7          & \greycell 42.1          & \greycell 73.2          & \greycell 76.8          & \greycell 91.8          & \greycell 81.1          & \greycell 77.6          & \greycell 67.8          & \greycell 43.4          & \greycell 73.1         \\
\hline
\end{tabular}
}
\end{table}

%% file: Tables/chap-6/eagle/openvocab-seenunseen.tex
\begin{table*}[!t]
\begin{minipage}[t]{0.6\textwidth}

  \setlength{\tabcolsep}{2.3pt}
\centering
\caption{Comparisons with Open-vocab \mbox{Semantic Segmentation}.}
\label{tab:eagle-open_vocab_exp}
\resizebox{1.0\textwidth}{!}{  
\begin{tabular}{|c|l|cccccc|ccccccc|}
\hline
\multirow{2}{*}{}          & \multirow{2}{*}{Method} & \multicolumn{6}{c|}{SYNTHIA $\to$ UAVID}           & \multicolumn{7}{c|}{GTA $\to$ UAVID}       \\
\cline{3-15}
          & & Road          & Building      & Car           & Tree          & Person        & mIoU          & Road          & Building      & Car           & Tree          & Terrain       & Person        & mIoU          \\
\hline
      & DenseCLIP  & 14.6 & 27.2 & 14.7 & 32.6 & 7.1  & 19.2 & 16.1 & 26.0 & 6.4  & 28.3 & 20.8 & 5.9  & 17.3 \\
      & DenseCLIP + AdvEnt                 & 27.7 & 62.0 & 48.6 & 40.2 & 18.1 & 39.3 & 25.5 & 39.4 & 20.6 & 41.4 & 38.7 & 14.9 & 30.1 \\
      & DenseCLIP + SAC                    & 28.6 & 63.5 & 51.5 & 43.4 & 18.3 & 41.1 & 17.2 & 52.3 & 30.8 & 35.7 & 41.9 & 15.3 & 32.2 \\
\multirow{-4}{*}{\rot{\begin{tabular}[c]{@{}c@{}}FPN\\ResNet\end{tabular}}}
& DenseCLIP + Cross-View             & \textbf{32.4}                & \textbf{67.0}                & \textbf{55.3}                & \textbf{50.2}                & \textbf{19.6}                & \textbf{44.9}                & \textbf{19.6} &\textbf{ 58.7} & \textbf{33.9} & \textbf{41.5} & \textbf{43.9} & \textbf{16.2} & \textbf{35.6} \\
\hline

      & DenseCLIP  & 17.2 & 28.9 & 16.9 & 37.3 & 8.6  & 21.8 & 17.7 & 28.3 & 8.9  & 33.1 & 23.5 & 6.3  & 19.6 \\
      & DenseCLIP + AdvEnt                 & 28.1 & 67.0 & 49.9 & 39.8 & 17.2 & 40.4 & 16.5 & 51.3 & 29.8 & 33.9 & 41.0 & 15.2 & 31.3 \\
      & DenseCLIP + SAC                    & 29.1 & 67.4 & 51.6 & 44.4 & 17.8 & 42.1 & 17.9	& 53.9	& 32.5	& 37.8	& 42.7	& 15.5	& 33.4 \\
\multirow{-4}{*}{\rot{\begin{tabular}[c]{@{}c@{}}FPN\\ViT\end{tabular}}}     & DenseCLIP + Cross-View             & \textbf{31.6}                & \textbf{71.4}                & \textbf{53.9}                & \textbf{50.1}                & \textbf{21.9}                & \textbf{45.8}                & \textbf{20.6} & \textbf{60.8} & \textbf{35.8} & \textbf{45.0} & \textbf{44.6} & \textbf{16.8} & \textbf{37.3} \\
\hline
      & FreeSeg    & 18.4 & 30.0 & 17.9 & 41.5 & 8.9  & 23.4 & 18.0 & 28.7 & 9.8  & 33.9 & 24.0 & 6.3  & 20.1 \\
      & FreeSeg + AdvEnt                   & 30.0 & 71.2 & 54.0 & 43.3 & 18.0 & 43.3 & 20.3 & 60.6 & 35.6 & 42.3 & 44.7 & 16.6 & 36.7 \\
      & FreeSeg + SAC                      & 32.0 & 73.3 & 56.6 & 50.4 & 19.2 & 46.3 & 22.1 & 62.5 & 38.1 & 45.7 & 45.6 & 17.4 & 38.6 \\
      & FreeSeg + Cross-View               &
      \textbf{36.4} & \textbf{76.5}	& \textbf{60.6}	& \textbf{60.5} & \textbf{22.6} & \textbf{51.3}                & 
      \textbf{25.7}	& \textbf{66.8} & 	\textbf{43.1}	& \textbf{57.2}	& \textbf{47.5}	& \textbf{26.2} & 	\textbf{44.4} \\
     \cdashline{2-15}
  & EAGLE                    &  36.8 & 75.5	& 61.3 & 60.8 & 21.2 & 51.1 & 27.3	& 66.8	& 42.3	& 55.5	& 47.1	& 25.1	& 44.0 \\ 
      & EAGLE + View Condition   & \textbf{38.4} & \textbf{76.1} & \textbf{62.8} & \textbf{62.1} & \textbf{21.8} & \textbf{52.2} & \textbf{29.2} & \textbf{67.1} & \textbf{45.2} & \textbf{56.6} & \textbf{48.5} & \textbf{27.9} & \textbf{45.7} \\
\multirow{-7}{*}{\rot{Mask2Former}}      & \greycell Supervised            & \greycell 79.8 & \greycell 92.6     & \greycell 82.9 & \greycell 79.1 & \greycell 48.0   & \greycell 76.5 & \greycell 80.5 & \greycell 93.3     & \greycell 82.7 & \greycell 79.2 & \greycell 71.3    & \greycell 49.9   & \greycell 76.1 \\

\hline
\end{tabular}
}
	   
\end{minipage} 
\begin{minipage}[t]{0.39\textwidth}

\centering
\caption{Comparisons with Open-vocab Segmentation on Seen (mIoU$^{S}$) and Unseen (mIoU$^{U}$) Classes.}
\label{tab:eagle-open_vocab_unseen_exp}
 \resizebox{1.0\textwidth}{!}{  
\begin{tabular}{|c|l|cc|cc|}
\hline
\multirow{2}{*}{}     & \multirow{2}{*}{Method} & \multicolumn{2}{c|}{SYNTHIA $\to$ UAVID} & \multicolumn{2}{c|}{GTA $\to$ UAVID} \\
\cline{3-6}
                             &                           & mIoU$^{S}$ & mIoU$^{U}$       & mIoU$^{S}$     & mIoU$^{U}$     \\
\hline
 \multirow{3}{*}{\rot{\begin{tabular}[c]{@{}c@{}}FPN\\ResNet\end{tabular}}} 
 & DenseCLIP + AdvEnt        & 54.7      & 30.4        & 40.9      & 30.3        \\
 & DenseCLIP + SAC           & 56.2      & 32.1        & 44.1      & 31.4        \\
 & DenseCLIP + Cross-View    & \textbf{58.3}      & \textbf{35.1}        & \textbf{46.3}      & \textbf{34.1}        \\
 \hline
 \multirow{3}{*}{\rot{\begin{tabular}[c]{@{}c@{}}FPN\\ViT\end{tabular}}} 
 & DenseCLIP + AdvEnt        & 55.2      & 31.2        & 44.5      & 31.7        \\
 & DenseCLIP + SAC           & 56.5      & 33.2        & 46.1      & 33.4        \\
 & DenseCLIP + Cross-View    & \textbf{59.0}      & \textbf{36.6}        & \textbf{48.5}      & \textbf{35.6}        \\
 \hline
 \multirow{6}{*}{\rot{Mask2Former}} 
 & FreeSeg + AdvEnt          & 55.8      & 31.1        & 46.5      & 33.5        \\
 & FreeSeg + SAC             & 58.0      & 34.8        & 48.6      & 36.1        \\
 & FreeSeg + Cross-View      & \textbf{60.2}      & \textbf{38.3}       & \textbf{50.7}      & \textbf{38.2}        \\
 \cdashline{2-6}
 & EAGLE                      & 60.6      & 37.7        & 50.5      & 37.5        \\
 & EAGLE + View Condition     & \textbf{61.6}      & \textbf{39.3}        & \textbf{51.4}      & \textbf{39.6}        \\
 & \greycell Fully Supervised                & \greycell 85.1      & \greycell 63.6        & \greycell 81.9      &  \greycell 64.5       \\
\hline
\end{tabular}
} 

\end{minipage}
\end{table*}

%% file: Tables/chap-6/eagle/real2real_exp.tex
\begin{wraptable}{r}{0.5\textwidth}
\caption{Comparison with Adaptation Methods and Open-Vocab Segmentation on Real-to-Real Setting.}
\label{tab:eagle-real2real}
\setlength{\tabcolsep}{2pt}
\resizebox{0.5\textwidth}{!}{

\begin{tabular}{|c|l|ccccccc|}
\hline
\multirow{2}{*}{Setting}                                                                          & \multirow{2}{*}{Method}    & \multicolumn{7}{c|}{BDD $\to$ UAVID}                                                                                                                                                      \\ \cline{3-9} 
                                                                                                  &                            &  {Road} &  {Building} &  {Car}  &  {Tree} &  {Terrain} &  {Person} & mIoU \\ 
                                                                                                  \hline

\multirow{5}{*}{\begin{tabular}[c]{@{}c@{}}Unsupervised\\Domain\\Adaptation\end{tabular}} & No Adaptation              &  {19.2} &  {8.5}      &  {34.6} &  {18.4} &  {13.6}    &  {4.0}    & 16.4 \\

       & 

    BiMaL \cite{dat2021bimal_iccv} &  {19.5} &  {52.4}     &  {35.1} &  {50.4} &  {46.0}    &  {10.2}   & 35.6 \\ 
  
  & EAGLE (DeepLab)            &  \textbf{24.0} &  \textbf{53.8}     &  \textbf{39.0} &  \textbf{52.2} &  \textbf{48.3}    &  \textbf{16.9}   & \textbf{39.0} \\ 
  \cline{2-9} 
  
  & DAFormer \cite{hoyer2022daformer} &  {25.8} &  {65.4}     &  {38.7} &  {54.5} &  {51.3}    &  {14.8}   & 41.8 \\ 
  & EAGLE (DAFormer)          &  \textbf{29.0} &  \textbf{66.1}     &  \textbf{41.5} &  \textbf{55.6} &  \textbf{53.3}    &  \textbf{21.5}   & \textbf{44.5} \\ 
\hline
\multirow{3}{*}{\begin{tabular}[c]{@{}c@{}}Open-Vocab\\Seg\end{tabular}} & 
        DenseCLIP + Cross-View          &  {25.9} &  {60.9}     &  {39.5} &  {35.5} &  {47.1}    &  {33.9}   & 40.5 \\ 
      & FreSeg + Cross-View        &  {32.6} &  {67.3}     &  {47.9} &  {51.8} &  {50.3}    &  {37.2}   & 47.9 \\ 
      & EAGLE                      &  \textbf{35.4} &  \textbf{68.9}     &  \textbf{50.6} &  \textbf{59.2} &  \textbf{51.7}    &  \textbf{38.6}   & \textbf{50.7} \\ 
\hline
\end{tabular}
}
\end{wraptable}

%% file: Chapters/Sections/chap-5/cvar.tex
\section{Cross-view Video Understanding From Exocentric to Egocentric Perspective}
\label{sec:paper-cvar}

\setcounter{propositioncounter}{0}
\setcounter{remarkcounter}{0}

\begin{wrapfigure}{r}{0.5\textwidth}
    \centering
    \includegraphics[width=0.5\textwidth]{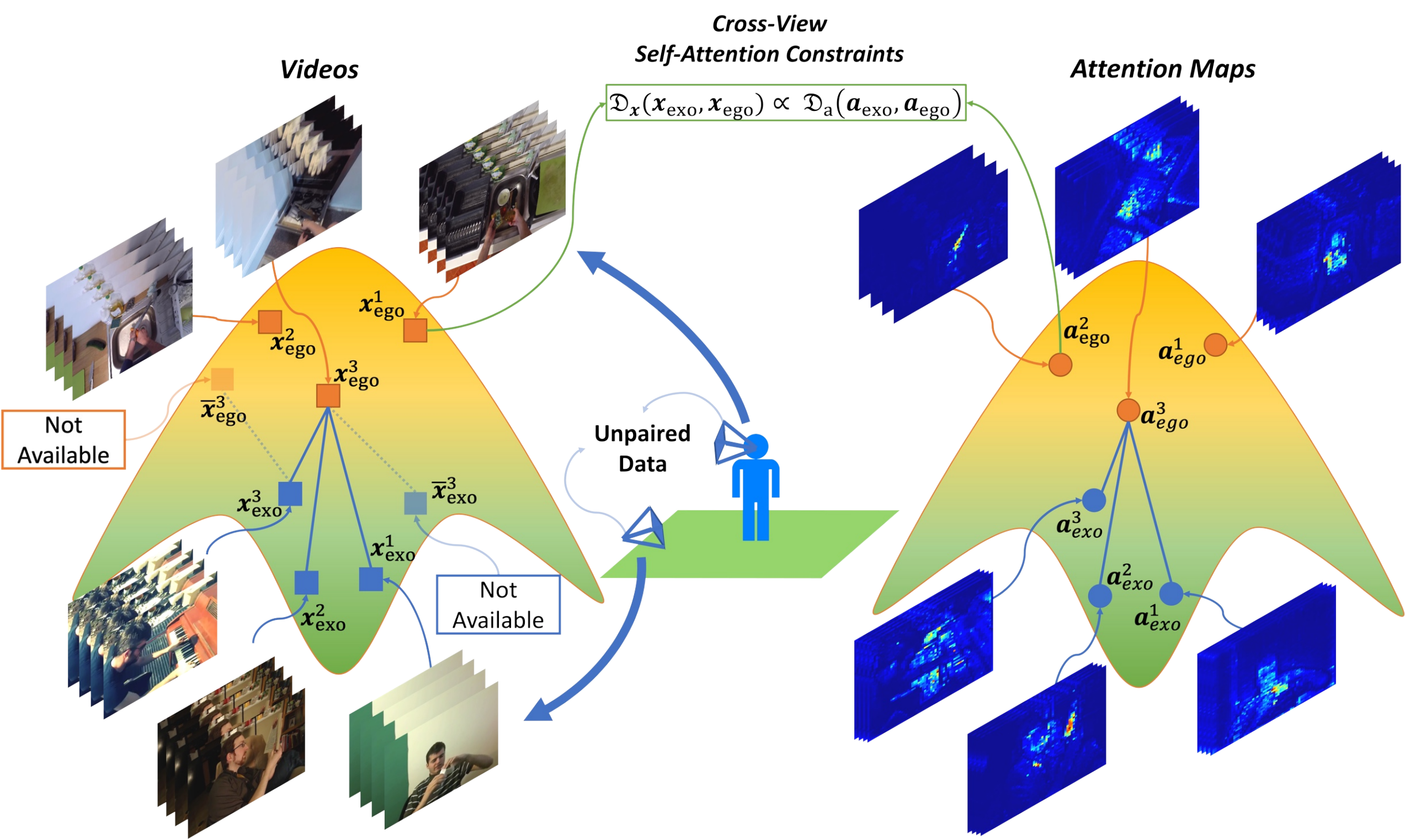}
    \caption{\textbf{The Cross-view Self-attention Constraints}. 
    Although under the setting of cross-view unpaired data where the corresponding video and its attention in the opposite view are inaccessible, our cross-view self-attention loss is proven to impose the cross-view constraints via unpaired samples based on the geometric properties between two camera positions.}
    \label{fig:cvar-figure_one}
\end{wrapfigure}
The properties of egocentric videos also bring new challenges to video analysis tasks. One of the main issues is the scale of datasets. It is well known that learning robust video models, e.g., action recognition models, usually requires a large amount of video data~\cite{swin, MViTv2, egovlp}. For example, the third-view action models are learned on the large-scale Kinetics-700~\cite{kinetic700} data comprising 650K videos over 700 classes. Meanwhile, the scale of egocentric video data is relatively small compared to third-view datasets, e.g., EPIC Kitchens~\cite{epic-100} only consists of 90K clips or Charades-Ego~\cite{charades-ego}  includes 68K clips of 157 classes. 
In addition, the egocentric video data lack variation, e.g., videos only in kitchens~\cite{epic-100} or daily indoor activities~\cite{charades-ego}. These problems pose a considerable challenge for learning robust video models on the first-view data.

Many prior works~\cite{egovlp, slowfast}  have improved the performance of action recognition models by adopting the pre-trained model on large-scale third-view datasets and fine-tuning it on the first-view dataset. However, these methods often ignore the unique characteristics of egocentric videos. Thus, they could meet the unaligned domain problems. Another method~\cite{ego-exo} has tried to alleviate this domain mismatch problem by introducing several additional egocentric tasks during the pre-training phase on the third-view datasets. However, this approach requires the labels of egocentric tasks on third-view data or relies on the off-the-shelf specific-task models. Domain adaptation methods~\cite{ego-exo, ganin2016domain, sigurdsson2018actor} have also been utilized to transfer the knowledge from the third-view to first-view data. Nevertheless, these methods still need to model the camera-view changes during the adaptation phase.

With the recent success of Vision Transformer, the self-attention mechanism is fundamental to building an efficient action recognition model. Still, fewer prior works have focused on leveraging self-attention to model action recognition from the third-view to first-view data.
Moreover, modeling the change in camera positions across views is also one of the primary factors in sufficiently developing a learning approach from the exocentric to egocentric view. 
Therefore, considering these characteristics, we introduce a novel cross-view learning approach to effectively model the self-attention mechanism to transfer the knowledge learned on third-view to first-view data.
Our proposed approach first considers the geometric correlation between two camera views. Then, the cross-view geometric correlation constraint is further embedded into the self-attention mechanism so that the model can generalize well from the exocentric to the egocentric domain. Figure~\ref{fig:cvar-figure_one} illustrates the cross-view self-attention constraint.

In this chapter, we introduce a novel \textbf{C}ross-\textbf{V}iew learning approach to \textbf{A}ction \textbf{R}ecognition (CVAR) via effectively transferring knowledge from the exocentric video domain to the egocentric one. By analyzing the role of the self-attention mechanism and the change of camera position across views, we introduce a new geometric cross-view constraint for correlation between videos and attention maps. Then, from the proposed cross-view restriction, we present a novel cross-view self-attention loss that models the cross-view learning into the self-attention mechanism. Our proposed loss allows the Transformer-based model to adapt knowledge and generalize from the third-view to first-view video data. The cross-view correlations of videos and attention maps are further enhanced using the deep metric and the Jensen-Shannon divergence metric, respectively, that capture deep semantic information.

\subsection{Cross-view Learning in Action Recognition}

Let $\mathbf{x}_{exo} \in \mathbb{R}^{T \times H \times W \times 3}$ be a third-view (exocentric) video and $\mathbf{y}_{exo} \in \mathcal{Y}_{exo}$ be its corresponding ground-truth class, $\mathcal{Y}_{exo}$ is the set of classes in the exocentric dataset. Similarly, $\mathbf{x}_{ego} \in \mathbb{R}^{T \times H \times W \times 3}$ be a first-view (egocentric) video and $\mathbf{y}_{ego} \in \mathcal{Y}_{ego}$ be its corresponding ground-truth class, $\mathcal{Y}_{exo}$ is the set of classes in the egocentric dataset. Let $F: \mathbb{R}^{T \times H \times W \times 3} \to \mathbb{R}^{D}$ be the backbone network that maps a video into the deep representation, $C_{exo}$ and $C_{ego}$ are the classifier of exocentric and egocentric videos that predict the class probability from the deep representation. Then, the basic learning approach to learning the action model from the exocentric to the egocentric view can be formulated as a supervised objective, as in Eqn.~\eqref{eqn:cvar-general_opt}.
\begin{equation} \label{eqn:cvar-general_opt}
\small
\begin{split}
    \arg\min_{\theta_F, \theta_{C_{exo}}, \theta_{C_{ego}}}[\mathbb{E}_{\mathbf{x}_{exo}, \mathbf{y}_{ego}}\mathcal{L}_{ce}(C_{exo}(F(\mathbf{x}_{exo})), \mathbf{y}_{exo}) \\ 
    + \mathbb{E}_{\mathbf{x}_{ego}, \mathbf{y}_{ego}}\mathcal{L}_{ce}(C_{ego}(F(\mathbf{x}_{ego})), \mathbf{y}_{ego})]
\end{split}
\end{equation}
where $\theta_F, \theta_{C_{exo}}, \theta_{C_{ego}}$ are the network parameters, $\mathcal{L}_{ce}$ is the supervised loss (i.e., cross-entropy loss). Several prior approaches~\cite{ego-exo, sigurdsson2018actor} have adopted this learning approach to learn a cross-view action recognition model. Then, other prior methods have further improved the performance of models by using a large pretrained model ~\cite{swin, slowfast}, 
domain adaptation ~\cite{ganin2016domain}, learning a joint embedding between two views~\cite{sigurdsson2018actor}, learning auxiliary egocentric tasks ~\cite{ego-exo}.

Although these prior approaches~\cite{ego-exo, swin, MViTv2, slowfast} showed their potential in improving performance, they have not effectively addressed the problem of cross-view learning. In particular, domain adaptation methods~\cite{ganin2016domain} are often employed in the context of environment changes (e.g., simulation to real data), and the camera views are assumed on the same position (either third view or first view). However, there is a vast difference in videos between the third view and the first view. Thus, domain adaptation is considered less effective in the cross-view setting.
Meanwhile, fine-tuning the first-view action model on the large pretrained models~\cite{swin, slowfast} usually relies on the deep representation learned from the large-scale third-view data. However, these deep representations do not have any guarantee mechanism well generalized in the first-view video. Also, learning the join embedding or auxiliary egocentric tasks~\cite{ego-exo} suffer a similar problem due to their design of learning approaches without considering camera changes. In addition, it requires a pair of views of video data during training.
Therefore, to effectively learn the cross-view action recognition model, the learning approach should consider the following properties:
(1) the geometric correlation between the third view and the first view has to be considered during the learning process,
(2) the mechanism that guarantees the knowledge learned is well generalized from the third view to the first view.

\subsubsection{Cross-view Geometric Correlation in Attentions}

With the success of Vision Transformer in action recognition~\cite{MViTv2, swin, vit},  the self-attention mechanism is the key to learning the robust action recognition models. Therefore, our work proposes explicitly modeling cross-view learning in action recognition models through the self-attention mechanism. First, we revise the geometric correlation of the exocentric and egocentric views in obtaining the videos. Let us assume that $\mathbf{\bar{x}}_{ego}$ is the corresponding egocentric video of the exocentric video $\mathbf{x}_{exo}$,
and $\mathbf{K}_{exo}, [\mathbf{R}_{exo}, \mathbf{t}_{exo}]$ and $\mathbf{K}_{ego}, [\mathbf{R}_{ego}, \mathbf{t}_{ego}]$ are the camera (intrinsic and extrinsic) parameters of third and first views, respectively. Then, the procedure of obtaining the videos can be formed as a rendering function, as in Eqn.~\eqref{eqn:cvar-render}.
\begin{equation} \label{eqn:cvar-render}
\small
\begin{split}
    \mathbf{x}_{exo} = \mathcal{R}(\mathbf{K}_{exo}, [\mathbf{R}_{exo},\mathbf{t}_{exo}]), \quad
    \mathbf{\bar{x}}_{ego} = \mathcal{R}(\mathbf{K}_{ego}, [\mathbf{R}_{ego},\mathbf{t}_{ego}])
\end{split}
\end{equation}
where $\mathcal{R}$ is a rendering function that obtains the video with the given corresponding camera matrix and position.
Inspired our our prior analysis of cross-view camera transformation in \textbf{Remark \ref{rmk:camera_camera}}, there exists a geometric transformation $\mathcal{T}$ of videos (images) between two camera views as in Eqn.~\eqref{eqn:cvar-xego-t-exo}.
\begin{equation}\label{eqn:cvar-xego-t-exo}
\small
    \mathbf{\bar{x}}_{ego} = \mathcal{T}(\mathbf{x}_{exo}; \mathbf{T_K}, \mathbf{T_{Rt}})
\end{equation}

In our proposed method, we consider the action recognition backbone model $F$, which is designed as a Transformer with self-attention layers. 
Given a video, the input of the Transformer is represented by $N+1$ tokens, including $N = \frac{T}{K}\frac{H}{P}\frac{W}{P}$ non-overlapped patches ($K \times P \times P$ is the patch size of the token) of a video and a single classification token. 
Let $\mathbf{a}_{exo}, \mathbf{\bar{a}}_{ego} \in \mathbb{R}^{\frac{T}{K} \times \frac{H}{P} \times \frac{W}{P}}$ be an attention map of the video frames w.r.t the classification token extracted from the network $F$ on the inputs $\mathbf{x}_{exo}$ and $\mathbf{\bar{x}}_{exo}$, respectively. The attention maps $\mathbf{a}_{exo}$ and $\mathbf{\bar{a}}_{ego}$ represent the focus of the model on the video over time w.r.t to the model predictions.
It should be noted that the video and its attention map could be considered a pixel-wise correspondence. Even though the patch size is greater than 1 ($K, P > 1$), a single value in the attention map always corresponds to a group of pixels in its patch. 
Therefore, without a lack of generality, with the changes of cameras from the exocentric view to the eccentric view, we argue that the focuses of the model (the attention maps) also change correspondingly to the transitions of the videos across views because both videos are representing the same action scene from different camera views. As a result, the transformation between two attention maps, i.e., $\mathbf{a}_{exo}$ and $\mathbf{\bar{a}}_{ego}$, can also be represented by a transformation 
$\mathcal{T}'$ w.r.t. the camera transformation matrices $\mathbf{T_K}$ and $\mathbf{T_{Rt}}$.

\begin{tcolorbox}[colback=black!5!white,colframe=black!75!black,title=\textbf{Remark \showremarkcounter\label{rmk:video_attention}}]
\textbf{Remark 2: Cross-view Equivalent Transformation of Videos and Attentions}
{We argue that the transformations $\mathcal{T}$ and $\mathcal{T'}$ remain similar ($\mathcal{T} \equiv \mathcal{T'}$) as they are both the transformation interpolation based on the camera transformation matrices $\mathbf{T_K}$ and $\mathbf{T_{Rt}}$. Hence, the transformation $\mathcal{T}$ could be theoretically adopted to the attention transformation.}
\begin{equation}
\small
    \begin{split}
        \mathbf{\bar{a}}_{ego} = \mathcal{T}'(\mathbf{a}_{exo}; \mathbf{T_K}, \mathbf{T_{Rt}}) \equiv \mathcal{T}(\mathbf{a}_{exo}; \mathbf{T_K}, \mathbf{T_{Rt}})
    \end{split}
\end{equation}
\end{tcolorbox}

Then, we further consider the cross-view correlation between the videos and the attention maps.
Let $\mathcal{D}_x(\mathbf{x}_{exo}, \mathbf{\bar{x}}_{ego})$ and $\mathcal{D}_a(\mathbf{a}_{exo}, \mathbf{\bar{a}}_{ego})$ be the metrics measure the cross-view correlation in videos ($\mathbf{x}_{exo}, \mathbf{\bar{x}}_{ego}$) and attention maps ($\mathbf{a}_{exo}, \mathbf{\bar{a}}_{ego}$), respectively. 
Adopting our analysis of camera transformation and the equivalent of image and segmentation in Section \ref{sec:paper-eagle} and Remark \ref{rmk:video_attention}, in this work, the proportion between $\mathcal{D}_x(\mathbf{x}_{exo}, \mathbf{\bar{x}}_{ego})$ and $\mathcal{D}_a(\mathbf{a}_{exo}, \mathbf{\bar{a}}_{ego})$ can also be theorized as a linear relation and modeled by a linear scale $\alpha$ as in Eqn.~\eqref{eqn:cvar-cross_view_condition}.
\begin{equation} \label{eqn:cvar-cross_view_condition}
\small
\begin{split}
    \mathcal{D}_x(\mathbf{x}_{exo}, \mathbf{\bar{x}}_{ego}) &\propto \mathcal{D}_a(\mathbf{a}_{exo}, \mathbf{\bar{a}}_{ego})
    \Leftrightarrow
    \mathcal{D}_x(\mathbf{x}_{exo}, \mathbf{\bar{x}}_{ego})  = \alpha \mathcal{D}_a(\mathbf{a}_{exo}, \mathbf{\bar{a}}_{ego})
\end{split}
\end{equation}

\subsubsection{Unpaired Cross-View Self-Attention Loss}

\begin{figure*}
    \centering
    \includegraphics[width=1.0\textwidth]{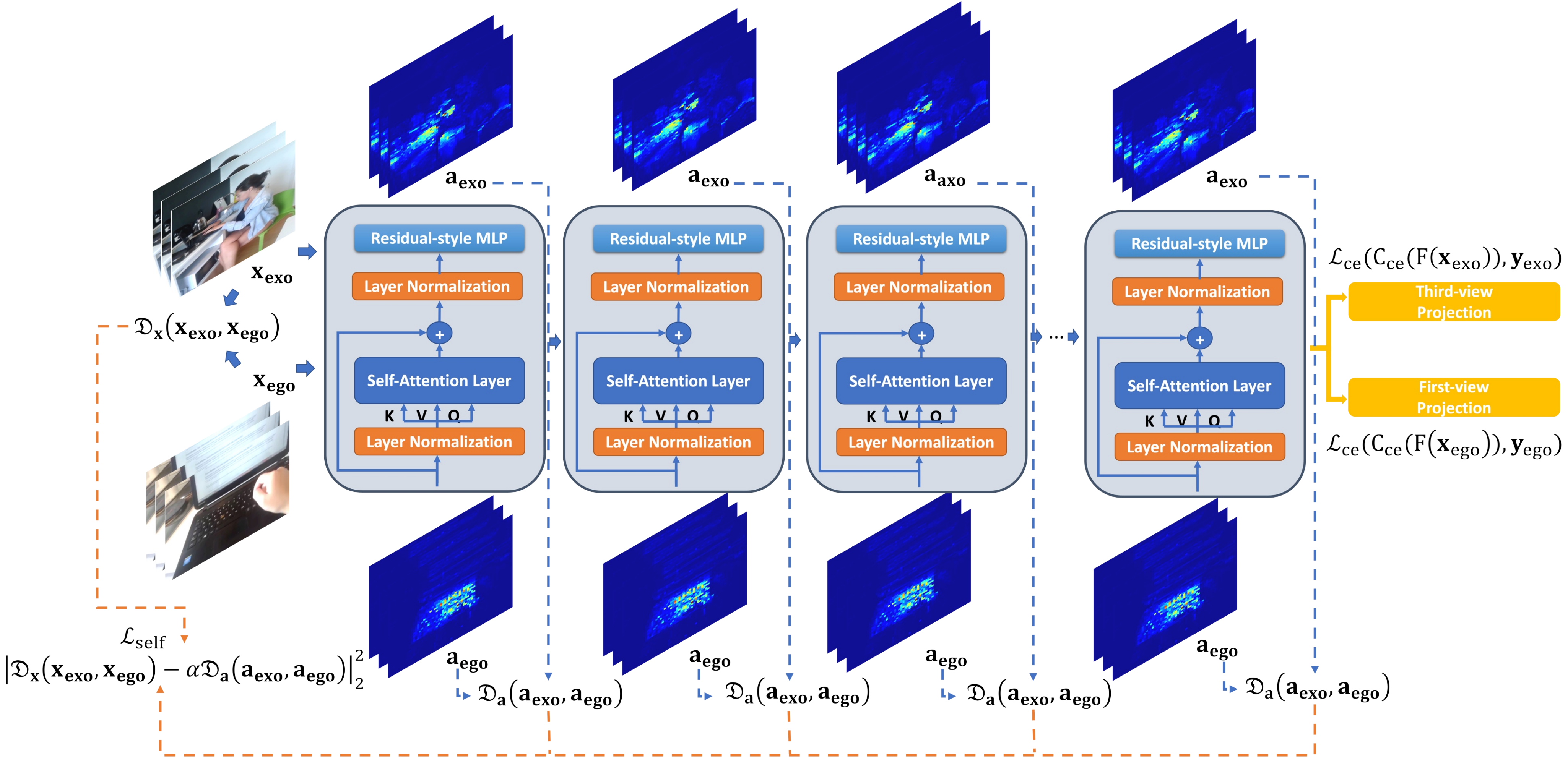}
    \caption{\textbf{The Proposed Framework.} The input videos $\mathbf{x}_{exo}$ and $\mathbf{x}_{ego}$ are first forwarded to Transformer $F$ followed by the corresponding classifiers $C_{exo}$ and $C_{ego}$, respectively. Then, the supervised cross-entropy loss $\mathcal{L}_{ce}$ is applied to the predictions produced by the model.
    Meanwhile, the attention maps of video inputs, i.e., $\mathbf{a}_{exo}$ and $\mathbf{a}_{ego}$,  are imposed by the cross-view self-attention loss $\mathcal{L}_{self}$. 
    }
    \label{fig:cvar-framework}
\end{figure*}

Eqn.~\eqref{eqn:cvar-cross_view_condition} defines a condition that explicitly models the self-attention correlation based on the geometric transformation across views. Thus, to efficiently learn the action recognition model from the exocentric to the egocentric view, Eqn \eqref{eqn:cvar-general_opt} can be optimized w.r.t the condition in Eqn.~\eqref{eqn:cvar-cross_view_condition}  and presented as in Eqn.~\eqref{eqn:cvar-general_opt_with_cond}.
\begin{equation} \label{eqn:cvar-general_opt_with_cond}
\small
\begin{split}
    \arg\min_{\theta_F, \theta_{C_{exo}}, \theta_{C_{ego}}}[\mathbb{E}_{\mathbf{x}_{exo}, \mathbf{y}_{ego}}\mathcal{L}_{ce}(C_{exo}(F(\mathbf{x}_{exo})), \mathbf{y}_{exo}) \\ 
    + \mathbb{E}_{\mathbf{x}_{ego}, \mathbf{y}_{ego}}\mathcal{L}_{ce}(C_{ego}(F(\mathbf{x}_{ego})), \mathbf{y}_{ego})] \\
    s.t. \quad\quad \mathcal{D}_x(\mathbf{x}_{exo}, \mathbf{\bar{x}}_{ego})  = \alpha \mathcal{D}_a(\mathbf{a}_{exo}, \mathbf{\bar{a}}_{ego})
\end{split}
\end{equation}

Then, similar to our cross-view learning of semantic segmentation presented in Section \ref{sec:paper-eagle}, our \textbf{\textit{Cross-view Self-Attention Loss}} on unpaired data can be formulated as in Eqn.~\eqref{eqn:cvar-loss_for_unpair}.
\begin{equation}\label{eqn:cvar-loss_for_unpair} 
\small
\begin{split}
    \mathcal{L}_{self} = \mathbb{E}_{\mathbf{x}_{exo}, \mathbf{x}_{ego}} \lambda||\mathcal{D}_{x}(\mathbf{x}_{exo}, \mathbf{x}_{ego}) - 
    \alpha\mathcal{D}_{a}(\mathbf{a}_{exo}, \mathbf{a}_{ego})||_2^2
\end{split}
\end{equation}
where $\lambda$ is the hyper-parameter controlling the relative importance of $\mathcal{L}_{self}$.
Intuitively, even though the pair samples between exocentric and egocentric views are inaccessible, the cross-view constraints between videos and attention maps can still be imposed by modeling the topological constraint among unpaired samples.

\noindent
\textbf{Cross-view Topological Preserving Property.} The proposed loss defined in Eqn.~\eqref{eqn:cvar-loss_for_unpair} to impose the cross-view correlation over all unpaired samples is a special case of the Gromov-Wasserstein~\cite{GW_distance} distance between the video and the attention map distributions where the association matrix has been pre-defined. As a result, our loss inherits these Gromov-Wasserstein properties to preserve the topological distributions between the video and attention space.
Remarkably, the cross-view topological structures of video distributions are preserved in cross-view attention distributions.

\subsubsection{The Choices of Correlation Metrics}

As shown in Eqn.~\eqref{eqn:cvar-loss_for_unpair}, the choice of correlation metric $\mathcal{D}_{x}$ and $\mathcal{D}_{a}$ is one of the primary factors directly influencing the performance of the action recognition models. The direct metrics, i.e., $\ell_2$, could be straightforwardly adopted for the correlation metric $\mathcal{D}_x$ and $\mathcal{D}_a$. However, this direct approach is ineffective because the deep semantic information of videos is not well modeled in the direct Euclidean metric $\ell_s$. To overcome this limitation, we propose designing $\mathcal{D}_x$ as the correlation metric on the deep latent spaces defined as in Eqn.~\eqref{eqn:cvar-dx_G}.
\begin{equation}
\label{eqn:cvar-dx_G}
\small
\mathcal{D}_{x}(\mathbf{x}_{exo}, \mathbf{x}_{ego}) = \mathcal{D}^G_{x}(\mathbf{x}_{exo}, \mathbf{x}_{ego}) = || G(\mathbf{x}_{exo}) - G(\mathbf{x}_{ego}) ||^2_2
\end{equation}
where $G: \mathbb{R}^{T \times H \times W \times 3} \to \mathbb{R}^{K}$ be the deep network trained on the large-scale dataset. %
Intuitively, measuring the correlation between two videos provides a higher level of semantic information since the deep representation extracted by the large pre-trained model $G$ captures more contextual information about the videos~\cite{johnson2016perceptual, duong2020vec2face}.

As $\mathcal{D}_a$ measures the correlation between two attention maps where, each of which is in the form of the probability distribution, $\mathcal{D}_a$ should be defined as the statistical distance to measure the correlation between two probabilistic attention maps comprehensively. Thus, we propose designing $\mathcal{D}_{a}$ as the Jensen-Shannon divergence defined in Eqn.~\eqref{eqn:cvar-da_JS}.
\begin{equation} \label{eqn:cvar-da_JS}
\small
\begin{split}
    \mathcal{D}_{a}(\mathbf{a}_{exo}, \mathbf{a}_{ego}) &= \mathcal{D}^{JS}_a(\mathbf{a}_{exo}, \mathbf{a}_{ego}) = \frac{1}{2}(\mathcal{D}_{KL}(\mathbf{a}_{exo} || \mathbf{a}_{ego}) + \mathcal{D}_{KL}(\mathbf{a}_{ego} || \mathbf{a}_{exo})) 
\end{split}
\end{equation}
where $\mathcal{D}_{KL}$ is the  Kullback–Leibler divergence.
To satisfy the cross-view distribution shift assumption aforementioned, 
the correlation metrics $\mathcal{D}_{x}$ and $\mathcal{D}_{a}$ are constrained by the threshold $\beta$, i.e., $\mathcal{D}_{x}(\mathbf{x}_{exo}, \mathbf{x}_{ego}) = \min\left(\mathcal{D}^G_{x}(\mathbf{x}_{exo}, \mathbf{x}_{ego}), \beta\right)$ and $\mathcal{D}_{a}(\mathbf{a}_{exo}, \mathbf{a}_{ego}) = \min\left(\mathcal{D}^{JS}_{a}(\mathbf{a}_{exo}, \mathbf{a}_{ego}), \beta\right)$. In our experiments, the value of $\beta$ is set to $200$.

\subsection{Experimental Results}

This section first briefly presents the datasets and the implementation details in our experiments. Then, we analyze the effectiveness of the approach in ablative experiments, followed by comparing results with prior methods on the standard benchmarks of first-view action recognition.

\subsubsection{Datasets and Implementation Details}

Following the common practice in action recognition~\cite{swin, slowfast, timesformer}, Kinetics has been used as the third-view dataset in our experiment due to its large scale and diverse actions. 
To evaluate the effectiveness of our approach, we use EPIC-Kitchens and Charades-Ego as our first-view datasets. These two datasets are currently known as large-scale and challenging benchmarks in egocentric action recognition.

\input{Tables/chap-6/cvar/scale-ablation}
\noindent
\textbf{Datasets.} Charades-Ego~\cite{charades-ego} is a first-view action recognition dataset that consists of 157 action classes with 68K clips. EPIC-Kitchens-55~\cite{epic} is a large-scale multi-task egocentric dataset of daily activities in kitchens. 
The action recognition task includes 55 hours of 39K clips and is annotated by interactions between 352 nouns and 125 verbs.
EPIC-Kitchens-100~\cite{epic-100} is a larger version of the EPIC-Kitchens-55 where it is extended to 100 hours of 90k action clips. Each single action segment is annotated by an action of 97 verbs and 300 nouns. NTU RGB+D~\cite{shahroudy2016ntu} is the RGB-D human action recognition dataset. The dataset consists of $56,880$ samples of $60$ action classes collected from $40$ subjects. Each action is captured using three cameras with different angles, i.e., $-45^o$, $0^o$, and $+45^o$.

\noindent
\textbf{Evaluation Metrics.} Our experiments follow the standard benchmarks of the Charades-Ego and EPIC-Kitchens for action recognition. We report the mean average precision (mAP) in the Charades-Ego~\cite{charades-ego} experiments and Top 1 and Top 5 accuracy of verb, noun, and action predictions of the validation set in EPIC-Kitchens~\cite{epic-100, epic} experiments.

\input{Tables/chap-6/cvar/metric-ablation}
\noindent
\textbf{Implementation.}
In our work, we adopt the design of the Vision Transformation Base model (ViT-B)~\cite{vit} for our Transformer backbone. Our model is implemented in Python using the PyTorch and PySlowFast~\cite{pyslowfast} frameworks. The input video of our network consists of $T = 16$ frames sampled at the frame rate of $1/4$, and the input resolution of each video frame is $H\times W = 224\times 224$. Each video is tokenized by the non-overlapping patch size of $K \times P \times P = 2 \times 16 \times 16$. Each token is projected by an embedding where the dimension length of the embedding is set to $768$. 
Our model has $12$ Transformer layers, and the number of heads in each self-attention layer is set to $8$. 
The SGD and cosine learning policy is utilized in our training, where the base learning rate is set to $0.00125$ with $50$ epoches. 
All of our models are trained on the four 40GB-VRAM A100 GPUs, and the batch size in each GPU is set to $4$.
Swin-B~\cite{swin} pre-trained on the Kinetics-400 dataset has been adopted for our network $G$ in Eqn.~\eqref{eqn:cvar-dx_G}. Since we do not want the gradients produced by the supervised loss $\mathcal{L}_{ce}$ being suppressed by the cross-view loss $\mathcal{L}_{self}$, the hyper-parameter $\lambda$ is set to $5.10^{-3}$.

\subsubsection{Ablation Studies}

Our ablative experiments report the results of our CVAR method with different settings trained on the Kinetics-400 $\to$ Charades-Ego and Kinetics-400 $\to$ EPIC-Kitchens-55 benchmarks. All the models are trained with the same configuration for fair comparisons. %

\input{Tables/chap-6/cvar/layer-ablation}
\noindent
\textbf{Effectiveness of the scale $\alpha$.} 
We study the effectiveness of the linear scale $\alpha$ to the performance of the model. 
In this experiment, 
the value $\alpha$ ranges from 0.0 to 2.0. When $\alpha = 0.0$, it is equivalent to ViT simultaneously trained on both third-view and first-view datasets.
As shown in Table~\ref{tab:cvar-alpha_ab}, the mAP performance on the Charades-Ego benchmark is consistently improved when the value of $\alpha$ increases from $0.1$ to $0.75$ and achieves the best performance at the value of $\alpha = 0.75$ and the mAP performance is $31.95\%$. 
Similarly, on the EPIC-Kitchen-55 benchmarks, the Top 1 and Top 5 accuracy is gradually improved w.r.t the increasing of $\alpha$ and reaches the maximum performance when the value of $\alpha$ is $1.50$ in which the Top 1 accuracy on EPIC Verb and EPIC Noun are $73.52\%$ and $68.19\%$. 
Then, the performance on both benchmarks steadily decreases when the value of $\alpha$ keeps increasing over the optimal point. 
Indeed, the variation in the video space is typically higher than in the attention maps due to the higher complexity of video data where the video data contains much more information, e.g., objects, humans, and interactions, etc.; meanwhile, the attention maps represent the focus of the models w.r.t model decisions. Thus, if the value of $\alpha$ is small, it could not represent the correct proportion of changes between videos and attention maps.
Meanwhile, the higher value of $\alpha$ inclines to exaggerate the model focuses which results in lower performance.

\noindent
\textbf{Effectiveness of the metrics.} 
This experiment studies the effectiveness of correlation metrics on the performance of the action recognition models on first-view videos.
For each metric correlation, we study its effect by comparing the performance of action recognition models using our metric in Eqn.~\eqref{eqn:cvar-dx_G} and Eqn.~\eqref{eqn:cvar-da_JS} against the Euclidean distance $\ell_2$.
As our results in Table~\ref{tab:cvar-metric_ab}, by measuring the correlation of videos on the deep latent spaces, i.e., $\mathcal{D}_x^G$, the performance of the action recognition model has been improved, e.g., $28.77\%$ to $31.95\%$ (results using $\mathcal{D}_a^{JS}$). This improvement is gained thanks to the deep semantic representation extracted by deep network $G$.
Besides, the probability metric used to measure the correlation between attention maps, i.e., $\mathcal{D}_a^{JS}$, has illustrated its significant role. For example, the performance of the model has been promoted by $+2.84\%$ from $29.11\%$ ($\ell_2$) to $31.95\%$ ($\mathcal{D}_a^{JS}$). As the attention map is the probability distribution, using the Jensen-Shannon divergence as the correlation metric provides the informative difference of the model's focus over the videos. Meanwhile, $\ell_2$ tends to rely on the difference of the magnitude of the attention, which provides less correlation information between two attentions.

\begin{figure}[!t]
    \centering
    \includegraphics[width=1.0\textwidth]{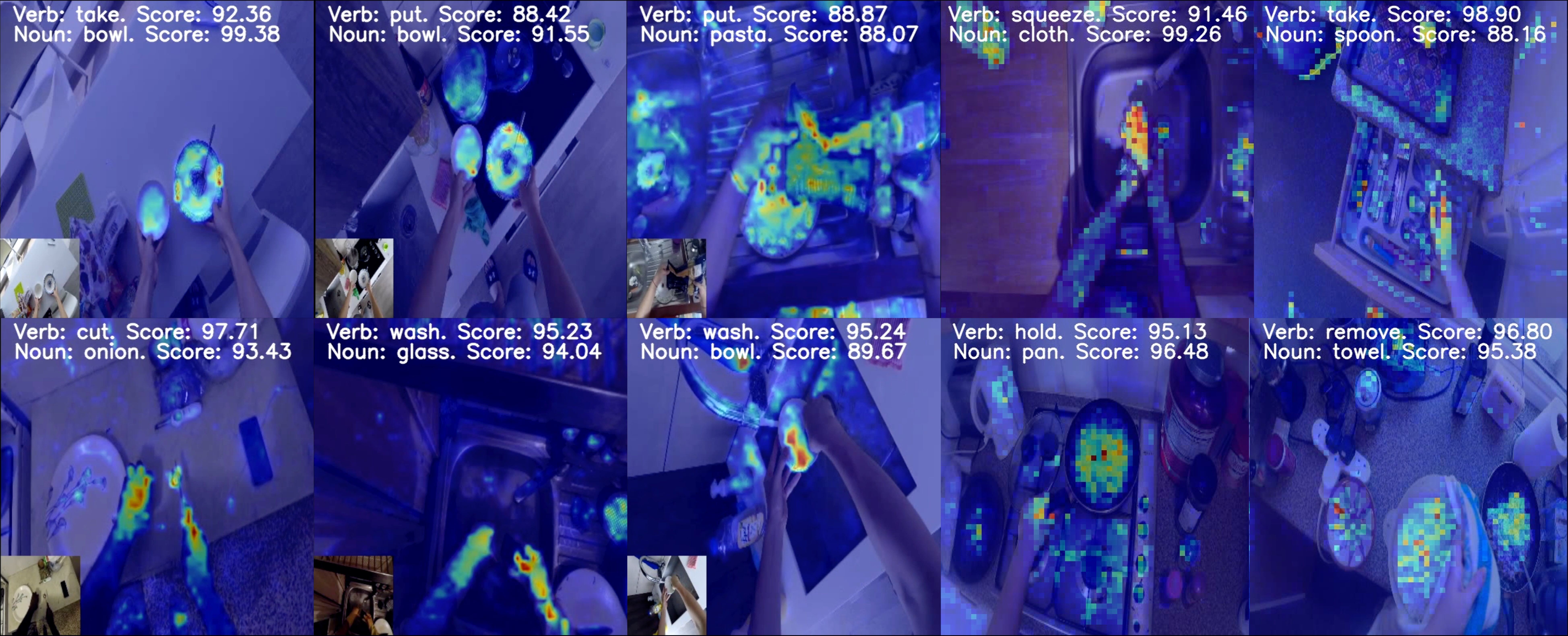}
    \caption{Attention Visualization of Model Prediction on EPIC Kitchen Videos.}
    \label{fig:cvar-att_vis}
\end{figure}

\input{Tables/chap-6/cvar/network-ablation}
\noindent
\textbf{Effectiveness of Transformer Layers.} 
In this experiment, 
we consider four groups of Transformer layers, each consisting of three consecutive layers, i.e., Layer 1-3, Layer 4-6, Layer 7-9, and Layer 10-12. As experimental results in Table~\ref{tab:cvar-attention_ab}, the later Transformer layers of our model play an important role than the initial ones. In particular, when imposing the cross-view loss on only the first three Transformer layers, the performance of Charades-Ego has achieved 25.65\% and the Top 1 accuracy of verb and noun predictions in EPIC-Kitchens-55 is 60.47\% and 57.85\%. Meanwhile, enforcing the cross-view self-attention loss into all attention layers brings better performance and achieves the best performance, i.e., the mAP of 31.95\% on Charades-Ego and Top 1 accuracy of 73.52\% and 68.19\% on EPIC-Kitchens-55.
Figure~\ref{fig:cvar-att_vis} visualizes the attention maps of the model predictions.

\input{Tables/chap-6/cvar/charades-ego-results}
\noindent
\textbf{Effectiveness of Different Network Backbone.}
To further illustrate the robustness of CVAR against the network backbone and ensemble models, we further evaluate CVAR with different network backbones.
We report the experimental results of ViT and TimeSFormer~\cite{timesformer} with and without our proposed geometric cross-view constraint.
We also report the result of the ensemble model with SwinB to illustrate the robustness of our approach compared to the ensemble approach.
The results in Table~\ref{tab:cvar-exp_net} have proved our proposed loss has robustly and consistently improved the performance of action recognition models. Moreover, our CVAR significantly outperforms the ensemble model.
Our approach emphasizes that our novel geometric cross-view metric can be applied to other Transformers.

\subsubsection{Comparisons with State-of-the-Art Results}

\noindent
\textbf{Kinetics-400 $\to$ Charades-Ego.}
Table~\ref{tab:cvar-charades_ego} presents results of our CVAR compared to prior methods, i.e., ActorObserverNet~\cite{sigurdsson2018actor}, SSDA~\cite{choi2020unsupervised}, I3D~\cite{choi2020unsupervised}, DANN~\cite{ganin2016domain}, SlowFast~\cite{slowfast}, Frozen~\cite{frozen}, MViT-V2~\cite{MViTv2}, Swin-B~\cite{swin}, and Ego-Exo~\cite{ego-exo}, on the Charades-Ego benchmark. 
Our results in Table~\ref{tab:cvar-charades_ego} have gained SOTA performance where our mAP accuracy in our approach has achieved $31.95\%$.
Compared to direct training approaches~\cite{vit, frozen, swin, mvit, choi2020unsupervised}, our method achieves better performance than other methods by a large margin, e.g., higher than Swin-B~\cite{swin} by $3.18\%$.
Compared with the prior pre-training method using additional egocentric tasks, our result is higher than Ego-Exo~\cite{ego-exo} by $+2.76\%$.
Meanwhile, compared with domain adaptation methods~\cite{ganin2016domain, choi2020unsupervised}, our methods outperform DANN by $+8.33\%$.

\input{Tables/chap-6/cvar/epic-55-results}
\noindent
\textbf{Kinetics-400 $\to$ EPIC-Kitchens-55.}
Table~\ref{tab:cvar-epic_55} presents the results of our approach compared to prior methods, i.e., ResNet-50~\cite{slowfast}, DANN~\cite{ganin2016domain}, SlowFast~\cite{slowfast}, MViT-V2~\cite{MViTv2}, Swin-B~\cite{swin}, and Ego-Exo~\cite{ego-exo}, on the EPIC-Kitchens-55 benchmark. 
Our CVAR has gained the SOTA performance where our Top 1 accuracy on EPIC Verb and Noun has achieved $73.52\%$ and $68.19\%$, respectively.
Our proposed approach outperforms the traditional direct training methods~\cite{slowfast, MViTv2, swin} by a large margin.
In addition, our result is higher than the pre-training methods using additional egocentric tasks, i.e., Ego-Exo~\cite{ego-exo}, by $+7.09\%$ and $+18.4\%$ on Top 1 accuracy of verb and noun predictions. Our method also outperforms the domain adaptation method~\cite{ganin2016domain}.

\input{Tables/chap-6/cvar/epic-100-results}

\input{Tables/chap-6/cvar/ntu-results}
\noindent
\textbf{Kinetics-400 $\to$ EPIC-Kitchens-100.}
Table~\ref{tab:cvar-epic_100} compares our results with TSN~\cite{tsn}, TRN~\cite{trn}, TBN~\cite{epic-fusion}, TSM~\cite{lin2019tsm}, SlowFast~\cite{slowfast}, MViT-V2~\cite{MViTv2}, Ego-Exo using SlowFast-R50~\cite{ego-exo}, and Swin-B~\cite{swin} on the EPIC-Kitchens-100 benchmark. 
Overall, our proposed CVAR has achieved the SOTA performance where the Top 1 accuracy of verb, noun, and action predictions are $69.37\%$, $61.03\%$, and $46.15\%$, respectively. Also, CVAR has gained competitive performance on the sets of unseen participants and tail classes.
Compared to prior direct training methods~\cite{swin, MViTv2, vit}, out method outperforms these approaches by a notable margin, i.e., higher than Swin-B by $+1.44\%$ and $+2.34\%$ on Top 1 Accuracy of Verb and Noun predictions in overall.
Also, our results outperform Ego-Exo in overall accuracy and unseen participants and tail classes.

\noindent
\textbf{Cross-view NTU RGB+D Action Recognition.} To further illustrate the effectiveness of our approach when the domain gap is not that large, we conducted an experiment on the NTU RGB+D action recognition dataset.
In this experiment, we use the videos captured from the $0^o$ angle as the source view, while the two other angles ($\pm45^o$) are considered as the target view. 
As shown in Table~\ref{tab:cvar-ntu_res}, our proposed CVAR has outperformed the other methods by a large margin.  In particular, while the performance of Swin B~\cite{swin} achieved 93.72\%, our CVAR approach gained the Top 1 accuracy of 95.95\%. 
These results have shown the effectiveness of our proposed approach in modeling action recognition across views.

%% file: Tables/chap-6/cvar/scale-ablation.tex
\begin{wraptable}{r}{0.5\textwidth}
\caption{\textbf{Effectiveness of the Scale $\alpha$ in the Linear Relation to the Charades-Ego (E-Ego) and EPIC-Kitchen-55 (EPIC) Action Recognition Benchmarks.}}
\label{tab:cvar-alpha_ab}

\centering
\resizebox{0.5\textwidth}{!}{
\begin{tabular}{|c|c|c|c|c|c|}
\hline
\multirow{2}{*}{$\alpha$}      &       C-Ego       & \multicolumn{2}{c|}{EPIC Verb}         & \multicolumn{2}{c|}{EPIC Noun}         \\
\cline{2-6}
 & mAP & Top  1 & Top 5 & Top 1 & Top 5 \\
 \hline
0.00  & 20.70        & 41.94  & 67.31 &	43.19 &	60.14 \\
0.25  & 25.09        & 55.96  & 89.37 &	55.96 & 80.65 \\
0.50  & 28.97        & 58.84  & 87.24 & 54.75 & 75.27 \\
0.75  & \textbf{31.95}        & 60.80  & 89.62 & 57.42 & 77.77 \\
1.00  & 30.68        & 68.97  & 89.53 & 44.87 & 70.98 \\
1.50  & 29.51        & \textbf{73.52}	& \textbf{92.22}	& \textbf{68.19}	& \textbf{84.93} \\ 
2.00  & 27.80     & 69.60	& 92.54	& 61.60	& 81.22 \\
\hline
\end{tabular}
}
\end{wraptable}

%% file: Tables/chap-6/cvar/metric-ablation.tex
\begin{wraptable}{r}{0.5\textwidth}
\centering
\caption{\textbf{Effectiveness of the Choices of Correlation Metrics to the Charades-Ego (E-Ego) and EPIC-Kitchen-55 (EPIC) Action Recognition Benchmarks.}}
\label{tab:cvar-metric_ab}
\setlength{\tabcolsep}{3pt}
\def\arraystretch{1.1}
\resizebox{0.5\textwidth}{!}{
\begin{tabular}{|cc|cc|c|cc|cc|}
\hline

\multicolumn{2}{|c|}{$\mathcal{D}_{x}$}                       & \multicolumn{2}{c|}{$\mathcal{D}_{a}$}                       & C-Ego &  \multicolumn{2}{|c}{EPIC Verb}       & \multicolumn{2}{|c|}{EPIC Noun}      \\
\cline{1-4}\cline{5-9}
$\ell_2$                    & $\mathcal{D}_x^{G}$                   & $\ell_2$                    & $\mathcal{D}_a^{JS}$                   & mAP & Top  1 & Top 5 & Top 1 & Top 5 \\
\hline
\cmark &                       & \cmark &                       & 27.80        & 60.97  & 89.95 & 58.05 & 78.07 \\
\cmark &                       &                       & \cmark & 28.77        & 61.13  & 90.16 & 58.05 & 78.40 \\
                      & \cmark & \cmark &                       & 29.11  & 63.13	& 90.12	& 59.68	& 80.03 \\
                      & \cmark &                       & \cmark & \textbf{31.95}        & \textbf{73.52}  & \textbf{92.22} & \textbf{68.19}	& \textbf{84.93} \\
\hline
\end{tabular}
}
\end{wraptable}

%% file: Tables/chap-6/cvar/layer-ablation.tex
\begin{wraptable}{r}{0.5\textwidth}
\centering
\caption{\textbf{Effectiveness of the Transformer Layers to the Charades-Ego (E-Ego) and EPIC-Kitchen-55 (EPIC) Action Recognition Benchmarks.}}
\label{tab:cvar-attention_ab}
\setlength{\tabcolsep}{2.5pt}
\resizebox{0.5\textwidth}{!}{
\begin{tabular}{|cccc|c|cc|cc|}
\hline

\multicolumn{4}{|c|}{Transformer Layers} & C-Ego &  \multicolumn{2}{|c}{EPIC Verb}       & \multicolumn{2}{|c|}{EPIC Noun}      \\
\cline{5-9}
 1-3 & 4-6 & 7-9 & 10-12 & mAP & Top  1 & Top 5 & Top 1 & Top 5 \\
\hline
\cmark & & & & 25.65 & 60.47 & 90.26 & 57.85 & 78.58 \\
\cmark & \cmark & & & 28.19 & 68.46 & 91.08 & 66.54 & 83.36 \\
\cmark & \cmark & \cmark & & 30.60 & 69.27 & \textbf{92.58} & 68.09 & \textbf{85.02} \\
\cmark & \cmark & \cmark & \cmark & \textbf{31.95}        & \textbf{73.52}  & 92.22 & \textbf{68.19}	& 84.93 \\
\hline
\end{tabular}
}
\end{wraptable}

%% file: Tables/chap-6/cvar/network-ablation.tex
\begin{wraptable}{r}{0.5\textwidth}
\centering
\caption{Effectiveness of Different Networks.}\label{tab:cvar-exp_net}
\resizebox{0.5\textwidth}{!}{
\begin{tabular}{|l|cc|cc|}
\hline
\multirow{2}{*}{Backbone}                 & \multicolumn{2}{c|}{EPIC-55 Verb} & \multicolumn{2}{c|}{EPIC-55 Noun} \\
\cline{2-5}
            & Top 1         & Top 5         & Top 1         & Top 5         \\
\hline
Swin-B \cite{swin} & 56.40 & 85.84 & 47.68 & 71.02 \\
\hline
ViT \cite{vit}               & 41.76         & 69.49         & 44.19         & 60.52         \\
Ensemble ViT + SwinB & 58.97	& 88.61 &	49.21	& 74.27\\
ViT+CVAR           & \textbf{73.52}         & 92.22         & \textbf{68.19}         & \textbf{84.93}         \\
\hline
TimeSFormer \cite{timesformer}        & 41.37         & 67.40         & 42.44         & 59.55         \\
Ensemble TimeSFormer + SwinB & 59.07 & 88.95	& 50.11 & 77.27 \\
TimeSFormer+CVAR & 72.17         & \textbf{95.19}         & 62.83         & 83.49        \\
\hline
\end{tabular}
}
\end{wraptable}

%% file: Tables/chap-6/cvar/charades-ego-results.tex
\begin{wraptable}{r}{0.4\textwidth}
\centering
\caption{\textbf{Comparisons on Charades-Ego.}}
\label{tab:cvar-charades_ego}
\resizebox{0.4\textwidth}{!}{
\begin{tabular}{|l|c|}
		\hline
		Method	& mAP	\\
		\hline
		ActorObserverNet~\cite{sigurdsson2018actor} & 20.00 \\
		SSDA~\cite{choi2020unsupervised} & 23.10 \\
        I3D~\cite{choi2020unsupervised} & 25.80 \\

        DANN \cite{ganin2016domain}  & 23.62 \\
        SlowFast~\cite{slowfast} & 25.93 \\
        Frozen \cite{frozen} & 28.80 \\
        MViT-V2 & 25.65 \\
        Swin-B \cite{swin} & 28.77 \\
        Ego-Exo  + SlowFast \cite{ego-exo}  & 28.04 \\
        Ego-Exo* + SlowFast \cite{ego-exo}  & 29.19 \\
        \textbf{CVAR (Ours)} & \textbf{ 31.95} \\
        \hline
\end{tabular}
}
\end{wraptable}

%% file: Tables/chap-6/cvar/epic-55-results.tex
\begin{wraptable}{r}{0.5\textwidth}
\centering
\caption{\textbf{Comparisons on EPIC-Kitchen-55.}}
\label{tab:cvar-epic_55}
\label{tb:different_baseline}
\resizebox{0.5\textwidth}{!}{
	\begin{tabular}{|l|cc|cc|}
		\hline
		\multirow{2}{*}{Method}   & \multicolumn{2}{c|}{EPIC verbs} & \multicolumn{2}{c|}{EPIC nouns} \\
		\cline{2-5}
		  & Top 1 & Top 5 & Top 1 & Top 5	\\
		\hline
        ResNet-50 \cite{slowfast}  & 61.19 & 87.49 & 46.18 & 69.72 \\
        MViT-V2 \cite{MViTv2} & 55.17	& 89.87	& 56.59	& 79.40 \\
        Swin-B \cite{swin} & 56.40 & 85.84 & 47.68 & 71.02 \\
        DANN \cite{ganin2016domain}  & 61.27 & 87.49 & 45.93 & 68.73 	  \\
        Joint-Embed \cite{sigurdsson2018actor} & 61.26 & 87.17 & 46.55 & 68.97	  \\
        Ego-Exo + SlowFast \cite{ego-exo} & 65.97	& 88.91	& 49.42	& 72.35 \\
        Ego-Exo* + SlowFast \cite{ego-exo} & 66.43	& 89.16	& 49.79	& 71.60 \\
        \textbf{CVAR (Ours)} & \textbf{73.52}	& \textbf{92.22}	& \textbf{68.19}	& \textbf{84.93} \\
		\hline				
	\end{tabular}
 }
\end{wraptable}

%% file: Tables/chap-6/cvar/epic-100-results.tex
\begin{table}[!t]
    \centering
    \caption{\textbf{Comparisons to Prior Methods on the EPIC-Kitchen-100 Action Recognition Benchmark.}}
    \label{tab:cvar-epic_100}
\setlength{\tabcolsep}{2pt}
\resizebox{1.0\textwidth}{!}{
\begin{tabular}{|l|ccc|ccc|ccc|ccc|}
\hline
    & \multicolumn{6}{c|}{Overall}                           & \multicolumn{3}{c|}{Unseen Participants} & \multicolumn{3}{c|}{Tail Classes} \\
\cline{2-13}
 Method & \multicolumn{3}{c|}{Top-1 Accuracy} & \multicolumn{3}{c|}{Top-5 Accuracy} & \multicolumn{3}{c|}{Top-1 Accuracy}           & \multicolumn{3}{c|}{Top-1 Accuracy}    \\
\cline{2-13}
                               & \multicolumn{1}{c}{Verb} & \multicolumn{1}{c}{Noun} & \multicolumn{1}{c|}{Action} & \multicolumn{1}{c}{Verb} & \multicolumn{1}{c}{Noun} & \multicolumn{1}{c|}{Action}& \multicolumn{1}{c}{Verb} & \multicolumn{1}{c}{Noun} & \multicolumn{1}{c|}{Action} & \multicolumn{1}{c}{Verb} & \multicolumn{1}{c}{Noun} & \multicolumn{1}{c|}{Action} \\ 
 \hline
                                            TSN~\cite{tsn}                    & 60.18 & 46.03 & 33.19 & 89.59 & 72.90 & 55.13 & 47.42 & 38.03 & 23.47 & 30.45 & 19.37 & 13.88 \\
                                            TRN~\cite{trn}                    & 65.88 & 45.43 & 35.34 & 90.42 & 71.88 & 56.74 & 55.96 & 37.75 & 27.70 & 34.66 & 17.58 & 14.07 \\
                                            TBN~\cite{epic-fusion}                & 66.00 & 47.23 & 36.72 & 90.46 & 73.76 & 57.66 & 59.44 & 38.22 & 29.48 & 39.09 & 24.84 & 19.13 \\
                                            TSM~\cite{lin2019tsm}                     & 67.86 & 49.01 & 38.27 & 90.98 & 74.97 & 60.41 & 58.69 & 39.62 & 29.48 & 36.59 & 23.37 & 17.62 \\
                                            SlowFast~\cite{slowfast} & 65.56 & 50.02 & 38.54 & 90.00 & 75.62 & 58.60 & 56.43 & 41.50 & 29.67 & 36.19 & 23.26 & 18.81 \\
                                            MViT-V2 \cite{MViTv2} & 67.13 & 60.89 & 45.79 & 91.13 & 83.93 & 66.83 & 57.75 & 50.52 & 34.84 & 40.85 & 38.47 & 25.35\\
                                            Ego-Exo \cite{ego-exo} & 66.61 & 59.51 & 44.89 & 91.13 & 82.03 & 65.05 & 56.57 & 48.87 & 33.71 & 40.91 & 38.26 & 25.23 \\
                                            Swin-B \cite{swin} & 67.93 & {58.69} & 46.05 & 90.96 & \textbf{83.77} & 65.23 & 58.69 & \textbf{50.89} & 35.02 & 41.08 & 37.21 & 25.41 \\
                                            \textbf{CVAR (Ours)} & \textbf{69.37} & \textbf{61.03} & \textbf{46.15} & \textbf{91.51} & 81.03 & \textbf{67.05} & \textbf{59.91} & 48.36	& \textbf{35.12} & \textbf{41.93} &	\textbf{38.58} & \textbf{25.99} \\
\hline
\end{tabular}%
}
\end{table}

%% file: Tables/chap-6/cvar/ntu-results.tex
\begin{wraptable}{r}{0.3\textwidth}
\centering
\caption{Comparision on NTU RGB+D Action Recognition}\label{tab:cvar-ntu_res}
\resizebox{0.3\textwidth}{!}{
\begin{tabular}{|l|c|}
\hline
Method & Top 1  \\
\hline
SlowFast~\cite{slowfast} & 90.22 \\
DANN \cite{ganin2016domain}  & 90.41           \\
MViT-V2 \cite{MViTv2}  & 91.35 \\
Ego-Exo \cite{ego-exo} & 92.14 \\
Swin B \cite{swin}  & 93.72         \\
\textbf{CVAR}   & \textbf{95.93}          \\
\hline
\end{tabular}
}
\end{wraptable}

%% file: Chapters/Chaps/chap-6-multimodal-tempoeral-learning.tex
\chapter{From Efficient Multimodal and Temporal Learning to Generalization in Large-Scale Foundation Models}\label{chap:multimodal-temporal}

The rapid increase of videos and multimodal data requires the development of efficient multimodal and temporal learning and has also promoted the development of the foundation model.
The applications of multi-modality and temporal learning are diverse, e.g., human action recognition, audio-visual localization, etc.
Although some prior methods have been introduced to approach these tasks, there still remain some limitations.  
First, a direct association of multimodal features may limit the correlations to be extracted due to different modalities.
Second, the relationship across temporal segments helping to maintain the consistency of correlation and contexts is not effectively exploited. 
Third, how does the temporal ordering of video frames affect the temporal understanding of the model?
Last, understanding the robustness and improving the generalizability of foundation models are becoming more critical.
Therefore, in this chapter, we first introduce a novel multimodal learning approach to audio-visual understanding.
Then, we propose a novel end-to-end Transformer-based Directed Attention (DirecFormer) framework for robust video understanding.
Finally, to analyze the robustness of the foundation model, we introduce a new simple but efficient Diffusion Sampling approach to Domain Generalization (EDSAM) to improve the generalizability of the foundation model.

\input{Chapters/Sections/chap-6/right2talk}

\input{Chapters/Sections/chap-6/direcformer}

\input{Chapters/Sections/chap-6/edsam}

\section{Summary}

This chapter has presented novel approaches to multimodal, temporal learning, and foundation models.
First,  we presented the Audio-Visual Transformer, a new multimodal learning approach, for Main Speaker Localization and Audio Separation. 
Thanks to the attention mechanisms in the spatial-temporal dimension and the domain alignment for better synchronization, our approach can effectively localize and highlight the main speaker in both visual and audio channels on multi-speaker conversation videos.
Second, we presented a new and simple DirecFormer method with a Directed Attention mechanism to improve temporal learning in video understanding.
The presented Directed Temporal-Spatial Attention not only learns the magnitude of the correlation between frames and tokens but also exploits the direction of attention. Moreover, the self-supervised guided loss further enhances the directed learning capability of the Directed Temporal Attention.
Last, under our theoretical analysis, we introduced a new efficient sampling to generate new diffusion-based adversarial samples based on our proposed transport transformation to improve the generalizability of the vision-language foundation model. 
The experiments on large-scale datasets have confirmed the state-of-the-art performance of our proposed approaches.

%% file: Chapters/Sections/chap-6/right2talk.tex
\section{The Transformer Approach to Multimodal Learning in Audio-Visual Understanding}
\label{eqn:paper-right2talk}

\setcounter{propositioncounter}{0}
\setcounter{remarkcounter}{0}

\input{Tables/chap-5/right2talk/multimodal-comparison}

\begin{figure}[t]
	\centering \includegraphics[width=1\columnwidth]{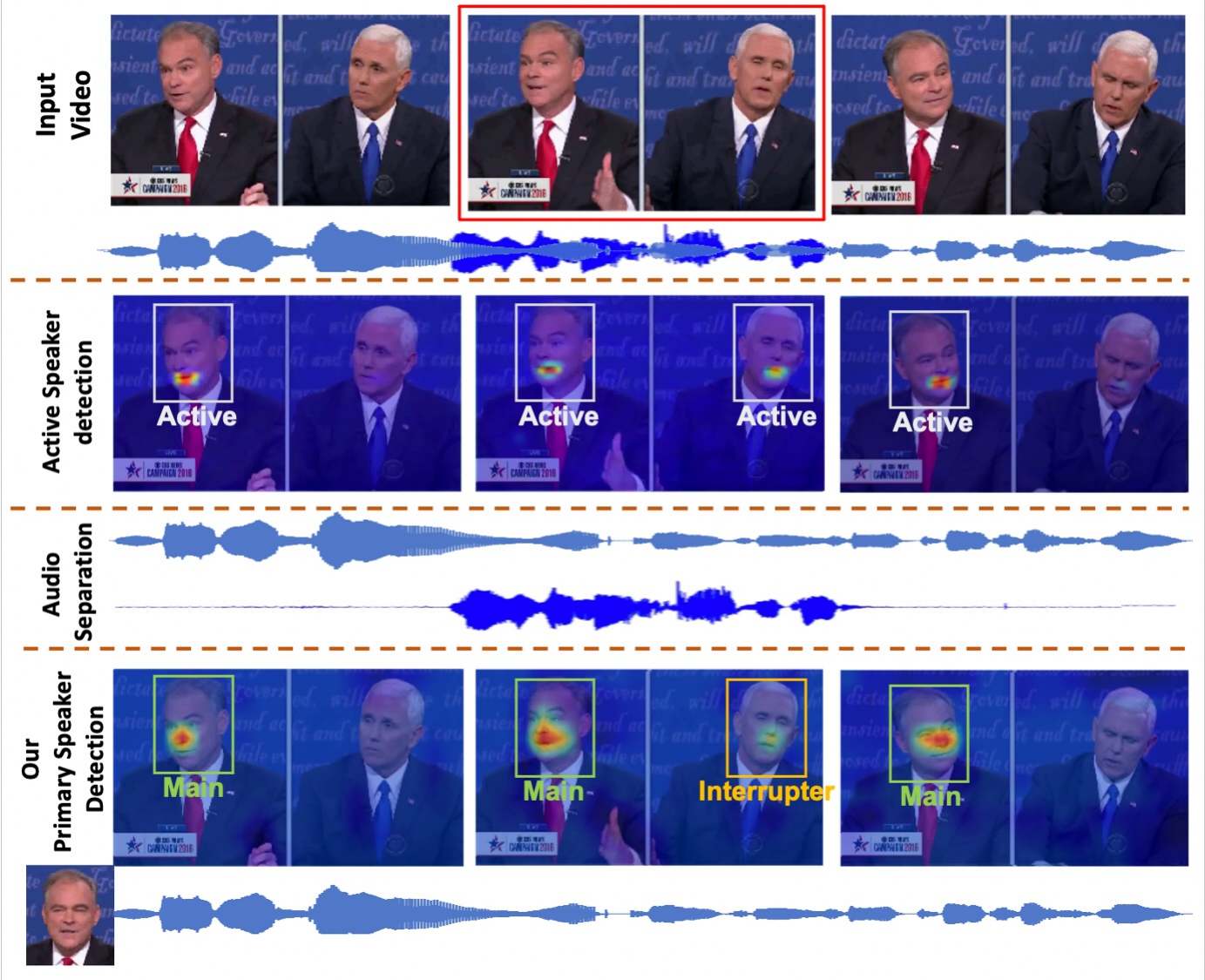}
	\small
	\caption{Given a multi-speaker video, our Audio-Visual Transformer can localize and highlight the main speaker in both visual and audio channels. }
	\label{fig:right2talk-PrimarySpeakerDetection}
\end{figure}
Although human beings possess capabilities of localizing and separating sounds from noisy environments, we still have trouble following a conversation with noises, background voices, or interruptions from other speakers. 
Either with blind audio separation~\cite{jin2009supervised,makino2007blind,reddy2007soft, radfar2007single,wang2005video} or visual-aid audio separation
\cite{afouras2019my, arandjelovic2017look, chung2019said, gabbay2018seeing, gao20192, harwath2018jointly, hu2019deep, liu2013source,khan2013speaker, owens2018learning, ramaswamy2020see, senocak2018learning,tian2018audio} approaches, this outlier separation task still remains a challenge in the wide conditions beyond the lab settings. The problem becomes especially harder when dealing with unknown numbers of speakers in an audio. Nachmani et al.~\cite{nachmani2020voice} make a comparison between methods and show how hard it is to separate voices when the number of sound sources increases. 
Existing methods achieve high performance 
with inputs from multiple microphones. 
Some methods assume a clean set of single source audio examples are available for supervision~\cite{afouras2018conversation, ephrat2018looking, zhao2019sound,Zhao_2018_ECCV}.
In practice, rather than solely trying to separate voices of all speakers in a conversation and determining ``\textit{who-spoke-when}'',
we tend to give more attentions to the \textbf{\textit{main speaker}}, i.e. \textit{who is on his/her turn of speaking and his/her talk is the main channel of communication}, and ignore the voices of remaining speakers, i.e. \textbf{\textit{interrupters}} or \textbf{\textit{listener}}, or background noises. 
Thus, 
an approach that highlights the main speaker in visual and audio channels would give new opportunities to popular applications such as auto-muting in a tele-conference or main speaker refocusable video generation.

Given a video of 
multi-speaker conversation, our goal is to learn an audio-visual model 
that enables the capabilities of both (1) localizing the main speaker; (2) true cancellation of audio sources of interrupters or background noises; and (3) automatically switching to a new subject when the speakers change their roles. The interruptions from other subjects and the background are considered as noises and removed.
In the scope of this work, we focus on \textbf{\textit{turn-taking conversation}} as the turn-taking mechanism has been commonly adopted for structuring conversation in social interactions. A subject is considered as the main speaker when he/she properly takes the turn of solo speaking and will continue the talk even after a simultaneous speech occurs~\cite{cutrone2019profiling}.%

Previous approaches have partially addressed 
this problem and can be divided into two categories, i.e.  \textit{audio-visual synchronization}~\cite{arandjelovic2018objects, chung2016out,cutler2000look,owens2018audio,khosravan2019attention} and \textit{mix-and-separate}~\cite{afouras2018conversation, 9054376, ephrat2018looking,gao2018learning,gao2019co, NEURIPS2020_7288251b, korbar2018co, zhao2019sound, Zhao_2018_ECCV, 10.1007/978-3-030-58610-2_4}. 
The former exploits the synchronization between audio and video frames within a specific time window to localize the image regions that are more sensitive to audio changes. Meanwhile, the latter 
learns to separate the speakers’ voices from a mix utterance based on audio and visual features. %
In both cases, there remain some limitations. Firstly, 
audio-visual relationships 
are extracted via a concatenation operator or the cosine distance metric. 
However, as audio and visual features distribute in two different latent spaces by their nature, 
these methods may not maximize correlations between the two feature domains.
Secondly, audio-visual relationships 
are only considered within a video segment, i.e. a short time window, while ignoring the ones across temporal segments, which helps to maintain the consistency of localization and separation, and \textit{contextual} main-subject switching.
Finally, the interactions between subjects in the temporal dimension to deliver accurate tracking and anticipatory decisions about transition to a new target are still ignored.

To address these aforementioned limitations, this chaptper introduces a novel Audio-Visual Transformer approach, a new multimodal learning approach, to highlight the main speaker in both audio and visual channels %
(Figure~\ref{fig:right2talk-PrimarySpeakerDetection}).
First, the proposed approach exploits various correlations presented in visual and audio signals 
including ``virtual'' interactions between speakers in a video scene and relationships between visual and auditory modalities.
Second, rather than extracting audio-visual correlations within a video segment, relationships across segments are further exploited via a temporal self-attention mechanism in the proposed Transformer structure. 
This helps to engage the contextual information and enhance attentions with longer context so that the main speaker can be robustly identified.
Third, a Cycle Synchronization Loss is introduced to learn the main speaker localization in a self-supervised manner.
To the best of our knowledge, it is one of the first works that is able to automatically localize and highlight the main speaker in multi-speaker conversation videos on both visual and audio channels (Table~\ref{tab:right2talk-TenMethodSumm}).

\subsection{The Proposed Multimodal Learning Approach}

\begin{figure*}[t]
	\centering \includegraphics[width=1.0\textwidth]{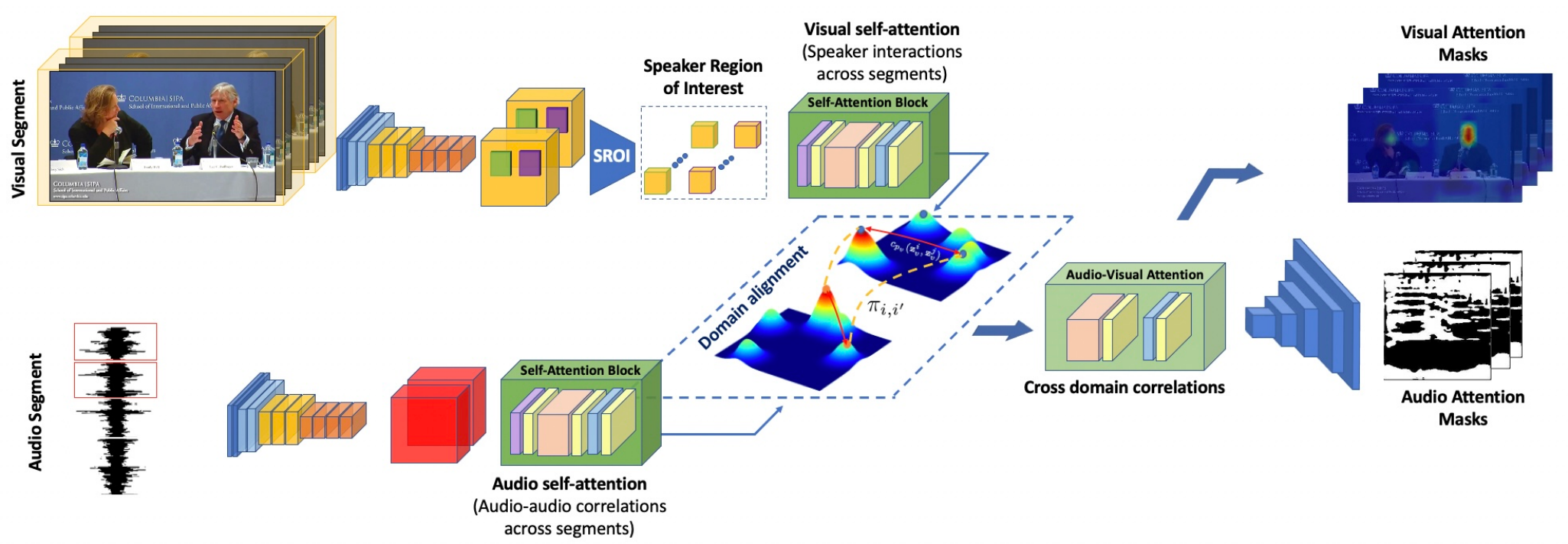}
	\small
	\caption{\textbf{Our Proposed Audio-Visual Transformer Framework.} Given a video segment set, the 
	context in the conversation is captured with
	three types of correlations, i.e. visual-visual, audio-audio, and audio-visual attentions. Then, the audio-visual attentional features are adopted for the main speaker localization and audio separation.
	}
	\label{fig:right2talk-ProposedFramework}
\end{figure*}

This work focuses on turn-taking conversations composing talking turns. %
The length of each turn is flexible according to the conversation's context and contents.
Let $\mathbf{x} \in \mathcal{X}$ be a multi-speaker conversation video consisting of a \textit{visual component} $\mathcal{V}$ (a sequence of RGB frames) and an \textit{audio component} $ \mathcal{A}$ (a mixed audio of one or multiple speakers).

\subsubsection{The Turn-taking Conversation} \label{sec:InTurnConversation}
With turn-taking regulations, a conversation $\mathbf{x}$ can be decomposed into speaking turns, i.e. $\mathbf{Turn}^h, h=1,...,H$ where $H$ is the number of turns in $\mathbf{x}$. Although many speakers can have their voices overlapped during a speaking turn in either cooperative or competitive manner, the role of each speaker can be classified into two groups.

\noindent
\textbf{{Main Speaker} $S_m^h$. } A subject $S$ is %
the main speaker of $\mathbf{Turn}^h$ when he/she carries the conversation and drives it forward~\cite{cutrone2019profiling}. Even an interruption (i.e. simultaneous speech) occurs, the subject will continue to solo speak until the end of the turn. Thus, a long solo speaking of $S$ during $\mathbf{Turn}^h$ can provide an  indication for the main speaker role. 

\begin{figure}[t]
	\centering \includegraphics[width=1\columnwidth]{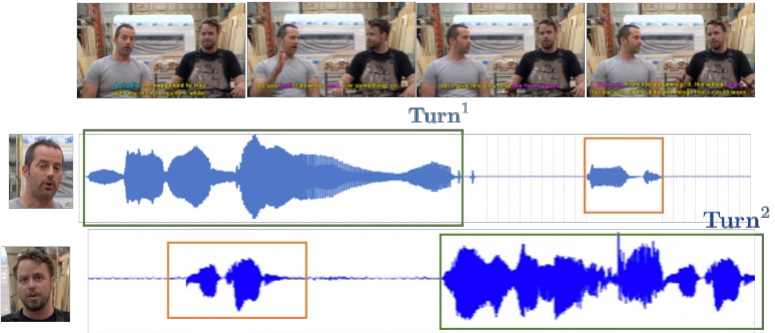}
	\small
	\caption{\textbf{Turn-taking Conversation.} The conversation is decomposed in to turns where each speaker acts as the main speaker (green box) of a turn, and the other speakers are considered as interrupter or listener (orange box) during that turn.}
	\label{fig:right2talk-TurnTakingConversation}
\end{figure}

\noindent
\textbf{{Interrupter or Listener} $S_I^h$. } %
An interrupter or Listener is the one who has reactions or comments to the main speaker's utterances. These reactions usually occur in a short time window during $\mathbf{Turn}^h$ and end up with the continuation of the main speaker's talk.
When the interrupter continues to solo speak after  a simultaneous speech, a turn changing of the main speaker occurs. Figure~\ref{fig:right2talk-TurnTakingConversation} illustrates an example of speakers' roles in a turn-taking conversation.

\subsubsection{Problem Definition}
Rather than decomposing $\mathbf{x}$ into $\{\mathbf{Turn}^h\}_1^H$, we propose to present  $\mathbf{x}$ as a composition of $K$ segments $\mathbf{Seg}^k = \{\mathbf{v}^k,\mathbf{a}^k \}, k=1,...,K$, $\mathbf{v}^k \in \mathcal{V}$ and $\mathbf{a}^k \in \mathcal{A}$. The $h$-th speaking turn $\mathbf{Turn}^h$ consists of one or multiple segments, i.e. $\mathbf{Turn}^h = \{\mathbf{Seg}^k\}_{h_{start}}^{h_{end}}$ where $h_{start}$ and $h_{end}$ mark the indices of the starting and ending time of $h$-th turn, respectively. 
Let $S_m^k$ be the main speaker and $S_I^k$ be the interrupters of $\mathbf{Seg}^k$ in the conversation. We have $S_m^k \equiv S_m^h$ and $S_I^k \equiv S_I^h$ when $\mathbf{Seg}^k \in \mathbf{Turn}^h$. Then, the goal is to extract the location and clean voice of $S_m^k$ for each $\mathbf{Seg}^k$ in the conversation.
Formally, the objectives are to learn the visual location map $\mathbf{M}_v^k$ and audio mask $\mathbf{M}_a^k$ of $S^k_m$ as in Eqn.~\eqref{eqn:right2talk-object_learning_visual}.
\begin{equation} \label{eqn:right2talk-object_learning_visual}
\scriptsize
\begin{split}
    \mathbf{M}_v^{k*} &= \arg \min_{\mathbf{M}_v^{k}}
   \Big[ -\log P\left(\mathbf{M}_v^{k}[Loc(S_m^k, \mathbf{v}^k)]|\mathbf{Seg}^{1:k}\right) + \log P\left(\mathbf{M}_v^{k}[Loc(S_I^k, \mathbf{v}^k)]|\mathbf{Seg}^{1:k}\right) \Big] \\
    \mathbf{M}_a^{k*} &= \arg \min_{\mathbf{M}_a^{k}} \| \mathbf{M}_a^{k} \odot Spec(\mathbf{a}^k) - Spec(\mathbf{a}_{S_m^k})\|_1
\end{split}
\end{equation}
where $\textit{Loc}(S_m^k, \mathbf{v}^k)$ is the location of $S_m^k$ in $\mathbf{v}^k$;
$Spec(\cdot)$ is the spectrogram conversion operator; $\odot$ is the Hadamard product;
and $\mathbf{a}_{S_m^k}$ is the clean voice of  $S_m^k$. %
$\mathbf{Seg}^{1:k}$ denotes the temporal information provided from the beginning to the $k$-th segment of the video $\mathbf{x}$.

To effectively estimate  $\mathbf{M}_v^k$ and  $\mathbf{M}_a^k$, we propose an Audio-Visual Transformer approach (see Figure~\ref{fig:right2talk-ProposedFramework}) consisting of three learning stages: (1) Learning the context with visual and audio self-attention; (2) Audio-Visual Correlation Learning; and (3) Main speaker localization and audio separation with Conversation Grammar. The proposed Audio-Visual Transformer is formulated via  $\{D,\phi, E_v, E_a\}$ as in Eqn.~\eqref{eqn:right2talk-GEVA}.
\begin{equation} \label{eqn:right2talk-GEVA}
\begin{split}
    G &= [D \circ \phi] (\mathbf{z}_v ,\mathbf{z}_a), \quad
    \mathbf{z}^k_v = E_v(\mathbf{v}^k| \mathbf{Seg}^{1..k}), \quad
    \mathbf{z}^k_a = E_a(\mathbf{a}^k| \mathbf{Seg}^{1..k})
\end{split}
\end{equation}
where $E_v$ and $E_a$ map $\mathbf{v}^k$ and $\mathbf{a}^k$ to their latent representations; and $\circ$ is the functional composition.  $\phi$ is the projection function to the shared representation space where these modalities are comparable.  $D$ maps these deep representations to the audio-visual mask of the main speaker.

\subsubsection{Visual and Audio Contextual Learning}
Besides prior works on temporal learning~\cite{8237665, truong2021direcformer}, in this work, given a conversation, contextual information can be extracted from visual and audio signals, i.e. \textit{visual-visual} and \textit{audio-audio} correlations \textit{across video segments}. While the former assists to track behaviors and interactions of each speaker over the spatial-temporal dimension, the latter provides more cues about the conversation flow,  i.e. when and how the main speaker switch his/her role. 
For example, the higher audio-audio correlation between two or more segments is, the lower the possibility is of the main speaker being switched. 
Thus, these cross-segment correlations can implicitly embed turn changing signals of the main speaker, and help to avoid the stage of pre-decomposing $\mathbf{x}$ into speaking turns $\mathbf{Turn}^h$. Even when interrupters dominate the main speaker's voice in a certain segment, this ``temporal-based'' correlation can exploit the relations to previous segments and identify the main speaker. %
We %
model the contextual correlations via two encoder structures with a self-attention mechanism before embedding their cross-domain correlations.

\subsubsection{Visual-Visual Self-Attention} \label{sec:Visual_Self_Att}

Given a sequence of segments $\{\mathbf{v}^k\}_1^K$, the visual encoder $E_v$ consists of three main functions, i.e. feature embedding, self-attention, and feature refinement with attention. Particularly, each $\mathbf{v}^k$ is firstly embedded into a deep feature embedding via $F_{v}: \mathcal{V} \mapsto \mathcal{F}_v$ as $\mathbf{f}^k_v = F_v(\mathbf{v}^k)$.

\noindent
{\textbf{Speaker Region of Interest (SROI).}}
Rather than embedding each $\mathbf{v}^k$ into a single feature for attention computation, we propose to project $\mathbf{f}^k_v$ into regions of interest where each region represents a speaker's location 
in a visual segment and learn the correlations among them. Particularly, let $\mathbf{b}=\{ \mathbf{b}_i^k\}, i=1..N, k=1..K$ where $\mathbf{b}_i^k \in \mathcal{B}  \subset \mathbb{R}^4$ denotes the location of $i$-th speaker in $\mathbf{Seg}^k$, and $N$ is the number of speakers. The projection function $R: \mathcal{F}_v \times \mathcal{B} \mapsto \mathcal{F}_{v}$ is defined as $\mathbf{f}^{k,i}_{v} = R(\mathbf{b}^k_i, \mathbf{f}_v^k)$. We adopt ROI Align~\cite{He_2017_ICCV} for the function $R$.
There are two approaches to obtain $\mathbf{b}$ and $N$, i.e. face detection and block decomposition. The former adopts a face detection %
to extract faces 
in all segments. %
The latter uniformly decomposes a visual segment into $N = n \times n$ blocks for $\mathbf{b}$. While face detection approach tends to give more direct focus on
face regions, our experiments show that block decomposition can provide attentions to regions of face track of the same speaker across segments.

\begin{figure}[t]
    \centering
    \includegraphics[width=1.0\textwidth]{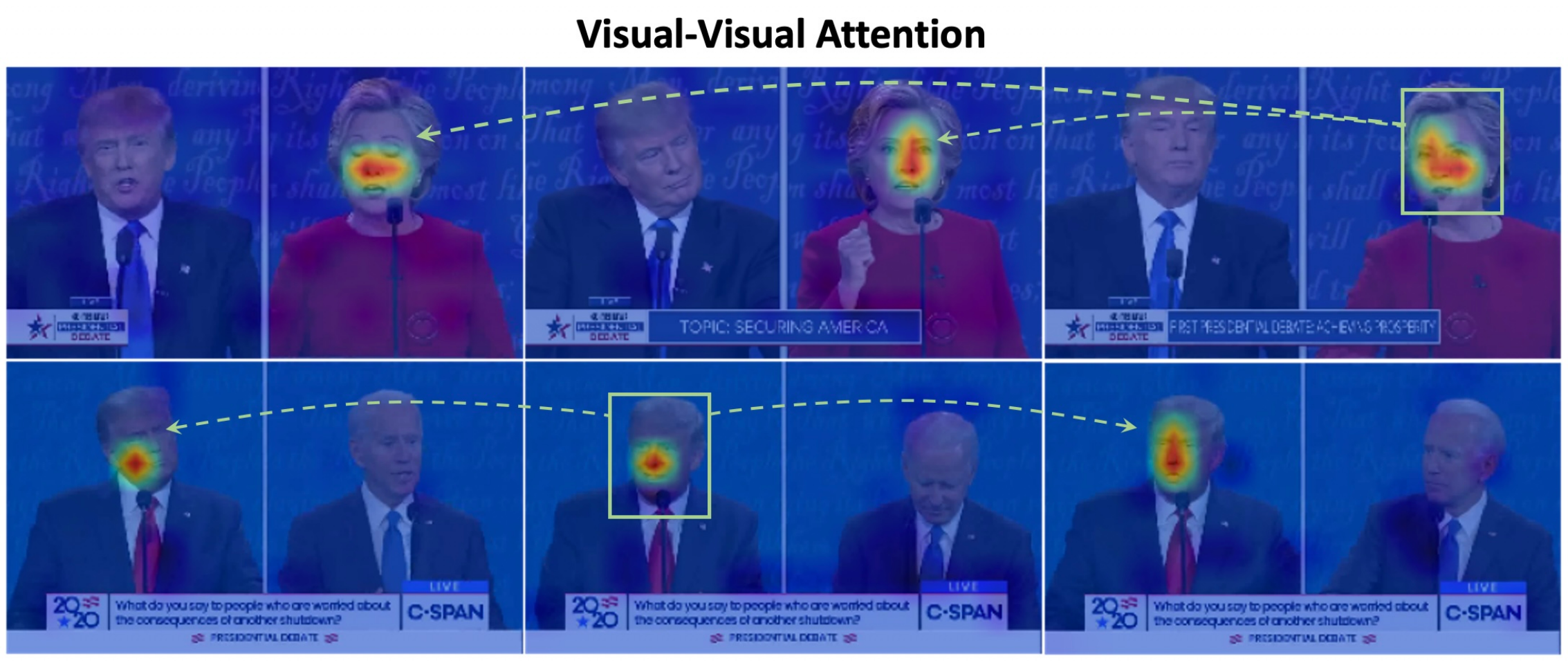}
    \caption{\textbf{Visual-Visual Attention.} The attention masks across video segments corresponding to the speaker in the green box. This type of attention can help to track the behaviors and interactions of each speaker over the  spatial-temporal  dimension.} %
    \label{fig:right2talk-Visual_Visual_attention}
\end{figure}

\noindent
{\textbf{Virtual Interaction Attention.}}
Given a feature set $\mathbf{f}^{k,i}_{v}$, the visual-visual context across the spatial-temporal dimension can be expressed as building a dynamic dictionary per feature set with three basic attention based elements
\cite{quach2021dyglip}, i.e. \textit{key, query, value}. While \textit{key} and \textit{query} are trained to support the dictionary look-up process where query feature is highly correlated to its matching key and dissimilar to others, \textit{value} represents a discriminative feature for each speaker. Particularly, the self-attention set $\{\mathbf{k}^{k,i}_v, \mathbf{q}^{k,i}_v, \mathbf{v}^{k,i}_v \}$ is extracted via three learnable projections $\{\Omega^Q_{v}, \Omega^K_{v}, \Omega^V_{v}\}$ as ${\mathbf{q}^{k,i}_{v}} = \Omega^Q_{v}(\mathbf{f}^{k,i}_{v}); {\mathbf{k}^{k,i}_{v}} = \Omega^K_{v}(\mathbf{f}^{k,i}_{v}); {\mathbf{v}^{k,i}_{v}} = \Omega^V_{v}(\mathbf{f}^{k,i}_{v})$. 
The visual correlation among speakers can be defined as $\alpha_{v}^{ki,k'j} = \sigma\left(\mathbf{q}^{k, i}_v ({\mathbf{k}^{k', j}_v})^{\top}/ \sqrt{d}\right)$, 
where $d$ is the feature dimension, $k'$ is a segment indexing variable. 
We consider the attention as a probability distribution that illustrates the responsive attention among speakers. Therefore, the softmax function can be adopted for $\sigma (\cdot)$.

\noindent
\textbf{Feature Refinement with attention.}
With these correlations, the visual self-attention
among speakers 
allows every speaker correlates to all other speakers through the spatial-time dimension. Then, the virtual interaction over speakers is explicitly embedded to their representations as $\mathbf{z}^{k, i}_v = \eta_{v}\left(\mathbf{f}^{k, i}_v + \sum_{k'=1}^K\sum_{j=1}^N\alpha^{ki, k'j}_v\mathbf{v}^{k,i}_v\right)$, 
where $\eta_v$ is the a residual-style MLP.
Throughout this process, the features of each speaker in one visual segment can interactively embed in their latent representation the correlations with those of the same speaker in other segments as well as other speakers of the same segment. Figure~\ref{fig:right2talk-Visual_Visual_attention} illustrates the attention mask across video segments corresponding to the speaker in the green box.

\noindent
\textbf{Audio-Audio Self-Attention.}
Similar to the virtual interaction, the audio self-attention is modeled as the correlation among audio segments. 
Particularly, let $F_a: \mathcal{A} \to \mathcal{F}_a$ be an audio embedding function that extracts audio latent representation for audio segments $\{\mathbf{a}^k\}_1^K$ and $\mathbf{f}^k_a = F_a(\mathbf{a}^k)$.
The audio self-attention correlation among segments can be computed as in Eqn.~\eqref{eqn:right2talk-att_aud_aud}.
\begin{equation} \label{eqn:right2talk-att_aud_aud}
    \begin{split}
        {\mathbf{q}^{k}_{a}} &= \Omega^Q_{a}(\mathbf{f}^{k}_{a}), \quad
        {\mathbf{k}^{k}_{a}} = \Omega^K_{a}(\mathbf{f}^{k}_{a}),  \quad
        {\mathbf{v}^{k}_{a}} = \Omega^V_{a}(\mathbf{f}^{k}_{a}), \\
        \alpha_{a}^{k,k'} &= \sigma\left(\mathbf{q}^{k}_a({\mathbf{k}^{k'}_a})^{\top}/\sqrt{d}\right), \quad
    \mathbf{z}^k_a = \eta_a(\mathbf{f}^k_a + \sum_{k'=1}^K \alpha_a^{k, k'}\mathbf{v}^{k'}_a)
    \end{split}
\end{equation}
The extracted audio feature of each segment is able to embed the correlation with other audio segments through time.

\subsubsection{Audio-Visual Correlation Learning}
\begin{figure}[!t]
    \centering
    \includegraphics[width=1.0\textwidth]{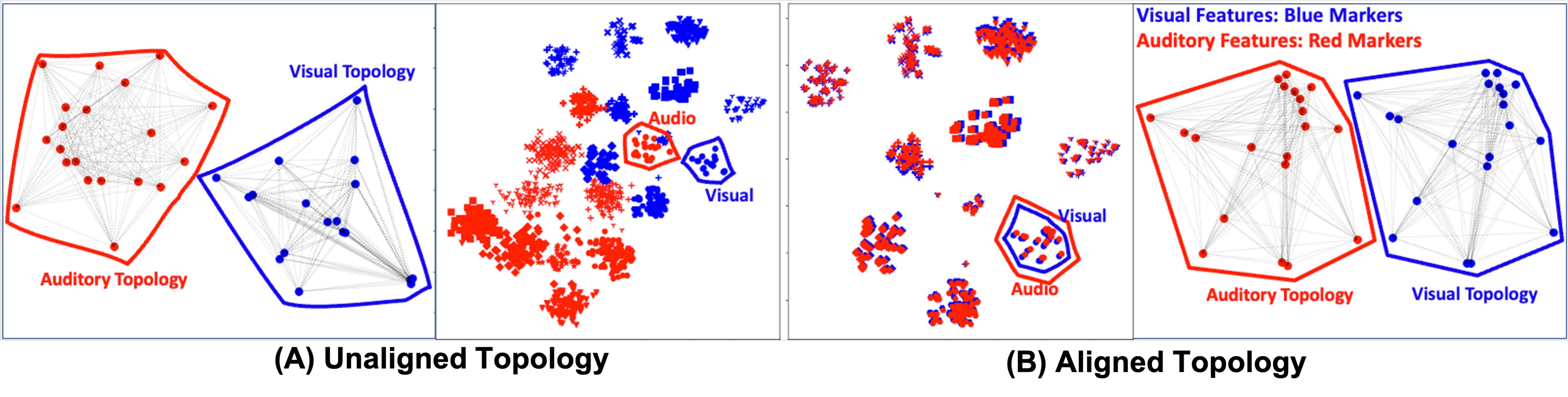}
    \caption{Topology of audio and visual feature domains.}
    \label{fig:right2talk-topo}
\end{figure}

The audio-visual correlations 
are computed from features of two different domains. As \textit{\textbf{topologies}}, i.e. \textit{how 
features distributed in the latent space and the correlations among its features} (see Figure~\ref{fig:right2talk-topo}), of these modalities may differ significantly, directly associating these features for correlation learning is not efficient.
One solution is to set up two encoders $E_v$ and $E_a$ extracting features from latent spaces of the same dimension to leverage the domain differences. However, the topology difference between these modalities may still present. To mitigate this issue, we 
align two domains' topologies before learning the correlations between their features. By this way, $\mathbf{z}^k_a$ and $\mathbf{z}^k_v$ are well aligned and their correlations can be fully exploited.

\noindent
\textbf{Cross Domain Alignment as Optimal Transport (OT) Problem.}
We present the distributions of visual and audio features by two distributions $p_v$ and $p_a$ where $\mathbf{z}_v \sim p_v(\mathbf{z}_v)$, and $\mathbf{z}_a \sim p_a(\mathbf{a}_a)$; and propose a two-stage alignment process: (1) Sample association between visual and audio samples via transport function $\pi$ and (2) Topology synchronization.
Formally, let $\pi$ be the transport function where $\pi_{i,i'}=\pi(\mathbf{z}_v^i, \mathbf{z}_a^{i'})$ indicates the probability of association between a visual sample $\mathbf{z}_v^i$ and an audio sample $\mathbf{z}_a^{i'}$. In addition, let $c_{p_v}(\cdot, \cdot)$ and $c_{p_a}(\cdot, \cdot)$ be the cost functions defined as the distance between two samples in visual and audio spaces, respectively. %
The alignment process is formulated with Gromov-Wasserstein distance as shown in Eqn.~\eqref{eqn:right2talk-GW_distance}.
\begin{equation} \label{eqn:right2talk-GW_distance} %
\begin{split}
    \mathcal{L}_{align} = GW^2(c_{p_v}, c_{p_a}, p_v, p_a) = \min_{\pi \in \Pi(p_v, p_a)}J(c_{p_v}, c_{p_a}, \pi)\\
    J(c_{p_v}, c_{p_a}, \pi) = \sum_{i,j,i',j'}|c_{p_v}(\mathbf{z}_v^i, \mathbf{z}_v^j) - c_{p_a}(\mathbf{z}_{a}^{i'}, \mathbf{z}_{a}^{j'})|^2\pi_{i,i'}\pi_{j,j'}
\end{split}
\raisetag{30pt}
\end{equation}
Intuitively, minimizing $J(c_{p_v}, c_{p_a}, \pi)$ aims at finding an appropriated association (i.e. via $\pi$) between samples in the two domains as well as minimizing the topology difference between them (i.e. via $c_{p_v}, c_{p_a}$). Notice that directly solving Eqn.~\eqref{eqn:right2talk-GW_distance} is costly due to the non-convex Quadratic Problem with the time complexity is $O(n^3)$. Therefore, we adopt the the sliced approach~\cite{vay_sgw_2019} for a fast computation of $\mathcal{L}_{align}$. Figure~\ref{fig:right2talk-topo} illustrates visual (blue points) and auditory features (red points) extracted from 500 clip segments of 10 different speakers (e.g. denoted by different markers) and projected into the 2D space using t-SNE method. %
Thanks to $\mathcal{L}_{align}$ in the alignment stage, visual and auditory features are brought into similar distributions (Figure~\ref{fig:right2talk-topo} (B) (left)) with more aligned feature distributions (Figure~\ref{fig:right2talk-topo} (B) (right)).

\noindent
\textbf{Audio-Visual Correlation.}
With the aligned visual and audio features, we further adopt similar attention mechanism to learn the associations between the visual features of each SROI $\mathbf{z}^{k,i}_{v}$  and the audio features $\mathbf{z}^k_a$ in each segment as in Eqn.~\eqref{eqn:right2talk-visual_att_aud_aud}.
\begin{equation} \label{eqn:right2talk-visual_att_aud_aud}
\small
    \begin{split}
        {\mathbf{q}^{k}} &= \Omega^Q(\mathbf{z}^{k}_{a}), \quad
        {\mathbf{k}^{k, i}} = \Omega^K(\mathbf{z}^{k, i}_{v}), \quad
        {\mathbf{v}^{k, i}} = \Omega^V(\mathbf{z}^{k, i}_{v}) \\
        \alpha^{k,k'i} &= \sigma\left(\mathbf{q}^{k}({\mathbf{k}^{k', i}})^{\top}/\sqrt{d}\right), 
        \quad \mathbf{z}^k = \phi(\mathbf{z}^k_a, \mathbf{z}^k_v) = \eta\left(\mathbf{z}^k_a + \sum_{k'=1}^K\sum_{i=1}^N \alpha^{k, k'i}\mathbf{v}^{k', i}\right)
    \end{split}
\end{equation}

\begin{figure}[!b]
    \centering
    \includegraphics[width=1.0\textwidth]{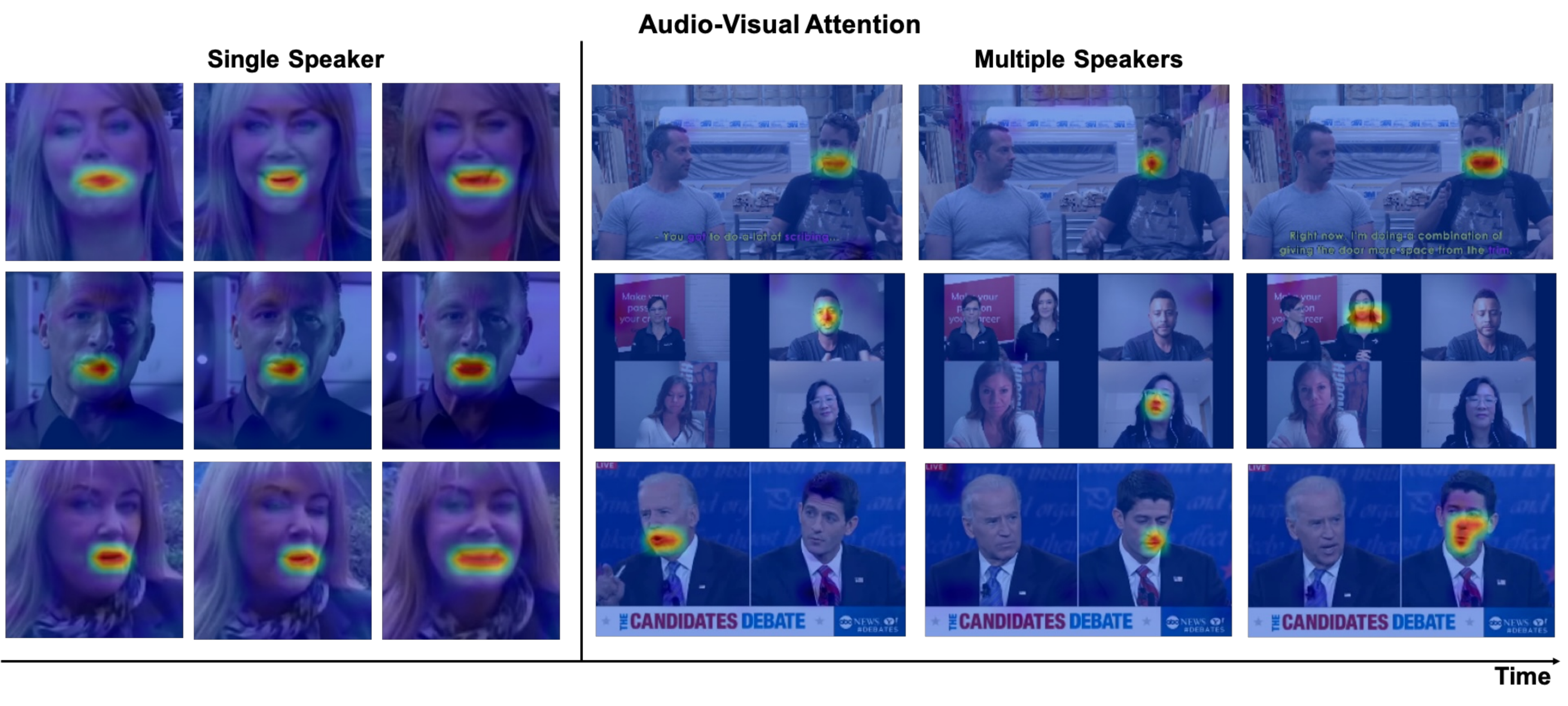}
    \caption{\textbf{Audio-Visual Attention To The Main Speaker.} The audio-visual attention mask across segments illustrates the response of audio to the visual.
    }
    \label{fig:right2talk-attnetion_visual_audio_speaker}
\end{figure}

The attention matrix 
assesses how much an audio responds to an SROI in the spatial-temporal dimension. A high response indicates a high correlation between the audio and the speaker associated with that SROI. This association embeds the probability of a speaker to be an active speaker of the audio segment. Figure~\ref{fig:right2talk-attnetion_visual_audio_speaker} illustrates audio-visual attentions in both single and multiple speaker conversation.

\subsubsection{Main Speaker Localization and Audio Separation with Conversation Grammar} \label{sec:LearningProcess}
Given the audio-visual attentional features $\mathbf{z}^k$ from previous step, the audio-visual masks are computed as in Eqn.~\eqref{eqn:right2talk-audio_visual_mask}.
\begin{equation} \label{eqn:right2talk-audio_visual_mask}
\small
\begin{split}
    \mathbf{M}_v^k(x,y) &= \begin{cases}
    \alpha^{k,ki}       &  \text{if } (x,y) \in \mathbf{b}_i^k, i=1..N\\
    0  & \text{otherwise}
  \end{cases} \\
  \mathbf{M}_a^k &= D(\mathbf{z}^k)
\end{split}
\end{equation}
where $D$ is a learnable decoder that maps $\mathbf{z}^k$ to the target audio mask; and $\alpha^{k,ki}$ denotes the correlation score between the audio and SROI of $i$-th speaker in the $k$-th segment.
The objective functions of Eqn.~\eqref{eqn:right2talk-object_learning_visual} to learn $\mathbf{M}_a^k$ and $\mathbf{M}_v$ can be reformulated as in Eqn.~\eqref{eqn:right2talk-objective_learning_mask}.
\begin{equation} \label{eqn:right2talk-objective_learning_mask}
\small
\begin{split}
    \mathcal{L}_{visual} &= \mathbb{E} \left[-\log \frac{e^{\alpha^{k,ki}}}{e^{\alpha^{k,ki}} + \sum_{j\neq i}e^{\alpha^{k,kj}}}\right], \quad
    \mathcal{L}_{audio} = \mathbb{E} \left[|| \mathbf{M}^k_a \odot Spec(\mathbf{a}^k) - Spec(\mathbf{a}_{S^k_m})||_1\right]
\end{split}
\end{equation}
Intuitively, on one hand, $\mathcal{L}_{audio}$ optimizes the model toward voice of the target (Main) speaker. On the other hand, $\mathcal{L}_{visual}$ aims at increasing the correlations between the audio and SROI of the target speaker, while reducing the correlation with other SROIs in spatial-temporal dimensions.

\begin{wrapfigure}{r}{0.5\textwidth}
	\centering \includegraphics[width=0.5\columnwidth]{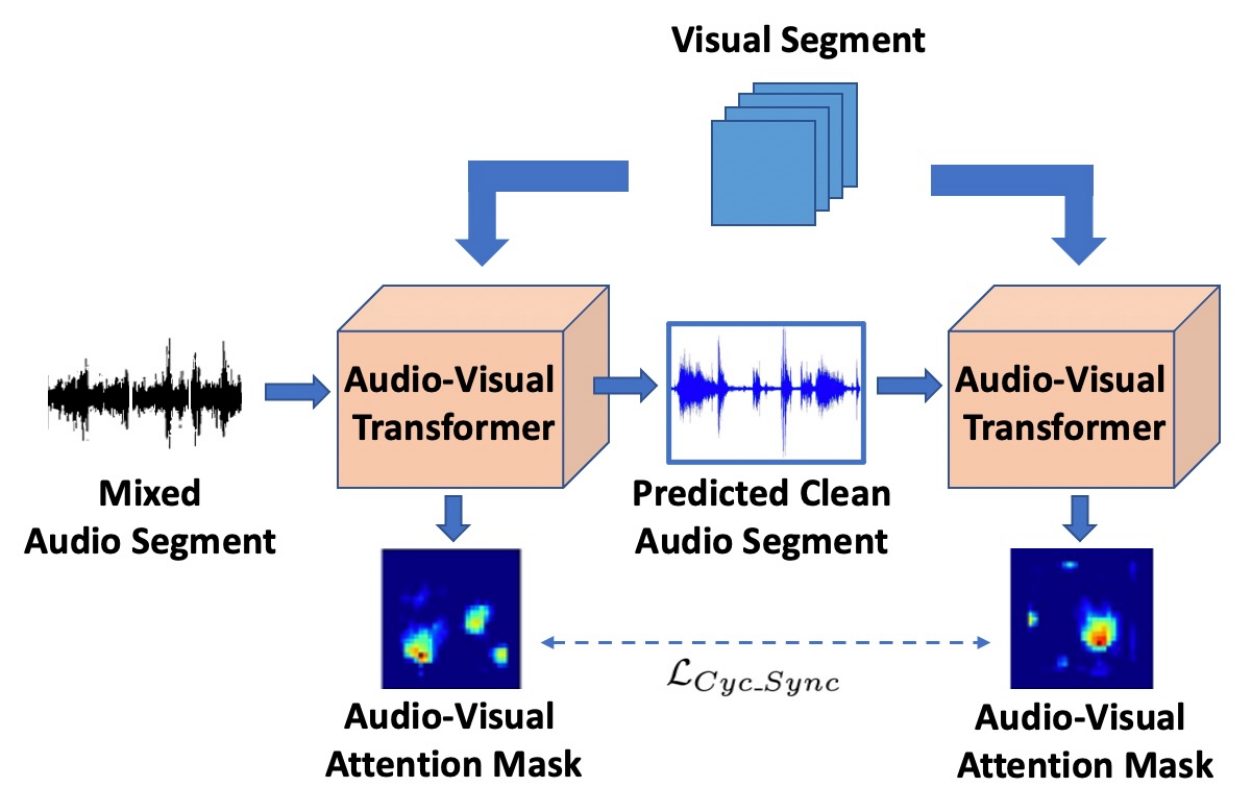}
	\small
	\caption{\textbf{Cycle Synchronization Loss.}}
	\label{fig:right2talk-Loss_Cyc_Sync}
\end{wrapfigure}

\noindent
\textbf{Self-supervised Learning.}
While our goal is to develop a self-supervised model that learns to localize the main speaker, the ground truth location for Main speaker is absent during the training stage. Therefore, we further propose a self-supervised version of $\mathcal{L}_{visual}$ , namely \textit{Cycle Synchronization Loss}, defined as in Eqn.~\eqref{eqn:right2talk-cycle_loss}.
\begin{equation} \label{eqn:right2talk-cycle_loss}
\begin{split}
    \mathcal{L}_{Cyc\_Sync} &= \mathbb{E} \left[ ||\alpha^{k,ki} - \hat{\alpha}^{k,ki}_{\text{max}}||_1\right]\\
    \hat{\alpha}^{k,ki}_{\text{max}} &= \begin{cases}
    \hat{\alpha}^{k,ki}       &  \text{if } (x,y) \in \mathbf{b}_{i^*}^k\\
    0  & \text{otherwise}
  \end{cases}; 
  i^* = \arg \max_i \hat{\alpha}^{k,ki}
\end{split}
\end{equation}
where $\hat{\alpha}^{k,ki}$ is the correlation between the predicted clean voice of the target speaker. The intuition of $\mathcal{L}_{Cyc\_Sync}$ is illustrated in Figure~\ref{fig:right2talk-Loss_Cyc_Sync} where the goal is to penalize the consistency between two terms: (1) the correlations of the input (mix) voices $\mathbf{a}^k$ and the visual component; and (2) the maximum correlations of the predicted (clean) voice of the target and the visual component.
As clean voice is of a single speaker and it reflects similar linguistic content as visual features of the target speaker, its correlations with the visual component can efficiently act as the guidance for localization process.
Moreover, by considering only the maximum $\hat{\alpha}^{k,ki}$ in $\hat{\alpha}^{k,ki}_{\text{max}}$, the correlation between the visual of other speakers (i.e. interrupter) and audio is also minimized.%

\noindent
\textbf{Learning with Conversation Grammar.} 
We adopt the mix-and-separate strategy~\cite{ephrat2018looking, Zhao_2018_ECCV} to obtain the ground truth for audio separation task of $\mathcal{L}_{audio}$ and further extend it with two types of Conversation Grammar, i.e. cooperative and competitive modes.  
In the first type, each speaker takes turn to speak during the conversation and the role changing happens when a speaker finishes his/her speech. 
In the second type, mixing voices happen during the interruption of other speakers. 
From these grammars, we synthesize a video training set containing multiple speakers by (1) randomly selecting different videos in the single subject training set; (2) concatenating these videos sequentially (i.e. cooperative mode); (3) mixing their voices in a short time window and vertically concatenating the video frames (i.e. competitive mode).
In all cases, $S^k_m$ is set to the one who occupies the audio segment or the all segments of the whole video, accordingly.
The Audio-Visual Transformer is optimized as in Eqn.~\eqref{eqn:right2talk-final_object_r2t}.
\begin{equation}\label{eqn:right2talk-final_object_r2t}
    \mathcal{L} = \alpha_{align} \mathcal{L}_{align} + \alpha_{visual} \mathcal{L}_{Cyc\_Sync} + \alpha_{audio} \mathcal{L}_{audio}
\end{equation}
where parameters $\{\alpha_{align}, \alpha_{visual}, \alpha_{audio}\}$ control the relative importance.

\subsection{Experimental Results}

\noindent
\textbf{Data Setting.} Our training data include 29 hours of training videos from Lip Reading Sentences 2 (LRS2)~\cite{LRS2}, and synthetic videos obtaining as presented in Sect. \ref{sec:LearningProcess}.
The length of synthetic segments varies from 4s to 8s decomposed into 2s short segments.
The overlapped ratio of the mixing voice is set to $\frac{1}{3}$.
For validation, we adopt the testing set of LRS2, Lip Reading Sentences 3 (LRS3)~\cite{LRS3}, Columbia~\cite{Columbia}, and the main speaker dataset~\cite{truong2021right2talk}.
While LRS2 and LRS3 include 0.5-hour to 1-hour testing videos, Columbia includes an 86-minute panel discussion. %
We adopt the ground truth for each active speaker in Columbia while annotating
bounding boxes using face detection~\cite{deng2019retinaface} for LRS2 and LRS3.

\noindent
\textbf{Audio Data Preprocess.} We employ a Short-Time Fourier Transform (STFT) to represent the audio signal. Our STFT uses the Han window function, which generates the magnitude and phase of spectrograms. %
We set the hop length of 10 ms with a window length of 40ms at a sample rate of 16000Hz.

\noindent
\textbf{Visual Data Preprocess.} All training videos are re-sampled to a resolution of $160 \times 160$ pixels at $25$ FPS. This chosen resolution results in a feature map composing $N = 6 \times 6$ blocks.
During the testing phase, we only re-sample the input video to $25$ FPS and retain the original resolution.

\noindent
\textbf{Implementation.} Our framework is implemented in PyTorch~\cite{paszke2019pytorch} and all the models are trained on a machine with four NVIDIA P6000 GPUs. The batch size is set to 32 for each GPU. We use RMSProp optimizer with the started learning rate of 0.0001. 
We set the control parameters to $1.0$, i.e $\alpha_{align} = \alpha_{visual} = \alpha_{audio} = 1.0$. 
We employ 3D VGG-style network for visual deep feature embedding $F_v$, and 2D VGG-style network for audio embedding $F_a$.
The linear projections $\{\Omega_v^Q, \Omega_v^K, \Omega_v^V, \Omega_a^Q, \Omega_a^K, \Omega_a^V, \Omega^Q, \Omega^K, \Omega^V \}$ are implemented as the fully connected layers that project features to $512-D$ spaces. The mapping functions $\{\eta_v, \eta_a, \eta\}$ are implemented as residual-style MLP
consisting of 2 fully connected layers followed by the normalization layer (the dimension of hidden layers is set to $1024$).
The audio-visual mask generator $D$ is implemented by a stack of 2 fully connected layers, which predicts both the magnitude mask and the phase mask of the spectrogram. We use the RetinaFace~\cite{deng2019retinaface} 
for face detection widely used in face recognition 
\cite{Chen_FG2011, duong2019shrinkteanet, 9185981, Duong_ICASSP2011, Le_JPR2015, Luu_BTAS2009, Luu_FG2011, Luu_IJCB2011, 9108692}.

\begin{figure}[t]
    \centering
    \includegraphics[width=0.8\textwidth]{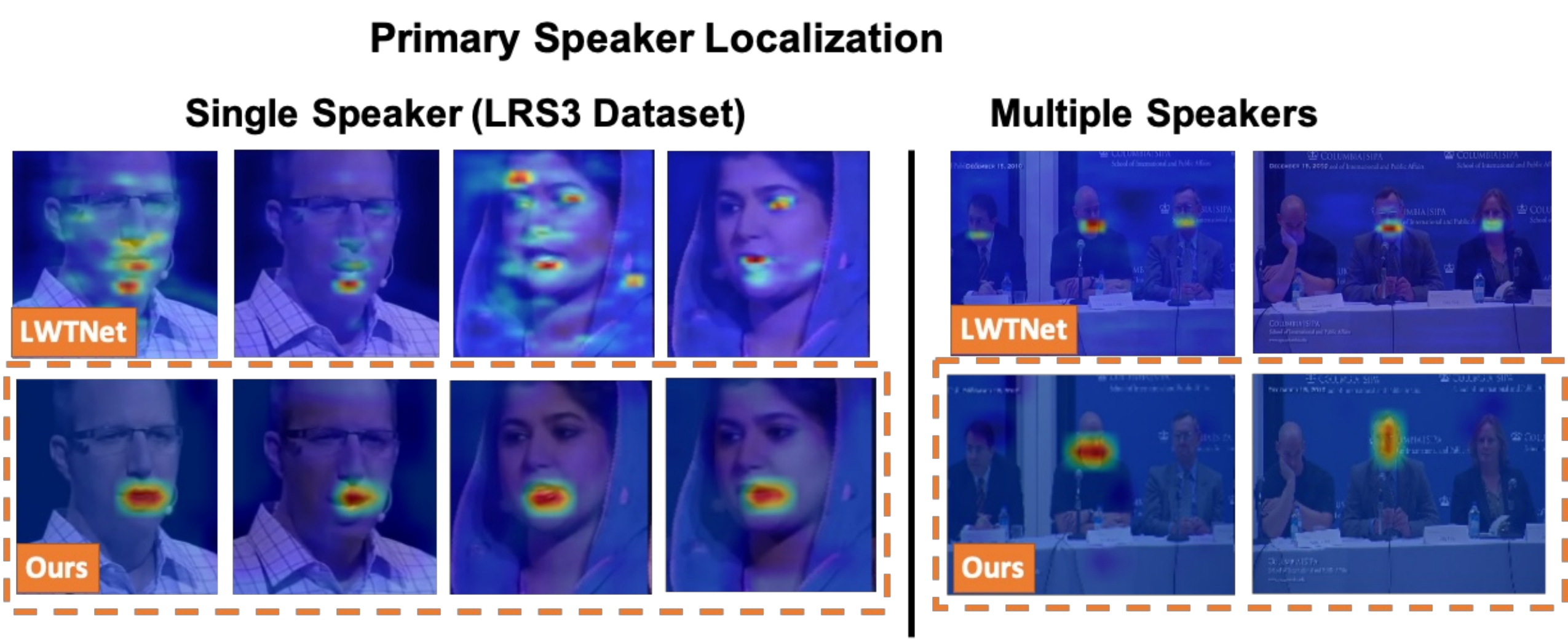}
    \caption{\textbf{Main Speaker Localization.}
    Visualization of attention mask localizing the main speaker. }
    \label{fig:right2talk-visualize_primary_speaker_localization}
\end{figure}

\input{Tables/chap-5/right2talk/alignment-ablation}
\noindent
\textbf{Evaluation Metrics.} To compare against prior methods, we adopt four common metrics for localization and audio separation tasks. 
For single speaker videos, a localization is correct if its lies in the ground-truth bounding box of Main speakers. For multiple-speaker videos, F1 score is adopted for validation. 
To evaluate Main speaker separation, we adopt the protocol of multi-source speaker audio separation, and estimate the Signal-to-Distortion-Ratio (SDR)~\cite{fevotte2005bss_eval} and Perceptual Evaluation of Speech Quality (PESQ)~\cite{rix2001perceptual}.%

\input{Tables/chap-5/right2talk/interupting-conversation}
\noindent
\textbf{Ablation Study.}
To study the effectiveness of our proposed cross-domain alignment method, we employ an ablation study with audio separation task 
on LRS2 using two configurations: without and with $\mathcal{L}_{align}$. 
We create synthetic testing video samples from LRS2 by combining audios from multiple videos. Three use-cases are evaluated including a primary  voice with background noise (1S+N); a primary voice mixed another speaker's voice (2S), and a primary voice mixed another speaker's voice plus background noise (2S+N). We report SDR (dB) and PESQ metrics for these cases in Table~\ref{tab:right2talk-AudioSeparation_Align}. By aligning the features of the two domains, the audio-visual correlations can be efficiently extracted and help to consistently improve SDR in all cases.%

\subsubsection{Main Speaker Localization}

\noindent
\textbf{{Competitive Turn-Taking Conversation.}}
This type is more challenging as two speakers may speak at the same time. 
Therefore, although the two speakers can be both active speakers, only one of them is considered as the main speaker while the other one is the interrupter. %
For this task, beside the Random Pixel and Center Pixel baselines, we consider two additional localization strategies. We firstly employ LWTNet~\cite{Afouras20b} to localize all active speakers of each video segment and then choose the main speaker as the one with (1) larger audio magnitude (i.e. \textit{large\_mag}), and (2) maximal audio-visual correlation  (i.e. \textit{high\_corr}).
Table~\ref{tab:right2talk-InterruptingConversation} reports the localization accuracy on the main speaker dataset~\cite{truong2021right2talk} in terms of F1 score against the four baseline approaches.
These results again emphasizes the advantages of 
our proposed approach in the capability of automatically and robustly localize the main speaker in a conversation. 
The achieved improvements comes from three properties of the proposed model: (1) the present of the contextual attention from both visual and audio domains; (2) the domain feature alignment, and (3) the Cycle Synchronization Loss $\mathcal{L}_{Cyc\_Sync}$ that minimizes the disparity between the localization masks obtained from mixed voices and clean voice. %

\input{Tables/chap-5/right2talk/inturn-localization}

\noindent
\textbf{{Cooperative Turn-Taking Conversation.}} In this type, as each speaker takes his/her turn to join the conversation, the main speaker is also the one who is actively speaking during the conversation. The localization accuracy for both single-speaker and multiple-speaker conversations in comparison to previous Active Speaker Detection approaches is reported in Table~\ref{tab:right2talk-InturnLocalization}. 
For each training mode of our model, we also include the configurations that take into account the correlations within and across segments. 
As can be seen, with the attention mechanisms as well as the domain alignment process, the visual and audio features are better correlated and provide more accurate locations of the main speaker. Moreover, when the spatial-temporal dimension is adopted in configuration (B) and (D), the performance is further boosted. Thanks to the correlations across segments (shown in Figure~\ref{fig:right2talk-Visual_Visual_attention}), the location of each speaker is highly correlated with face of the same subject in other segments and, therefore, enable the tracking consistency of that speaker during the conversation. 
Our approach outperforms LWTNet~\cite{Afouras20b} in all datasets with the margins from $0.1\%$ to $4.1\%$.
Figure~\ref{fig:right2talk-visualize_primary_speaker_localization} shows our  %
localization results compared to LWTNet.

\subsubsection{Main Speaker Audio Separation}

\input{Tables/chap-5/right2talk/audio-separation}
To quantitatively evaluate the capability of audio separation for the proposed 
approach, we employ the evaluation protocol of~\cite{Afouras20b} and use SDR and PESQ as the validation metrics. Similar to the previous section, we create synthetic testing videos from LRS2 on three cases, i.e. 1S + N, 2S, and 2S + N, and evaluate different configurations of our approach in comparison to previous methods as shown in Table~\ref{tab:right2talk-AudioSeparation}. 
With the spatial-temporal attentions, all configurations that take into account the cross-segment correlations get improvements from $0.3$ to $1.2$dB of SDR when separating voices of two speakers. Furthermore, the audio-visual attentions also give more cues to improve the separation process. 
We validate the roles of SROI by adopting two strategies (see Sect. \ref{sec:Visual_Self_Att}), i.e. block decomposition and face detection. Although the use of face detection can give more focus on face regions and produce further improvements, the block decomposition approach can still attend to the track of the same speaker across segments and give competitive performance. Moreover, our approach with both configurations outperforms LWTNet~\cite{Afouras20b} in SDR and PESQ.

%% file: Tables/chap-5/right2talk/multimodal-comparison.tex
\begin{table} [t]
    \small
	\centering
	\caption{Comparisons of our proposed approach and other modeling methods. Sound Source Localization (SSL).}\label{tab:right2talk-TenMethodSumm}
 \resizebox{1.0\textwidth}{!}{
\begin{tabular}{ | >{\arraybackslash}  m{3.2cm} | c c c c c|}
  \hline
		& \textbf{Ours} & LWTNet\cite{Afouras20b} & SyncNet \cite{chung2016out}  & SoundOfPixel \cite{Zhao_2018_ECCV}  & CocktailParty \cite{ephrat2018looking}  \\
		
  \hline

		\textbf{Goal} & \begin{tabular}{@{}c@{}} \textbf{Main Speaker} \\ \textbf{Highlight} \end{tabular} & \begin{tabular}{@{}c@{}} Active Speaker \\Highlight\end{tabular}   & SSL & Audio Separation & Audio Separation \\ %
		\hdashline
		\textbf{Temporal Model} & \textbf{Across-Segments} & Within-Segment  & Within-Segment & Within-Segment & Within-Segment \\ %
		\hdashline
		\textbf{People-Independent} & \cmark & \cmark & \xmark & \xmark & \cmark \\
  \hdashline
		\textbf{Visual Context modeling (Visual-visual attention)} & \cmark & \xmark & \xmark & \xmark & \xmark \\
        \hdashline
		\textbf{Audio-Visual Correlation} & \begin{tabular}{@{}c@{}} \textbf{Audio-Visual} \\ \textbf{Transformer} \end{tabular} & \begin{tabular}{@{}c@{}}Cosine \\ Distance\end{tabular}  & \begin{tabular}{@{}c@{}} Audio-Visual \\Synchronization \end{tabular} & \begin{tabular}{@{}c@{}} Feature \\Concatenation \end{tabular}& \begin{tabular}{@{}c@{}} Feature \\Concatenation \end{tabular} \\
  \hline
	\end{tabular}
 }
\end{table}

%% file: Tables/chap-5/right2talk/alignment-ablation.tex
\begin{wraptable}{r}{0.5\textwidth}
    \centering
    \caption{\textbf{Main Speaker Audio Separation on LRS2 with and without Domain Alignment}. The higher value is better.}
    \label{tab:right2talk-AudioSeparation_Align}
    \resizebox{0.5\textwidth}{!}{
    \begin{tabular}{|c|c|c|c|}
    \hline
    \multicolumn{2}{|c|}{}         & \textbf{Ours w/o $\mathcal{L}_{align}$} &\textbf{ Ours W $\mathcal{L}_{align}$} \\
    \hline
    \multirow{3}{*}{\textbf{SDR (dB)}}  & \textbf{1S+N} &    14.4     &    \textbf{15.8}       \\
    \cline{2-4}
                      & \textbf{2S }  &    9.8       &      \textbf{10.3}     \\
    \cline{2-4}                    
                      & \textbf{2S+N} &    7.0      &     \textbf{7.2}      \\
    \hline
    \multirow{3}{*}{\textbf{PESQ}} & \textbf{1S+N} &     3.1        &    \textbf{3.3}       \\
    \cline{2-4}
                      & \textbf{2S }  &      2.7       &    \textbf{2.9}       \\
    \cline{2-4}
                      &\textbf{ 2S+N }&       2.5      &    \textbf{2.5}      \\
    \hline
    \end{tabular}
    }
\end{wraptable}

%% file: Tables/chap-5/right2talk/interupting-conversation.tex
\begin{wraptable}{r}{0.5\textwidth}
    \centering
    \caption{\textbf{Main Speaker Localization Accuracy in Competitive Turn-Taking Conversation.}} 
    \label{tab:right2talk-InterruptingConversation}
    \resizebox{0.5\textwidth}{!}{
     \begin{tabular}{|l|ccc|}
        \hline
        \multirow{1}{*}{} & \multirow{1}{*}{\textbf{Discussion}} & \multirow{1}{*}{\textbf{Tele}} & \multirow{2}{*}{\textbf{Debate}} \\
          \multirow{1}{*}{} & \multirow{1}{*}{\textbf{Panel}} &  \textbf{Conf}& \\
        
        \hline
        Baseline (Random) &4.8\%  &3.0\%  & 9.5\%\\
a            Baseline (Center) &1.3\%  &1.1\%  & 5.9\%\\
        \hline
       
         LWTNet\cite{Afouras20b}+large\_mag & 62.8\% & 55.07\%  & 58.7\% \\
         LWTNet\cite{Afouras20b}+high\_corr & 88.0\% & 80.0\%  &63.3\% 
        \\
        \hline
        \textbf{Ours} & \textbf{90.2\%} & \textbf{83.3\%} & \textbf{69.4\%}\\
         \hline
    \end{tabular}
    }
\end{wraptable}

%% file: Tables/chap-5/right2talk/inturn-localization.tex
\begin{table} [!t]
	\centering
	\caption{\textbf{Main Speaker Localization Accuracy with Cooperative Turn-Taking Conversation} (\%). For LRS2 and LRS3, a localization is considered correct if it lies within the true bounding box. For Columbia, F1 score is adopted.} \label{tab:right2talk-InturnLocalization}
    \resizebox{1.0\textwidth}{!}{
	 \begin{tabular}{|l|cc|c|ccccc|}
    \hline
     \multirow{3}{*}{}&\multicolumn{2}{c|}{\textbf{Single Speaker}} & \multicolumn{6}{c|}{\textbf{Multiple Speakers}}\\
     \cline{2-9}
      \multirow{2}{*}{} & \multirow{2}{*}{\textbf{LRS2}} & \multirow{2}{*}{\textbf{LRS3}} & \multirow{1}{*}{\textbf{Columbia}} & \multicolumn{5}{c|}{\textbf{Columbia} (Per subject)}\\
      & \multicolumn{2}{c|}{}& \textbf{(Avg)} & \textbf{Bell} & \textbf{Boll} & \textbf{Lieb} & \textbf{Long} & \textbf{Sick}\\
    
    \hline
    Baseline (Random Pixel) & 2.8\% & 2.9\% & 8.5\% &7.8\%  &8.6\% &9.9\% &7.9\% &8.7\%\\
    Baseline (Center Pixel) & 23.9\% & 25.9\% & 14.9\% &13.0\%  &11.5\%  &21.4\% &19.2\% &17.9\%\\
    \hline
    Multisensory \cite{owens2018audio} & 99.3\% & 24.8\% & 52.7\% &  52.0\%& 43.8\% &62.3\% &64.8\% &60.9\%\\
    Chakravarty et al.\cite{Columbia}  & $-$ & $-$& 80.2\% & 82.9\%& 65.8\%& 73.6\%& 86.9\%& 81.8\%\\
    
    SyncNet \cite{chung2016out}& $-$ & $-$ & 89.5\% &  93.7\%& 83.4\%& 86.8\%& \textbf{97.7\%}& 86.1\%\\
	LWTNet \cite{Afouras20b} & 99.6\%& 99.7\%& 90.8\%& 92.6\%& 82.4\%& 88.7\%& 94.4\%&95.9\%\\
	\hline
	\hline
    (A) \textbf{Ours - Block attention} & 99.7\%& 99.8\%& 92.7\% & 93.7\% & 85.0\% & 87.5\% & 92.8\% &  97.2\%\\
    (B) \quad +  across segments& \textbf{99.8\%}  & \textbf{99.9\%} & 93.4\% & 95.8\% &  85.0\% & 87.5\% & 92.8\% & 97.2\% \\
     \hline
    (C) \textbf{Ours - Speaker Attention \footnotemark} & 100\% & 100\% & 93.8\% &  95.8\% & 85.0\% & 91.6\% & 92.8\% & 97.2\%\\
	(D) \quad +  across segments & \textbf{100\%} & \textbf{100\%} & \textbf{94.9\%} &  \textbf{95.8\%} & \textbf{88.5\%} & \textbf{91.6\%} & 96.4\% & \textbf{97.2\%} \\
     \hline
\end{tabular}
}
	
\end{table}

%% file: Tables/chap-5/right2talk/audio-separation.tex
\begin{wraptable}{r}{0.6\textwidth}
	
	\centering
    	\caption{\textbf{Main Speaker Audio Separation on LRS2.}}
    	 \label{tab:right2talk-AudioSeparation}
      \resizebox{0.6\textwidth}{!}{
    	 \begin{tabular}{|l|ccc|ccc|} %
        \hline
         \multirow{3}{*}{}&\multicolumn{3}{c|}{\textbf{SDR (dB)$\boldsymbol\uparrow$}}&\multicolumn{3}{c|}{\textbf{PESQ}$\boldsymbol\uparrow$} \\ 
         \cline{2-7}
          \multirow{1}{*}{} & \multirow{1}{*}{\textbf{1S+N}} & \multirow{1}{*}{\textbf{2S}} & \multirow{1}{*}{\textbf{2S+N}} & \multirow{1}{*}{\textbf{1S+N}} & \multirow{1}{*}{\textbf{2S}} & \multirow{1}{*}{\textbf{2S+N}} \\
        \hline
        Mix input &1.3&1.31 &0.6 &1.1&1.1 &1.0  \\ %
        \hline
        SoundOfPixel \cite{Zhao_2018_ECCV} &9.4  & 1.5  &0.5 &1.2  &1.1  &1.0  \\ %
        Deep-Clustering \cite{hershey2016deep} & 9.0 & 6.0 & 3.2 & 2.3 & 2.3 & 1.9 \\ %
        Conv-TasNet \cite{PMID:31485462} & $-$ & 10.7 & $-$ & $-$ & $-$ & $-$ \\
        LWTNet \cite{Afouras20b} & $-$ &10.8  & $-$ & $-$ &3.0  & $-$  \\
    	\hline
    	\hline
    	\textbf{Ours (Audio Only)} & 11.1  & 9.1 & 7.0 & 2.8 & 2.8 & 2.5 \\ %
    	\quad +  across segments & 11.2  & 9.4 & 7.1 & 2.8 & 2.9 & 2.6  \\
        \hline
    	(A) \textbf{Ours - Block Attention} & 15.8 & 10.3 & 7.2 & 3.3 & 2.9 & 2.5  \\
    	(B) \quad +  across segments & 16.6 & 11.5 & 8.1 & \textbf{3.6} & 3.1 & 2.7  \\
    
    	(C) \textbf{Ours - Speaker Attention} & 16.5 & 10.5 & 7.5 & 3.4 & 3.0 &  3.0  \\
    	(D) \quad +  across segments & \textbf{16.7} & \textbf{11.6} & \textbf{8.2} & 3.4 & \textbf{3.1} & \textbf{3.1} \\
         \hline
        \end{tabular}
        }

\end{wraptable}

%% file: Chapters/Sections/chap-6/direcformer.tex
\section{A Directed Attention in Transformer Approach to Efficient Temporal Learning}

\setcounter{propositioncounter}{0}
\setcounter{remarkcounter}{0}

Video understanding has recently become one of the popular research topics in the computer vision community. Video data has become ubiquitous and occurs in numerous daily activities and applications, e.g., movies and camera surveillance~\cite{Duong_2017_ICCV, le2018segmentation, truong2021fastflow, duong2029cvpr_automatic}. 
In the field of video understanding~\cite{dat2021bimal_iccv}, action recognition has become a fundamental problem. 
In action recognition, there is a need to pay more attention to the temporal structures of the video sequences. Indeed, emphasis on temporal modeling is a common strategy among most methods. It can be considered as the main difference between video and images. These works include long-short term dependencies~\cite{duong2019learning, duong2019dam_ijcv}, temporal structure, low-level motion, and action modeling as a sequence of events or states.

The current methods in video action recognition utilize 3D or pseudo-3D convolution to extract the spatio-temporal features~\cite{c3d, i3d, s3d, p3d}. However, these 3D CNN-based methods suffer from intensive computation with many parameters to be learned. Others attempt to adopt two-stream structures~\cite{2streams_simonyan, 2streams_feichtenhofer, spatiotemporal_resnet, slowfast} for accurate action recognition since information from one branch could be fused to the other one. Some methods in this category require computing the optical flow first, which could be time-consuming and require a large amount of storage. Others apply 3D convolution to avoid computing the optical flow. Nonetheless, this approach also requires a large amount of computational resources to implement.

Although prior methods~\cite{crosstransformer,noframe,actionnet,vatt,tdn} have achieved remarkable performance, they have several limitations related to the robustness of the models. In this chapter, we, therefore, address two fundamental questions for current action recognition models.
In the first question, given a set of video frames \textit{shuffled in a random order} and different from the original one, will it be classified as the same label as the original recognition result?
If this is the case, these models have been clearly overfitted or biased to other factors (e.g., scene background) rather than learning semantic information about the actions. 
In the second question, we want to understand whether these action recognition models are able to \textit{correct the incorrectly-ordered frames} to the right ones and provide an accurate prediction? Finally, we introduce a new theory to improve the robustness and generalization of the action recognition models.

Toward our desired goal, we present a new end-to-end Transformer-based Directed Attention (DirecFormer) approach to robust action recognition. Our method takes a simple but novel perspective of the Transformer-based approach to learning the right order of a sequence of actions. This work makes three contributions. 
First, we introduce the problem of ordered temporal learning in action recognition. Second, a new Directed Attention mechanism is introduced to provide human action attention in the right order. Third, we introduce the conditional dependency in action sequence modeling that includes orders and classes.

\begin{figure*}
    \centering
    \includegraphics[width=0.85\textwidth]{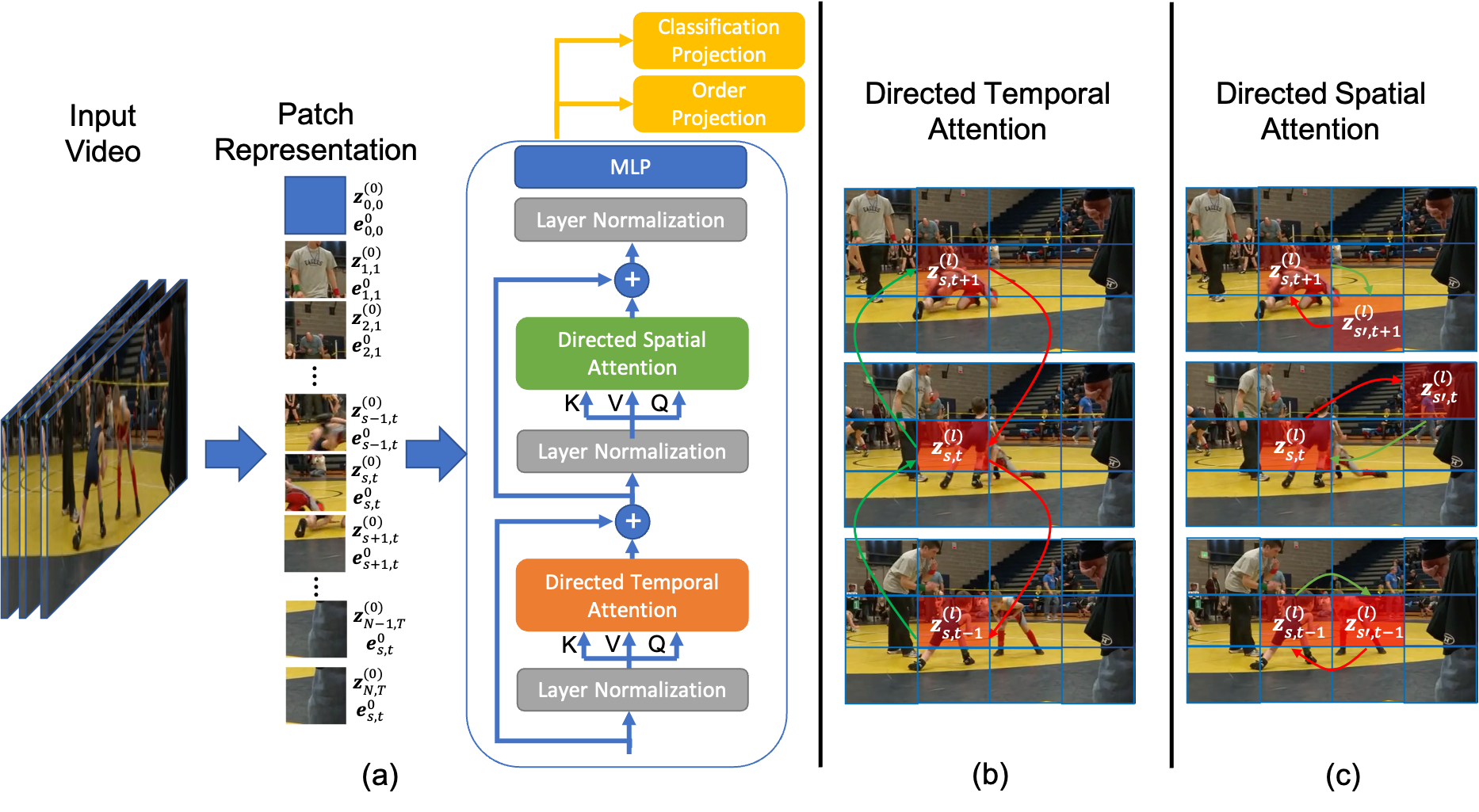}
    \caption{\textbf{The Proposed Framework.} (a) The Proposed DirecFormer. (b) Directed Temporal Attention. (c) Directed Spatial Attention. The green arrows in (b) and (c) denotes for positive correlation and the red arrows denotes for negative correlation.}
    \label{fig:direcformer-framework}
\end{figure*}

\subsection{The Proposed Efficient Temporal Learning Approach}

Let $\mathbf{x} \in \mathbb{R}^{T \times H \times W \times 3}$ be the input video and $\mathbf{y}$ be the corresponding label of the video $\mathbf{x}$. $H, W$ and $T$ are the height, the width and the number of frames of a video, respectively. Let $\mathbf{o} \in \mathbb{N}^{T}$ be the permutation representing the reordering of video frames and $\mathbf{i}$ be the indexing associated with the permutation. Our goal is to learn a deep network to classify the actions and infer the permutation simultaneously as in Eqn.~\eqref{eqn:direcformer-MaxE}.
\begin{equation} \label{eqn:direcformer-MaxE}
\small
\begin{split}
  \arg\max_{\theta} \mathbb{E}_{\mathbf{x, y, o, i}} \left(\log(p(\mathbf{y} | \mathbf{x}; \theta)) + \log(p(\mathbf{i} | \mathcal{T}(\mathbf{x}, \mathbf{o}); \theta))\right)
\end{split}
\end{equation}
where $\theta$ is the parameters of the deep neural network, %
and $\mathcal{T}$ is the permutation function. Given a video $\mathbf{x}$ and the permutation $\mathbf{o}$, the goal is to learn the class label $\mathbf{y}$ of the ordered video and learn the ordering $\mathbf{i}$ of the video after permutation $\mathcal{T}\mathbf{(x, o)}$.

To effectively predict the class label $\mathbf{y}$ and the indexing of the permutation $\mathbf{i}$, a Transformer with Directed Attention is introduced to learn the directed attention in both spatial and temporal dimensions. The proposed DirecFormer is therefore formulated as in Eqn.~\eqref{eqn:direcformer-direcformer}.
\begin{equation} \label{eqn:direcformer-direcformer}
\small
\begin{split}
    \mathbf{\hat{y}} &= \phi_{cls} \odot \mathcal{G} (\mathbf{x}), \quad
    \mathbf{\hat{i}} = \phi_{ord} \odot \mathcal{G} (\mathcal{T}(\mathbf{x}, \mathbf{o}))
\end{split}
\end{equation}
where $\mathcal{G}$ is the proposed DirecFormer; $\phi_{cls}$ and $\phi_{ord}$ are the projections that map the token outputted from DirecFormer to the predicted class label $\mathbf{\hat{y}}$ and the predicted ordering index $\mathbf{\hat{i}}$, respectively; and $\odot$ is the functional composition. Figure~\ref{fig:direcformer-framework} illustrates our proposed framework.
The proposed DirecFormer method will be described in detail in the following section. 

\subsubsection{Patch Representation}

Given a video frame, it is represented by $N$ non-overlapped patches of $P \times P$ ($N = \frac{HW}{P^2}$) as in~\cite{vit}. 
Let us denote $\mathbf{x}_{s, t} \in \mathbb{R}^{3P^2}$ as a vector representing the patch $s$ of the video frame $t$, where $s$ ($1 \leq s \leq N$) denotes the spatial position and $t$ represents the temporal dimension ($1 \leq t \leq T$).
To embed the temporal information into the representation, the raw patch representation is projected to the latent space with additive temporal representation as $\mathbf{z}^{(0)}_{s,t} = \alpha(\mathbf{x}_{s,t}) + \mathbf{e}_{s, t}$, 
where $\alpha$ is the embedding network and $\mathbf{e}_{s, t}$ is the spatial-temporal embedding added into the patch representation. The output sequences $\{\mathbf{z}^{(0)}_{s,t}\}_{s=1,t=1}^{N, T}$ represent the input tokens fed to our DirecFormer 
network. We also add one more learnable token $\mathbf{z}_{0,0}$ in the first position, as in BERT~\cite{bert} to represent the classification token.

\subsubsection{Directed Attention Approach}

The proposed DirecFormer consists of $L$ encoding blocks. In particular, the current block $l$ takes the output tokens of the previous block $l-1$ as the input and decomposes the token into the key $\mathbf{k}^{(l)}_{s,t}$, value $\mathbf{v}^{(l)}_{s,t}$, and query $\mathbf{q}^{(l)}_{s,t}$ vectors as in Eqn.~\eqref{eqn:direcformer-key_value_query}.
\begin{equation} \label{eqn:direcformer-key_value_query}
\small
\begin{split}
    \mathbf{k}^{(l)}_{s,t} &= \beta^{(l)}_k\left(\tau^{(l)}_k\left(\mathbf{z}^{(l-1)}_{s,t}\right)\right), \;
    \mathbf{v}^{(l)}_{s,t} = \beta^{(l)}_v\left(\tau^{(l)}_v\left(\mathbf{z}^{(l-1)}_{s,t}\right)\right), \;
    \mathbf{q}^{(l)}_{s,t} = \beta^{(l)}_q\left(\tau^{(l)}_q\left(\mathbf{z}^{(l-1)}_{s,t}\right)\right) \\
\end{split}
\end{equation}
where $\beta^{(l)}_{k}$, $\beta^{(l)}_{v}$ and $\beta^{(l)}_{q}$ represent the key, value, and query embedding, respectively; $\tau^{(l)}_k$, $\tau^{(l)}_v$ and $\tau^{(l)}_q$ are the layer normalization~\cite{layer_norm}. 

In the traditional self-attention approach, the attention matrix is computed by the scaled dot multiplication between key and query vectors. 
Although scaled dot attention has shown its potential performance in video classification, this attention is non-directed because it is unable to illustrate the direction of attention. 
In particular, the scaled dot attention simply indicates the correlations among tokens and ignores the temporal or spatial ordering among tokens. 
It is noticed that the ordering of frames in a video sequence does matter. The recognition of actions in a video is highly dependent on the ordering of video frames. For example,
the same group of video frames,  if ordered differently in time, may result in different actions, e.g., walking might become running. 
However, traditional Softmax attention can not fully exploit the ordering of video frames because it does not contain the directional information of the correlation. 
Therefore, we propose a new Directed Attention using the cosine similarity. Formally, the attention weights $\mathbf{a}^{(l)}_{(s,t)}$ for a query $\mathbf{q}^{(l)}_{s,t}$ can be formulated as in Eqn.~\eqref{eqn:direcformer-DAeqn}.
\begin{equation} \label{eqn:direcformer-DAeqn}
\small
\begin{split}
    \mathbf{a}^{(l)}_{(s,t)} = \left[\operatorname{cos}\left(\frac{\mathbf{q}^{(l)}_{s,t}}{\sqrt{D}}, \mathbf{k}^{(l)}_{0,0}\right) \; \left\{\operatorname{cos}\left(\frac{\mathbf{q}^{(l)}_{s,t}}{\sqrt{D}}, \mathbf{k}^{(l)}_{s',t'}\right)\right\}_{s'=1,t'=1}^{N,T}\right]
\end{split}
\end{equation}
where $D$ is the dimensional length of the query vector $\mathbf{q}^{(l)}_{s,t}$, $\mathbf{a}^{(l)}_{p,t} \in \mathbb{R}^{NT+1}$ denotes the directed attention weights. 
This attention is computed over the spatial and temporal dimensions. As a result, this operator suffers a heavy computational cost.
We, therefore, divide and conquer the Directed Attention in the spatial dimension and temporal dimension sequentially as in~\cite{timesformer}. 
Formally, we first implement the attention mechanism over the time dimension ($\mathbf{a}^{(l)-time}_{(s,t)}$) as in Eqn.~\eqref{eqn:direcformer-AttTime}.
\begin{equation} \label{eqn:direcformer-AttTime}
\small
\begin{split}
    \mathbf{a}^{(l)-time}_{(s,t)} = \left[\operatorname{cos}\left(\frac{\mathbf{q}^{(l)}_{s,t}}{\sqrt{D}}, \mathbf{k}^{(l)}_{0,0}\right) \; \left\{\operatorname{cos}\left(\frac{\mathbf{q}^{(l)}_{s,t}}{\sqrt{D}}, \mathbf{k}^{(l)}_{s,t'}\right)\right\}_{t'=1}^{T}\right]
\end{split}
\end{equation}
Then, the directed temporal attention information is accumulated to the current token representations as in Eqn.~\eqref{eqn:direcformer-time_att_vec}.
\begin{equation} \label{eqn:direcformer-time_att_vec}
\small
\begin{split}
    \mathbf{s}^{(l)-time}_{s,t} &= \mathbf{a}^{(l)-time}_{(s, t), (0, 0)}\mathbf{v}^{(l)}_{0,0} + \sum_{t'=1}^T \mathbf{a}^{(l)-time}_{(s, t), (s, t')}\mathbf{v}^{(l)}_{s,t'} \\
    \mathbf{z'}^{(l)-time}_{s,t} &= \mathbf{z}^{(l-1)}_{s,t} +  \gamma^{(l)-time}\left(\mathbf{s}^{(l)-time}_{s,t}\right) 
\end{split}
\end{equation}
where $\gamma^{(l)-time}$ denotes the temporal projection. Secondly, the temporally attentive vector $\mathbf{z'}^{(l)-time}_{s,t}$ is projected to the new key, value, and query to drive Spatial Directed Attention as in Eqn.~\eqref{eqn:direcformer-SDA2}.
\begin{equation} \label{eqn:direcformer-SDA2}
\scriptsize
\begin{split}
    \mathbf{k'}^{(l)}_{s,t} &= \beta'^{(l)}_k\left(\tau'^{(l)}_k\left(\mathbf{z'}^{(l)-time}_{s,t}\right)\right), \;
    \mathbf{v'}^{(l)}_{s,t} = \beta'^{(l)}_v\left(\tau'^{(l)}_v\left(\mathbf{z'}^{(l)-time}_{s,t}\right)\right), \;
    \mathbf{q'}^{(l)}_{s,t} = \beta'^{(l)}_q\left(\tau'^{(l)}_q\left(\mathbf{z'}^{(l)-time}_{s,t}\right)\right)
\end{split}
\end{equation}
Next, the Directed Attention over the spatial dimension ($\mathbf{a}^{(l)-space}_{(s,t)}$) can be computed as in Eqn.~\eqref{eqn:direcformer-DAtt3}.
\begin{equation} \label{eqn:direcformer-DAtt3}
\small
\begin{split}
    \mathbf{a}^{(l)-space}_{(s,t)} = \left[\operatorname{cos}\left(\frac{\mathbf{q'}^{(l)}_{s,t}}{\sqrt{D}}, \mathbf{k'}^{(l)}_{0,0}\right) \; \left\{\operatorname{cos}\left(\frac{\mathbf{q'}^{(l)}_{s,t}}{\sqrt{D}}, \mathbf{k'}^{(l)}_{s,t'}\right)\right\}_{t'=1}^{T}\right]
\end{split}
\end{equation}
The Spatial Directed Attention is then embedded to the temporal attentive features $\mathbf{z'}^{(l)-time}_{s,t}$ to obtain a new spatial attentive feature $\mathbf{z'}^{(l)-space}_{s,t}$ as in Eqn.~\eqref{eqn:direcformer-SDAtt3}.
\begin{equation} \label{eqn:direcformer-SDAtt3}
\small
\begin{split}
    \mathbf{s}^{(l)-space}_{s,t} &= \mathbf{a}^{(l)-space}_{(s, t), (0, 0)}\mathbf{v}^{(l)}_{0,0} + \sum_{s'=1}^N \mathbf{a}^{(l)-space}_{(s, t), (s', t)}\mathbf{v}^{(l)}_{s',t} \\
    \mathbf{z'}^{(l)-space}_{s,t} &= \mathbf{z'}^{(l)-times}_{s,t} +  \gamma^{(l)-space}\left(\mathbf{s}^{(l)-space}_{s,t}\right)  \\
\end{split}
\end{equation}
Finally, the Spatial-Temporal Attentive features $\mathbf{z'}^{(l)-space}_{s,t}$ are projected to the output token, getting ready for the next transformer block.
Formally, the output of the current transformer block ($\mathbf{z}^{(l)}_{s,t}$) can be formed as in Eqn.~\eqref{eqn:direcformer-transblock}.
\begin{equation} \label{eqn:direcformer-transblock}
\small
\begin{split}
    \mathbf{z}^{(l)}_{s,t} &= \varphi^{(l)}\left(\tau^{(l)}\left(\mathbf{z'}^{(l)-space}_{s,t}\right)\right) + \mathbf{z'}^{(l)-space}_{s,t}
\end{split}
\end{equation}
where $\varphi^{(l)}$ is a projection mapping implemented using a multi-layer perception network, and $\tau^{(l)}$ denotes the layer normalization~\cite{layer_norm}.

\subsubsection{Classification Embedding}

The final representation is obtained in the final block of DirecFormer. Then, the class index and the order index of the video are predicted using linear projections as in Eqn.~\eqref{eqn:direcformer-final-prediction}.
\begin{equation}\label{eqn:direcformer-final-prediction}
\small
\begin{split}
    \mathbf{\hat{y}} = \phi_{cls}\left(\tau_{cls}\left(\mathbf{z}^{(L)}_{0,0}\right)\right), \quad
    \mathbf{\hat{i}} = \phi_{odr}\left(\tau_{odr}\left(\mathbf{z}^{(L)}_{0,0}\right)\right) 
\end{split}
\end{equation}
where $\phi_{cls}$ and $\phi_{ord}$ are the classification projection and order projection, respectively; $\tau_{cls}$ and $\tau_{ord}$ are the layer normalization~\cite{layer_norm}.

\subsubsection{Self-supervised Guided Loss For Directed Temporal Attention Loss}

In this stage, we are given the permutation of the current input video. To further reduce the burden of the network when learning the temporal attention, we propose a new self-supervised guided loss to enforce the temporal attention learning from the prior order knowledge. Formally, the self-supervised loss can be formulated as in Eqn.~\eqref{eq:newloss}.
\begin{equation} \label{eq:newloss}
\small
    \mathcal{L}_{self} = \frac{1}{LNT^2} \sum_{l=1}^{L}\sum_{s=1,t=1}^{N,T}\sum_{t'=1}^{T}\left(1 - \mathbf{a}^{(l)-time}_{(s,t),(s,t')}\right) \varsigma(\mathbf{o}_t, \mathbf{o}_{t'})
\end{equation}
where $\varsigma(\mathbf{o}_t, \mathbf{o}_{t'}) = 1$ if the index $\mathbf{o}_t < \mathbf{o}_{t'}$, otherwise $\varsigma(\mathbf{o}_t, \mathbf{o}_{t'}) = -1$. The guided loss $\mathcal{L}_{self}$ helps to indicate the attention learning the correct direction during the training process.
Finally, the total loss function of DirecFormer is defined as in Eqn.~\eqref{eq:finalloss}.
\begin{equation} \label{eq:finalloss}
    \mathcal{L} = \lambda_{cls}\mathcal{L}_{cls} + \lambda_{ord}\mathcal{L}_{ord} + \lambda_{self}\mathcal{L}_{self}
\end{equation}
where $\mathcal{L}_{cls}$ and $\mathcal{L}_{ord}$ are the cross-entropy losses of the classification projection ($\phi_{cls}$) and order projection ($\phi_{ord}$), respectively;
$\{\lambda_{cls}, \lambda_{ord}, \lambda_{self}\}$ are the parameters controlling their relative importance.

\subsection{Experimental Results}

In this section, we present the evaluation results with DirecFormer on three popular action recognition benchmarking datasets, i.e. Jester~\cite{jester}, Something-Something V2~\cite{ssv2}, and Kinetics 400~\cite{kinetics}. Firstly, we describe our implementation details and datasets used in our experiments. Secondly, we analyze our results with different settings shown in the ablation study on the Jester dataset. Lastly, we present our results on Something-Something V2 and Kinetics compared to prior state-of-the-art methods.

\subsubsection{Implementation and Datasets}

\noindent
\textbf{Implementation.} 
The architecture of DirecFormer consists of $L=12$ blocks. The input video consists of $T=8$ frames sampled at a rate of $1/32$ and the resolution of each frame is $224 \times 224$ ($H = W = 224$).
The patch size is set to $18 \times 18$; therefore, there are $N = \frac{224^2}{16^2} = 196$ patches in total for each frame.
The embedding network $\alpha$ is implemented by a linear layer in which the output dimension is set to $768$. All values ($\beta^{(l)}_v, \beta'^{(l)}_v$), key ($\beta^{(l)}_k, \beta'^{(l)}_k$), query ($\beta^{(l)}_q, \beta'^{(l)}_q$) embedding networks, and projections ($\gamma^{(l)-time}, \gamma^{(l)-space}$) are also implemented by the linear layers.
Similar to~\cite{vit,timesformer}, we adopt the multi-head attention in our implementation, where the number of heads is set to $12$. 
The network $\varphi^{(l)}$ is implemented as the residual-style multi-layer perceptron consisting of two fully
connected layers followed by a normalization layer.
Finally, the classification projection ($\phi_{cls}$) and the order projection ($\phi_{ord}$) are implemented as the linear layer.  We set the control parameters of loss to $1.0$, i.e. $\lambda_{cls} = \lambda_{ord} = \lambda_{self} = 1.0$. 

There will be a total of $T!$ permutations of the video frames. Therefore, learning with all permutations is ineffective. Moreover, the permutation set plays an important role. If these two permutations are very far from each other, the network may easily predict the order since the two permutations have significant differences.
However, if all the permutations are close to each other, learning the temporal attention is more challenging since the two permutations have minor differences in order. Therefore, we select 1,000 random permutations from $T! = 8!$ permutations so that the Hamming distance between permutations is as minimum as possible. Similar to~\cite{domain_jigsaw_puzzle}, we use a greedy algorithm to generate the set of permutations.

In the evaluation, following the protocol of other studies~\cite{slowfast, timesformer,x3d}, the single clip is sampled in the middle of the video. We use three spatial crops (top-left, center, and bottom-right) from the temporal clip and obtain the final result by averaging the prediction scores for these three crops.

\noindent
\textbf{Jester.}~\cite{jester} This dataset is a large-scale gesture recognition real-world video dataset that includes $148,092$ videos of $27$ actions. Each video is recorded for approximately 3 seconds. 

\noindent
\textbf{Something-in-Something V2.}~\cite{ssv2} The dataset is a large-scale dataset to show humans performing predefined basic actions with everyday objects, which includes $174$ classes. It contains $220,847$ videos, with $168,913$ videos in the training set, $24,777$ videos in the validation set, and $27,157$ videos in the testing set. Similar to other work~\cite{msnet,trg,timesformer,slowfastmg}, we report the accuracy on the validation set. 
The licenses of the Something-Something V2 and Jester datasets are registered by the TwentyBN team and are publicly available for academic research purposes.

\input{Tables/chap-5/direcformer/jester-ablation}
\noindent
\textbf{Kinetics-400.}~\cite{kinetics} The dataset contains 400 human action classes, with at least 400 videos for each action. In particular, Kinetics-400 contains $234,619$ training videos and $19,761$ validation videos. 
The videos were downloaded from youtube and each video lasts for 10 seconds.
There are different types of human actions: \textit{Person Actions} (e.g. singing, smoking, sneezing, etc.); \textit{Person-Person Actions} (e.g. wrestling, hugging, shaking hands, etc.); and \textit{Person-Object Actions} (e.g. opening a bottle, walking the dog, using a computer, etc.).  
In our experiment, following the protocol of other studies~\cite{slowfastmg,x3d,timesformer,x3d,slowfast}, we report the accuracy of the validation set.
The license of Kinetics is registered by Google Inc. under a Creative Commons Attribution 4.0 International License.

\subsubsection{Ablation Study}

\noindent
\textbf{Effectiveness Of Directed Attention.}
To show the effectiveness of our proposed Directed Attention, we consider three different types of the temporal-spatial attention: (i) Softmax Temporal Attention followed by Cosine Spatial Attention (DirecFormer $S-C$), (ii) Cosine Temporal Attention followed by Softmax Spatial Attention (DirecFormer $C-S$), and (iii) Cosine Temporal Attention followed by Cosine Spatial Attention (DirecFormer $C-C$). The method is also compared with TimeSformer where the softmax attention is applied for both time and space. 
Table~\ref{tab:direcformer-jester_exp} illustrates the results of the DirecFormer with different settings compared to TimeSFormer and other approaches. In all configurations, our proposed DirecFormer outperforms the prior methods.

\input{Tables/chap-5/direcformer/order-correction}
Considering the effectiveness of directed attention in time and space,
the direction of the attention over the spatial dimension is important in some cases. For example, if $A$ performs an action to $B$, then $B$ receives an action from $A$. 
Considering the mentioned example, spatial attention should involve directions so that the model can learn the actor(s) performing actions in a video.
However, the order of the temporal dimension plays a more important role in a video compared to the spatial dimension, since the order of the frames represents how the action is happening. 
As in Table~\ref{tab:direcformer-jester_exp}, the results of DirecFormer $C-S$ are better than DirecFormer $S-C$ confirming our hypothesis about the importance of time and space.
When the Directed Attention is deployed in both temporal and spatial dimensions, the results of DirecFormer $C-C$ were significantly improved and achieved the SOTA performance on the Jester dataset.

\input{Tables/chap-5/direcformer/ssv2-results}
\noindent
\textbf{Effectiveness Of Losses.}
With the order prediction loss $\mathcal{L}_{ord}$, the performance of the DirecFormer in all settings has been improved since the prediction loss influences the way that the network learns the Directed Temporal Attention. Moreover, the performance of DirecFormer is improved by employing the self-supervised guided loss $\mathcal{L}_{self}$. This self-supervised loss further enhances the directed temporal attention learning during the training. Consequently, the performance of DirecFormer is consistently improved by using our proposed losses, as in Table~\ref{tab:direcformer-jester_exp}.

\noindent
\textbf{Order Correction.} To illustrate the ability of order learning of DirecFormer, we conduct an experiment in which, given a random temporal order video,  we show our approaches can retrieve back the correct order of the video from the directed temporal attention. 
In this experiment, we use the temporal attention of the last block and average this temporal attention over the spatial dimension. Then, we perform a search algorithm to find the Hamiltonian path on the temporal attention to find the correct order. 
In particular, we consider temporal attention to be the graph's adjacency matrix, where each frame is a node. 

\input{Tables/chap-5/direcformer/k400-results}
The Hamilton path is the path that goes through each node exactly once (no revisit). Since our attention represents both direction and correlation among the frames, the higher the correlation, the higher the possibility of correct order between frames. Therefore, the Hamilton path with maximum total weight is going to represent the order in the video. 
Let $\mathbf{\hat{o}}$ be the order obtained by the Hamilton algorithm, the accuracy of the order retrieval can be defined as $\operatorname{OrderAcc} = \frac{\operatorname{LCS}(\mathbf{\hat{o}}, \mathbf{o})}{T} \times 100$, where $\operatorname{LCS}(\mathbf{\hat{o}}, \mathbf{o})$ is the longest common subsequence between $\mathbf{\hat{o}}$ and $\mathbf{o}$.
In this evaluation, for each video, we randomly select a permutation of $\{1,...,N\}$ as the order of the input video. To be fair between benchmarks, we set the same random seed value at the beginning of the testing script so that every time we conduct the evaluation, we obtain the same permutation for each video.

As shown in Table~\ref{tab:direcformer-order_exp}, we use the Softmax attention of TimeSFomer to retrieve the order of the video. The order accuracy of the TimeSFormer is only $52.84$. In other words, the Softmax attention of TimeSFormer can only predict the correct order of approximately 4 frames over 8 frames. With the support of order prediction loss, the order accuracy of TimeSFormer is improved to $72.57\%$. 
However, %
without the order prediction loss, our DirecFormer $C-S$ and DirecFormer $C-C$ have already correctly predicted the order of approximately 6 frames over 8 frames ($75.04\%$ and $76.16\%$).
When we further employ the order prediction and self-supervised guided losses, the performance of DirecFormer is significantly improved. Particularly, with the order prediction loss only, DirecFormer in all settings gains more than $87.0\%$ (which is approximately 7 frames over 8 frames). When both losses ($\mathcal{L}_{ord}$ and $\mathcal{L}_{self}$) are employed, the order accuracy of both DirecFormer $C-S$ and DirecFormer $C-C$ is improved to $90.02\%$ and $90.19\%$, respectively.
It should be noted that the performance of DirecFormer $C-C$ is only minorly greater than DirecFormer $C-S$ as the directed attention over the space does not largely affect the temporal order predictions.

\subsubsection{Comparison with State-of-the-Art Results}

\input{Tables/chap-5/direcformer/network-size}
\noindent
\textbf{Something-Something V2.}
Table~\ref{tab:direcformer-ssv2_exp} illustrates the performance of our proposed approaches evaluated on Something-Something V2 compared to prior SOTA approaches. In this experiment, similar to other approaches~\cite{timesformer}, we use the DirecFormer pretrained on ImageNet-1K~\cite{imagenet15russakovsky}.
As in Table~\ref{tab:direcformer-ssv2_exp}, our results in all settings outperform other candidates. 
With the simple design of the Transformer network with the directed attention mechanisms over time and space, our approaches achieve SOTA performance compared to traditional 3D CNN approaches ~\cite{slowfast, msnet} and other Transformer approaches~\cite{timesformer, vidtr} by a competitive margin.

\noindent
\textbf{Kinetics 400.} 
We conduct the experiments on Kinetics 400 and compare our results with prior SOTA methods.
The pretrained model on ImageNet-21K~\cite{imagenet15russakovsky} for our DirecFormer is used, similar to~\cite{timesformer}.
It is noted that the prior methods~\cite{slowfast,x3d} use 10 temporal clips with 3 spatial crops of a video in the evaluation phase.
However, TimeSFormer and our DirecFormer use only 3 spatial crops of a video with a single clip to achieve solid results.
In particular, our method achieves the SOTA performance compared to prior methods as shown in Table~\ref{tab:direcformer-k400_exp}. 
The Top 1 accuracy of the best model is approximately $2\%$ higher than TimeSFormer-L~\cite{timesformer} sitting at $82.75\%$.
The effectiveness of the proposed directed attention has been also proved in these experiments, as the performance of DirecFormer is consistently improved when we deploy the directed attention over time and space.

\noindent
\textbf{Network Size Comparison.} As shown in Table~\ref{tab:direcformer-networksize},
 although the number of parameters and the GFLOPS of single view used in our DirecFormer are higher than the traditional 3D-CNN approaches~\cite{i3d, slowfast, x3d}, we only use 3 views compared to 30 views of prior approaches and maintain competitive performance. 
In comparison with TimeSFormer, we gain the same performance in terms of network size and inference flops; however, we achieve better accuracy on three large-scale %
benchmarks as shown in Tables \ref{tab:direcformer-jester_exp}, \ref{tab:direcformer-ssv2_exp}, and \ref{tab:direcformer-k400_exp}.

%% file: Tables/chap-5/direcformer/jester-ablation.tex
\begin{wraptable}{r}{0.5\textwidth}
    \centering
    \caption{\textbf{Ablation Study On Jester.} $X-Y$ denotes for the attention types of temporal and spatial dimension, respectively.  $X$ (and $Y$) could be either $S$: Softmax or $C$: Cosine.} 
    \resizebox{0.5\textwidth}{!}{
    \begin{tabular}{|l|c|c|c|c c|}
    \hline
        \textbf{Models}      & \begin{tabular}{@{}c@{}} \textbf{Attention}\\\textbf{Time-Space} \end{tabular}  & $\mathcal{L}_{ord}$ & $\mathcal{L}_{self}$& \textbf{Top 1} & \textbf{Top 5} \\ \hline
        I3D \cite{i3d}          & $-$   & $-$ & $-$ & 91.46 & 98.67 \\
        3D SqueezeNet \cite{squeezenet} & $-$   & $-$ & $-$ & 90.77 & $-$     \\ 
        ResNet 50 \cite{resnet}     & $-$   & $-$ & $-$ & 93.70 & $-$     \\ 
        ResNet 101 \cite{resnet}    & $-$   & $-$ & $-$ & 94.10 & $-$     \\ 
        ResNeXt \cite{resnext}       & $-$   & $-$ & $-$ & 94.89 & $-$     \\ 
        PAN \cite{pan}           & $-$   & $-$ & $-$ & 96.70 & $-$     \\ 
        STM \cite{stm}           & $-$   & $-$ & $-$ & 96.70 & $-$     \\ 
        ViViT-L/16x2 320 \cite{vivit} & $-$ & $-$ & $-$ & 81.70 & 93.80 \\
        TimeSFormer \cite{timesformer}   & $S-S$ & $-$ & $-$ & 94.14 & 99.19 \\ \hline
        \hline
        DirecFormer   & $S-C$ &  &  & 94.52 & 99.26 \\
        DirecFormer   & $S-C$ & \cmark &  & 94.65 & 99.25 \\
        \hline
        DirecFormer   & $C-S$ &  &  & 95.52 & 99.20 \\
        DirecFormer   & $C-S$ & \cmark &  & 96.28 & 99.45 \\
        DirecFormer   & $C-S$ & \cmark & \cmark & 97.55 & 97.54 \\ 
        \hline
        DirecFormer   & $C-C$ &  &  & 96.15 & 99.38 \\ 
        DirecFormer   & $C-C$ & \cmark &  & 97.48 & 99.48 \\ 
        DirecFormer   & $C-C$ & \cmark & \cmark & 98.15 & 99.57  \\
        \hline
    \end{tabular}
    }
    \label{tab:direcformer-jester_exp}
\end{wraptable}

%% file: Tables/chap-5/direcformer/order-correction.tex
\begin{wraptable}{r}{0.5\textwidth}
    \small
    \centering
    \caption{\textbf{Order Correction By Hamilton Algorithm Performance On Jester.} $X-Y$ denotes for the attention types of temporal and spatial dimension, respectively.  $X$ (and $Y$) could be either $S$: Softmax or $C$: Cosine.} %
    \resizebox{0.5\textwidth}{!}{
    \begin{tabular}{|l|c|c|c|c|}
    \hline
        \textbf{Models}      & \begin{tabular}{@{}c@{}} \textbf{Attention}\\\textbf{Time-Space} \end{tabular}  & $\mathcal{L}_{ord}$ & $\mathcal{L}_{self}$& \textbf{OrderAcc} \\ \hline
        TimeSFormer \cite{timesformer}   & $S-S$ & $-$    & $-$ &  52.84 \\
        TimeSFormer \cite{timesformer}   & $S-S$ & \cmark & $-$ &  72.57 \\
        \hline
        \hline
        DirecFormer   & $C-S$ &  &  & 75.04 \\
        DirecFormer   & $C-S$ & \cmark &  & 87.16 \\
        DirecFormer   & $C-S$ & \cmark & \cmark & 90.02 \\
        \hline
        DirecFormer   & $C-C$ &  &  & 76.16 \\ 
        DirecFormer   & $C-C$ & \cmark &  & 88.96 \\ 
        \textbf{DirecFormer}   & $C-C$ & \cmark & \cmark & \textbf{90.19} \\ 
        \hline
    \end{tabular}
    }
    \label{tab:direcformer-order_exp}
\end{wraptable}

%% file: Tables/chap-5/direcformer/ssv2-results.tex
\begin{wraptable}{r}{0.5\textwidth}
    \centering
    \caption{\textbf{Comparison with the SOTA methods on Something-Something V2.} $X-Y$ denotes for the attention types of temporal and spatial dimension, respectively.  $X$ (and $Y$) could be either $S$: Softmax or $C$: Cosine.}
    \resizebox{0.5\textwidth}{!}{
    \begin{tabular}{|l | c | c c|}
    \hline
        \textbf{Models}         & \begin{tabular}{@{}c@{}} \textbf{Attention}\\\textbf{Time-Space} \end{tabular} &  \textbf{Top 1} & \textbf{Top 5} \\ \hline
        MSNet \cite{msnet}                   & $-$ & 63.00          & 88.40          \\ 
        SlowFast \cite{slowfast}                & $-$ & 63.00          & 88.50          \\ 
        SlowFast Multigrid \cite{slowfastmg}      & $-$ & 63.50          & 88.70 \\ 
        TRG \cite{trg}                     & $-$ & 62.20	       & \textbf{90.30}	 \\
        VidTr-L \cite{vidtr}                 & $-$ & 60.20           & $-$ \\
        TimeSFormer \cite{timesformer}             & $S-S$ & 59.10          & 85.60          \\ 
        TimeSFormer $-$ HR \cite{timesformer}     & $S-S$ & 61.80          & 86.90          \\ 
        TimeSFormer $-$ L \cite{timesformer}       & $S-S$ & 62.00          & 87.50          \\ 
        \hline \hline
        DirecFormer            & $S-C$ & 61.70          & 85.20     \\ 
        DirecFormer            & $C-S$ & 63.85          & 85.92          \\ 
        \textbf{DirecFormer}   & $C-C$ & \textbf{64.94} & 87.90          \\ 
        \hline
    \end{tabular}
    }
    \label{tab:direcformer-ssv2_exp}
\end{wraptable}

%% file: Tables/chap-5/direcformer/k400-results.tex
\begin{wraptable}{r}{0.5\textwidth}
\centering
    \caption{\textbf{Comparison with the SOTA methods on Kinetics 400.} $X-Y$ denotes for the attention types of temporal and spatial dimension, respectively.  $X$ (and $Y$) could be either $S$: Softmax or $C$: Cosine.} %
    \resizebox{0.5\textwidth}{!}{
    \begin{tabular}{|l | c | c c|}
        \hline
         \textbf{Models}         & \begin{tabular}{@{}c@{}} \textbf{Attention}\\\textbf{Time-Space} \end{tabular} &  \textbf{Top 1} & \textbf{Top 5} \\
        \hline
        I3D NLN \cite{i3d}            & $-$ & 74.00 & 91.10          \\
        ip-CSN-152 \cite{ipcsn152}         & $-$ & 77.80 & 92.80          \\
        LGD-3D-101 \cite{lgd3d}         & $-$ & 79.40 & 94.40          \\
        SlowFast \cite{slowfast}           & $-$ & 77.00 & 92.60   \\
        SlowFast Multigrid \cite{slowfastmg} & $-$ & 76.60 & 92.70   \\
        X3D-M \cite{x3d}              & $-$ & 75.10 & 91.70   \\
        X3D-L \cite{x3d}              & $-$ & 76.90 & 92.50   \\
        X3D-XXL \cite{x3d}            & $-$ & 80.40 & 94.60   \\
        MViT \cite{mvit}               & $-$ & 78.40 & 93.50    \\
        TimeSFormer \cite{timesformer}        & $S-S$ & 77.90 & 93.20 \\
        TimeSFormer $-$ HR \cite{timesformer} & $S-S$ & 79.70 & 94.40 \\
        TImeSFormer $-$ L \cite{timesformer}  & $S-S$ & 80.70 & 94.70 \\
        \hline
        \hline
        DirecFormer          & $S-C$ & 80.16 & 94.55 \\ 
        DirecFormer          & $C-S$ & 81.69 & 94.62 \\ 
        \textbf{DirecFormer} & $C-C$ & \textbf{82.75} & \textbf{94.86} \\ 
        \hline
        \end{tabular}
        }
    \label{tab:direcformer-k400_exp}
\end{wraptable}

%% file: Tables/chap-5/direcformer/network-size.tex
\begin{wraptable}{r}{0.5\textwidth}
\centering
\caption{\textbf{Network Size Comparison.} We report the computational cost of the inference phase with a single ``view'' (temporal clip with spatial crop) $\times$ the numbers of such views used (GFLOPs $\times$ views). ``N\/A'' indicates the number is not available for us.}
\label{tab:direcformer-networksize}
    \resizebox{0.5\textwidth}{!}{
    \begin{tabular}{|l|c|c|}
    \hline
        \textbf{Model}    & \textbf{GFLOPS x Views} & \textbf{Params} \\ 
        \hline
        I3D \cite{i3d}               & $108 \times $N\/A             & 12.0M             \\ 
        SlowFast 8x8 R50 \cite{slowfast}  & $36.1 \times 30$              & 34.4M           \\ 
        SlowFast 8x8 R101 \cite{slowfast} & $106 \times 30$               & 53.7M           \\ 
        Nonlocal R50 \cite{non_local}      & $282 \times 30$               & 35.3M           \\ 
        X3D-XL \cite{x3d}            & $35.8 \times 30$              & 11.0M           \\ 
        X3D-XXL \cite{x3d}           & $143.5 \times 30$             & 20.3M           \\ 
        \hline
        ViViT-L/16x2 320 \cite{vivit}           & $3980 \times 3$ &  310.8M \\ 
        TimeSformer \cite{timesformer}       & $196 \times 3$                & 121.4M          \\ 
        DirecFormer       & $196 \times 3$                & 121.4M          \\ 
        \hline
        \end{tabular}
        }
\end{wraptable}

%% file: Chapters/Sections/chap-6/edsam.tex
\section{Domain Generalization in Vision-Language Foundation Models}

\setcounter{propositioncounter}{0}
\setcounter{remarkcounter}{0}

The vision-language foundation models trained based on contrastive learning and exemplified by CLIP~\cite{radford2021learning}, have gained more attention due to their outstanding performance on various tasks.
Although the vision-language foundation models have shown advantages on various downstream visual tasks, limited studies investigate their generalizability. 
Meanwhile, the generalizability of the foundation models still majorly relies on the large-scale pre-training datasets.
While many prior studies~\cite{guo2023domaindrop, hu2023dandelionnet, yao2022pcl, li2023simple, long2023rethinking, zhang2023adanpc, chang2023domain, wang2023sharpness, chen2023metacausal} have been introduced to domain generalization for classification~\cite{lee2023decompose, yao2022pcl, volpi2018generalizing, bui2021exploiting, li2023intra}, detection~\cite{vidit2023clip, lin2021domain}, semantic segmentation~\cite{zhong2022adversarial, lee2022wildnet, ding2023hgformer, huang2023style}, there are limited studies that address the domain generalization problem in the vision-language foundation model.

\begin{wrapfigure}{r}{0.5\textwidth}
    \centering
    \includegraphics[width=0.5\textwidth]{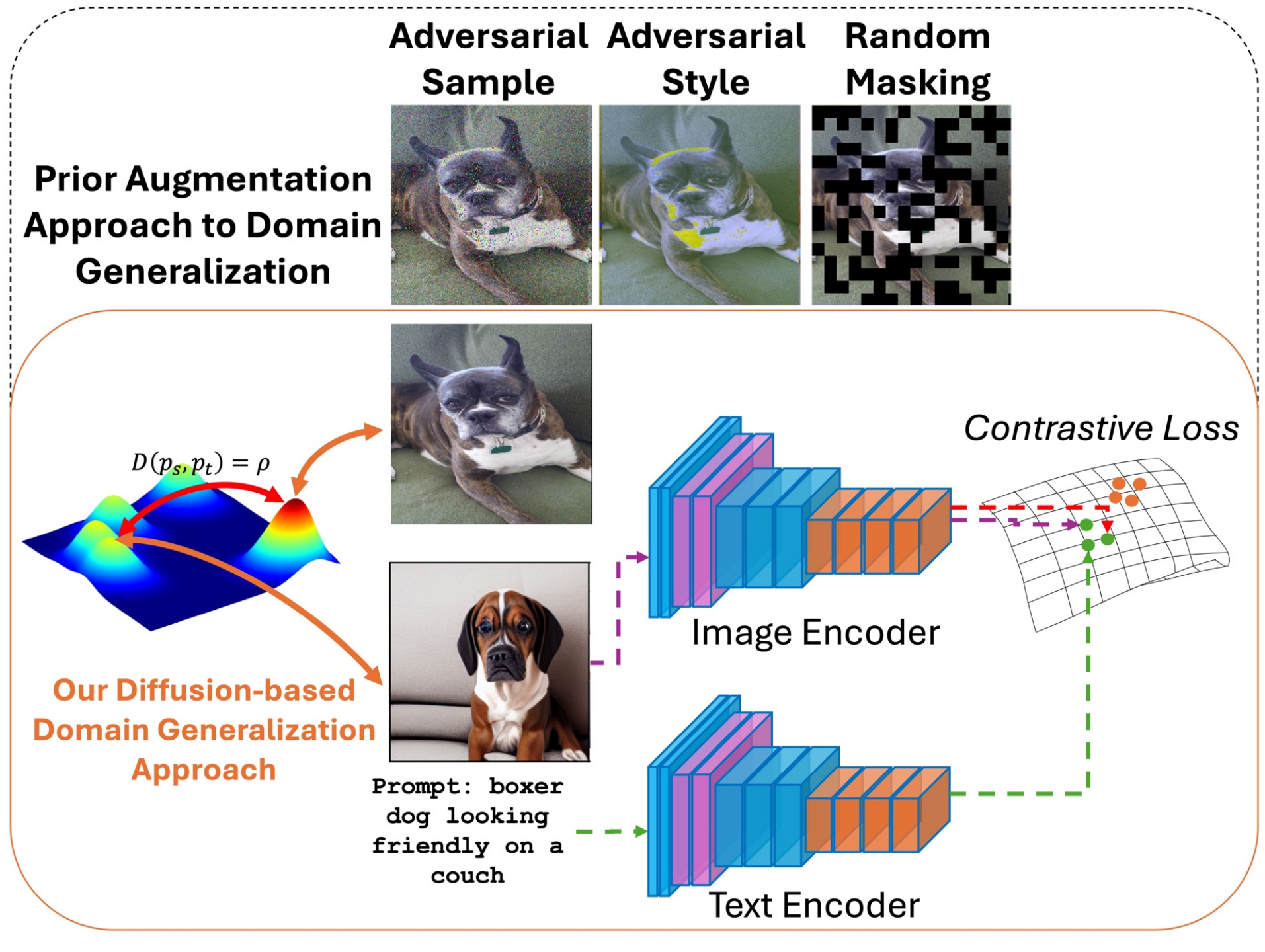}
    \caption{Comparison between Our Proposed Diffusion-based Domain Generalization with Prior Methods~\cite{volpi2018generalizing, zhong2022adversarial, li2023scaling}.}\label{fig:edsam-abstract}
\end{wrapfigure}
Despite being trained on a large-scale dataset, the generalizability of the vision-language foundation model has to be considered because it is a key factor in guaranteeing the performance of models against unknown data distributions.
The domain generalization approaches are urgently needed for foundation model training to ensure optimal performance and generalizability.
The current vision-language foundation models trained using contrastive learning often rely on data augmentations to improve their robustness and prevent overfitting.
However, these methods are not effective enough to improve the generalization of the foundation model. 
In particular, to improve the performance of CLIP models, most of the prior visual foundation models perform the data augmentation on visual inputs~\cite{li2023scaling, mu2022slip, radford2021learning, nguyen2023micron, jia2021scaling} to increase the number of training samples and create challenging samples. 
These augmentation methods aim to increase the diversity of the data, thus enhancing the generalization of the foundation models. 
However, these visual augmentations concentrate on pixel-level modification like masking, adversarial perturbations, adversarial styles, or color jittering, which have a limited impact on enriching the semantic information of visual concepts.
Therefore, the generalizability to unknown data distributions of vision-language models remains limited.

In recent years, in parallel with the development of vision-language models, the diffusion model has shown its outstanding performance in data distribution modeling and generative AI. The diffusion approach, designed based on the nonequilibrium thermodynamics~\cite{ho2020denoising}, is able to model the data distribution via the parameterized Markov chain trained using variational inference. Hence, the diffusion models can synthesize novel, high-quality, and complex data.
Moreover, the diffusion models are also able to efficiently model the conditional data distributions, e.g., text-to-image diffusion~\cite{rombach2022high}.
Inspired by the success of diffusion, this paper fundamentally investigates its role and relation to the generalizability of the vision-language foundation models. 
In particular, we introduce a novel \textbf{\textit{Diffusion-based Domain Generalization}} approach, a simple yet effective approach to improving the generalizability of the vision-language model, i.e., CLIP, by exploiting the power of the diffusion model (Figure~\ref{fig:edsam-abstract}). 
First, we form the domain generalization problem of the vision-language model via the worst-case formula over the training data distribution. By modeling the data conditional distribution via the diffusion model, we further provide a complete theoretical analysis of the relation of the diffusion model to adversarial augmentation.
Second, we introduce a new simple yet efficient \textbf{\textit{Transport Transformation}} to diffusion sampling that can synthesize adversarial samples to improve the generalizability of the vision-language model.
Thanks to our proposed Transport Transformation, our approach efficiently expands the training data distributions, therefore improving the ability to generalize to unseen data distributions of the vision-language model.
Our theoretical analysis has shown the proposed approach is robust and well-generalized. It also has a better domain generalization compared to prior methods~\cite{volpi2018generalizing, zhong2022adversarial}.

\subsection{Theoretical Analysis of Generalizability in Foundation Model}

\subsubsection{Preliminary}
\label{sec:edsam-diffusion_background}

\noindent
\textbf{Diffusion Model.}~\cite{ho2020denoising}
formulates the data distribution $p(\mathbf{x})$ by gradually denoising a normally distributed variable via the reverse process of a fixed Markov Chain of length $T$, i.e., $p(\mathbf{x}_0) = \int p(\mathbf{x}_{0:T})d\mathbf{x}_{1:T}$,  with a Gaussian transition starting at $p(\mathbf{x}_T) = p(\mathbf{z}) = \mathcal{N}(\mathbf{z}; \mathbf{0}, \mathbf{I})$
The diffusion model includes the forward and backward processes. The forward diffusion process, i.e., $q(\mathbf{x}_i|\mathbf{x}_{i-1})$ is defined as in Eqn.~\eqref{eqn:edsam-forward-diff}.
\begin{equation} \label{eqn:edsam-forward-diff}
\small
    q(\mathbf{x}_{1:T}|\mathbf{x}_0) = \prod_{i=1}^{T} q(\mathbf{x}_i |\mathbf{x}_{i-1}) \quad q(\mathbf{x}_{i} | \mathbf{x}_{i-1}) = \mathcal{N}(\mathbf{x}_i, \sqrt{1-\beta_i}\mathbf{x}_{t-1}, \beta_i\mathbf{I})
\end{equation}
where $\beta_i$ is a variance schedule. Then, the backward process, i.e., $p(x_{k-1}|x_{k-1})$, is defined as in Eqn.~\eqref{eqn:edsam-diffusion_backward}.
\begin{equation} \label{eqn:edsam-diffusion_backward}
\small
    p(\mathbf{x}_{0:T}) = p(\mathbf{x}_T) \prod_{i=1}^{T} p(\mathbf{x}_{i-1}|\mathbf{x}_i) \quad p(\mathbf{x}_{i-1}|\mathbf{x}_i) = \mathcal{N}(\mathbf{x}_{i-1}; \boldsymbol{\mu}_{\theta}(\mathbf{x}_i, i), \boldsymbol{\Sigma}_{\theta}(\mathbf{x}_i, i)) 
\end{equation}
The backward process adopts
a denoising model $\epsilon_{\theta}$ to predict the denoised variant from $\mathbf{x}_i$.
Then, 
the model is learned via 
the usual variational bound on negative log-likelihood is as in Eqn.~\eqref{eqn:edsam-logllk-diff}.
\begin{equation} \label{eqn:edsam-logllk-diff}
\small
    \theta^* = \arg\min_{\theta} \mathbb{E}_{\mathbf{x}, \epsilon \in \mathcal{N}(\mathbf{0, I}), i} \left[|| \epsilon - \epsilon_{\theta}(\mathbf{x}_i, i) ||_2^2 \right]
\end{equation}
where $\theta$ is the parameter of $\epsilon$, $\mathbf{x}_i = \sqrt{\overline{\alpha}_i}\mathbf{x} + \sqrt{1 - \overline{\alpha}_i}\epsilon$, $\alpha_i = 1 -\beta_i$, $\overline{\alpha}_i = \prod^i_{s=1}\alpha_s$, and $i$ is uniformly sampled from $1$ to $T$, i.e., $i \in \mathcal{U}(1, T)$.
The diffusion model is capable of modeling the conditional distribution, i.e., $p(\mathbf{x}|\mathbf{p})$ where $\mathbf{p}$ is the condition (e.g., a text prompt).
This ability can be done by implementing a conditional denoising model $\epsilon_{\theta}(\mathbf{x}_i, i, \mathbf{p})$.

\noindent
\textbf{Contrastive Language-Image Pretraining (CLIP).}~\cite{radford2021learning} has shown its outstanding performance in the training vision-language foundation model using language supervision. Formally, let $\mathbf{x}, \mathbf{p} \sim p(\mathbf{x}, \mathbf{p})$ be the source training data of the CLIP model where $\mathbf{x}$ is the image, and $\mathbf{p}$ is the corresponding prompt,  $F^{\mathbf{x}}$ and $F^{\mathbf{p}}$ be the vision and language encoder, and $\mathbf{f}^{\mathbf{x}}$ and $\mathbf{f}^{\mathbf{p}}$ be the features extracted by the vision and language encoder, respectively, i.e., $\mathbf{f}^{\mathbf{x}} = F^{\mathbf{x}}(\mathbf{x})$ and $\mathbf{f}^{\mathbf{p}} = F^{\mathbf{p}}(\mathbf{p})$. The CLIP model is learned via contrastive loss, where the pairs of images and corresponding texts are the positive pairs. The CLIP model can formulated as in Eqn.~\eqref{eqn:edsam-clip_model_learning}.
\begin{equation}\label{eqn:edsam-clip_model_learning}
\small
\theta^*_{F^{\mathbf{x}}}, \theta^*_{F^{\mathbf{p}}} = \arg\min_{\theta_{F^{\mathbf{x}}}, \theta_{F^{\mathbf{p}}}} \mathbb{E}_{\mathbf{x}, \mathbf{p}_t \sim p(\mathbf{x}, \mathbf{p})} -\log \frac{\exp({\operatorname{sim}(F^{\mathbf{x}}(\mathbf{x}), F^{\mathbf{p}}(\mathbf{p}))}/{\tau})}{\sum_{k}\exp({\operatorname{sim}(F^{\mathbf{x}}(\mathbf{x}), F^{\mathbf{p}}(\mathbf{p}^k))}/{\tau})}
\end{equation}
where $\theta_{F^{\mathbf{x}}}, \theta_{F^{\mathbf{p}}}$ are parameters of $F^{\mathbf{x}}$ and $F^{\mathbf{p}}$, $\mathbf{p}^k$ is the negative text sample of $\mathbf{x}$, $\tau$ is the a temperature to scale logits, $\operatorname{sim}$ is the dot product to measure distance between features. For simplicity, Eqn.~\eqref{eqn:edsam-clip_model_learning} only illustrates the contrastive loss over images. In practice, a symmetrical loss over texts is also applied, and the loss is the average of the contrastive loss over images and texts.

\subsubsection{Domain Generalization of Contrastive Language-Image Pre-Training}

In our paper, we aim to develop a domain generalization approach to CLIP that is able to better generalize to new unknown data distributions. In this work, we consider the training data of CLIP drawn from a single source data~\cite{volpi2018generalizing}, i.e., $\mathbf{x}_s, \mathbf{p}_s \in p(\mathbf{x}_s, \mathbf{p}_s)$. Inspired by prior work in robust optimization, we propose to model the domain generalization of CLIP via the worst-case problem around the source data distribution $p(\mathbf{x}_s, \mathbf{p}_s)$ as in Eqn.~\eqref{eqn:edsam-dg_general}.
\begin{equation} \label{eqn:edsam-dg_general}
\small
    \theta^*_{F^{\mathbf{x}}}, \theta^*_{F^{\mathbf{p}}} = \arg\min_{\theta_{F^{\mathbf{x}}}, \theta_{F^{\mathbf{p}}}} \sup_{p_t:\mathcal{D}(p_t, p_s)\leq \rho} \mathbb{E}_{\mathbf{x}_t, \mathbf{p}_t \sim p_t(\mathbf{x}_t, \mathbf{p}_t)} -\log \frac{\exp({\operatorname{sim}(F^{\mathbf{x}}(\mathbf{x}_t), F^{\mathbf{p}}(\mathbf{p}_t))}/{\tau})}{\sum_{k}\exp({\operatorname{sim}(F^{\mathbf{x}}(\mathbf{x}_t), F^{\mathbf{p}}(\mathbf{p}^k))}/{\tau})}
\end{equation}
where $\mathbf{x}_t, \mathbf{p}_t$ are images and prompt sampled from $p_t$, 
$\mathcal{D}(p_s, p_t)$ is the Wasserstein metric measure the distance between two data distributions $p_s$ and $p_t$, $\rho$ is the distance constraint, $p_t$ is $\rho$-away unknown data distributions from $p_s$, i.e., $\mathcal{D}(p_s, p_t) \leq \rho$.
Eqn.~\eqref{eqn:edsam-dg_general} aims to guarantee good performance of the CLIP model against the unknown data distribution.

\noindent
\textbf{Domain Generalization of CLIP.} In our paper, we are interested in the problem of domain generalization of CLIP where we aim to improve the performance of CLIP, especially when using the CLIP model for downstream tasks, e.g., zero-shot classification, linear probing, or fine-tuning. In this learning scenario, since the target data distribution $p_t$ is completely unknown, the hyper-parameter $\rho$ plays an important role since it will indicate the generalizability of the CLIP model to new data (or test domains). 
To solve the Eqn \eqref{eqn:edsam-dg_general}, the Lagrange multiplier can be adopted to reform Eqn \eqref{eqn:edsam-dg_general} as in Eqn.~\eqref{eqn:edsam-adv_sample}.
\begin{equation}\label{eqn:edsam-adv_sample}
\small
    \mathbf{x}^*_t =\arg \max_{\mathbf{x}_t} \Big\{ \mathcal{L}_{CLIP}(\mathbf{x}_t,\mathbf{p}_s) - \lambda \mathcal{D}(p_t(\mathbf{x}_t, \mathbf{p}_s), p_s(\mathbf{x}_s, \mathbf{p}_s)) \Big\}
\end{equation}
where $\mathbf{x}^*_t$ is the \textbf{\textit{adversarial sample}} (corresponding to prompt $\mathbf{p}_s$) to improve the generalization and robustness of the CLIP model,
$\mathcal{L}_{CLIP}$ is the contrastive language-image pretraining loss defined in Eqn.~\eqref{eqn:edsam-clip_model_learning}, 
$\lambda$ is the hyper-parameter that is inverse proportional of $\rho$, $\mathcal{D}(p_t(\mathbf{x}_t, \mathbf{p}_s), p_s(\mathbf{x}_s, \mathbf{p}_s))$ is the transportation cost that moving from the $\mathbf{x}_s, \mathbf{p}_s \sim p_s(\mathbf{x}_s, \mathbf{p}_s)$ to the distribution $p_t$.
Since our paper aims to improve the generalizability of the CLIP model on the downstream vision tasks, the scope of this work focuses on the adversarial sample in the vision domain. 
Eqn.~\eqref{eqn:edsam-adv_sample} aims to create augmented samples so that the distribution of augmented samples is $\rho$-away from the original one and increases contrastive learning loss. Then, using these augmented samples will potentially improve the generalizability of CLIP. 

\begin{figure}[!t]
    \centering
    \includegraphics[width=1.0\textwidth]{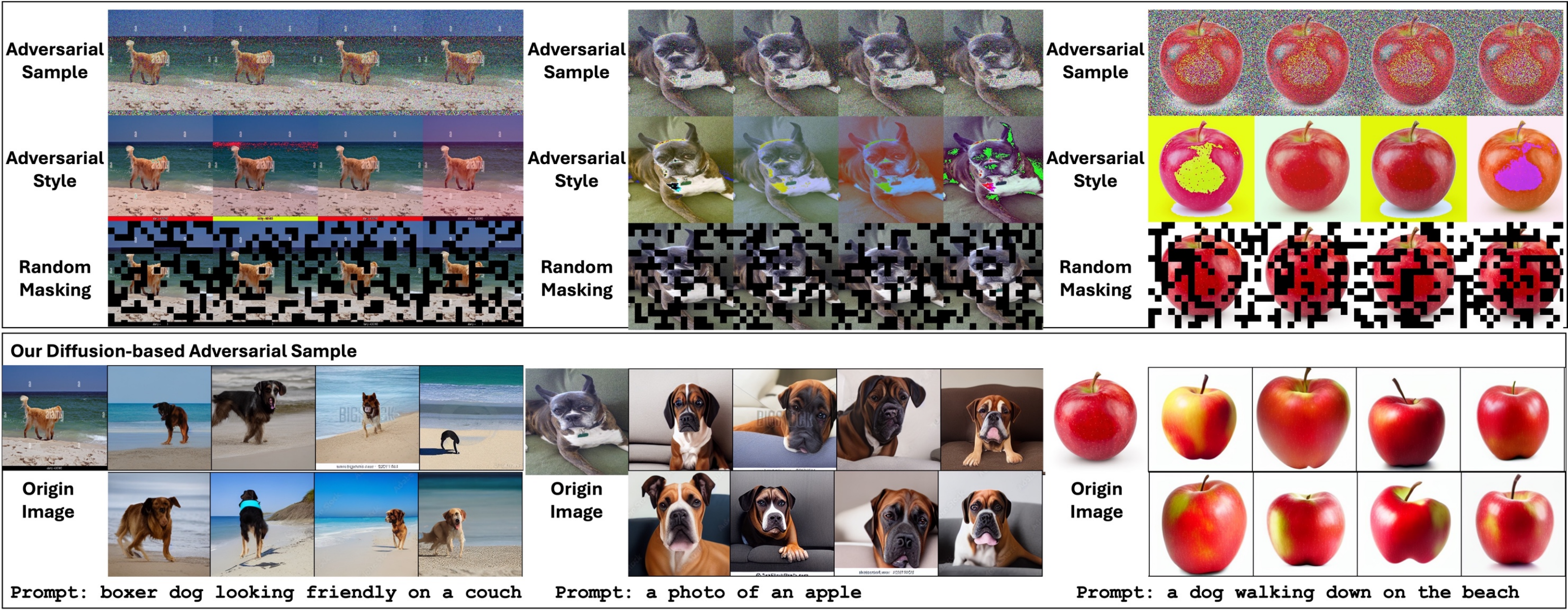}
    \caption{The Comparison of Our Diffusion-based Adversarial Sample and Prior Methods (Adversarial Sample~\cite{volpi2018generalizing}, Adversarial Style~\cite{zhong2022adversarial}, Masking Sample~\cite{li2023scaling}).}
    \label{fig:edsam-adv_sample_compare}
\end{figure}

\noindent
\textbf{Limitation of Prior Work.}
Prior work adopts adversarial training~\cite{volpi2018generalizing}, augmentation methods~\cite{li2023scaling}, or adversarial style augmentation~\cite{zhong2022adversarial} to generate adversarial/augmented samples to improve the generalizability.
Although prior results have shown the potential performance improvement,
these approaches remain limited in terms of expanding their generalizability to unknown distributions.
Indeed, adversarial learning~\cite{volpi2018generalizing, zhong2022adversarial} tries to add the perturbation via maximizing loss or adversarial styles into images.
Meanwhile, the augmentation methods create different variations of images by performing heuristic pixel-wise image operations (e.g., masking, cropping, color jittering, etc).
However, the data distributions of augmented samples generated by prior methods~\cite{volpi2018generalizing, zhong2022adversarial, li2023scaling} 
remain unchanged or little changed compared to the original data distribution.
This can be explained since, despite the different variations of augmented samples, the content information, e.g., object appearances, shapes, etc., and the semantic background information remained the same.
For example, as shown in Figure~\ref{fig:edsam-adv_sample_compare}, the target object of augmented samples created by~\cite{volpi2018generalizing, zhong2022adversarial, li2023scaling} remains unchanged. The semantic background, in general, is similar to the original image, with noise added.

\subsubsection{The Relation of Diffusion to Adversarial Augmentation}

As aforementioned, the goal of the adversarial sample in Eqn.~\eqref{eqn:edsam-adv_sample} is to move the data sample from the source training $\mathbf{x}_s$ to the $\mathbf{x}^*_t$ in the $\rho$-away distribution so that maximize the contrastive language-image pretraining loss $\mathcal{L}_{CLIP}(\mathbf{x}_t, p_s)$. As shown in Eqn.~\eqref{eqn:edsam-adv_sample}, the sample $\mathbf{x}^*_t$ is depending on the source training sample $\mathbf{x}_s$, the text prompt $\mathbf{p}_s$, and the distance between two distributions $\rho$. Therefore, 
in our work, we consider the adversarial sample $\mathbf{x}^*_t$ is draw from a $\rho$-away distribution conditioned on $\mathbf{x}_s$, $\mathbf{p}_s$, and $\rho$, i.e., $\mathbf{x}^*_t \in p(\mathbf{x}^*_t | \mathbf{x}_s, \mathbf{p}_s, \rho)$.

\begin{wrapfigure}[10]{r}{0.25\textwidth}
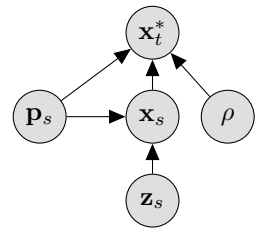

    \label{eqn:edsam-graphical_model}
    \centering
    \tikz{
         \node[obs] (xt) {$\mathbf{x}^*_t$};
         \node[obs,below=of xt,xshift=-0cm, yshift=0.6cm] (xs) {$\mathbf{x}_s$}; 
         \node[obs,below=of xs,xshift=-0cm, yshift=0.6cm] (zs) {$\mathbf{z}_s$}; 
         \node[obs,below=of xt,xshift=-1.5cm, yshift=0.6cm] (ps) {$\mathbf{p}_s$}; 
         \node[obs,below=of xt,xshift=1cm, yshift=0.6cm] (rho) {$\rho$}; 
         \edge {xs,ps,rho} {xt}
         \edge {zs,ps} {xs}
    }
    \caption{The Relation Between Adversarial Sample and Source Data.}
    \label{fig:edsam-graphic_model}
\end{wrapfigure}
\noindent
\textbf{The Source Data Distribution.} Since the image $\mathbf{x}_s$ and $\mathbf{p}_s$ is a pair of image and text, without a strict argument, we could assume that the image $\mathbf{x}_s$ is conditioned on the text prompt $\mathbf{p}_s$, i.e., $\mathbf{x}_s \in p(\mathbf{x}_s | \mathbf{p}_s)$. 
As presented in Sec. \ref{sec:edsam-diffusion_background}, the conditional distribution $p(\mathbf{x}_s | \mathbf{p}_s)$ could be efficiently modeled by the diffusion. Let $\mathbf{z}_s \in \mathcal{N}({\mathbf{0}, \mathbf{I}})$ be the latent variable of image $\mathbf{x}_s$. 
Then, the image $\mathbf{x}_s$ can be modeled via the backward process of diffusion conditioned on $\mathbf{p}_s$ and $\mathbf{z}_s$ as in Eqn.~\eqref{eqn:edsam-diffusion_backward}
For simplicity, we rewrite the data distribution $\mathbf{x}_s \sim p(\mathbf{x}_{0:T} | \mathbf{p}_s)$ via the latent variable $\mathbf{z}_s$ as $\mathbf{x}_s \sim p(\mathbf{x}_s | \mathbf{z}_s, \mathbf{p}_s)$. 

\noindent
\textbf{The Diffusion-based Adversarial Augmentation.}
Figure~\ref{fig:edsam-graphic_model} illustrates the graphical model that define the relation among $\mathbf{x}_s$, $\mathbf{p}_s$, $\mathbf{z}_s$, $\mathbf{x}^*_t$, and $\rho$. The relation in this graphical model is established based on two conditions of the adversarial sample $p(\mathbf{x}^*_t | \mathbf{x}_s, \mathbf{p}_s, \rho)$ and the conditional diffusion model $p(\mathbf{x}_s | \mathbf{z}_s, \mathbf{p}_s)$.
As shown in the graphical model, we have observed that the adversarial sample $\mathbf{x}^*_t$ depends on $(\mathbf{x}_s, \mathbf{p}_s, \rho)$ while the image $\mathbf{x}_s$ is conditioned on $(\mathbf{z}_s, \mathbf{p}_s)$. Therefore, for simplicity, without a strict argument, we assume that the adversarial sample $\mathbf{x}^*_t$ is equivalently depending on $(\mathbf{z}_s, \mathbf{p}_s, \rho)$ defined as in Eqn.~\eqref{eqn:edsam-adv_sample_dist}.
\begin{equation}\label{eqn:edsam-adv_sample_dist}
\begin{split}
\small
    \mathbf{x}^*_t &\sim p(\mathbf{x}^*_t | \mathbf{x}_s, \mathbf{p}_s, \rho) \Rightarrow \mathbf{x}^*_t \sim p(\mathbf{x}^*_t | \mathbf{z}_s, \mathbf{p}_s, \rho) \Rightarrow \mathbf{x}^*_t \sim p(\mathbf{x}^*_t | \mathbf{z}^*_t, \mathbf{p}_s) \; \text{where} \; \mathbf{z}^*_t = \mathcal{T}(\mathbf{z}_s, \rho) \\
\end{split}
\end{equation}
where $\mathcal{T}$ is the transport transformation on the latent space.
Intuitively, instead of moving the image $\mathbf{x}_s$ in the image space to $\mathbf{x}^*_t$ in the new distribution with a transportation cost of $\mathcal{D}(p_t(\mathbf{x}^*_t, \mathbf{p}_s), p_s(\mathbf{x}_s, \mathbf{p}_s))$ as in Eqn.~\eqref{eqn:edsam-adv_sample} which is a challenging problem,
we are going to move the latent variable $\mathbf{z}_s$ to $\mathbf{z}^*_t$ via the transport function $\mathcal{T}$ controlled by $\rho$.
Since the latent space of the diffusion model is tractable (as it is a Gaussian distribution), moving $\mathbf{z}_s$ to $\mathbf{z}^*_t$ on latent space is controllable and easier than moving samples on the image space.
Then, the adversarial sample of $\mathbf{x}^*_t$ can be achieved by the reverse process of the diffusion model.
With our proposed diffusion-based augmentation approach, thanks to the power of the diffusion model~\cite{rombach2022high}, our approach is able to synthesize novel adversarial samples that still maintain the semantic conditions on the prompt $\mathbf{p}_s$ while being effectively used to improve the generalizability in training CLIP model.
As shown in Figure~\ref{fig:edsam-adv_sample_compare}, our proposed approach can generalize a new sample with the sample condition prompt, but the content and semantic background of the image have been changed significantly. This helps to strongly expand the data distribution during training to improve the generalizability of unknown data distribution.

\subsubsection{The Proposed Transport Transformation} 

It is important to design a transformation $\mathcal{T}$ that satisfies the condition of domain generalization, i.e., $\mathcal{D}(p_s(\mathbf{x}_s, \mathbf{p}_s), p_t(\mathbf{x}_t, \mathbf{p}_t)) \leq \rho$ in Eqn.~\eqref{eqn:edsam-dg_general}, to guarantee the generalizability defined in Eqn.~\eqref{eqn:edsam-dg_general}.
Since the data distribution in our approach is displaced in the latent space of $\mathbf{z}_s$, with a strict argument, the condition of domain generalization via the latent space could be written as in Eqn.~\eqref{eqn:edsam-requirement}.
\begin{equation}\label{eqn:edsam-requirement}
    \small
    \mathcal{D}(p_s(\mathbf{x}_s, \mathbf{p}_s), p_t(\mathbf{x}^*_t, \mathbf{p}_t)) \propto \mathcal{D}(p_s(\mathbf{z}_s), p_t(\mathbf{z}^*_t)) \leq \rho
\end{equation}

In our proposed approach, in order the meet the requirement as defined in Eqn.~\eqref{eqn:edsam-requirement}, the transport transformation $\mathcal{T}$ can be defined as in Eqn.~\eqref{eqn:edsam-transform_function}.
\begin{equation} \label{eqn:edsam-transform_function}
    \small
    \mathbf{z}^*_t = \mathcal{T}(\mathbf{z}_s, \rho) = \frac{\mathbf{z}_s + \mathcal{N}(\alpha\sqrt{2}, \mathbf{I})}{\sqrt{2}} \quad \text{where } \alpha \sim \mathcal{U}(-\rho, \rho) 
\end{equation}
where $\alpha$ is controllable hyper-parameter uniformly sampled from $\mathcal{U}(-\rho, \rho)$.

\begin{tcolorbox}[colback=blue!5!white,colframe=blue!75!black,title=\textbf{Proposition \showpropositioncounter\label{pro:domain_generalizatino_edsam}}]
Given $\mathbf{z}_s \in \mathcal{N}(\mathbf{0, I})$ and $\alpha$ ($-\rho \leq \alpha \leq \rho$), the condition of distance between distributions $\mathcal{D}(p_s(\mathbf{z}_s), p_t(\mathbf{z}^*_t)) \leq \rho$ holds if the transport transformation $\mathcal{T}$ is defined as in Eqn.~\eqref{eqn:edsam-proposition-condition}.
\begin{equation}\label{eqn:edsam-proposition-condition}\small
    \mathbf{z}^*_t = \mathcal{T}(\mathbf{z}_s, \rho) = \frac{\mathbf{z}_s + \mathcal{N}(\alpha\sqrt{2}, \mathbf{I})}{\sqrt{2}}
\end{equation}
\end{tcolorbox}

The proof of our \textbf{Proposition \ref{pro:domain_generalizatino_edsam}} can be found in our preliminary work~\cite{truong2024edsam}.
While there could be multiple transport transformations that satisfy the condition of the distance between two distributions, i.e.,  $\mathcal{D}(p_s, p_t) \leq \rho$, we have observed that our proposed metric in Eqn.~\eqref{eqn:edsam-transform_function} provides a better mechanism to move the sample on the latent spaces. This could be explained since our metric is able to expand the training data distribution by moving the original latent vectors in the latent space while still maintaining the important property as mentioned in \textbf{Proposition \ref{pro:domain_generalizatino_edsam}}. In addition, by moving the latent vector $\mathbf{z}_s$ in the latent space with a controlled parameter $\rho$, our metric can guarantee the semantic content information compared to the original one while creating the diverse semantic variations of the images. This also encourages the diffusion model to avoid synthesizing useless random images with uncontrolled latent vectors.

\begin{figure}[!t]
    \centering
    \includegraphics[width=1.0\textwidth]{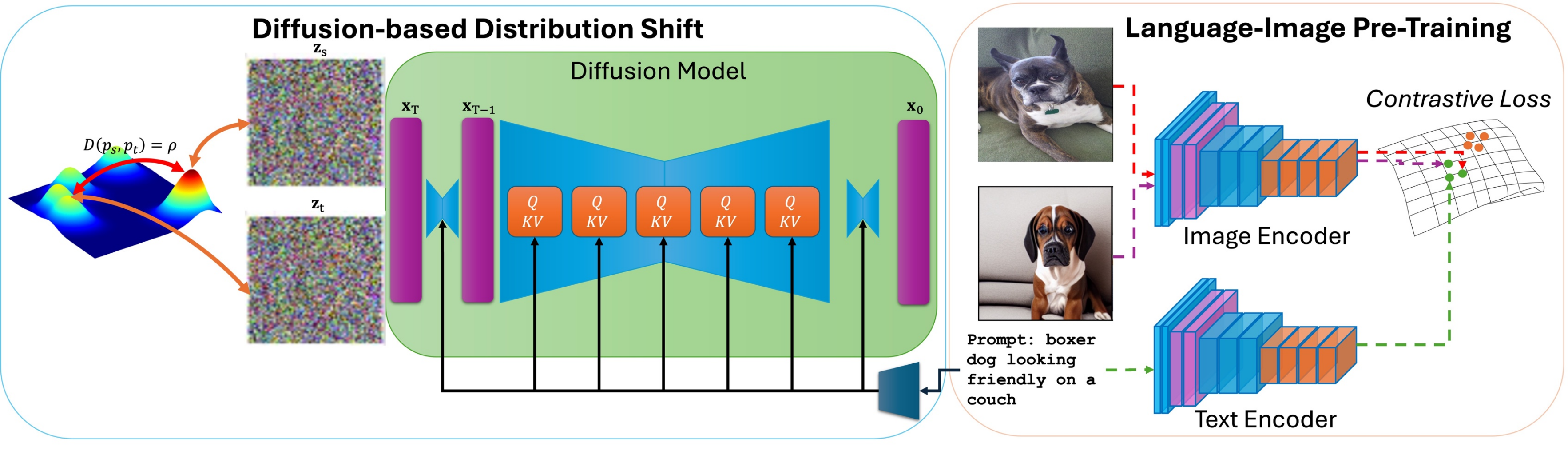}
    \caption{The Proposed Diffusion-based Domain Generalization Framework}
    \label{fig:edsam-da_framework}
\end{figure}

\subsection{The Proposed Diffusion-based Domain Generalization Training Approach}

\noindent
\textbf{Large Scale Diffusion-based Augmentation Sample Generation.}
As shown in our theoretical analysis, generating our diffusion-based adversarial samples does not require alternative training steps with the CLIP training procedure. We have empirically observed that retraining the text-to-image diffusion model is unnecessary because the pre-trained diffusion model has been well learned on extreme-scale datasets, can model the data distribution well, and generates diverse synthetic data. Therefore, in our approach, we adopt the pretrained Latent Diffusion model~\cite{rombach2022high} to generate the adversarial samples in advance to save the training time of CLIP. Formally, for each image $\mathbf{x}_s$ and its corresponding prompt $\mathbf{p}_s$, we are going to generate $M$ different augmented samples $\mathbf{x}^*_t$ via the latent diffusion model by the following process as in Eqn.~\eqref{eqn:edsam-adv-gen-process}.
\begin{equation}\label{eqn:edsam-adv-gen-process}
\small
\begin{split}
    \mathbf{z}_s &= \operatorname{LDMForward}(\mathbf{x}_s), \;
    \mathbf{z}^*_t = \frac{\mathbf{z}_s + \mathcal{N}(\alpha\sqrt{2}, \mathbf{I})}{\sqrt{2}} \text{ where } \alpha \sim \mathcal{U}(-\rho, \rho), \;
    \mathbf{x}^*_t = \operatorname{LDMBackward}(\mathbf{z}_t)
\end{split}
\end{equation}
where $\operatorname{LDMForward}$ and $\operatorname{LDMBackward}$ are the forward and backward processes of the latent diffusion model. Generating the adversarial samples during training will result in a longer training time for CLIP, which is unnecessary. Therefore, we propose to generate the adversarial samples via diffusion in advance, followed by using them to train the CLIP model, which is more time-efficient.

\noindent
\textbf{The Diffusion-based Domain Generalization Training.}
Figure~\ref{fig:edsam-da_framework} illustrates our proposed domain generalization framework. After the generation steps, each real image has $M$ different adversarial samples. Then, we are able to improve the generability of the CLIP model by training on the real and adversarial samples together. Formally, learning the CLIP model can be re-written as in Eqn.~\eqref{eqn:edsam-domain_adaptation}. 
\begin{equation} \label{eqn:edsam-domain_adaptation}
\small
        \theta^*_{F^{\mathbf{x}}}, \theta^*_{F^{\mathbf{p}}} = \arg\min_{\theta_{F^{\mathbf{x}}}, \theta_{F^{\mathbf{p}}}} \mathbb{E}_{\mathbf{x}_s, \mathbf{p}_s, \mathbf{x}^*_t} \left[\mathcal{L}_{CLIP}(\mathbf{x}_s, \mathbf{p}_s) + \mathcal{L}_{CLIP}(\mathbf{x}^*_t, \mathbf{p}_s) \right]
\end{equation}

\subsection{Experimental Results}

\subsubsection{Datasets, Implementations, and Evaluations}

\input{Tables/chap-5/edsam/rho_ablation}
\noindent
\textbf{Datasets.}
We trained our foundation model on three different image-text datasets at different scales: Conceptual Captions 3M (CC3M) ~\cite{sharma2018conceptual}, Conceptual Captions 12M (CC12M)~\cite{changpinyo2021cc12m}, and LAION400M~\cite{schuhmann2021laion}.
Due to the hardware constraints, our ablation studies are mainly conducted on CC3M and CC12M. 
We evaluate our models on ImgageNet~\cite{imagenet15russakovsky} and six common datasets, including, STL-10~\cite{coates2011analysis}, Country-211~\cite{thomee2016yfcc100m},	Caltech-101~\cite{fei2006one}, Flowers~\cite{nilsback2006visual}, Pets~\cite{parkhi2012cats}, and SUN-397~\cite{xiao2016sun}.

\noindent
\textbf{Implementation.}
We adopt the implementation of OpenCLIP~\cite{cherti2023reproducible} and Latent Diffusion~\cite{rombach2022high} in our experiments. 
For the CLIP model, we use the ViT-B/16 architecture.
For a fair comparison, our model is trained for 32 epochs with a similar hyper-parameter setting as~\cite{radford2021learning, cherti2023reproducible}. 
We utilize $32$ NVIDIA A100 GPUs (40GB), and the batch size of our experiments is set to $320$ per GPU. 
For image synthesis, we use the text-to-image latent diffusion model~\cite{rombach2022high} to generate images at the resolution of $256 \times 256$ with $10$ DDIM steps.
For each real image, we generate $M=10$ different synthetic images.
The controlling hyper-parameter of the distance between distributions $\rho$ is set to $0.5$ in our experiments.
Due to time and hardware constraints, we choose to use only $10$ DDIM generation steps. This offers image quality that meets acceptable standards~\cite{rombach2022high} while maintaining efficient data generation time on large-scale datasets (e.g., approximately 7.5 hours to generate 12M adversarial samples of CC12M on 32 GPUs).

\input{Tables/chap-5/edsam/number_of_images_ablation}
\noindent
\textbf{Evaluation Setup.}
In our experiments, we consider three different evaluation metrics, i.e., Zero-shot Classification Accuracy, Linear Probing Accuracy, and Fine-tuning Accuracy. 
For zero-shot classification, we adopt the template of prompts and evaluation protocol as described in CLIP~\cite{radford2021learning}. 
For linear probing, following the common practices~\cite{radford2021learning, li2023scaling, mu2022slip}, we use our frozen pre-trained image encoder to extract features followed by training a linear classifier. 
For a fair comparison, we adopt the hyper-parameter setting from~\cite{radford2021learning, cherti2023reproducible}. 
For fine-tuning evaluation, we fine-tune the end-to-end image encoder with a linear classifier on the ImageNet 1K dataset. We adopt the implementation and learning hyper-parameter setting from~\cite{cherti2023reproducible} for fair comparisons.
The majority of our experiments are evaluated on the ImageNet 1K dataset.
To further illustrate the generability of our model, we also perform the zero-shot evaluation on six different zero-shot benchmarks STL-10, Country-211, Caltech-101, Flowers, Pets, and SUN-397.

\subsubsection{Ablation Studies}

\noindent
\textbf{Effectiveness of Distribution Moving $\rho$.}
The results in Table~\ref{tab:edsam-rho_ablation} illustrate the effectiveness of the distance between distribution $\rho$. When the value of $\rho$ is small, i.e., $\rho = 0.05$, the CLIP gains a little improvement due to the small distribution shift. Then, the performance is gradually improved when the value of $\rho$ increases from $0.05$ to $0.5$.
When the value of $\rho$ is increased, the CLIP model can improve its generalizability to unknown distributions. 
However, if we keep increasing the value of $\rho$, the performance tends to drop. This is because if we shift the new data distribution in the latent space far away from the original data distribution ($\mathcal{N}(\mathbf{0, I})$), the quality of synthetic images generated by the latent diffusion model will dramatically drop in both realism and content information. Our best performance of CLIP is achieved at $\rho$ of $0.5$. 

\input{Tables/chap-5/edsam/transformation_ablation}
\noindent
\textbf{Effectiveness of Number of Augmented Images.}
As shown in Table~\ref{tab:edsam-number_of_image_abl}, the performance of our domain generalization approach evaluated on ImageNet1K is gradually increased when the number of adversarial images is increased.
When we use only $3$ adversarial images, the CLIP model gains a minor performance.
Meanwhile, when we use the $10$ adversarial images during training, the zero-shot classification performance of CLIP trained on CC3M and CC12M archives up to $20.33\%$ and $39.34\%$. The performance of linear probing and fine-tuning is also significantly improved when the number of adversarial images is increased.
However, if we keep increasing the number of images, we have observed that the performance of the CLIP model is becoming stable.
Therefore, generating $10$ adversarial images for each real image is a good trade-off between performance and time efficiency.

\input{Tables/chap-5/edsam/ldm_ablation}

\noindent
\textbf{Effectiveness of Transport Transformation.}
To illustrate the effectiveness of our Transport Transformation $T$, we 
compared it with another transformation. We define another random transformation by sampling $\mathbf{z}^*_t$ from the normal distribution $\mathbf{z}^*_t \sim \mathcal{N}(\rho, \mathbf{I})$.
For a fair comparison, this transformation also satisfies the condition of $\mathcal{D}(p(\mathbf{z}_s), p(\mathbf{z}^*_t)) \leq \rho$. Then, the image $\mathbf{x}^*_t$ is generated via the diffusion model with $\mathbf{z}^*_t \sim \mathcal{N}(\rho, \mathbf{I})$ and the original prompt $\mathbf{p}_s$. 
As shown in Table~\ref{tab:edsam-transport_abl}, our transport transformation $T$ significantly outperforms the random transformation. Indeed, using the random transformation even downgrades the performance of the CLIP model since the generation of the diffusion model by using random transformation is uncontrolled.
Meanwhile, by controlling the latent variable $\mathbf{z}^*_t$ via $\mathbf{z}_s$ and $\rho$, as defined in Eqn.~\eqref{eqn:edsam-transform_function}, the generation of adversarial samples is oriented and significantly improves the CLIP's performance.

\noindent
\textbf{Effectiveness of Pre-trained and Re-trained Diffusion Model.}
We compared the pre-trained LDM with a re-trained latent diffusion model on CC3M and CC12M.
We only re-train the second stage of the LDM while we adopt the pre-trained VQ-VAE of LDM~\cite{rombach2022high} for the first stage.
As shown in Table~\ref{tab:edsam-ldm_ablation}, the experimental results show that using adversarial samples generated via our transport transformation has significantly improved the performance in both cases of using re-trained and pre-trained LDM.
However, practically, the performance of using the pre-trained diffusion model outperforms re-training the diffusion model on the corresponding dataset.
This is because the pre-trained latent diffusion model was trained on the large-scale dataset and is able to model the data distribution better than the re-trained latent diffusion on the specific datasets. Therefore, using the pre-trained latent diffusion model is beneficial in terms of not only time efficiency but also performance improvement.

\input{Tables/chap-5/edsam/different_dataset_clip_laclip_slip}

\noindent
\textbf{Effectiveness of Our Domain Generalization on Different Datasets and CLIP-based Models.}
Table~\ref{tab:edsam-ablation_diffent_data_model} illustrates the results of our proposed approach on three datasets at different scales and CLIP-based models, i.e., CLIP~\cite{radford2021learning}, LaCLIP~\cite{fan2023improving}, and SLIP~\cite{mu2022slip}.
The zero-shot classification results have illustrated the generalizability of our proposed approach on different dataset scales. In particular, our proposed approach improves the zero-shot results of CLIP by $+3.23\%$, $+2.48\%$, and $3.11\%$ on CC3M, CC12M, and LAION400M, respectively. 
Further fine-tuning the model via linear probing or end-to-end fine-tuning significantly improves the performance of the CLIP model. 
The results of fine-tuned models on ImageNet achieved $81.12\%$, $84.67\%$, and $86.98\%$ on CC3M, CC12M, and LAION400M, respectively. 
Our proposed approach is effective not only on different datasets but also with different CLIP-based approaches. By further using better CLIP-based training approaches, i.e., LaCLIP or SLIP, the performance of zero-shot results is significantly improved, up to $75.58\%$ trained LAION-400M using SLIP. By further fine-tuning the SLIP model, our proposed approach achieved state-of-the-art performance on ImageNet1K, i.e., $87.49\%$.
The results in Table  \ref{tab:edsam-ablation_diffent_data_model} have confirmed the scalability and generalizability of our approach across the training datasets and CLIP-based models.

\subsubsection{Comparisons With State-of-the-Art Approaches}

\input{Tables/chap-5/edsam/dg_comparison}

In this section, we present the results of our approach compared with other augmentation and domain generalization approaches, i.e., ADA~\cite{volpi2018generalizing}, AdvStyle~\cite{zhong2022adversarial}, and Masking Augmentation (FLIP)~\cite{li2023scaling}. 

\noindent
\textbf{Zero-shot Classification.}
Table~\ref{tab:edsam-dg_comparison} compares our approach with other augmentation and domain generalization methods.
Our proposed approach consistently improves the performance of zero-shot classification.
While the masking augmentation generates masked augmented samples, ADA~\cite{volpi2018generalizing} and AdvStyle~\cite{zhong2022adversarial} generate the adversarial samples via adversarial training. However, the distribution shift in these methods remains limited compared to our diffusion-based approach.
As a result, our proposed approach significantly outperforms other augmentation and domain generalization approaches. In particular, by pre-training on the large-scale LAION400M dataset, our model achieves the state-of-the-art zero-shot classification performance, i.e., $70.11\%$ and $72.53\%$ by using CLIP and SLIP training.
The results have shown our advantages in improving the generalizability of vision-language models against unknown data distributions.

\input{Tables/chap-5/edsam/zeroshot_others}
\noindent
\textbf{Linear-Probing and End-to-end Fine-tuning Classification.}
Table~\ref{tab:edsam-dg_comparison} illustrates the results of our linear probing and fine-tuning experiments.
Similar to the zero-shot classification results, our linear probing and end-to-end fine-tuning results consistently improve the performance of CLIP~\cite{radford2021learning} and SLIP~\cite{mu2022slip} and outperform other augmentation approaches.
By pre-training on LAION-400M and further fine-tuning on ImageNet-1K, our training approach achieved state-of-the-art performance, with the accuracy of CLIP and SLIP improved to $86.98\%$ and $87.49\%$.
These results have further confirmed the effectiveness of our approach across evaluation settings and pre-training datasets.

\noindent
\textbf{Other Zero-shot Classification Benchmarks.} Table~\ref{tab:edsam-zeroshot_other} illustrates the results of our proposed approach (pretrained on LAION400M)
on six different zero-shot benchmarks. 
Our approach consistently improves the performance of CLIP and SLIP on all zero-shot classification benchmarks which have illustrated the generalizability of our approach to unseen domains. 
Thanks to our generalization approach, the vision-language foundation model is able to learn better visual representation against the data distribution shift. 
Therefore, the vision-language model can be later well generalized to various downstream tasks.

%% file: Tables/chap-5/edsam/rho_ablation.tex
\begin{wraptable}{r}{0.4\textwidth}
\centering
\caption{The Effectiveness of Distribution Moving $\rho$.}
\label{tab:edsam-rho_ablation}
\resizebox{0.4\textwidth}{!}{
    \begin{tabular}{|l|c|ccc|}
    \hline
                           & $\rho$ & Zeroshot & Linear Prob & Fine-Tune \\
    \hline
    \multirow{5}{*}{\rotatebox{90}{CC3M}}  & 0.05          & 17.28    & 54.13       & 80.08     \\
                           & 0.20          & 18.79    & 55.11       & 80.61     \\
                           & 0.50          & \textbf{20.33}    & \textbf{56.14}       & \textbf{81.12}     \\
                           & 0.70          & 19.82    & 55.44       & 80.09     \\
                           & 1.00          & 16.68    & 52.92       & 79.14     \\
    \hline
    \multirow{5}{*}{\rotatebox{90}{CC12M}} & 0.05          & 36.44    & 69.27       & 80.28     \\
                           & 0.20          & 38.37    & 71.17       & 83.11     \\
                           & 0.50          & \textbf{39.34}    & \textbf{72.12}       & \textbf{84.67}     \\
                           & 0.70          & 37.34    & 69.12       & 82.89     \\
                           & 1.00          & 35.19    & 68.94       & 81.74    \\
    \hline
    \end{tabular}
}
\end{wraptable}

%% file: Tables/chap-5/edsam/number_of_images_ablation.tex
\begin{wraptable}{r}{0.4\textwidth}
\centering
\caption{The Effectiveness of Number of Generated Samples.}
\label{tab:edsam-number_of_image_abl}
\resizebox{0.4\textwidth}{!}{
    \begin{tabular}{|l|c|ccc|}
    \hline
                           & {$M$} & {Zeroshot} & {Linear Prob} & {Fine-Tune} \\
    \hline
    \multirow{6}{*}{\rotatebox{90}{CC3M}}  & 
    0                   & 17.10          & 53.50          & 79.50  \\
    
    & 3                     & 18.36                        & 54.05                           & 79.90                         \\
                           & 5                     & 19.05                        & 55.17                           & 80.82                         \\
                           & 10                    & \textbf{20.33}               & \textbf{56.14}                  & \textbf{81.12}                \\
                           & 15                    & 20.40                        & 57.26                           & 81.17                         \\
                           & 20                    & 20.28                        & 56.18                           & 81.11                         \\
    \hline
    \multirow{6}{*}{\rotatebox{90}{CC12M}} & 0 & 36.50          & 69.00          & 82.10 \\
    & 3                     & 37.34                        & 70.25                           & 82.84                         \\
                           & 5                     & 38.10                        & 71.44                           & 83.58                         \\
                           & 10                    & \textbf{39.34}               & \textbf{72.12}                  & \textbf{84.67}                \\
                           & 15                    & 39.21                        & 72.18                           & 84.65                         \\
                           & 20                    & 39.49                        & 72.15                           & 84.68        \\
        \hline
    \end{tabular}
}
\end{wraptable}

%% file: Tables/chap-5/edsam/transformation_ablation.tex
\begin{wraptable}{r}{0.4\textwidth}
\centering
\setlength{\tabcolsep}{3pt}
\caption{The Effectiveness of Transport Transformation.}
\label{tab:edsam-transport_abl}
\resizebox{0.4\textwidth}{!}{
\begin{tabular}{|c|l|ccc|}
\hline
                       &                    & Zeroshot & Linear Prob & Fine-Tune \\
\hline
\multirow{3}{*}{\rotatebox{90}{CC3M}}  & CLIP & 17.10          & 53.50          & 79.50  \\
                      & Random              & 15.34    & 50.10       & 77.87     \\
                       & $\mathcal{T}$       & \textbf{20.33}    & \textbf{56.14}      & \textbf{81.12}     \\
\hline
\multirow{3}{*}{\rotatebox{90}{CC12M}} & CLIP & 36.50          & 69.00          & 82.10 \\
& Random              & 34.90    & 67.35       & 80.61     \\
                       & $\mathcal{T}$       & \textbf{39.34}    & \textbf{72.12}       & \textbf{84.67}    \\
\hline
\end{tabular}
}
\end{wraptable}

%% file: Tables/chap-5/edsam/ldm_ablation.tex
\begin{wraptable}{r}{0.4\textwidth}
\centering
\caption{The Effectiveness of Pre-trained Latent Diffusion Model.}
\label{tab:edsam-ldm_ablation}
\resizebox{0.4\textwidth}{!}{
\begin{tabular}{|c|l|ccc|}
\hline
                       &                    & Zeroshot & Linear Prob & Fine-Tune \\
\hline
\multirow{3}{*}{\rotatebox{90}{CC3M}}  & CLIP & 17.10          & 53.50          & 79.50  \\
& Retrained-LDM  & 18.77	& 55.12	& 80.18 \\
                       & Pretrained-LDM    & \textbf{20.33}    & \textbf{56.14}      & \textbf{81.12}     \\
\hline
\multirow{3}{*}{\rotatebox{90}{CC12M}} & CLIP & 36.50          & 69.00          & 82.10 \\
& Retrained-LDM & 38.26	& 71.11	& 83.06 \\
                       & Pretrained-LDM    & \textbf{39.34}    & \textbf{72.12}       & \textbf{84.67}    \\
\hline
\end{tabular}
}
\end{wraptable}

%% file: Tables/chap-5/edsam/different_dataset_clip_laclip_slip.tex
\begin{table}[!b]
\centering
\caption{The Effectiveness of Our Proposed Approach on Different Datasets and Different Language-Image Pretraining Models.}
\label{tab:edsam-ablation_diffent_data_model}
\setlength{\tabcolsep}{3pt}
\resizebox{1.0\textwidth}{!}{
\begin{tabular}{|l | ccc | ccc | ccc|}
\hline
                       & \multicolumn{3}{c|}{CC3M}                         & \multicolumn{3}{c|}{CC12M}                        & \multicolumn{3}{c|}{LAION400M}                    \\
                       & Zeroshot       & Linear Prob    & Fine-Tune      & Zeroshot       & Linear Prob    & Fine-Tune      & Zeroshot       & Linear Prob    & Fine-Tune      \\

\hline

CLIP                   & 17.10          & 53.50          & 79.50          & 36.50          & 69.00          & 82.10          & 67.00          & 78.60          & 84.70          \\
\textbf{Ours + CLIP}   & \textbf{20.33} & \textbf{56.14} & \textbf{81.12} & \textbf{39.34} & \textbf{72.12} & \textbf{84.67} & \textbf{70.11} & \textbf{80.74} & \textbf{86.98} \\
$\Delta$                  & +3.23           & +2.64           & +1.62           & +2.84           & +3.12           & +2.57           & +3.11           & +2.14           & +2.28           \\

\hline

LaCLIP                 & 21.50          & 56.50          & 81.15          & 48.40          & 72.30          & 82.53          & $-$ & $-$ & $-$ \\
\textbf{Ours + LaCLIP} & \textbf{24.12} & \textbf{58.03} & \textbf{83.11} & \textbf{51.16} & \textbf{74.34} & \textbf{84.68} & 
$-$ & $-$ & $-$ \\
$\Delta$                  & +2.62           & +1.53           & +1.95           & +2.76           & +2.04           & +2.15           & $-$ & $-$ & $-$ \\

\hline

SLIP                   & 23.00          & 65.40          & 81.40          & 40.70          & 73.70          & 83.10          & 70.21          & 80.34          & 85.83          \\
\textbf{SLIP+ Our}     & \textbf{26.97} & \textbf{67.60} & \textbf{83.18} & \textbf{43.13} & \textbf{75.58} & \textbf{84.95} & \textbf{72.53} & \textbf{83.21} & \textbf{87.49} \\
$\Delta$                  & +3.97           & +2.20           & +1.78           & +2.43           & 1.88           & +1.85           & +2.33           & +2.87           & +1.67           \\

\hline
\end{tabular}
}
\end{table}

%% file: Tables/chap-5/edsam/dg_comparison.tex
\begin{table}[t]
\centering
\caption{The Comparison With Other Augmentation and Generalization Approaches.}
\label{tab:edsam-dg_comparison}
\setlength{\tabcolsep}{3pt}
\resizebox{1.0\textwidth}{!}{
\begin{tabular}{|l|ccc|ccc|ccc|}
\hline
                & \multicolumn{3}{c|}{CC3M}           & \multicolumn{3}{c|}{CC12M}          & \multicolumn{3}{c|}{LAION400M}     \\
                & Zeroshot & Linear Prob & Fine-Tune & Zeroshot & Linear Prob & Fine-Tune & Zeroshot & Linear Prob & Fine-Tune \\
\hline
CLIP            & 17.10     & 53.50       & 79.50     & 36.50    & 69.00       & 82.10     & 67.00    & 78.60       & 84.70     \\
CLIP + Masking  & 17.69 & 54.13 & 80.08 & 37.34 & 70.56 & 82.28 & 68.06 & 78.95 & 85.03 \\
CLIP + ADA      & 18.36 & 55.75 & 80.43 & 38.10 & 70.95 & 82.93 & 68.59 & 79.54 & 85.23 \\
CLIP + AdvStyle & 19.01 & 55.55 & 80.40 & 38.77 & 71.22 & 81.21 & 69.47 & 79.90 & 85.57 \\
CLIP + Ours     & \textbf{20.33} & \textbf{56.14} & \textbf{81.12} & \textbf{39.34} & \textbf{72.12} & \textbf{84.67} & \textbf{70.11} & \textbf{80.74} & \textbf{86.98} \\
\hline

SLIP                 & 23.00          & 65.40          & 81.40          & 40.70          & 73.70          & 83.10          & 70.21          & 80.34          & 85.83          \\
SLIP + Masking       & 24.13          & 65.98          & 81.91          & 41.01          & 73.97          & 83.29          & 70.47          & 80.92          & 86.05          \\
SLIP + ADA           & 24.89          & 66.26          & 82.09          & 41.64          & 74.02          & 83.56          & 70.95          & 81.49          & 86.23          \\
SLIP + AdvStyle      & 25.50          & 66.55          & 82.59          & 42.30          & 74.46          & 84.01          & 71.37          & 81.74          & 86.58          \\
\textbf{SLIP + Ours} & \textbf{26.97} & \textbf{67.60} & \textbf{83.18} & \textbf{43.13} & \textbf{75.58} & \textbf{84.95} & \textbf{72.53} & \textbf{83.21} & \textbf{87.49} \\
\hline
\end{tabular}
}
\end{table}

%% file: Tables/chap-5/edsam/zeroshot_others.tex
\begin{wraptable}{r}{0.5\textwidth}
\centering
\caption{Zero-shot Classification Results on Six Benchmarks, i.e., STL-10, Country-211, Caltech-101, Flowers, Pets, and SUN-397.}
\label{tab:edsam-zeroshot_other}
\setlength{\tabcolsep}{2pt}
\resizebox{0.5\textwidth}{!}{
\begin{tabular}{|l|cccccc|}
\hline
                      & \rotatebox{0}{STL-10}         & \rotatebox{0}{Coun-211}     & \rotatebox{0}{Cal-101}    & \rotatebox{0}{Flowers}        & \rotatebox{0}{Pets}           & \rotatebox{0}{SUN-397}         \\
\hline
CLIP                  & 97.30          & 17.80          & 91.20          & 63.90          & 90.10          & 66.80          \\
\textbf{CLIP + Our}   & \textbf{97.58} & \textbf{18.34} & \textbf{93.14} & \textbf{77.12} & \textbf{91.74} & \textbf{68.85} \\
\hline
SLIP & 97.50 & 19.90 & 92.10 & 75.62 & 91.00 & 67.40 \\
\textbf{SLIP + Our} & \textbf{98.87} & \textbf{21.73} & \textbf{94.63} & \textbf{81.35} & \textbf{94.67} & \textbf{70.41} \\
\hline

\end{tabular}
}
\end{wraptable}

%% file: Chapters/Chaps/chap-7-conclusion.tex
\chapter{Conclusions and Future Work}\label{chap:conclusions}

\noindent
\textbf{Conclusions.}
This thesis has comprehensively investigated the research towards robustness and fairness in machine vision learning.
While the true robust and fair vision learning framework still requires more effort and studies, the research in this dissertation has presented several learning approaches and promoted vision learning toward robustness and fairness.
In this dissertation, we have identified two critical points in developing a robust and fair vision learning framework, including Learning Approach and Feature Representation.
The challenges of the learning approach rely on four problems of 
(1) Large-scale Labeled Data Requirement,
(2) Biased Prediction,
(3) Novel Classes Handling,
and (4) Limited Open-world Deployment.
Meanwhile, the robust feature representation relies on four perspectives of 
(1) Unexplainable Feature Representations,
(2) Multimodal Feature Representations, 
(3) Temporal Feature Representations,
and (4) Cross-view Feature Representations.
To address these challenges, this dissertation has presented four major research directions toward fairness and robustness.
First, Chapter~\ref{chap:adaptation} has presented the Fairness Domain Adaptation approaches to address the problem of large-scale labeled data and biased prediction.
Then, Chapter~\ref{chap:continual-learning} has introduced a novel Open-world Fairness Continual Learning approach to address the problem of novel class handling and open-world deployment.
Chapter~\ref{chap:cross-view} has presented new Geometry-based Approaches to learning robust cross-view feature representations.
Later, to study the robust representations, Chapter~\ref{chap:multimodal-temporal} has proposed new Transformer-based approaches to improve the robustness of multimodal and temporal feature representations.
In addition, this chapter has introduced  new Generalization Approaches to improve the robustness of large-scale vision language models. 
The theoretical analysis and experimental results presented in each chapter have confirmed the effectiveness of our proposed approaches, demonstrating their state-of-the-art performance compared to prior studies.
I believe that the contributions in this dissertation have contributed toward the improvement of fairness and robustness in machine vision learning.

\noindent
\textbf{Future Work.}
Developing a fair and robust vision learning framework is long-term research that requires more studies to focus on this direction. The research studies outlined in this dissertation are only the first few steps in the direction of fairness and robustness.
Recently, the rapid rise of large-scale multimodal models (LMMs) has promoted their advancement in various machine vision problems.
The future research of this dissertation will focus on addressing the current limitations of LMMs related to fairness and robustness.
In recent years, the demand for large-scale models, including large-scale language and multimodal models, e.g., large vision-language models (LVLMs) \cite{liu2024visual, liu2024improved}, has dramatically increased in academics and industries. 
More than 50+ large-scale multimodal models, including Gemini (Google), Chameleon (Meta), and ChatGPT (OpenAI) have been introduced within two recent years since 2023 \cite{yue2024mmmu}. 
However, the rapid development of LMMs has also raised concerns about their reliability of robustness, and fairness and poses new research challenges.
Then, the lack of robustness of fairness will result in \textbf{\textit{hallucination}} in LMMs. In general, \textit{hallucination refers to the problem of the model producing inaccurate, irrelevant, or unintelligible outputs}. 
In particular, the following research directions should be potentials to be explored in LMMs.

\begin{itemize}
    \item \textbf{Hallucinations Under Biased Data.} The imbalance in data distributions may be reflected via their generated outputs. As a result, the LMMs will hallucinate results consistent with the bias but not with reality.
    In future research, inspired by the success in fairness learning in this dissertation \cite{Truong:CVPR:2023FREDOM, truong2023fairness}, I will focus on addressing the hallucinations under biased data.

    \item \textbf{Hallucinations Under Limited Data.} The success of LMMs relies on large-scale data. However, in some cases, collecting large-scale data is impossible. The LMMs trained on this limited data scale may have hallucinations due to the lack of diversity.
    Therefore, given the success of the proposed unsupervised domain adaptation in this dissertation \cite{dat2021bimal_iccv, Truong:CVPR:2023FREDOM}, my future research will develop a novel adaptation learning that can train LMMs on limited while avoiding the hallucinations caused by limited data.

    \item \textbf{Hallucinations Under Misalignment Feature Representations.} Learning the feature alignment across input modalities is crucial in LMMs. LMMs are usually biased to language preferences since the large language model has been pre-trained on the exascale data. Meanwhile, due to the data complexity, the LMMs tend to overlook the information of other modalities, e.g., visual input. Hence, modality misalignment may lead to hallucinations.
    In future research, following the success of multimodal and temporal learning in this dissertation \cite{truong2021direcformer, truong2021right2talk}, I will primarily focus on developing robust representation learning in multimodal models to improve the alignment of features across input modalities.
\end{itemize}